%% file: main.tex
\pdfoutput=1
\documentclass[11pt]{article}
\usepackage[T1]{fontenc}
\usepackage[utf8]{inputenc}
\usepackage[margin=1in]{geometry}
\usepackage{placeins}
\input{preamble}

\usepackage{titletoc}
\usepackage{caption}
\usepackage{floatrow}
\usepackage{subcaption}
\usepackage{color}
\usepackage{tikz}
\usetikzlibrary{arrows.meta}

\usepackage[
  colorlinks=true,
  linkcolor=blue,
  citecolor=cyan,
  urlcolor=blue,
  pdfview=XYZ,
  pdfstartview=Fit,
  linktoc=all
]{hyperref}

\title{\bf Statistical Analysis of Synthetic Data with Verification: A Linear Regression Framework}

\author{Heejune Sheen \  and Harrison H. Zhou \\~\\
Department of Statistics and Data Science\\
Yale University\\
New Haven, CT 06511 USA \\
}

\date{}

\begin{document}
\makeatletter
\let\orig@hyper@refstepcounter\hyper@refstepcounter

\def\hyper@refstepcounter#1{%
  \edef\eq@counter@name{#1}%
  \ifx\eq@counter@name\name@of@eq
    \HyCnt@ProvideTheHCounter{#1}%
    \hyper@makecurrent{#1}%
    \Hy@ifnotmeasuring{%
      \hbox to 0pt{%
        \kern-\displaywidth
        \Hy@raisedlink{%
          \hyper@anchorstart{\@currentHref}%
          \hyper@anchorend
        }%
        \hss
      }%
    }%
  \else
    \orig@hyper@refstepcounter{#1}%
  \fi
}
\makeatother
\maketitle
\begingroup
\renewcommand{\thefootnote}{}
\footnotetext[0]{Email: heejune.sheen@yale.edu, harrison.zhou@yale.edu}
\endgroup

\input{sections/abstract.tex}

\input{sections/1introduction.tex}
\input{sections/2problem_setting.tex}
\input{sections/3main_results.tex}
\input{sections/4proof_sketch.tex}
\input{sections/5iterative_training.tex}
\input{sections/6empirical_studies.tex}
\input{sections/7conclusion.tex}

\bibliographystyle{ims}
\bibliography{reference}

\newpage
\appendix


\startcontents[appendices]
\section*{Contents of the Appendix}
\printcontents[appendices]{}{1}[1]{}

\input{Unified_appendix/app_00_preamble.tex}
\input{Unified_appendix/app_01A_probability_tools.tex}
\input{Unified_appendix/app_02B_pretraining_covariance.tex}
\input{Unified_appendix/app_03C_foundational_proofs.tex}
\input{Unified_appendix/app_04D_variance_framework.tex}
\input{Unified_appendix/app_07E_accepted_population_moments.tex}
\input{Unified_appendix/app_08F_empirical_comparisons.tex}
\input{Unified_appendix/app_09aG_conditional_risk_mean.tex}
\input{Unified_appendix/app_09bG5_conditional_risk_variance.tex}
\input{Unified_appendix/app_09cG6_weighted_population_variance.tex}
\input{Unified_appendix/app_10H1_small_u_analysis.tex}
\input{Unified_appendix/app_11H2_large_u_analysis.tex}
\input{Unified_appendix/app_12aI_regime_i_ii_results.tex}
\input{Unified_appendix/app_12bI_regime_iii_results.tex}
\input{Unified_appendix/app_12cI_regime_iv_results.tex}
\input{Unified_appendix/app_15J_moment_tools.tex}
\input{Unified_appendix/app_16K_iterative_training.tex}
\input{Unified_appendix/app_17L_empirical_details.tex}

\newpage
\end{document}

%% file: preamble.tex
\usepackage{amsmath, amsfonts, amssymb, amsthm}
\usepackage{mathtools}
\mathtoolsset{showonlyrefs=true}
\usepackage{bm}  
\usepackage{bbm} 
\usepackage{xcolor}
\usepackage[round]{natbib}
\usepackage[expansion=false]{microtype} 
\usepackage{array}
\usepackage{booktabs} 
\usepackage{enumitem} 

\newtheorem{theorem}{Theorem}[section]
\newtheorem{proposition}[theorem]{Proposition}
\newtheorem{lemma}[theorem]{Lemma}
\newtheorem{corollary}[theorem]{Corollary}
\newtheorem{definition}[theorem]{Definition}
\newtheorem{assumption}[theorem]{Assumption}

\DeclareMathOperator{\Var}{Var}
\DeclareMathOperator{\Cov}{Cov}
\DeclareMathOperator{\tr}{tr}

\DeclareMathOperator{\argmin}{argmin}

\allowdisplaybreaks 
\DeclareMathSizes{11}{11}{8}{6} 

\newcounter{regimectr}

\DeclareRobustCommand{\regimeid}[1]{%
  \begingroup
  \setcounter{regimectr}{#1}%
  \ensuremath{\mathrm{\Roman{regimectr}}}%
  \endgroup
}

\definecolor{highlight}{HTML}{0064E0}



%% file: sections/abstract.tex
\begin{abstract}

\noindent
Synthetic data are widely used for training modern AI systems. However,
their statistical understanding remains limited.  
It is well recognized in statistics that repeated self-training on synthetic data
can amplify errors inherited from pre-trained models, leading to model collapse.
Verification plays a crucial role in selecting high-quality synthetic data to help
prevent model collapse and improve model performance.
We study verified self-training in a linear regression framework,
where a pre-trained estimator generates synthetic data with injected generation noise
and a verifier selects samples according to a verification threshold.
Post-training on the accepted samples updates either all coordinates (full fine-tuning)
or only an active subset (partial fine-tuning).
Our analysis reveals three potentially counterintuitive effects.
We establish that tighter verification does not always lead to lower prediction risk.
We also find that a suitable level of injected generation noise can reduce prediction risk,
while excessive noise can increase it.
Moreover, full and partial fine-tuning
can respond differently to the same verification rule, 
since pre-training error in the frozen coordinates affects both sample selection and post-training risk.
We analyze these effects by deriving finite sample risk bounds for four separate regimes determined by
the relative scales of the verification threshold, 
generation noise, and pre-training error.
The bounds precisely characterize how verification affects prediction risk across these regimes.
In particular, we quantify how the generation noise, the effective accepted sample size, and
the covariance of the accepted covariates affect prediction risk.
The analysis extends to iterative verified full fine-tuning, where sufficiently tight verification
reduces the initial error contribution at a geometric rate. \\

\noindent
\textbf{Keywords:} Synthetic data, Fine-tuning, Verification, Prediction risk, Model collapse.
\end{abstract}

%% file: sections/1introduction.tex
\section{Introduction}

Synthetic data generated by models are increasingly used to supplement human-written training data.
Pre-trained models can produce large numbers of  responses and thereby increase the data available for post-training.
This approach is particularly valuable when additional labeled data are difficult to obtain.
However, a fundamental statistical difficulty arises since the model used to generate
 the additional training samples is itself estimated from finite data.
 Thus, synthetic data can carry errors from the pre-trained model, which can be propagated or even amplified through subsequent updates.
Without an adequate verification procedure, repeated self-training may lead to progressive performance degradation, 
a phenomenon known as model collapse
\citep{shumailov2024ai, dohmatob2024model, alemohammad2024self, gerstgrasser2024model,dey2024universality}.

Verification provides a practical approach by selecting only generated
samples that satisfy an accuracy criterion based on ground truth labels
\citep{zelikman2022star}, automated checks such as code execution
\citep{zhao2026absolute}, or judge models
\citep{zheng2023judging, gu2026survey}.
The effectiveness of fine-tuning with synthetic data depends on verification quality 
\citep{feng2025beyond, el2025maximizing, yi2026escaping}.
However, the effects of verification are not limited to the accuracy of the retained responses.
It also 
affects the number of accepted samples and the distribution of the data used for post-training. 
Thus, response accuracy by itself does not fully characterize the statistical effect of verification.

A theoretical understanding of training with verified synthetic data is important,
as this form of training  has not been fully characterized despite its growing use.
Three natural questions arise.
Under what conditions can post-training on verified self-generated data reduce prediction risk 
relative to the pre-trained model, and to what extent?
How does the verification threshold affect the accuracy and number
of synthetic responses available for post-training?
How does pre-training error propagate when only part of the parameter vector is updated, as in parameter efficient fine-tuning?

We study these questions in a linear regression framework and 
 quantify  how pre-training error, generation noise, verification,
and partial fine-tuning jointly determine prediction risk.
An ordinary least squares estimator of $\beta^*\in\mathbb R^d$ is constructed from $n$ real samples, which we refer to as the 
pre-trained estimator, $\hat\beta^{\mathrm{pre}}$.  
We independently draw $m$ downstream 
covariates and use the pre-trained estimator to generate
synthetic responses satisfying $\tilde y_j=\tilde x_j^\top\hat\beta^{\mathrm{pre}}+\tilde\varepsilon_j$, 
where $\tilde x_j$ is a downstream covariate and $\tilde\varepsilon_j$ is the injected generation noise with scale $\tilde\sigma$.
An oracle verifier accepts a generated pair if $|\tilde y_j-\tilde x_j^\top\beta^*|\le\delta$, where $\delta$ is the verification threshold.
This process is motivated by language model fine-tuning methods that filter
generated responses based on the correctness of their final answers
\citep{zelikman2022star, yuan2023scaling, setlur2024rl}.
In practice, verification can exploit known true answers, executable checks,
or judgments from a more capable large language model.
In our setting, the verifier evaluates response errors exactly, while the verification threshold $\delta$ determines the accuracy requirement for acceptance.
It provides only acceptance decisions, without revealing $\beta^*$ or the true regression values.
The accepted samples are used to construct the post-training estimator
$\hat\beta^{\mathrm{post}}$, which either updates all coordinates, referred to as full fine-tuning, or updates only
 an active set of $s\le d$
coordinates while keeping the remaining coordinates, called the frozen
coordinates, at their pre-trained values, referred to as partial fine-tuning.

 \begin{figure}[t]
  \centering
  \includegraphics[width=\textwidth]{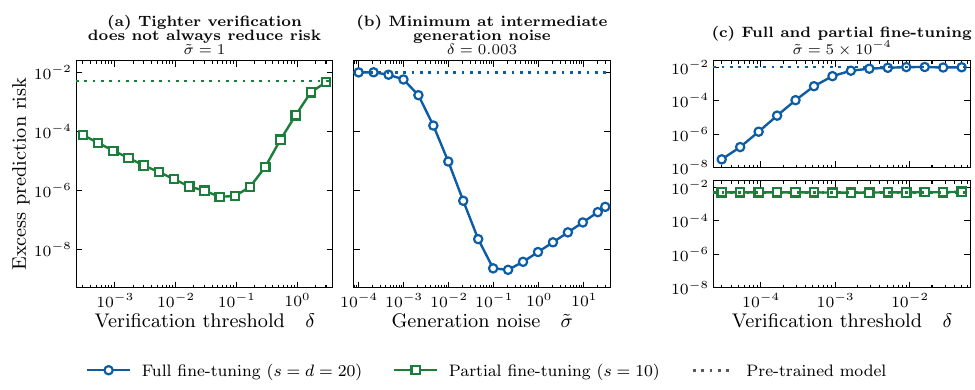}
  \caption{Excess prediction risk under verified post-training.
  Panel~(a) displays risk against the verification threshold under partial
  fine-tuning, panel~(b)  shows risk against generation noise under full
  fine-tuning. In panel~(c), we compare full and partial fine-tuning across
  verification thresholds when generation noise is small.  For partial
  fine-tuning, the excess risk excludes the direct contribution from the
  frozen coordinates.}
  \label{fig:intro_verification}
\end{figure}

We observe three potentially counterintuitive phenomena in verified post-training.
A natural expectation is that tighter verification always improves prediction risk and that additional 
generation noise can only worsen it.
However,
this is not always true, as we observe in
Figure~\ref{fig:intro_verification}.  
First, in panel~(a), we notice that tighter verification can
increase excess prediction risk under partial fine-tuning.
As the verification threshold decreases, risk first decreases and
then increases, producing a U-shaped curve.
Second, panel~(b) illustrates 
that small injected generation noise is not always preferable, since prediction
risk is minimized at an intermediate injected noise level.
Third, we observe the distinct behavior of full and partial fine-tuning in panel~(c). 
Under suitable conditions,
a verification threshold below the generation noise scale sharply reduces excess
prediction risk for full fine-tuning, whereas the corresponding risk
contribution remains close to its pre-training scale under partial fine-tuning.
These phenomena motivate a
more precise analysis of how the verification threshold, generation noise,
and the choice between full and partial fine-tuning jointly affect the resulting
prediction risk.

In the following informal theorem, we make the phenomena underlying Figure~\ref{fig:intro_verification}
explicit 
and establish a unified leading risk rate for partial and full fine-tuning.
The formal results and precise sample conditions are given in
Section~\ref{section:main_results}.
Let $\mathrm{ExcessRisk}$ denote the difference between the prediction risk of
the post-training estimator and the minimum prediction risk over the active
coordinates, with the frozen coordinates fixed at their pre-trained values.

\begin{theorem}[Informal Fine-Tuning Rate]
Suppose that $n\ge Cd$ and $1\le s\le d$.  Define the acceptance probability scale
$p:=\min(1,\delta/\sqrt{\tilde\sigma^2+d/n})$.  If the effective accepted
sample size $mp$ is sufficiently large, then, with high probability, $\mathrm{ExcessRisk}$ is of order
\begin{align*}
\underbrace{\frac{s}{n}\biggl(
\left(\frac{d-s}{n\tilde\sigma^2+d-s}\right)^2
+\min(1,(\delta/\tilde\sigma)^4)
\biggr)}_{\text{pre-training bias}}
+
\underbrace{\frac{s g_{d-s,d}(\sqrt{d/n}/\max(\tilde\sigma,\delta))}{mp}\left(
\min(\delta^2,\tilde\sigma^2)
+\frac{\tilde\sigma^2(d-s)}{n\tilde\sigma^2+d-s}
\right)}_{\text{post-training variance}}.
\end{align*}
\end{theorem}

From the first term, which represents the expected squared conditional bias inherited from pre-training,
we see that, when $\delta/\tilde\sigma$ is small, generation noise can mitigate
the error from pre-training among the accepted samples by a factor of
$(\delta/\tilde\sigma)^4$, while the error in the frozen coordinates limits
the extent of this reduction through the term
$((d-s)/(n\tilde\sigma^2+d-s))^2$.
From the second term, which represents the expected conditional post-training variance induced by injected generation noise, we see
a verification threshold tradeoff in minimizing the excess risk.
The factor $g_{d-s,d}$ reflects the change in the covariate distribution induced by verification and equals one when $d-s\ge3$.
For full fine-tuning, $s=d$, the frozen coordinate terms in both parts disappear, while the variance factor becomes $g_{0,d}(\sqrt{d/n}/\max(\tilde\sigma,\delta))=1+1/(n\max(\tilde\sigma^2,\delta^2))$.

In the formal results, we present this characterization across four regimes determined by the relative scales of the
verification threshold~$\delta$, the injected generation noise scale~$\tilde\sigma$,
and the typical pre-training error scale $\sqrt{d/n}$.
The four regimes reveal a common principle.
 We show that the effect of verification
depends on  the threshold relative to generation noise and on the covariate
variation 
preserved among the accepted samples.
 If the accepted covariates remain representative of the downstream population, the threshold is sufficiently 
small relative to the injected generation noise scale, and sufficiently many samples are accepted, then
post-training can reduce the leading contribution from the pre-trained estimator.
By contrast, when the pre-training error dominates the injected generation noise, the improvement can be less substantial, 
as verification tends to select samples already well fitted by the pre-trained estimator.
The reduction is further limited if the injected generation noise 
does not exceed the verification threshold 
and the threshold lies below the pre-training error scale. In this regime, the leading contribution inherited 
from pre-training remains of order $s/n$.

We also characterize the difference in prediction risk between partial and full fine-tuning. Under partial fine-tuning
 ($s<d$), 
 the pre-training error in the frozen coordinates affects both verification and prediction risk, 
 and its interaction with the updated coordinates can persist, while this interaction does not appear under full fine-tuning ($s=d$).
Moreover, we extend the analysis to iterative verified full fine-tuning and characterize how estimation error propagates across generations.
Finally, we  complement our  theoretical results with Gaussian simulations and real data
 experiments that illustrate the effects of the verification threshold, generation noise, and the choice between partial and full fine-tuning.

\subsection{Contributions}

Our main contributions are:

\begin{itemize}
\item \textbf{Nonasymptotic risk characterization.}
We establish nonasymptotic prediction risk bounds in four regimes
determined by the relative scales of the verification threshold, the
generation noise, and the pre-training error.  
The bounds identify the underlying effects of accepted response error,
effective sample size, and covariate variation that lead to counterintuitive phenomena.


\item \textbf{Partial and full fine-tuning.}
We show that verification can have fundamentally different effects  under
partial and full fine-tuning.
When the generation noise dominates the
pre-training error, tighter verification reduces the active coordinate
contribution inherited from pre-training but can increase the contribution
induced by the frozen coordinates through the smaller effective number of
accepted samples.  This tradeoff can lead to a U-shaped risk under partial
fine-tuning and does not appear under full fine-tuning.  However, when the
pre-training error dominates the generation noise, the reduced variation of
the accepted covariates can have a stronger effect under full fine-tuning.

\item \textbf{Iterative verified full fine-tuning.}
We extend the analysis to repeated verified full fine-tuning and establish
upper and lower bounds for the propagation of estimation error across
generations. When the verification threshold is sufficiently small relative
to the  generation noise, the contribution from the initial error contracts
geometrically.  The lower bound further shows that the initial error cannot
decrease rapidly when the verification threshold is large relative to the
generation noise.

\end{itemize}

\subsection{Related Work}

\textbf{Model collapse.}
A growing literature studies model collapse, the degradation that can occur when models are repeatedly trained on self-generated data \citep{shumailov2024ai}.
The phenomenon has been analyzed theoretically in linear regression 
\citep{gerstgrasser2024model, dey2024universality, garg2026preventing}, 
distribution estimation \citep{kanabar2025model, xu2025probabilistic}, 
and through changes in scaling laws \citep{dohmatob2025strong}. Prior work shows 
that estimation error inherited from earlier generations can persist through subsequent 
rounds of synthetic training. 
To mitigate the resulting degradation, several strategies have been proposed.
One approach retains or accumulates real data across generations 
\citep{gerstgrasser2024model, dey2024universality, kanabar2025model}. 
Another approach mixes synthetic data with fresh real data with an optimized 
mixing weight \citep{he2026recursive, garg2026preventing}. In contrast, we study 
whether self-generated samples themselves can be made useful through verification, 
without relying on additional real training responses.

\textbf{Self-distillation and self-generated data.}
A related line of work studies when models can benefit from training on their own predictions. 
Empirically, self-generated data have been shown to improve performance 
in a range of settings \citep{zelikman2022star, setlur2024rl, song2025mind, zhao2026self, zhao2026absolute}. 
Theoretical analyses consider self-distillation in classification 
\citep{frei2022self, das2023understanding, das2024retraining, dang2026prediction} 
and regression \citep{das2023understanding, dang2026prediction, lecoiu2026self, 
wuself, moniri2026mechanisms}. 
These results show that training on model predictions can improve performance over the initial estimator under suitable conditions.
Motivated by these results, we study how external verification of
self-generated responses affects prediction risk under partial and full
fine-tuning.

\textbf{Verification and data selection.}
Verification has become an important mechanism for selecting useful synthetic data
 in practice \citep{zelikman2022star, setlur2024rl, song2025mind, zhao2026absolute}, 
 and empirical evidence shows that a verified subset can substantially outperform synthetic data 
 without verification and, in some settings, even the full human-written dataset \citep{feng2025beyond, amin2026escaping}. 
 Recently, for theoretical understanding, \citet{feng2025beyond} characterize when an imperfect verifier 
 can recover Bayes optimal classification through its acceptance rates for correct and incorrect labels, 
 while \citet{el2025maximizing}  quantify the effect of 
 generator estimation error even under perfect label verification. Related work considers filtering 
 in recursive training \citep{fu2026self}, weak external supervision \citep{amin2026escaping}, 
 preference-based curation \citep{ferbach2024self}, and the scaling of verified sample sizes 
 across rounds \citep{yang2026spend}. These analyses focus on classification or iterative generative learning.
For linear regression, covariance based synthetic data selection
without verification has been studied \citep{rezaei2026high}.
 
 Most closely related, \citet{yi2026escaping} study
 synthetic retraining with verification in linear regression and characterize both the improvement
 from one round and the behavior under repeated retraining. Their theoretical setting uses fixed
 orthogonal covariate directions, which reduces linear regression to separate one-dimensional
 Gaussian mean estimation problems with a fixed verification threshold and a fixed generation noise scale.
 We instead draw synthetic covariates from the downstream distribution and allow the verification
 threshold $\delta$ and generation noise scale $\tilde\sigma$ to vary relative to the pre-training
 error scale $\sqrt{d/n}$. Under this general setting, our nonasymptotic analysis yields four distinct risk regimes
 arising from the interaction between the effective accepted sample size and the covariance of the
 accepted covariates. These rates do not follow from a direct extension of the fixed orthogonal
 setting. Moreover, we also study partial fine-tuning, where the contribution from the frozen
 coordinates leads to a different risk characterization.

\subsection{Organization.}
In Section~\ref{sec:problem_setting}, we introduce the data model,
verification rule, and post-training estimator.
In Section~\ref{section:main_results}, we establish the
nonasymptotic risk bounds for partial and full fine-tuning across the
four regimes.  In Section~\ref{sec:proof_sketch}, we provide a proof sketch of the main results.
In Section~\ref{sec:iterative_training}, we extend the
analysis to repeated verified full fine-tuning.  In
Section~\ref{sec:empirical_studies}, we present Gaussian simulations and real
data experiments.  In Section~\ref{sec:conclusion_main}, we conclude the
paper.  The appendices contain the
detailed proofs, auxiliary results, and further details for the experiments.

\subsection{Notation.}
We denote the standard normal distribution function and density by $\Phi$ and $\phi$, respectively.
For vectors, $\|x\|=\|x\|_2$, and for a positive semidefinite matrix $M$, $\|x\|_M^2=x^\top Mx$.
The trace of $A$ is denoted by $\tr(A)$.
For $S\subseteq\{1,\ldots,d\}$, let $-S$ denote its complement and let $A_{SS}$ and $A_{S,-S}$ denote the corresponding matrix blocks.
We denote the indicator function by $\mathbf 1\{\cdot\}$.
Unless stated otherwise, $c,C$ denote positive constants independent of $n,m,d,s,\delta,\tilde\sigma$.
The relations $a\lesssim b$ and $a\gtrsim b$ denote the bounds $a\le Cb$ and $a\ge cb$, respectively.
For nonnegative $a,b$, the notation $a\asymp b$ denotes the bounds $cb\le a\le Cb$.
The notation $a=O(b)$ denotes the bound $|a|\le Cb$.

%% file: sections/2problem_setting.tex
\section{Problem Setting}\label{sec:problem_setting}

\subsection{Pre-Training Data and the Pre-Trained Estimator}

We consider pre-training data with sample size $n$ and true coefficient vector $\beta^*\in\mathbb R^d$.
Let $X_{\mathrm{pre}}\in\mathbb R^{n\times d}$ be a matrix with rows $x_{\mathrm{pre},i}^{\top}$, and let $Y_{\mathrm{pre}}\in\mathbb R^n$ and $\varepsilon_{\mathrm{pre}}\in\mathbb R^n$ denote the response and noise vectors, respectively.  They satisfy
\begin{align*}
Y_{\mathrm{pre}}=X_{\mathrm{pre}}\beta^*+\varepsilon_{\mathrm{pre}},\qquad
x_{\mathrm{pre},i}\sim\mathcal N(0,\Sigma),\qquad
\varepsilon_{\mathrm{pre}}\sim\mathcal N(0,I_n).
\end{align*}
We assume that $x_{\mathrm{pre},1},\ldots,x_{\mathrm{pre},n}$ are independent and that $\varepsilon_{\mathrm{pre}}$ is independent of $X_{\mathrm{pre}}$.
For the downstream task, the covariate and response satisfy
\begin{align*}
y=x^\top\beta^*+\varepsilon,\qquad
x\sim\mathcal N(0,\Sigma_1),\qquad
\mathbb E[\varepsilon\mid x]=0,\qquad
\Var(\varepsilon\mid x)=\sigma_1^2.
\end{align*}
Furthermore, we assume that the pre-training and downstream covariances are well conditioned.

\begin{assumption}[Covariance regularity]\label{ass:design}
There exist fixed constants $0<c_\lambda\le C_\lambda<\infty$ such that
\begin{align*}
c_\lambda I_d\preceq\Sigma\preceq C_\lambda I_d,\qquad
c_\lambda I_d\preceq\Sigma_1\preceq C_\lambda I_d.
\end{align*}
\end{assumption}
\noindent
For $n\ge d$, the ordinary least squares pre-trained estimator is $\hat\beta^{\mathrm{pre}}=(X_{\mathrm{pre}}^\top X_{\mathrm{pre}})^{-1}X_{\mathrm{pre}}^\top Y_{\mathrm{pre}}$.
Define the pre-training error by
$\Delta=\beta^*-\hat\beta^{\mathrm{pre}}$.  We then have
\begin{align*}
\Delta
&=-(X_{\mathrm{pre}}^\top X_{\mathrm{pre}})^{-1}
X_{\mathrm{pre}}^\top\varepsilon_{\mathrm{pre}},
&\Delta\mid X_{\mathrm{pre}}
&\sim\mathcal N(0,\Gamma/n),
&\Gamma
&=\bigl(n^{-1}X_{\mathrm{pre}}^\top X_{\mathrm{pre}}\bigr)^{-1}.
\end{align*}
For $n\ge Cd$, the normalized Gram matrix satisfies $(c_\lambda/2)I_d\preceq X_{\mathrm{pre}}^\top X_{\mathrm{pre}}/n\preceq2C_\lambda I_d$ with probability at least $1-2e^{-cn}$, as presented in Lemma~\ref{lem:pretraining_covariance_regularity}.

\subsection{Verified Synthetic Data}

Given the pre-trained estimator, we draw $m$ covariates from the
downstream distribution and generate the corresponding synthetic
responses.  In the language model interpretation, the synthetic covariates $\tilde x_j$ can be considered as the features of queries for the task,
and the synthetic responses $\tilde y_j$ represent the generated answers.  For $j=1,\ldots,m$,
\begin{align*}
\tilde y_j
&=\tilde x_j^\top\hat\beta^{\mathrm{pre}}+\tilde\varepsilon_j,
&\tilde x_j
&\sim\mathcal N(0,\Sigma_1),
&\tilde\varepsilon_j
&\sim\mathcal N(0,\tilde\sigma^2),
\end{align*}
where $\tilde\sigma>0$ denotes the scale of the injected generation noise.
The scale $\tilde\sigma$ represents the amount of injected randomness,
analogous to the role of sampling temperature in large language model
response generation \citep{renze2024effect}.  We assume
that the synthetic covariate $\tilde x_j$ and generation noise $\tilde\varepsilon_j$  are mutually
independent and independent of the pre-training data. Let
$\tilde X\in\mathbb R^{m\times d}$ denote the matrix with rows
$\tilde x_j^\top$, and let
$\tilde\varepsilon=(\tilde\varepsilon_1,\ldots,\tilde\varepsilon_m)^\top$.

Let $\delta\in(0,\infty]$ be the verification threshold.  An oracle
verifier accepts the $j$th pair when
$|\tilde y_j-\tilde x_j^\top\beta^*|\le\delta$.
The threshold $\delta$ specifies the accuracy criterion for verification,
corresponding to sample selection based on response correctness or reward
scores in large language model fine-tuning
\citep{zelikman2022star, gulcehre2023reinforced}.  Although the oracle verifier evaluates response errors exactly,
accepted responses may have error up to $\delta$.  Correspondingly, we define the
generated response error and acceptance indicator by
\begin{align*}
\xi_j
&:=\tilde y_j-\tilde x_j^\top\beta^*
=-\tilde x_j^\top\Delta+\tilde\varepsilon_j,
&
\mathbb I_j
&:=\mathbf 1\{|\xi_j|\le\delta\},
\qquad j=1,\ldots,m.
\end{align*}
The representation of $\xi_j$ reveals that verification depends jointly
on the pre-trained prediction error $-\tilde x_j^\top\Delta$ and the
injected generation noise $\tilde\varepsilon_j$.  We refer to samples $\{(\tilde x_j, \tilde y_j)\}_{j=1}^m$ with
$\mathbb I_j=1$ as \emph{accepted samples}.
Note that we consider Gaussian distributions for the synthetic covariates and generation noise,
since this allows us to obtain explicit expressions for the acceptance probability and conditional moments of accepted samples. 

\subsection{Post-Training Estimator}

Let $S\subseteq\{1,\ldots,d\}$ be a fixed active coordinate set,
independent of the data, with $|S|=s$ and $1\le s\le d$.  We refer to
its complement $-S$ as the frozen coordinate set.
Motivated by parameter efficient fine-tuning in practice \citep{houlsby2019parameter}, post-training
updates the coordinates in $S$ by least squares on the accepted samples
and keeps the coordinates in $-S$ at their pre-trained values, which leads to
$\hat\beta_{-S}^{\mathrm{post}}=\hat\beta_{-S}^{\mathrm{pre}}$.
The cases $s<d$ and $s=d$ correspond to partial and full fine-tuning,
respectively.
When the accepted active coordinate Gram matrix $\sum_{j=1}^m\mathbb I_j\tilde x_{j,S}\tilde x_{j,S}^\top$
is invertible, the
active coordinates of the post-training estimator are
\begin{align*}
\hat\beta_S^{\mathrm{post}}
&=
\bigg(\sum_{j=1}^m
\mathbb I_j\tilde x_{j,S}\tilde x_{j,S}^\top\bigg)^{-1}
\sum_{j=1}^m
\mathbb I_j\tilde x_{j,S}
\bigl(\tilde y_j-\tilde x_{j,-S}^\top
\hat\beta_{-S}^{\mathrm{pre}}\bigr),
\end{align*}
where $\tilde x_{j,S},\hat\beta_S^{\mathrm{post}}\in\mathbb R^s$.
Since the frozen coordinates retain their pre-training error, the
downstream prediction risk over the active coordinates is not necessarily
minimized at $\beta_S^*$.  Therefore, with
$\hat\beta_{-S}^{\mathrm{pre}}=\beta_{-S}^*-\Delta_{-S}$ fixed, we define
the corresponding downstream population target by
\begin{align*}
\beta_S^{\mathrm{corr}}
:=
\argmin_{\beta_S\in\mathbb R^s}
\mathbb E_x\bigl[
\bigl(
x_S^\top\beta_S
+x_{-S}^\top\hat\beta_{-S}^{\mathrm{pre}}
-x^\top\beta^*
\bigr)^2
\bigr] 
=
\beta_S^*
+\Sigma_{1,SS}^{-1}\Sigma_{1,S,-S}\Delta_{-S}.
\end{align*}
Thus, under partial fine-tuning, the pre-training error $\Delta_{-S}$
can shift the active coordinate target through the covariance between the
active and frozen coordinates.  Under full fine-tuning, the frozen
block is empty and $\beta_S^{\mathrm{corr}}=\beta_S^*$.
Figure~\ref{fig:verified_pipeline} summarizes the four steps of verified post-training.
\input{figures/fig_paper_pipeline.tex}

\subsection{Prediction Risk}

Let $\hat\beta$ be any estimator constructed from the pre-training and
synthetic data, and let $(x,y)$ be an independent test sample from the
downstream model.  Conditioned on the realized covariates
$(X_{\mathrm{pre}},\tilde X)$, define its prediction risk by
\begin{equation}\label{eq:R1_def}
\begin{aligned}
R_1\bigl(\hat\beta;X_{\mathrm{pre}},\tilde X\bigr)
:=
\mathbb E\bigl[(y-x^\top\hat\beta)^2
\big|X_{\mathrm{pre}},\tilde X\bigr]
=
\sigma_1^2+
\mathbb E\bigl[\|\hat\beta-\beta^*\|_{\Sigma_1}^2
\big|X_{\mathrm{pre}},\tilde X\bigr].
\end{aligned}
\end{equation}
The expectation is over the pre-training noise
$\varepsilon_{\mathrm{pre}}$, the generation noise $\tilde\varepsilon$,
and the independent test sample $(x,y)$. In the next section, we
decompose the risk and state the regime specific rates for
verified post-training.

%% file: figures/fig_paper_pipeline.tex
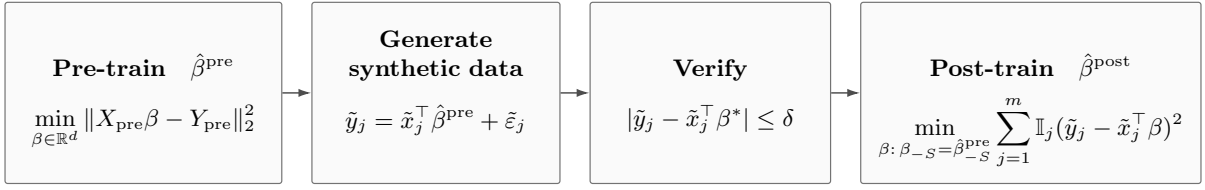
\begin{figure}[!t]
\centering
\begin{tikzpicture}[
  stage/.style={draw=black!55, fill=black!2, line width=0.45pt,
    rounded corners=1pt, minimum height=2.4cm,
    inner sep=4pt, align=center, font=\footnotesize},
  stage title/.style={anchor=south, align=center, font=\footnotesize, inner sep=0pt, yshift=0.15cm},
  stage formula/.style={anchor=north, align=center, font=\footnotesize, inner sep=0pt, yshift=-0.15cm},
  flow/.style={-{Latex[length=1.7mm,width=1.1mm]}, draw=black!70,
    line width=0.55pt}
]
\node[stage, text width=3.38cm] (pre) at (0,0) {};
\node[stage, text width=3.0cm] (gen) at (3.87,0) {};
\node[stage, text width=2.9cm] (ver) at (7.5,0) {};
\node[stage, text width=4.22cm] (post) at (11.74,0) {};
\node[stage title] at (pre.center) {{\bfseries Pre-train}\quad $\hat\beta^{\mathrm{pre}}$};
\node[stage title, text width=3.0cm] at (gen.center) {{\bfseries Generate\\synthetic data}};
\node[stage title] at (ver.center) {{\bfseries Verify}};
\node[stage title] at (post.center) {{\bfseries Post-train}\quad $\hat\beta^{\mathrm{post}}$};
\node[stage formula] at (pre.center) {$\displaystyle\min\limits_{\beta\in\mathbb R^d}\|X_{\mathrm{pre}}\beta-Y_{\mathrm{pre}}\|_2^2$};
\node[stage formula] at (gen.center) {$\tilde y_j=\tilde x_j^\top\hat\beta^{\mathrm{pre}}+\tilde\varepsilon_j$};
\node[stage formula] at (ver.center) {$|\tilde y_j-\tilde x_j^\top\beta^*|\le\delta$};
\node[stage formula, yshift=0.12cm] at (post.center) {$\displaystyle\min\limits_{\beta:\,\beta_{-S}=\hat\beta_{-S}^{\mathrm{pre}}}\sum_{j=1}^{m}\mathbb I_j(\tilde y_j-\tilde x_j^\top\beta)^2$};
\draw[flow] (pre.east) -- (gen.west);
\draw[flow] (gen.east) -- (ver.west);
\draw[flow] (ver.east) -- (post.west);
\end{tikzpicture}
\caption{Post-training with verified synthetic data.}
\label{fig:verified_pipeline}
\end{figure}

%% file: sections/3main_results.tex
\section{Main Results}\label{section:main_results}

We first decompose the prediction risk and present the baseline
without verification. Then, we present and compare the partial and full
fine-tuning rates across the four regimes.

\subsection{Risk Decomposition}
\label{ssec:risk_decomp_main}

The prediction risk consists of the downstream noise variance, the
active coordinate error, and the contribution from the frozen coordinates.
Let $\Sigma_{1,-S\mid S}:=\Sigma_{1,-S,-S}-\Sigma_{1,-S,S}\Sigma_{1,SS}^{-1}\Sigma_{1,S,-S}$
denote the Schur complement of $\Sigma_{1,SS}$ in $\Sigma_1$.  By the
notation, we decompose the prediction risk as follows.

\begin{proposition}[Risk Decomposition]\label{prop:risk_decomp}
Under the model in Section~\ref{sec:problem_setting}, let $\hat\beta$
be any post-training estimator satisfying
$\hat\beta_{-S}=\hat\beta_{-S}^{\mathrm{pre}}$.
Conditioned on $X_{\mathrm{pre}}$ and $\tilde X$,
\begin{align*}
R_1\bigl(\hat\beta;X_{\mathrm{pre}},\tilde X\bigr)
=
\sigma_1^2+\mathrm{Bias}_S^2+\Var_S+\Var_{-S},
\end{align*}
where $\mathrm{Bias}_S^2:=\|\mathbb E[\hat\beta_S-\beta_S^{\mathrm{corr}}\mid X_{\mathrm{pre}},\tilde X]\|_{\Sigma_{1,SS}}^2$, and
\begin{align*}
\Var_S
&:=\tr(\Sigma_{1,SS}\Cov(\hat\beta_S-\beta_S^{\mathrm{corr}}\mid X_{\mathrm{pre}},\tilde X)),
&
\Var_{-S}
&:=\tr(\Sigma_{1,-S\mid S}\Cov(\Delta_{-S}\mid X_{\mathrm{pre}})).
\end{align*}
\end{proposition}

The decomposition is proved in Appendix~\ref{sec:foundations}. 
 Let $G(\Delta):=\sum_{j=1}^m\mathbb I_j\tilde x_{j,S}\tilde x_{j,S}^\top$ denote the accepted active coordinate Gram matrix for the realized 
 synthetic covariates $\tilde X$.  Since $G(\Delta)$ can be singular, we consider the following extended estimator.

\begin{definition}[Extended Post-Training Estimator]\label{def:ext_estimator}
Fix $C_G\ge 2$.  Let $\Omega_{\mathrm{post}}:=\{\lambda_{\min}(G(\Delta)/m)\ge m^{-C_G}\}$.  We define the extended post-training estimator as the post-training estimator on $\Omega_{\mathrm{post}}$ and the pre-trained estimator on its complement:
\begin{align*}
\hat\beta_{S,\mathrm{ext}}^{\mathrm{post}} :=
\begin{cases}
\hat\beta_S^{\mathrm{post}},&\text{on }\Omega_{\mathrm{post}},\\
\hat\beta_S^{\mathrm{pre}},&\text{on }\Omega_{\mathrm{post}}^{\mathrm c},
\end{cases},\qquad
\hat\beta_{-S,\mathrm{ext}}^{\mathrm{post}}
 :=\hat\beta_{-S}^{\mathrm{pre}}.
\end{align*}
\end{definition}
To analyze the active coordinate contribution to the risk, let $r_S:=\hat\beta_S^{\mathrm{post}}-\beta_S^{\mathrm{corr}}$ denote the post-training residual on $\Omega_{\mathrm{post}}$.
We define $r_S\mathbf 1_{\Omega_{\mathrm{post}}}$ to be zero on $\Omega_{\mathrm{post}}^{\mathrm c}$.
The corresponding residual for the extended estimator is
\begin{equation}\label{eq:r_ext}
r_{S,\mathrm{ext}}
:=\hat\beta_{S,\mathrm{ext}}^{\mathrm{post}}-\beta_S^{\mathrm{corr}}
=r_S\mathbf 1_{\Omega_{\mathrm{post}}}
-\Sigma_{1,SS}^{-1}(\Sigma_1\Delta)_S
\mathbf 1_{\Omega_{\mathrm{post}}^{\mathrm c}}.
\end{equation}
We consider this extended estimator when evaluating the downstream
prediction risk.  It follows from
Lemma~\ref{lem:zero_active_bias} that $\mathrm{Bias}_S^2=0$.
Furthermore, since $\Cov(\Delta\mid X_{\mathrm{pre}})=\Gamma/n$, we also have
$\Var_{-S}=\tr(\Sigma_{1,-S\mid S}\Gamma_{-S,-S})/n$.
Thus, it remains to analyze the active coordinate variance $\Var_S$.

For the remainder of this section, $\Var_S$ denotes the
active coordinate variance of the extended estimator.  Conditioned on
$(X_{\mathrm{pre}},\tilde X)$, the law of total variance applied to
$r_{S,\mathrm{ext}}$ yields $\Var_S=\mathcal V_1+\mathcal V_2$, where
\begin{equation}\label{eq:V1V2_def}
\begin{aligned}
\mathcal V_1
&:=\tr(\Sigma_{1,SS}\Cov(\mathbb E_{\tilde\varepsilon}[r_{S,\mathrm{ext}}\mid\Delta,\tilde X]\mid X_{\mathrm{pre}},\tilde X)),\\
\mathcal V_2
&:=\mathbb E[\tr(\Sigma_{1,SS}\Cov_{\tilde\varepsilon}(r_{S,\mathrm{ext}}\mid\Delta,\tilde X))\mid X_{\mathrm{pre}},\tilde X].
\end{aligned}
\end{equation}
Here, $\mathcal V_1$ is the expectation of the squared \emph{conditional bias}
inherited from the pre-training error, while $\mathcal V_2$ is the
expectation of the \emph{conditional post-training variance} induced by
injected generation noise.
For fixed $\Delta$ and $\tilde X$, the conditional bias is the difference between $\mathbb E_{\tilde\varepsilon}[\hat\beta_{S,\mathrm{ext}}^{\mathrm{post}}\mid\Delta,\tilde X]$ and the downstream population target $\beta_S^{\mathrm{corr}}$.
The conditional variance describes the variation of $\hat\beta_{S,\mathrm{ext}}^{\mathrm{post}}$ around this conditional expectation,
as injected generation noise affects both the synthetic responses and which samples are accepted.
These two terms determine the risk rates under the four verification
regimes introduced in Section~\ref{sec:detailed_regime_results}.
Before presenting the main results, we first consider post-training without verification, which provides a baseline for evaluating the effect of verification.

\subsection{Baseline without Verification}\label{sec:no_filter}
For the baseline, we consider $\delta=\infty$, where every generated sample is accepted,
and post-training uses all $m$ synthetic samples.  In this case,
$G(\Delta)=G:=\sum_{i=1}^m\tilde x_{i,S}\tilde x_{i,S}^{\top}$, and the
variance components $\mathcal V_1$ and $\mathcal V_2$ admit explicit forms.

\begin{proposition}[Baseline without Verification]
\label{prop:unfiltered}
Let $c_1>0$ and $1\le s\le d$, and suppose that
Assumption~\ref{ass:design} holds and $\delta=\infty$.
If $n\ge Cd$ and $m\ge C(s+\log m)$, then, with probability
at least $1-2e^{-cn}-m^{-c_1}$ jointly over
$(X_{\mathrm{pre}},\tilde X)$, the post-training estimator satisfies
\begin{align*}
R_1-\sigma_1^2-\Var_{-S}
=\underbrace{\frac{1}{n}\tr\bigl(
\Sigma_{1,SS}^{-1}(\Sigma_1\Gamma\Sigma_1)_{SS}
\bigr)}_{\mathcal V_1}
+
\underbrace{\tilde\sigma^2\tr(\Sigma_{1,SS}G^{-1})}_{\mathcal V_2}
\asymp\frac{s}{n}+\frac{s\tilde\sigma^2}{m}.
\end{align*}
In particular, when $s=d$, we have
\begin{align*}
R_1-\sigma_1^2
=\underbrace{\frac{1}{n}\tr(\Sigma_1\Gamma)}_{\mathcal V_1}
+
\underbrace{\tilde\sigma^2\tr(\Sigma_1G^{-1})}_{\mathcal V_2}
\asymp\frac{d}{n}+\frac{d\tilde\sigma^2}{m}.
\end{align*}
\end{proposition}

The proposition is proved in Appendix~\ref{sec:foundations}.  To assess improvement over pre-training,
we note that the pre-trained estimator itself has conditional
prediction risk $\sigma_1^2+\tr(\Sigma_1\Gamma)/n$, with the excess risk rate
of order $d/n$.
Thus, post-training without verification leaves the contribution from the pre-training
error rate unchanged while adding post-training variance $\mathcal V_2\asymp s\tilde\sigma^2/m$ as
we inject generation noise during the synthetic data generation. 

\subsection{Verification Regimes}\label{sec:detailed_regime_results}

Now, we consider $0<\delta<\infty$, under which verification changes both
the number and distribution of samples employed for post-training.  Since
$\xi_j=-\tilde x_j^\top\Delta+\tilde\varepsilon_j$, the effect of
verification depends on the relative magnitudes of the verification
threshold $\delta$, the injected generation noise scale $\tilde\sigma$, and the
pre-training error scale $\sqrt{d/n}$.  Therefore, we distinguish the
following four regimes.
\begin{align*}
\begin{array}{lll}
\text{Regime~I}:&
\displaystyle\sqrt{d/n}\lesssim\delta,
&\text{loose verification},\\
\text{Regime~II}:&
\displaystyle\delta\lesssim\sqrt{d/n}\lesssim\tilde\sigma,
&\text{tight verification, noise-dominated},\\
\text{Regime~III}:&
\displaystyle\delta\lesssim\tilde\sigma\lesssim\sqrt{d/n},
&\text{tight verification, pre-training error-dominated},\\
\text{Regime~IV}:&
\tilde\sigma\lesssim\delta\lesssim\sqrt{d/n},
&\text{low injected generation noise}.
\end{array}
\end{align*}

\begin{figure}[t]
  \centering
  \includegraphics[width=\textwidth]{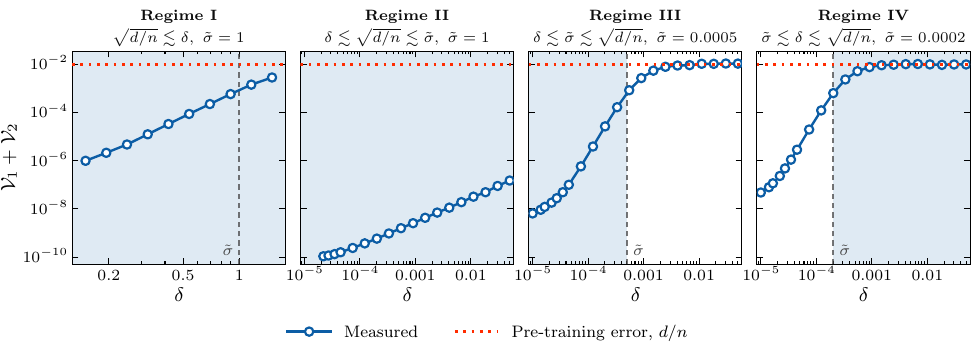}
  \caption{Full fine-tuning with $s=d=20$, $n=2000$, and
    $m=3\times10^6$. Simulation estimates of $\mathcal V_1+\mathcal V_2$
    are plotted against the verification threshold $\delta$, with
    $\tilde\sigma$ fixed in each panel. Tighter verification reduces the
    variance in Regime~I even when $\delta$ exceeds $\sqrt{d/n}$. The
    variance is smallest in Regime~II, increases in Regime~III, and remains
    near $d/n$ in Regime~IV, where tight verification provides little
    improvement when generation noise is too small.}
  \label{fig:full_delta}
\end{figure}

\begin{figure}[t]
  \centering
  \includegraphics[width=\textwidth]{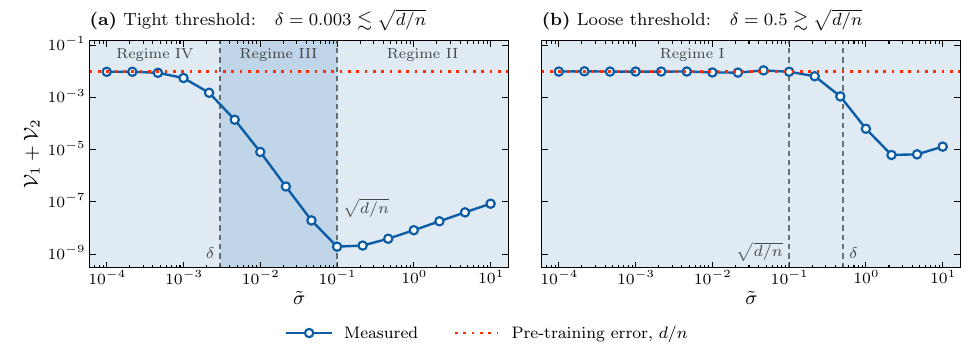}
  \caption{Full fine-tuning with $s=d=20$, $n=2000$, and
    $m=3\times10^6$. Simulation estimates of $\mathcal V_1+\mathcal V_2$
    are plotted against the generation noise $\tilde\sigma$ under
    (a) tight and (b) loose verification. In both cases, the variance is
    minimized at an intermediate generation noise level. Nearly deterministic
    generation leaves the variance near $d/n$, whereas excessive generation
    noise increases it.}
  \label{fig:full_sigma}
\end{figure}

\begin{figure}[t]
  \centering
  \includegraphics[width=\textwidth]{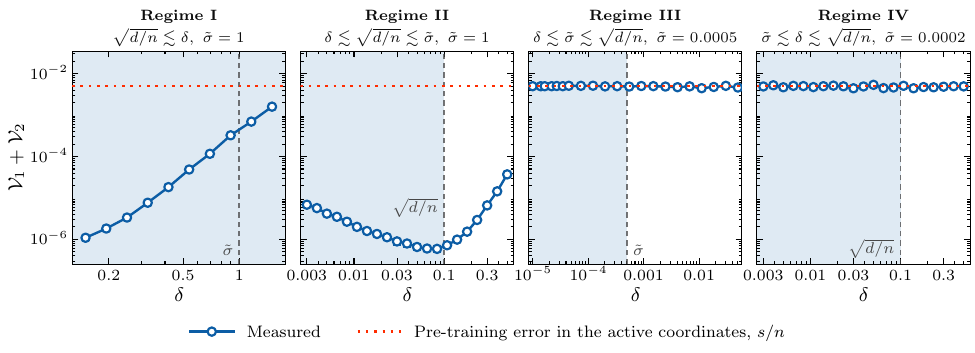}
  \caption{Partial fine-tuning with $s=10$, $d=20$, $n=2000$, and
    $m=3\times10^6$. Simulation estimates of $\mathcal V_1+\mathcal V_2$
    are plotted against $\delta$, with $\tilde\sigma$ fixed in each panel.
    Regime~I exhibits the same reduction as under full fine-tuning, while
    the frozen coordinates change the behavior in the Regimes~II and~III. 
    The U-shaped graph in Regime~II reflects the tradeoff between the
    error in the accepted responses and the frozen coordinate contribution,
    and the variance remains near $s/n$ in Regimes~III and~IV.}
  \label{fig:partial_delta}
\end{figure}

\begin{figure}[t]
  \centering
  \includegraphics[width=\textwidth]{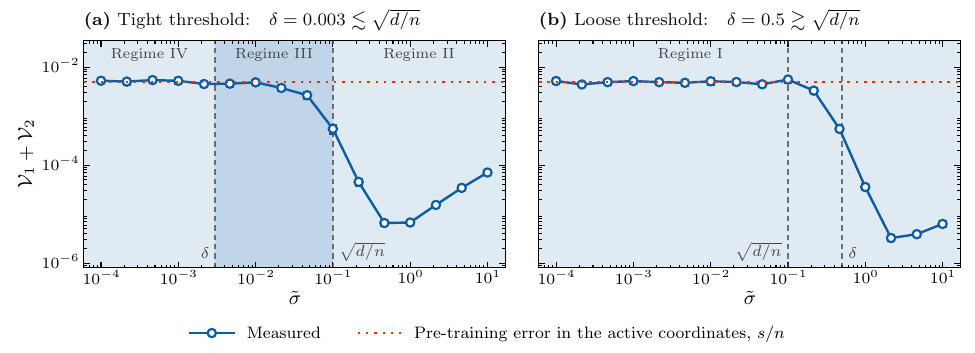}
  \caption{Partial fine-tuning with $s=10$, $d=20$, $n=2000$, and
    $m=3\times10^6$. Simulation estimates of $\mathcal V_1+\mathcal V_2$
    are plotted against $\tilde\sigma$ under (a) tight and (b) loose
    verification. In both cases, sufficiently large generation noise is
    needed to reduce the variance below $s/n$, while excessive generation
    noise increases it. The variance is therefore minimized at an
    intermediate generation noise level.}
  \label{fig:partial_sigma}
\end{figure}

These inequalities hold up to fixed constants, hence the adjacent
regimes overlap when the corresponding scales are comparable.  
We present the partial and full fine-tuning results
within each regime in the following subsections.
Figures~\ref{fig:full_delta}--\ref{fig:partial_sigma} illustrate the active
coordinate variance $\mathcal V_1+\mathcal V_2$ across the four verification
regimes under full and partial fine-tuning.

For the four regimes, we work under Assumption~\ref{ass:design} and
suppose that $n\ge Cd$ and $m\ge C$.  Each regime additionally imposes its
stated scale and accepted sample conditions.
We assume that $d+\log m$ is bounded
above by a polynomial in $m$.  For fixed $c_1>0$, the results
hold when the constants in the sample conditions are sufficiently
large.  
We provide a proof sketch in Section~\ref{sec:proof_sketch} and the detailed
proofs in Appendix~\ref{sec:regime_specializations}.

\setcounter{topnumber}{1}

\subsection{Regime~I: Loose Verification}\label{ssec:regime_rates}

In Regime~I, verification is loose relative to the pre-training error scale,
$\sqrt{d/n}\lesssim\delta$, and does not impose an ordering between
$\delta$ and $\tilde\sigma$. 
The scale condition is
\begin{equation}\label{eq:I_conditions}
\tag{I-scale}
\sqrt{d/n}\lesssim\delta.
\end{equation}
Under this condition, $p_{\regimeid{1}}:=\min(1,\delta/\tilde\sigma)$ is the acceptance scale, and $m p_{\regimeid{1}}$ is the effective accepted sample size.

\begin{theorem}[Partial Fine-Tuning in Regime~I]
\label{thm:regime_I_partial}
Suppose that Assumption~\ref{ass:design} holds and $1\le s<d$.
If \eqref{eq:I_conditions} holds and $m p_{\regimeid{1}}\ge C(s+\log m)(\log m)^{7/2}$, then, with probability
at least $1-2e^{-cn}-m^{-c_1}$ jointly over
$(X_{\mathrm{pre}},\tilde X)$, the extended estimator satisfies
\begin{align*}
R_1-\sigma_1^2-\Var_{-S} &\lesssim\underbrace{\frac{s}{n}\min\left(1,(\delta/\tilde\sigma)^4\right)+C\mathrm{Rem}_{\mathcal V_1}}_{\mathcal V_1}+\underbrace{\frac{s\min(\delta^2,\tilde\sigma^2)}{m p_{\regimeid{1}}}}_{\mathcal V_2},\\
R_1-\sigma_1^2-\Var_{-S} &\gtrsim\frac{s}{n}\min\left(1,(\delta/\tilde\sigma)^4\right)-C\mathrm{Rem}_{\mathcal V_1}+\frac{s\min(\delta^2,\tilde\sigma^2)}{m p_{\regimeid{1}}},
\end{align*}
where $\mathrm{Rem}_{\mathcal V_1}:=d(s+\log m)/(mn)+s(s+(\log m)^2)(\log m)^3\min(\delta^2,\tilde\sigma^2)/(m p_{\regimeid{1}})^2$.
\end{theorem}

We obtain the following corollary under an additional sample condition,
which allows $\mathrm{Rem}_{\mathcal V_1}$ to be absorbed into the sum of the leading $\mathcal V_1$ and $\mathcal V_2$ terms.

\begin{corollary}[Leading Risk Rate in Regime~I]
Suppose that the assumptions of Theorem~\ref{thm:regime_I_partial} hold and additionally assume that $m\min(1,(\delta/\tilde\sigma)^4)\ge Cd(s+\log m)/s$ for a sufficiently large constant $C$.
Then, with probability at least $1-2e^{-cn}-m^{-c_1}$ jointly over $(X_{\mathrm{pre}},\tilde X)$, the extended estimator satisfies
\begin{align*}
R_1-\sigma_1^2-\Var_{-S}
&\asymp
\underbrace{\frac{s}{n}\min\bigl(1,(\delta/\tilde\sigma)^4\bigr)}_{\mathcal V_1}
+\underbrace{\frac{s\min(\delta^2,\tilde\sigma^2)}{m p_{\regimeid{1}}}}_{\mathcal V_2}.
\end{align*}
\end{corollary}

Compared to the baseline without verification, verification changes $\mathcal V_1$
and $\mathcal V_2$ in different ways.  The leading $\mathcal V_1$ term has
order $(s/n)\min(1,(\delta/\tilde\sigma)^4)$.  Hence, the active coordinate
term inherited from pre-training falls below the baseline scale
$s/n$ when $\delta/\tilde\sigma$ is sufficiently small and the remainder
terms are of lower order.
For $\mathcal V_2$, the conditional variance induced by the generation noise has scale
$\min(\delta^2,\tilde\sigma^2)$, while the effective accepted sample size
has order $m p_{\regimeid{1}}$.  Its leading rate is
$s\min(\delta^2,\tilde\sigma^2)/(m p_{\regimeid{1}})$, equivalently
$s\tilde\sigma\min(\delta,\tilde\sigma)/m$.  When
$\delta\lesssim\tilde\sigma$, this is smaller than the baseline rate
$s\tilde\sigma^2/m$ by a factor of order $\delta/\tilde\sigma$.  


\begin{theorem}[Full Fine-Tuning in Regime~I]
Suppose that Assumption~\ref{ass:design} holds and $s=d$.  If
\eqref{eq:I_conditions} holds and $m p_{\regimeid{1}}\ge C(d+\log m)(\log m)^{7/2}$, then, with probability
at least $1-2e^{-cn}-m^{-c_1}$ jointly over $(X_{\mathrm{pre}},\tilde X)$, 
the extended estimator satisfies
\begin{align*}
&  R_1-\sigma_1^2 
\asymp
\underbrace{\frac{d}{n}\min\left(1,(\delta/\tilde\sigma)^4\right)}_{\mathcal V_1}
+
\underbrace{\frac{d\min(\delta^2,\tilde\sigma^2)}
{m p_{\regimeid{1}}}}_{\mathcal V_2}. 
\end{align*}
\end{theorem}

For the full fine-tuning, the leading $\mathcal V_1$ term has scale
$(d/n)\min(1,(\delta/\tilde\sigma)^4)$, while the leading $\mathcal V_2$ term
has scale $d\tilde\sigma\min(\delta,\tilde\sigma)/m$.  Therefore, when
$\delta/\tilde\sigma$ is sufficiently small, verification reduces the leading term inherited from
pre-training below the scale $d/n$.  Similar to partial fine-tuning, this reduction can occur even when the
verification threshold exceeds the pre-training error scale.

The key feature of Regime~I is that verification can reduce both
leading variance terms when $\delta/\tilde\sigma$ is small, even
though the threshold is loose relative to the pre-training error scale.
When $\delta/\tilde\sigma$ is small, verification retains samples for which
the injected generation noise offsets the error inherited from pre-training.  This
mechanism applies under both partial and full fine-tuning, and the leading
active coordinate rates have the same form, with $s$ replaced by $d$ under
full fine-tuning.

\subsection{Regime~II: Tight Verification, Noise-Dominated}

Regime~II considers a verification threshold smaller than the
pre-training error scale, while the pre-training error scale is
smaller than the injected generation noise scale $\tilde\sigma$.  In this
regime, the generation noise can offset the pre-training prediction error,
 while each accepted generated response has error at
most $\delta$.
The scale condition is
\begin{equation}\label{eq:II_conditions}
\tag{II-scale}
\delta\lesssim\sqrt{d/n}\lesssim\tilde\sigma.
\end{equation}
Under this condition, $p_{\regimeid{2}}:=\delta/\tilde\sigma$ is the
acceptance scale, and $m p_{\regimeid{2}}$ is the effective accepted sample
size.

\begin{theorem}[Partial Fine-Tuning in Regime~II]
\label{thm:regime_II_partial}
Suppose that Assumption~\ref{ass:design} holds and $1\le s<d$.  If
\eqref{eq:II_conditions} holds and $m p_{\regimeid{2}}\ge C(s+\log m)(\log m)^{7/2}$, then, with probability
at least $1-2e^{-cn}-m^{-c_1}$ jointly over $(X_{\mathrm{pre}},\tilde X)$, 
the extended estimator satisfies
\begin{align*}
R_1-\sigma_1^2-\Var_{-S} &\lesssim \underbrace{\frac{s}{n}\left(\left(\frac{d-s}{n\tilde\sigma^2+d-s}\right)^2+(\delta/\tilde\sigma)^4\right)+C\mathrm{Rem}_{\mathcal V_1}}_{\mathcal V_1} +\underbrace{\frac{s}{m p_{\regimeid{2}}}\left(\delta^2+\frac{\tilde\sigma^2(d-s)}{n\tilde\sigma^2+d-s}\right)}_{\mathcal V_2},\\*
R_1-\sigma_1^2-\Var_{-S} &\gtrsim \frac{s}{n}\left(\left(\frac{d-s}{n\tilde\sigma^2+d-s}\right)^2+(\delta/\tilde\sigma)^4\right)-C\mathrm{Rem}_{\mathcal V_1}+\frac{s}{m p_{\regimeid{2}}}\left(\delta^2+\frac{\tilde\sigma^2(d-s)}{n\tilde\sigma^2+d-s}\right),
\end{align*}
where $\mathrm{Rem}_{\mathcal V_1}:=d(s+\log m)/(mn)+s(s+(\log m)^2)(\log m)^3(\delta^2+\tilde\sigma^2(d-s)/(n\tilde\sigma^2+d-s))/(m p_{\regimeid{2}})^2$.
\end{theorem}

We obtain the following corollary under an additional sample condition,
which allows $\mathrm{Rem}_{\mathcal V_1}$ to be absorbed into the sum of the leading $\mathcal V_1$ and $\mathcal V_2$ terms.

\begin{corollary}[Leading Risk Rate in Regime~II]
Suppose that the assumptions of Theorem~\ref{thm:regime_II_partial} hold and additionally assume that $m\delta^2(\delta^2+(d-s)/n)/\tilde\sigma^4\ge Cd(s+\log m)/s$ for a sufficiently large constant $C$.
Then, with probability at least $1-2e^{-cn}-m^{-c_1}$ jointly over $(X_{\mathrm{pre}},\tilde X)$, the extended estimator satisfies
\begin{align*}
R_1-\sigma_1^2-\Var_{-S}
&\asymp
\underbrace{\frac{s}{n}\bigg(\bigg(\frac{d-s}{n\tilde\sigma^2+d-s}\bigg)^2+(\delta/\tilde\sigma)^4\bigg)}_{\mathcal V_1}
+\underbrace{\frac{s}{m p_{\regimeid{2}}}\bigg(\delta^2+\frac{\tilde\sigma^2(d-s)}{n\tilde\sigma^2+d-s}\bigg)}_{\mathcal V_2}.
\end{align*}
\end{corollary}

The leading $\mathcal V_1$ rate has two terms.  The first is
induced by the pre-training error in the frozen coordinates, while the term
proportional to $(\delta/\tilde\sigma)^4$ represents the threshold dependent active
coordinate contribution inherited from pre-training.  When $\delta/\tilde\sigma$ is small,
verification reduces the threshold dependent term.  The frozen
coordinate term in $\mathcal V_1$ can nevertheless limit this
reduction under partial fine-tuning.

For $\mathcal V_2$, the effective accepted sample size has order
$m p_{\regimeid{2}}$, while every accepted sample satisfies
$|\xi_j|\le\delta$.  Since $p_{\regimeid{2}}=\delta/\tilde\sigma$ and
$\tilde\sigma^2(d-s)/(n\tilde\sigma^2+d-s)\asymp(d-s)/n$, its leading
rate has the form
\begin{align*}
\frac{s\delta\tilde\sigma}{m}
+\frac{s\tilde\sigma(d-s)}{mn\delta}.
\end{align*}
The first term is induced by generation noise after verification, while the
second is induced by the pre-training error in the frozen coordinates. We
observe that there exists a tradeoff in $\delta$ and these two $\mathcal V_2$
terms balance at $\delta\asymp\sqrt{(d-s)/n}$.
Moreover, the $(\delta/\tilde\sigma)^4$
term in $\mathcal V_1$ also becomes smaller as $\delta$ decreases,
while the other $\mathcal V_1$ term is independent of $\delta$ and
provides a lower bound. 
These terms explain the U-shaped Regime~II plot
in Figure~\ref{fig:partial_delta}.


\begin{theorem}[Full Fine-Tuning in Regime~II]
Suppose that Assumption~\ref{ass:design} holds and $s=d$.  If
\eqref{eq:II_conditions} holds and $m p_{\regimeid{2}}\ge C(d+\log m)(\log m)^{7/2}$, then, with
probability at least $1-2e^{-cn}-m^{-c_1}$ jointly
over $(X_{\mathrm{pre}},\tilde X)$, the extended estimator satisfies
\begin{align*}
R_1-\sigma_1^2
& \asymp
\underbrace{
\frac{d}{n}(\delta/\tilde\sigma)^4}_{\mathcal V_1}
+
\underbrace{\frac{d\delta^2}{m p_{\regimeid{2}}}}_{\mathcal V_2}.
\end{align*}
\end{theorem}

 For the full fine-tuning, the leading $\mathcal V_1$ and $\mathcal V_2$ terms have
scales $(\delta/\tilde\sigma)^4d/n$ and $d\delta\tilde\sigma/m$, respectively.  Both terms are reduced when
$\delta/\tilde\sigma$ is small as we can observe in Figure~\ref{fig:full_delta}. 

The key feature of Regime~II is that tight verification reduces
the threshold dependent term in $\mathcal V_1$ and the generation
noise term in $\mathcal V_2$, while the frozen coordinate terms
create a verification tradeoff under partial fine-tuning.
Under partial fine-tuning, the frozen coordinates add a $\delta$-independent term to $\mathcal V_1$ and a term to
$\mathcal V_2$ that increases as $\delta$ decreases, producing the U-shaped
behavior.  Full fine-tuning removes both terms, and hence tighter
verification directly reduces the two leading terms when
$\delta/\tilde\sigma$ is small.  At the same time, the accepted covariates
retain variation across the downstream directions relevant to prediction.
This distinguishes Regime~II from Regimes~III and~IV, where reduced variation
along the pre-training error direction contributes additional factors to $\mathcal V_2$.

\subsection{Regime~III: Tight Verification, Pre-Training Error-Dominated}

Regime~III considers a verification threshold smaller than the injected generation
noise scale, while the generation noise scale is smaller than the
pre-training error scale.  In this regime, verification depends substantially
on $-\tilde x_j^\top\Delta$ and favors covariates for which the pre-training
prediction error $|\tilde x_j^\top\Delta|$ is small.  The scale condition is
\begin{equation}\label{eq:III_conditions}
\tag{III-scale}
\delta\lesssim\tilde\sigma\lesssim\sqrt{d/n}.
\end{equation}
Under this condition, $p_{\regimeid{3}}:=\delta/\sqrt{d/n}$ is
the acceptance scale, and $m p_{\regimeid{3}}$ is the effective accepted
sample size.
For the partial fine-tuning $\mathcal V_2$ rates in Regimes~III and~IV,
define, for $z>0$,
\begin{equation*}
g_{d-s,d}(z):=
\begin{cases}
1+z/\sqrt d, & d-s=1,\\
\log(e+z^2/d), & d-s=2,\\
1, & d-s\ge3.
\end{cases}
\end{equation*}

\begin{theorem}[Partial Fine-Tuning in Regime~III]
\label{thm:regime_III_partial}
Let Assumption~\ref{ass:design} hold and let $1\le s<d$.  Suppose that \eqref{eq:III_conditions} holds and
$m p_{\regimeid{3}}\ge C(1+d/(n\tilde\sigma^2))(s+\log m)(\log m)^{7/2}$.  Then, with probability
  at least $1-2e^{-cn}-m^{-c_1}$ jointly over
$(X_{\mathrm{pre}},\tilde X)$, the extended estimator satisfies
\begin{align*}
&R_1-\sigma_1^2-\Var_{-S} \lesssim \\
&\underbrace{\frac{s}{n}\left(\left(\frac{d-s}{n\tilde\sigma^2+d-s}
\right)^2+(\delta/\tilde\sigma)^4\right)+C\mathrm{Rem}_{\mathcal V_1}}_
{\mathcal V_1}+\underbrace{\frac{s}{m p_{\regimeid{3}}}\left(\delta^2
+\frac{\tilde\sigma^2(d-s)}{n\tilde\sigma^2+d-s}\right)g_{d-s,d}(\sqrt{d/
n}/\tilde\sigma)}_{\mathcal V_2},\\
&R_1-\sigma_1^2-\Var_{-S} \gtrsim \\
&\frac{s}{n}\left(\left(\frac{d-s}{n\tilde\sigma^2+d-s}\right)^2+(\delta/
\tilde\sigma)^4\right)-C\mathrm{Rem}_{\mathcal V_1}+\frac{s}{m p_
{\regimeid{3}}}\left(\delta^2+\frac{\tilde\sigma^2(d-s)}{n\tilde\sigma^2
+d-s}\right)g_{d-s,d}(\sqrt{d/n}/\tilde\sigma),
\end{align*}
where $\mathrm{Rem}_{\mathcal V_1}:=d(s+\log m)\cdot\sqrt{1+d/
(n\tilde\sigma^2)}\cdot(1+s/(n\tilde\sigma^2))/(mn)+s(1+d/
(n\tilde\sigma^2))\cdot(s+(1+d/(n\tilde\sigma^2))(\log m)^2)\cdot(\log m)
^3\cdot(\delta^2+\tilde\sigma^2(d-s)/(n\tilde\sigma^2+d-s))/(m p_
{\regimeid{3}})^2$.
\end{theorem}

We obtain the following corollary under an additional sample condition,
which allows $\mathrm{Rem}_{\mathcal V_1}$ to be absorbed into the sum of the leading $\mathcal V_1$ and $\mathcal V_2$ terms.

\begin{corollary}[Leading Risk Rate in Regime~III]
Suppose that the assumptions of Theorem~\ref{thm:regime_III_partial} hold and additionally assume that $mp_{\regimeid{3}}\bigl(((d-s)/(n\tilde\sigma^2+d-s))^2+(\delta/\tilde\sigma)^4\bigr)\ge Cd(s+\log m)(1+d/(n\tilde\sigma^2))\sqrt{\log m}/s$ for a sufficiently large constant $C$.
Then, with probability at least $1-2e^{-cn}-m^{-c_1}$ jointly over $(X_{\mathrm{pre}},\tilde X)$, the extended estimator satisfies
\begin{align*}
&R_1-\sigma_1^2-\Var_{-S}\asymp\\
&\underbrace{\frac{s}{n}\bigg(\bigg(\frac{d-s}{n\tilde\sigma^2+d-s}\bigg)^2+(\delta/\tilde\sigma)^4\bigg)}_{\mathcal V_1}
+\underbrace{\frac{s}{m p_{\regimeid{3}}}\bigg(\delta^2+\frac{\tilde\sigma^2(d-s)}{n\tilde\sigma^2+d-s}\bigg)g_{d-s,d}(\sqrt{d/n}/\tilde\sigma)}_{\mathcal V_2}.
\end{align*}
\end{corollary}

The leading $\mathcal V_1$ rate has a frozen coordinate term and a
threshold dependent term.  When
$(d-s)/(n\tilde\sigma^2)\gtrsim1$, the first has order $s/n$ and does not
decrease as $m$ grows, which is observed in the Regime~III plot of Figure~\ref{fig:partial_delta}.
The second term has order
$(s/n)(\delta/\tilde\sigma)^4$ and is reduced under tight verification as similar to
Regime~II.  

For $d-s\ge3$, by the definition, $g_{d-s,d}(\sqrt{d/n}/\tilde\sigma)=1$, and the leading
$\mathcal V_2$ rate differs from Regime~II primarily through the acceptance
scale.  In particular, the acceptance scales satisfy $p_{\regimeid{3}}/p_{\regimeid{2}}=\tilde\sigma/\sqrt{d/n}$.
Hence, the effective accepted sample size decreases relative to Regime~II as the pre-training error scale becomes larger relative to the generation noise scale.  Within the numerator of $\mathcal V_2$,
the factor $\delta^2$ is induced by generation noise after verification, while
$\tilde\sigma^2(d-s)/(n\tilde\sigma^2+d-s)$ is induced by the pre-training
error in the frozen coordinates.  When $d-s=1$ or $d-s=2$, the inverse
covariance of the accepted covariates additionally introduces the factors
$1+1/(\sqrt n\tilde\sigma)$ and $\log(e+1/(n\tilde\sigma^2))$,
respectively.
These factors can be large when the generation noise scale $\tilde\sigma$ is small compared to $1/\sqrt n$.


\begin{theorem}[Full Fine-Tuning in Regime~III]
Let Assumption~\ref{ass:design} hold and let $s=d$.  Suppose that
\eqref{eq:III_conditions} holds and $m p_{\regimeid{3}}\ge C(1+d/(n\tilde\sigma^2))(d+\log m)(\log m)^{7/2}$.  Then, with
probability at least $1-2e^{-cn}-m^{-c_1}$ jointly
over $(X_{\mathrm{pre}},\tilde X)$, the extended estimator satisfies
\begin{align*}
R_1-\sigma_1^2
& \asymp
\underbrace{ \frac{d}{n}(\delta/\tilde\sigma)^4}_{\mathcal V_1}
+
\underbrace{\frac{d\delta^2(1+1/(n\tilde\sigma^2))}{m p_{\regimeid{3}}}}_{\mathcal V_2}.
\end{align*}
\end{theorem}

Under full fine-tuning, the leading $\mathcal V_1$ term has scale
$(\delta/\tilde\sigma)^4d/n$.  Since verification favors covariates for 
which $|\tilde x_j^\top\Delta|$ is small, the accepted covariates are 
concentrated
toward the hyperplane orthogonal to the pre-training error.  In 
Regime~III,
$|\tilde x_j^\top\Delta|$ typically has scale $\tilde\sigma$ among 
accepted samples.  The resulting reduction in variation along the pre-training error
direction contributes the additional factor $1+1/(n\tilde\sigma^2)$ to
$\mathcal V_2$.  Moreover, this effect is stronger under full fine-tuning than
under partial fine-tuning.  Thus, full fine-tuning removes the
frozen coordinate term in $\mathcal V_1$, while the covariance of
the accepted covariates can increase $\mathcal V_2$ in Regime~III.
Figure~\ref{fig:full_sigma} illustrates this increase as $\tilde\sigma$
decreases from Regime~II into Regime~III.

The key feature of Regime~III is that tight verification can
reduce the threshold dependent term of $\mathcal V_1$, while the
smaller effective accepted sample size and inverse covariance factor can
increase $\mathcal V_2$.
However, under partial fine-tuning, the frozen coordinate term maintains
$\mathcal V_1$ at order $s/n$, while the inverse covariance effect in
$\mathcal V_2$ is weaker.  On the other hand, for the full fine-tuning, this term in
$\mathcal V_1$ disappears, while the reduced variation along the pre-training
error direction yields the stronger
factor $1+1/(n\tilde\sigma^2)$ in $\mathcal V_2$.

To interpret these results, regard $\tilde x_j$ as a query and
$\tilde y_j$ as its generated response.  Since the generation noise is
smaller than the pre-training error scale, verification favors queries on
which the pre-trained estimator already has a small prediction error.
This selection reduces the effective accepted sample size and distorts the
structure of the covariance of the accepted covariates.

\subsection{Regime~IV: Low Generation Noise}

Regime~IV considers an injected generation noise scale smaller than the verification
threshold, while the verification threshold is smaller than the pre-training
error scale.  In this regime, verification depends substantially on
$-\tilde x_j^\top\Delta$ and favors covariates for which
$|\tilde x_j^\top\Delta|\lesssim\delta$.  The scale condition is
\begin{equation}\label{eq:IV_conditions}
\tag{IV-scale}
\tilde\sigma\lesssim\delta\lesssim\sqrt{d/n}.
\end{equation}
Under this condition, $p_{\regimeid{4}}:=\delta/\sqrt{d/n}$ is the acceptance
scale, and $m p_{\regimeid{4}}$ is the effective accepted sample size.

\begin{theorem}[Partial Fine-Tuning in Regime~IV]
\label{thm:regime_IV_partial}
Let Assumption~\ref{ass:design} hold and let $1\le s<d$.  Suppose that
\eqref{eq:IV_conditions} holds and $m p_{\regimeid{4}}\ge C d(s+\log m)(\log m)^{7/2}/(n\delta^2)$.  Then, with probability
at least $1-2e^{-cn}-m^{-c_1}$ jointly over $(X_{\mathrm{pre}},\tilde X)$, the extended estimator satisfies
\begin{align*}
R_1-\sigma_1^2-\Var_{-S} &\lesssim\underbrace{\frac{s}{n}+C\mathrm{Rem}_{\mathcal V_1}}_{\mathcal V_1}+\underbrace{\frac{s\tilde\sigma^2}{m p_{\regimeid{4}}}g_{d-s,d}(\sqrt{d/n}/\delta)}_{\mathcal V_2},\\
R_1-\sigma_1^2-\Var_{-S} &\gtrsim\frac{s}{n}-C\mathrm{Rem}_{\mathcal V_1}+\frac{s\tilde\sigma^2}{m p_{\regimeid{4}}}g_{d-s,d}(\sqrt{d/n}/\delta),
\end{align*}
where $\mathrm{Rem}_{\mathcal V_1}:=d(s+\log m)\cdot\sqrt{d/n}\cdot(1+\min(s/(n\tilde\sigma^2),d/(n\delta^2)))/(mn\delta)+sd\cdot(s+d(\log m)^2/(n\delta^2))\cdot(\log m)^3\cdot\tilde\sigma^2/(n\delta^2(m p_{\regimeid{4}})^2)$.
\end{theorem}

We obtain the following corollary under an additional sample condition,
which allows $\mathrm{Rem}_{\mathcal V_1}$ to be absorbed into the sum of the leading $\mathcal V_1$ and $\mathcal V_2$ terms.

\begin{corollary}[Leading Risk Rate in Regime~IV]
Suppose that the assumptions of Theorem~\ref{thm:regime_IV_partial} hold and additionally assume that $mp_{\regimeid{4}}\ge Cd(s+\log m)\bigl(1+\min(s/(n\tilde\sigma^2),d/(n\delta^2))\bigr)\sqrt{\log m}/s$ for a sufficiently large constant $C$.
Then, with probability at least $1-2e^{-cn}-m^{-c_1}$ jointly over $(X_{\mathrm{pre}},\tilde X)$, the extended estimator satisfies
\begin{align*}
R_1-\sigma_1^2-\Var_{-S}
&\asymp
\underbrace{\frac{s}{n}}_{\mathcal V_1}
+\underbrace{\frac{s\tilde\sigma^2}{m p_{\regimeid{4}}}g_{d-s,d}(\sqrt{d/n}/\delta)}_{\mathcal V_2}.
\end{align*}
\end{corollary}

Observe that the leading $\mathcal V_1$ term has order $s/n$.  Since $\tilde\sigma\lesssim\delta$,
the threshold dependent factor is of constant order, and verification does not
yield the $(\delta/\tilde\sigma)^4$ rate reduction that appears in
Regimes~II and~III.  Thus, the leading active coordinate
term inherited from pre-training remains of order $s/n$, which
explains why the Regime~IV curve in Figure~\ref{fig:partial_delta} remains
near $s/n$.

For $d-s\ge3$, we have $g_{d-s,d}(\sqrt{d/n}/\delta)=1$, and the leading
$\mathcal V_2$ rate has scale $s\tilde\sigma^2/(m p_{\regimeid{4}})$.  Since
$p_{\regimeid{4}}=\delta/\sqrt{d/n}$, tighter verification reduces the
effective accepted sample size.  Since $\tilde\sigma\lesssim\delta$, the 
generation noise term in the $\mathcal V_2$ numerator has order 
$\tilde\sigma^2$,
while the frozen coordinate term $\tilde\sigma^2(d-s)/(n\tilde\sigma^2
+d-s)$ is bounded by $\tilde\sigma^2$.  On the other hand, when $d-s=1$ 
or $d-s=2$, the inverse covariance of the
accepted covariates additionally contributes the factors $1+1/(\sqrt 
n\delta)$ and $\log(e+1/(n\delta^2))$, respectively.
These factors arise from the change in the covariate distribution induced by verification.


\begin{theorem}[Full Fine-Tuning in Regime~IV]
Let Assumption~\ref{ass:design} hold and let $s=d$.  Suppose that
\eqref{eq:IV_conditions} holds and $m p_{\regimeid{4}}\ge C d(d+\log m)(\log m)^{7/2}/(n\delta^2)$.  Then, with
probability at least $1-2e^{-cn}-m^{-c_1}$ jointly
over $(X_{\mathrm{pre}},\tilde X)$, the extended estimator satisfies
\begin{align*}
R_1-\sigma_1^2
 & \asymp
\underbrace{\frac{d}{n}}_{\mathcal V_1}
+
\underbrace{\frac{d\tilde\sigma^2(1+1/(n\delta^2))}{m p_{\regimeid{4}}}}_{\mathcal V_2}.
\end{align*}
\end{theorem}

For the full fine-tuning, it follows that $\mathcal V_1$ remains of order $d/n$.  Hence,
removing the frozen coordinates does not restore the
$(\delta/\tilde\sigma)^4$ reduction.  In addition, as in Regime~III, verification favors
covariates near the hyperplane orthogonal to the pre-training error.  Especially, in
Regime~IV, $|\tilde x_j^\top\Delta|$ typically has scale $\delta$ among accepted
samples.  The resulting reduction in variation along the pre-training error
direction contributes the factor $1+1/(n\delta^2)$ to $\mathcal V_2$.
Furthermore, the order $d/n$ of
$\mathcal V_1$ explains why the Regime~IV curve in
Figure~\ref{fig:full_delta} remains near $d/n$.

The key feature of Regime~IV is that, when the generation noise
is smaller than the verification threshold, tight verification does not
improve the leading rates of either $\mathcal V_1$ or $\mathcal V_2$.
For the partial fine-tuning, $\mathcal V_1$ remains of order $s/n$, while
$\mathcal V_2$ retains the scale
$s\tilde\sigma^2/(m p_{\regimeid{4}})$, up to its covariance factor.
For the full fine-tuning, $\mathcal V_1$ remains of order $d/n$, while
the reduced variation along the pre-training error direction contributes the
factor $1+1/(n\delta^2)$ in $\mathcal V_2$.

Considering the query and generated response interpretation introduced in
Section~\ref{sec:problem_setting}, verification similarly favors queries on which the pre-trained
estimator is already accurate.  Since the generation noise is smaller than
the threshold in Regime~IV, this selection depends more directly on the
pre-training error. 

\subsection{Comparison Across Regimes}\label{sec:interpretation}

\begin{table}[t]
\centering
\scriptsize
\setlength{\tabcolsep}{3pt}
\caption{Leading active coordinate variance terms for partial
fine-tuning $(s<d)$.}
\label{tab:regime_rates}
\begin{tabular}{llll}
\toprule
Regime & Scales & $\mathcal V_1$ terms & $\mathcal V_2$ terms \\
\midrule
No verification &
all generated samples &
$\displaystyle s/n$ &
$\displaystyle s\tilde\sigma^2/m$ \\[6pt]
I &
$\sqrt{d/n}\lesssim\delta$ &
$\displaystyle \frac{s}{n}\min\left(1,(\delta/\tilde\sigma)^4\right)$ &
$\displaystyle \frac{s\min(\delta^2,\tilde\sigma^2)}
{m p_{\regimeid{1}}}$ \\[10pt]
II &
$\delta\lesssim\sqrt{d/n}\lesssim\tilde\sigma$ &
$\displaystyle \frac{s}{n}\left(
\left(\frac{d-s}{n\tilde\sigma^2+d-s}\right)^2+(\delta/\tilde\sigma)^4\right)$ &
$\displaystyle
\frac{s}{m p_{\regimeid{2}}}\left(\delta^2+
\frac{\tilde\sigma^2(d-s)}{n\tilde\sigma^2+d-s}\right)$ \\[10pt]
III &
$\delta\lesssim\tilde\sigma\lesssim\sqrt{d/n}$ &
$\displaystyle \frac{s}{n}\left(
\left(\frac{d-s}{n\tilde\sigma^2+d-s}\right)^2
+(\delta/\tilde\sigma)^4\right)$ &
$\displaystyle
\frac{s}{m p_{\regimeid{3}}}
\left(\delta^2+\frac{\tilde\sigma^2(d-s)}{n\tilde\sigma^2+d-s}\right)
g_{d-s,d}(\sqrt{d/n}/\tilde\sigma)$ \\[10pt]
IV &
$\tilde\sigma\lesssim\delta\lesssim\sqrt{d/n}$ &
$\displaystyle s/n$ &
$\displaystyle
\frac{s\tilde\sigma^2}{m p_{\regimeid{4}}}
g_{d-s,d}(\sqrt{d/n}/\delta)$ \\
\bottomrule
\end{tabular}
\end{table}

We summarize the leading active coordinate variance terms for partial
fine-tuning in Table~\ref{tab:regime_rates}.
When $\delta/\tilde\sigma$ is small, verification reduces
the threshold dependent term in $\mathcal V_1$ by the factor
$(\delta/\tilde\sigma)^4$ in Regimes~I--III.  In Regime~IV, the leading
$\mathcal V_1$ rate instead remains of order $s/n$ under partial fine-tuning
and $d/n$ under full fine-tuning.  For $\mathcal V_2$, the accepted
covariates retain variation across the downstream directions in
Regimes~I and~II.  In Regimes~III and~IV, verification favors covariates
for which $|\tilde x^\top\Delta|$ is small, and the resulting change in
their covariance can contribute additional factors to $\mathcal V_2$.

In addition, we find that partial and full fine-tuning differ through the
pre-training error retained in the frozen coordinates.  In Regimes~II
and~III, the frozen coordinate error contributes to both $\mathcal V_1$ and $\mathcal V_2$,
and the resulting leading term in $\mathcal V_1$ can remain of order $s/n$ even
when the verification threshold is sufficiently small.  In Regimes~III and~IV,
the change in the covariance of the accepted covariates has a weaker effect
on $\mathcal V_2$ under partial fine-tuning and contributes only a constant
order factor when $d-s\ge3$.  On the other hand, full fine-tuning removes the frozen coordinate
terms, while this covariance effect becomes stronger and results in the
factors $1+1/(n\tilde\sigma^2)$ in Regime~III and
$1+1/(n\delta^2)$ in Regime~IV.

For post-training of large language models, the generation noise scale can be interpreted as the
randomness in response sampling, which can be adjusted through the sampling
temperature.  The verification threshold corresponds to the strictness of
the criterion exploited by a verifier to accept generated responses,
such as a minimum quality score or a required proportion of
verification checks passed.
Our results suggest considering how these criteria affect response
accuracy and the number and distribution of accepted samples.
Overall, these results characterize when verified synthetic data can 
improve prediction risk and how its effects vary across the four regimes 
and between partial and full fine-tuning, providing useful insights for 
the 
employment of
synthetic data and verification in practice.

\FloatBarrier

%% file: sections/4proof_sketch.tex

\section{Proof Sketch of the Main Results}\label{sec:proof_sketch}

In Section~\ref{section:main_results}, we have characterized the prediction risk
and reduced the analysis of the active coordinate variance to
$\mathcal V_1+\mathcal V_2$.  In this section, we provide a high level proof
sketch of how the conditional mean and conditional covariance of the
post-training residual lead to bounds for these two terms.  
We focus on partial fine-tuning and briefly explain the differences under full fine-tuning.
Throughout this section, we fix $X_{\mathrm{pre}}\in\Omega_{\mathrm{pre}}$, where $\Omega_{\mathrm{pre}}$ is the event defined in~\eqref{eq:Omega_pre_def}
on which the pre-training sample covariance is well conditioned.  Expectations over $\Delta$ are conditioned on
$X_{\mathrm{pre}}$, and $\Delta\mid X_{\mathrm{pre}}\sim\mathcal N(0,\Gamma/n)$.

\paragraph{Step 1: Identify the conditional mean and conditional covariance.}
By the risk decomposition and $\mathrm{Bias}_S^2=0$, the active coordinate
contribution to excess prediction risk is
$\Var_S=\mathcal V_1+\mathcal V_2$.  Recall that the terms $\mathcal V_1$ and
$\mathcal V_2$ are determined by the conditional mean and conditional
covariance of $r_{S,\mathrm{ext}}$ over generation noise, respectively.
Since $\mathrm{Bias}_S^2=0$, their definitions imply
\begin{align*}
\mathcal V_1
=\mathbb E_\Delta\left[\left\|\mathbb E_{\tilde\varepsilon}
[r_{S,\mathrm{ext}}\mid\Delta,\tilde X]\right\|_{\Sigma_{1,SS}}^2\right], \quad
\mathcal V_2
=\mathbb E_\Delta\left[\tr\bigl(\Sigma_{1,SS}
\Cov_{\tilde\varepsilon}(r_{S,\mathrm{ext}}\mid\Delta,\tilde X)\bigr)\right].
\end{align*}
On $\Omega_{\mathrm{post}}$, we have
$r_{S,\mathrm{ext}}=r_S:=\hat\beta_S^{\mathrm{post}}-\beta_S^{\mathrm{corr}}$,
which we analyze below.

\paragraph{Common residual decomposition.}
To analyze $r_S$, we define the following quantities:
\begin{align*}
G(\Delta)
&:=\sum_{i=1}^m\mathbb I_i\tilde x_{i,S}\tilde x_{i,S}^{\top},
&
\mu(\Delta)
&:=\sum_{i=1}^m\mathbb I_i\tilde x_{i,S}
\bigl(\tilde x_{i,-S}^{\top}\Delta_{-S}+\xi_i\bigr),\\
\bar G(\Delta)
&:=\mathbb E_{\tilde\varepsilon}
[G(\Delta)\mid\Delta,\tilde X],
&
\bar\mu(\Delta)
&:=\mathbb E_{\tilde\varepsilon}
[\mu(\Delta)\mid\Delta,\tilde X].
\end{align*}
Since $G(\Delta)$ is invertible on $\Omega_{\mathrm{post}}$, whenever $\bar G(\Delta)$ is also invertible, 
we can decompose the residual as
\begin{align*}
r_S
&=G(\Delta)^{-1}\mu(\Delta)
-\Sigma_{1,SS}^{-1}\Sigma_{1,S,-S}\Delta_{-S}\\
&=\underbrace{
\bar G(\Delta)^{-1}\bar\mu(\Delta)
-\Sigma_{1,SS}^{-1}\Sigma_{1,S,-S}\Delta_{-S}
}_{\text{leading conditional mean}}
+\underbrace{
\bar G(\Delta)^{-1}
\bigl(\mu(\Delta)-G(\Delta)\bar G(\Delta)^{-1}\bar\mu(\Delta)\bigr)
}_{\text{centered term for conditional covariance}}\\
&\quad+\underbrace{
\bigl(G(\Delta)^{-1}-\bar G(\Delta)^{-1}\bigr)
\bigl(\mu(\Delta)-G(\Delta)\bar G(\Delta)^{-1}\bar\mu(\Delta)\bigr)
}_{\text{inverse remainder}}.
\end{align*}

By the definitions of $\bar G(\Delta)$ and $\bar\mu(\Delta)$, the centered
numerator in the last two terms has conditional mean zero over
$\tilde\varepsilon$.  
Hence, the leading contribution to $\mathcal V_1$ is the squared $\Sigma_{1,SS}$ norm of the   conditional mean,
while the leading contribution to  $\mathcal V_2$ is the conditional second moment of the  centered term in the same norm.
The final term is the error from replacing $G(\Delta)^{-1}$ by $\bar G(\Delta)^{-1}$.

\paragraph{Step 2: Analysis of the $\mathcal V_1$ rates.}
First, we analyze $\mathcal V_1$ through the conditional mean in Step~1.
For fixed $\Delta$, we approximate this conditional mean by its population
expression and then take expectation over $\Delta$.
For one synthetic sample, let $\tilde x$ and $\tilde\varepsilon$ denote the covariate and injected generation noise.
Recall that $\xi=-\tilde x^\top\Delta+\tilde\varepsilon$ and $\mathbb I=\mathbf 1\{|\xi|\le\delta\}$.
Conditioned on $\Delta$, the residual $\xi$ is centered Gaussian with variance $\sigma_\xi^2(\Delta):=\tilde\sigma^2+\|\Delta\|_{\Sigma_1}^2$.
By Lemma~\ref{lem:truncated_normal_moments}, its acceptance probability satisfies $p_\Delta:=\mathbb P(|\xi|\le\delta\mid\Delta)\asymp\min(1,\delta/\sigma_\xi(\Delta))$.
We define the population Gram matrix and residual numerator by
\begin{align*}
\bar G_{\mathrm{pop}}(\Delta):=\mathbb E_{\tilde x,\tilde\varepsilon}
[\mathbb I\tilde x_S\tilde x_S^\top\mid\Delta],\quad
\bar\mu_{\mathrm{pop}}(\Delta):=\mathbb E_{\tilde x,\tilde\varepsilon}
[\mathbb I\tilde x_S(\tilde x_{-S}^\top\Delta_{-S}+\xi)\mid\Delta].
\end{align*}
It follows that  $\mathbb E_{\tilde X}[\bar G(\Delta)\mid\Delta]=m\bar G_{\mathrm{pop}}(\Delta)$ and $\mathbb E_{\tilde X}[\bar\mu(\Delta)\mid\Delta]=m\bar\mu_{\mathrm{pop}}(\Delta)$.

We replace $\bar G(\Delta)$ and $\bar\mu(\Delta)$ in the leading conditional mean of Step~1 by $m\bar G_{\mathrm{pop}}(\Delta)$ and $m\bar\mu_{\mathrm{pop}}(\Delta)$, respectively.
Hence, the leading population contribution to $\mathcal V_1$ is
\begin{align*}
\mathbb E_\Delta\left[
\left\|\bar G_{\mathrm{pop}}(\Delta)^{-1}\bar\mu_{\mathrm{pop}}(\Delta)
-\Sigma_{1,SS}^{-1}\Sigma_{1,S,-S}\Delta_{-S}
\right\|_{\Sigma_{1,SS}}^2\right].
\end{align*}
For the population conditional mean, we employ the identity $\tilde x_{-S}^\top\Delta_{-S}+\xi=-\tilde x_S^\top\Delta_S+\tilde\varepsilon$ and obtain
\begin{align*}
\bar\mu_{\mathrm{pop}}(\Delta)
&=-\bar G_{\mathrm{pop}}(\Delta)\Delta_S
+\mathbb E_{\tilde x,\tilde\varepsilon}
[\mathbb I\tilde x_S\tilde\varepsilon\mid\Delta].
\end{align*}
To evaluate $\mathbb E_{\tilde x,\tilde\varepsilon}[\mathbb I\tilde x_S\tilde\varepsilon\mid\Delta]$, we condition the jointly Gaussian vector $(\tilde x_S,\tilde\varepsilon,\xi)$ on $\xi$ and then take expectation after multiplying by $\mathbb I$.
With $\theta_\Delta$ defined in Lemma~\ref{lem:common_population_mu}, we obtain $\bar G_{\mathrm{pop}}(\Delta)^{-1}\mathbb E_{\tilde x,\tilde\varepsilon}[\mathbb I\tilde x_S\tilde\varepsilon\mid\Delta]=\theta_\Delta\Sigma_{1,SS}^{-1}(\Sigma_1\Delta)_S$.
Therefore, the population conditional mean has the form
\begin{align*}
\bar G_{\mathrm{pop}}(\Delta)^{-1}\bar\mu_{\mathrm{pop}}(\Delta)
-\Sigma_{1,SS}^{-1}\Sigma_{1,S,-S}\Delta_{-S}
&=-(1-\theta_\Delta)\Sigma_{1,SS}^{-1}(\Sigma_1\Delta)_S,
\end{align*}
where $0<\theta_\Delta<1$.
By Lemma~\ref{lem:common_population_theta_bounds}, it follows that
\begin{align*}
1-\theta_\Delta &\asymp
\frac{\|\Delta_{-S}\|_{\Sigma_{1,-S\mid S}}^2}
{\tilde\sigma^2+\|\Delta_{-S}\|_{\Sigma_{1,-S\mid S}}^2}
+\min\bigl(1,(\delta/\tilde\sigma)^2\bigr).
\end{align*}
Note that $\hat\beta_S^{\mathrm{pre}}-\beta_S^{\mathrm{corr}}=-\Sigma_{1,SS}^{-1}(\Sigma_1\Delta)_S$, and the population conditional mean is $1-\theta_\Delta$ times this pre-training residual.
Since $\Delta\mid X_{\mathrm{pre}}\sim\mathcal N(0,\Gamma/n)$, covariance regularity implies $\mathbb E_\Delta[\|(\Sigma_1\Delta)_S\|_{\Sigma_{1,SS}^{-1}}^2]\asymp s/n$.
Thus, we take the squared $\Sigma_{1,SS}$ norm of the population conditional mean and then take expectation over $\Delta$ to obtain the leading population contribution to $\mathcal V_1$:
\begin{align}\label{eq:sketch_population_targets}
\mathbb E_\Delta\!\left[
(1-\theta_\Delta)^2
\|(\Sigma_1\Delta)_S\|_{\Sigma_{1,SS}^{-1}}^2\right]
&\asymp
\frac{s}{n}\min\bigl(1,(\delta/\tilde\sigma)^4\bigr)
+
\mathbb E_\Delta\!\left[
\|(\Sigma_1\Delta)_S\|_{\Sigma_{1,SS}^{-1}}^2
\left(
\frac{\|\Delta_{-S}\|_{\Sigma_{1,-S\mid S}}^2}
{\tilde\sigma^2+\|\Delta_{-S}\|_{\Sigma_{1,-S\mid S}}^2}
\right)^2 \right].
\end{align}
For $1\le s<d$, we condition on $\Delta_{-S}$ and apply Gaussian moment bounds to the second term 
on the right hand side of~\eqref{eq:sketch_population_targets}.
The corresponding contribution from  the pre-training error in the frozen coordinates is $(s/n)((d-s)/(n\tilde\sigma^2+d-s))^2$.
Hence, by Corollary~\ref{cor:common_population_partial_mean_scale}, we obtain
\begin{align*}
\mathbb E_\Delta\left[
(1-\theta_\Delta)^2\|(\Sigma_1\Delta)_S\|_{\Sigma_{1,SS}^{-1}}^2\right]
&\asymp\frac{s}{n}\left(
\left(\frac{d-s}{n\tilde\sigma^2+d-s}\right)^2
+\min\bigl(1,(\delta/\tilde\sigma)^4\bigr)\right).
\end{align*}
We obtain the rates in the four regimes by substituting their scale conditions, while the finite sample remainder is treated in Step~4.
When $s=d$, the frozen coordinate contribution is zero.
For the sharper finite sample bound, we condition on the acceptance indicators and obtain $\mathcal V_1\asymp\min(1,(\delta/\tilde\sigma)^4)d/n$ under the sample conditions in Corollary~\ref{cor:full_V1_direct}.

\paragraph{Step 3: Analysis of the $\mathcal V_2$ rates.}
We analyze $\mathcal V_2$ through the conditional covariance over generation noise.
By the decomposition in Step~1, the leading covariance contribution is determined by $\bar G(\Delta)^{-1}(\mu(\Delta)-G(\Delta)\bar G(\Delta)^{-1}\bar\mu(\Delta))$.
First, we evaluate its conditional covariance trace for fixed $(\Delta,\tilde X)$.
The centered numerator can be expressed as
\begin{align*}
\mu(\Delta)-G(\Delta)\bar G(\Delta)^{-1}\bar\mu(\Delta)
=\sum_{i=1}^m\mathbb I_i\tilde x_{i,S}
\left(
\tilde\varepsilon_i-\tilde x_{i,S}^\top
\left(\Delta_S+\bar G(\Delta)^{-1}\bar\mu(\Delta)\right)
\right).
\end{align*}
Since $\bar G(\Delta)^{-1}$ is fixed conditioned on $(\Delta,\tilde X)$ and the generation noises are independent across $i$, we obtain
\begin{align*}
&\tr\!\left(\Sigma_{1,SS}\Cov_{\tilde\varepsilon}\!\left(
\bar G(\Delta)^{-1}
\bigl(\mu(\Delta)-G(\Delta)\bar G(\Delta)^{-1}\bar\mu(\Delta)\bigr)
\middle|\Delta,\tilde X\right)\right) \\
&=\tr\!\Biggl(\Sigma_{1,SS}\bar G(\Delta)^{-1}
\Biggl[\sum_{i=1}^m\tilde x_{i,S}\tilde x_{i,S}^{\top}
\Var_{\tilde\varepsilon_i}\left(
\mathbb I_i\left(
\tilde\varepsilon_i-\tilde x_{i,S}^\top
\left(\Delta_S+\bar G(\Delta)^{-1}\bar\mu(\Delta)\right)
\right)
\middle|\Delta,\tilde X\right)\Biggr]
\bar G(\Delta)^{-1}\Biggr).
\end{align*}

For the population conditional covariance, we replace $\bar G(\Delta)$ by $m\bar G_{\mathrm{pop}}(\Delta)$ and $\bar G(\Delta)^{-1}\bar\mu(\Delta)$ by $\bar G_{\mathrm{pop}}(\Delta)^{-1}\bar\mu_{\mathrm{pop}}(\Delta)$.
By the population conditional mean identity in Step~2, we have $\Delta_S+\bar G_{\mathrm{pop}}(\Delta)^{-1}\bar\mu_{\mathrm{pop}}(\Delta)=\theta_\Delta\Sigma_{1,SS}^{-1}(\Sigma_1\Delta)_S$.
Thus, we evaluate the conditional variance over generation noise of $\mathbb I(\tilde\varepsilon-\theta_\Delta\tilde x_S^\top\Sigma_{1,SS}^{-1}(\Sigma_1\Delta)_S)$ and then take expectation over $\tilde x$.
By Lemma~\ref{lem:common_population_residual_covariance}, we obtain
\begin{align*}
c\tilde\sigma^2(1-\theta_\Delta)\bar G_{\mathrm{pop}}(\Delta)
&\preceq\mathbb E_{\tilde x}\!\left[
\tilde x_S\tilde x_S^\top
\Var_{\tilde\varepsilon}\!\left(
\mathbb I\left(
\tilde\varepsilon-\theta_\Delta\tilde x_S^\top
\Sigma_{1,SS}^{-1}(\Sigma_1\Delta)_S
\right)\middle|\Delta,\tilde x\right)
\middle|\Delta\right]
\preceq\tilde\sigma^2(1-\theta_\Delta)\bar G_{\mathrm{pop}}(\Delta).
\end{align*}
Therefore, the resulting population covariance is bounded above and below by constant multiples of
\begin{align*}
\left(m\bar G_{\mathrm{pop}}(\Delta)\right)^{-1}
\left(m\tilde\sigma^2(1-\theta_\Delta)\bar G_{\mathrm{pop}}(\Delta)\right)
\left(m\bar G_{\mathrm{pop}}(\Delta)\right)^{-1}
=\frac{\tilde\sigma^2(1-\theta_\Delta)}m
\bar G_{\mathrm{pop}}(\Delta)^{-1}.
\end{align*}
We take the trace weighted by $\Sigma_{1,SS}$ and then take expectation over $\Delta$.  The resulting population contribution to $\mathcal V_2$ is $m^{-1}\mathbb E_\Delta[\tilde\sigma^2(1-\theta_\Delta)\tr\bigl(\Sigma_{1,SS}\bar G_{\mathrm{pop}}(\Delta)^{-1}\bigr)]$.
To analyze this expectation, we note that it follows from Lemmas~\ref{lem:common_population_theta_bounds} and~\ref{lem:common_population_covariance} that 
\begin{align*}
\tilde\sigma^2(1-\theta_\Delta)
&\asymp\min(\delta^2,\tilde\sigma^2)
+\frac{\tilde\sigma^2\|\Delta_{-S}\|_{\Sigma_{1,-S\mid S}}^2}
{\tilde\sigma^2+\|\Delta_{-S}\|_{\Sigma_{1,-S\mid S}}^2},\\
\tr\bigl(\Sigma_{1,SS}\bar G_{\mathrm{pop}}(\Delta)^{-1}\bigr)
&\asymp\frac1{p_\Delta}\left(
s+\frac{\|(\Sigma_1\Delta)_S\|_{\Sigma_{1,SS}^{-1}}^2}
{\tilde\sigma^2+\delta^2+\|\Delta_{-S}\|_{\Sigma_{1,-S\mid S}}^2}\right).
\end{align*}
Then, we substitute these expressions and obtain
\begin{align}
&\frac1m\mathbb E_\Delta\left[
\tilde\sigma^2(1-\theta_\Delta)
\tr\bigl(\Sigma_{1,SS}\bar G_{\mathrm{pop}}(\Delta)^{-1}\bigr)
\right]\\
&\quad\asymp
\frac1m\mathbb E_\Delta\left[
\left(
\min(\delta^2,\tilde\sigma^2)
+\frac{\tilde\sigma^2
\|\Delta_{-S}\|_{\Sigma_{1,-S\mid S}}^2}
{\tilde\sigma^2+\|\Delta_{-S}\|_{\Sigma_{1,-S\mid S}}^2}
\right)
\frac1{p_\Delta}
\left(
s+\frac{\|(\Sigma_1\Delta)_S\|_{\Sigma_{1,SS}^{-1}}^2}
{\tilde\sigma^2+\delta^2+
\|\Delta_{-S}\|_{\Sigma_{1,-S\mid S}}^2}
\right)
\right].  \label{eq:sketch_population_V2_target}
\end{align}
Observe that ~\eqref{eq:sketch_population_V2_target} separates the contributions from generation noise after verification and pre-training error in the frozen coordinates, with $mp_\Delta$ as the effective accepted sample size.

For partial fine-tuning, we condition on $\Delta_{-S}$, as in Step~2.
By Gaussian conditioning and covariance regularity, we have $cs/n\le\mathbb E_\Delta[\|(\Sigma_1\Delta)_S\|_{\Sigma_{1,SS}^{-1}}^2\mid\Delta_{-S}]\le C(s/n+\|\Delta_{-S}\|_{\Sigma_{1,-S\mid S}}^2)$.
Since $\|\Delta_{-S}\|_{\Sigma_{1,-S\mid S}}^2/(\tilde\sigma^2+\delta^2+\|\Delta_{-S}\|_{\Sigma_{1,-S\mid S}}^2)\le1$ and $s\ge1$, it follows that
\begin{align*}
\mathbb E_\Delta\!\left[
s+\frac{\|(\Sigma_1\Delta)_S\|_{\Sigma_{1,SS}^{-1}}^2}
{\tilde\sigma^2+\delta^2+\|\Delta_{-S}\|_{\Sigma_{1,-S\mid S}}^2}
\middle|\Delta_{-S}\right]
&\asymp s\left(
1+\frac{1}{n(\tilde\sigma^2+\delta^2+
\|\Delta_{-S}\|_{\Sigma_{1,-S\mid S}}^2)}
\right).
\end{align*}
To analyze the inverse moment, we use the Gaussian distribution of $\Delta_{-S}$.
Let $k=d-s$, and $Z_1,\ldots,Z_k$ be independent standard normal variables,
and $X_k=\sum_{\ell=1}^kZ_\ell^2\sim\chi_k^2$. By the Gaussian distribution of
$\Delta$ and the covariance bounds, it holds that 
\begin{align*}
n\|\Delta_{-S}\|_{\Sigma_{1,-S\mid S}}^2
&\overset{\mathrm d}{=}\sum_{\ell=1}^ka_\ell Z_\ell^2,
\qquad cX_k\le\sum_{\ell=1}^ka_\ell Z_\ell^2\le CX_k.
\end{align*}
By Lemma~\ref{lem:common_weighted_population_inverse_trace}, the additional factor in the population rate is comparable to the following inverse moment.
With $t=1/(n(\tilde\sigma^2+\delta^2))$, we have
\begin{align*}
1+\frac1n\mathbb E_\Delta\left[
\frac1{\tilde\sigma^2+\delta^2+
\|\Delta_{-S}\|_{\Sigma_{1,-S\mid S}}^2}\right]
\asymp1+t\mathbb E[(1+tX_k)^{-1}]
\asymp
\begin{cases}
1+\sqrt t, & k=1,\\
\log(e+t), & k=2,\\
1, & k\ge3,
\end{cases}
=:g_{k,d}(\sqrt{dt}),
\end{align*}
where the last comparison follows from inverse moment bounds for the $\chi^2$ distribution in
Lemma~\ref{lem:chisq_denom_uniform}.
Hence, for $1\le s<d$, the population contribution in~\eqref{eq:sketch_population_V2_target} is comparable to
\begin{align*}
\frac{s}{m\min(1,\delta/\sqrt{\tilde\sigma^2+d/n})}
\left(\min(\delta^2,\tilde\sigma^2)
+\frac{\tilde\sigma^2(d-s)}{n\tilde\sigma^2+d-s}\right)
g_{d-s,d}\left(\sqrt{\frac{d}{n(\tilde\sigma^2+\delta^2)}}\right).
\end{align*}
The factor $g_{d-s,d}(\sqrt{d/(n(\tilde\sigma^2+\delta^2))})$ has constant order in Regimes~I and~II.
Since $\tilde\sigma^2+\delta^2\asymp\tilde\sigma^2$ in Regime~III and $\tilde\sigma^2+\delta^2\asymp\delta^2$ in Regime~IV, the corresponding factors are $g_{d-s,d}(\sqrt{d/n}/\tilde\sigma)$ and $g_{d-s,d}(\sqrt{d/n}/\delta)$, respectively.

When $s=d$, the frozen coordinate terms are zero, while the additional factor in the population contribution to $\mathcal V_2$ becomes $1+1/(n(\tilde\sigma^2+\delta^2))$.
We provide the details of these calculations in
Lemma~\ref{lem:common_weighted_population_inverse_trace} and the resulting
$\mathcal V_2$ rates for the four regimes in
Appendix~\ref{sec:regime_specializations}.
Next, we bound the differences between the empirical and population quantities and the inverse remainder to establish the finite sample upper and lower bounds.

\paragraph{Step 4: Establish the finite sample bounds.}
Finally, let $0<\eta\le1/2$ and fix $\Delta$ satisfying the sample condition~\eqref{eq:common_delta_empirical_sample}.
By the concentration inequalities over $\tilde X$,
with conditional probability at
least $1-\eta$, we have
\begin{align*}
\frac12m\bar G_{\mathrm{pop}}(\Delta)
\preceq\bar G(\Delta)\preceq\frac32m\bar G_{\mathrm{pop}}(\Delta), \quad 
\bar G(\Delta)^{-1}\bar\mu(\Delta) &
=\bar G_{\mathrm{pop}}(\Delta)^{-1}\bar\mu_{\mathrm{pop}}(\Delta)
+e_{\mu,m}(\Delta,\tilde X),\\
\mathbb E_{\tilde\varepsilon}\left[
\left\|\bar G(\Delta)^{-1}
\bigl(\mu(\Delta)-G(\Delta)\bar G(\Delta)^{-1}\bar\mu(\Delta)\bigr)
\right\|_{\Sigma_{1,SS}}^2\middle|\Delta,\tilde X\right]
&\asymp\frac{\tilde\sigma^2(1-\theta_\Delta)}m
\tr\bigl(\Sigma_{1,SS}\bar G_{\mathrm{pop}}(\Delta)^{-1}\bigr),
\end{align*}
where the $\Sigma_{1,SS}$ norm of $e_{\mu,m}(\Delta,\tilde X)$ has an $m^{-1/2}$
factor, with the complete finite sample bound stated in
Lemma~\ref{lem:common_empirical_prediction_mean}.
Under the stated sample conditions, the Gaussian bounds for $\Delta$ and concentration over $\tilde X$ establish the above relations with failure probability at most $\eta^2$ over $(\Delta,\tilde X)$.
Then, the tower property and Markov's inequality imply that, with probability at least $1-\eta$ over $\tilde 
X$, the above relations hold with conditional probability at least $1-\eta$ over $\Delta$.

Furthermore, conditioned on $(\Delta,\tilde X)$, concentration over $\tilde\varepsilon$ justifies replacing $G
(\Delta)^{-1}$ by $\bar G(\Delta)^{-1}$ in Step~1.
The errors from this replacement and the empirical approximations account
for the remainder terms.
We bound these errors and the contribution on the complementary event,
then combine the bounds with the population calculations in Steps~2
and~3 to obtain the finite sample upper and lower bounds for partial fine-tuning in Regimes~I--IV.
The complete argument is provided in
Appendices~\ref{sec:common_empirical} and~\ref{sec:common_conditional_risk}.

%% file: sections/5iterative_training.tex
\section{Iterative Full Fine-Tuning with Verification}\label{sec:iterative_training}

Up to this section, we have characterized the prediction risk after a
single verified post-training update.
For the iterative full fine-tuning, the current estimator generates the responses in the next synthetic batch, and the accepted 
samples determine the next estimator.  
In this section, we study how the estimation error propagates across generations.

Let $\hat\beta^{(0)}$ be an initial estimator and define $\Delta^{(0)}=\beta^*-\hat\beta^{(0)}$.  For each generation $t\ge1$, given $\hat\beta^{(t-1)}$, we draw $m$ covariates from the downstream distribution and generate the corresponding synthetic responses.  For $j=1,\ldots,m$, we consider
\begin{align*}
\tilde y_j^{(t)}
&=(\tilde x_j^{(t)})^\top\hat\beta^{(t-1)}+\tilde\varepsilon_j^{(t)},
&
\tilde x_j^{(t)}
&\sim\mathcal N(0,\Sigma_1),
&
\tilde\varepsilon_j^{(t)}
&\sim\mathcal N(0,\tilde\sigma^2),
\end{align*}
where the covariates and generation noises are mutually independent across $j$ and $t$ and independent of $\hat\beta^{(0)}$.
At every generation, we follow the verification procedure.  Let $\delta>0$ be the verification threshold, and define the 
generated response error and acceptance indicator by
\begin{align*}
\xi_j^{(t)}
&:=\tilde y_j^{(t)}-(\tilde x_j^{(t)})^\top\beta^*
=-(\tilde x_j^{(t)})^\top\Delta^{(t-1)}+\tilde\varepsilon_j^{(t)},
&
\mathbb I_j^{(t)}
&:=\mathbf 1\{|\xi_j^{(t)}|\le\delta\},
\qquad j=1,\ldots,m.
\end{align*}
For each $t\ge1$, using the samples with $\mathbb I_j^{(t)}=1$, let $\hat\beta^{(t)}$ be the extended estimator in 
Definition~\ref{def:ext_estimator} with $s=d$ and $\hat\beta^{(t-1)}$ in place of $\hat\beta^{\mathrm{pre}}$, and define $\Delta^{(t)}
=\beta^*-\hat\beta^{(t)}$.  We fix the batch size $m$, verification threshold $\delta$, injected generation noise scale $\tilde\sigma>0$, and 
covariance $\Sigma_1$ across generations.
Now, we present the iterative full fine-tuning results as follows.

\begin{theorem}[Iterative Full Fine-Tuning Upper Bound]
\label{thm:iterative_full_finetuning}
Under the iterative model above, suppose that Assumption~\ref{ass:design} holds.  There exist constants $C>0$ and $c\in(0,1]$, depending only 
on $c_\lambda$, $C_\lambda$, and $C_G$.
If $m\ge C$, $0<\delta/\tilde\sigma\le c$, and $m\delta/\tilde\sigma\ge C\bigl(d^2\log m+(d+\log m)(\log m)^2\bigr)$, then, for every initial error
$\Delta^{(0)}$ satisfying $\|\Delta^{(0)}\|_{\Sigma_1}^2\le\tilde\sigma^2$ and every integer $T\ge0$, it holds that
\begin{align*}
\mathbb E[\|\Delta^{(T)}\|_{\Sigma_1}^2\mid\Delta^{(0)}]
&\le\bigl(2(\delta/\tilde\sigma)^4/9\bigr)^T\|\Delta^{(0)}\|_{\Sigma_1}^2
+\frac{1-\bigl(2(\delta/\tilde\sigma)^4/9\bigr)^T}{1-2(\delta/\tilde\sigma)^4/9}\frac{Cd\delta\tilde\sigma}{m}.
\end{align*}
\end{theorem}

We note that the condition $\|\Delta^{(0)}\|_{\Sigma_1}^2\le\tilde\sigma^2$ is required only at initialization.  If $\hat\beta^{(0)}=\hat\beta^{\mathrm
{pre}}$, $n\ge Cd$, and $\tilde\sigma^2\ge C(d+\log m)/n$, Lemma~\ref{lem:pretraining_covariance_regularity} and Corollary~\ref
{cor:pretraining_error_concentration} imply that this condition holds with probability at least $1-2e^{-cn}-m^{-c}$ over the pre-training 
data.  By the Theorem~\ref{thm:iterative_full_finetuning}, we have that the term arising from the pre-training error decreases 
by the factor $2(\delta/\tilde\sigma)^4/9$ at each generation, while the accumulated sampling error remains of order $d\delta\tilde\sigma/m$ 
uniformly in $T$.
In the following theorem, we present the complementary lower bound for the iterative full fine-tuning. 

\begin{theorem}[Iterative Full Fine-Tuning Lower Bound]
\label{thm:iterative_full_finetuning_lower_bound}
Under the iterative model above, there exists a universal constant $C>0$ such that, for every initial error $\Delta^{(0)}\in\mathbb R^d$ and integer $T\ge0$, it holds that
\begin{align*}
\mathbb E[\|\Delta^{(T)}\|_{\Sigma_1}^2\mid\Delta^{(0)}]
&\ge\biggl(\frac{\delta^2}{\delta^2+C\tilde\sigma^2}\biggr)^{2T}\|\Delta^{(0)}\|_{\Sigma_1}^2.
\end{align*}
\end{theorem}

We observe from the lower bound that the initial error cannot decrease rapidly when the verification threshold is large relative to the generation noise.  In particular, reducing this error by a fixed proportion requires at least a constant multiple of $(\delta/\tilde\sigma)^2$ generations.  The proofs of Theorem~\ref{thm:iterative_full_finetuning} and Theorem~\ref{thm:iterative_full_finetuning_lower_bound} are given in Appendix~\ref{sec:iterative_training_proofs}.

%% file: sections/6empirical_studies.tex

\section{Empirical Experiments}\label{sec:empirical_studies}

\begin{figure}[t]
  \centering
  \includegraphics[width=\textwidth]{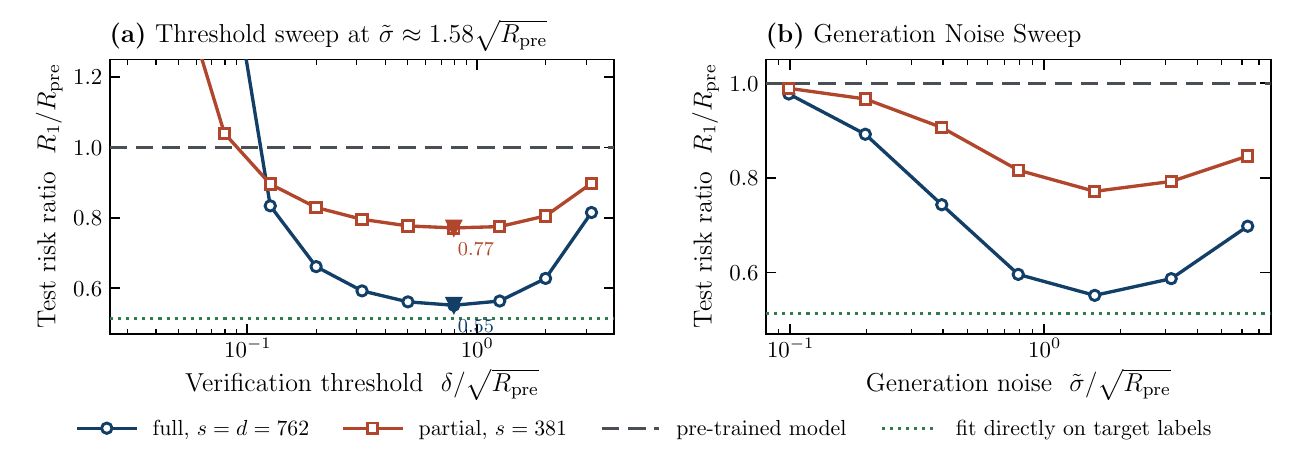}
  \caption{Test risk across verification thresholds and generation noise
    scales, with $n=2048$, $d=762$, and $m=32{,}000$.  Panel~(a) varies the
    threshold at $\tilde\sigma\approx1.58\sqrt{R_{\mathrm{pre}}}$, while
    panel~(b) presents the best average test risk ratio among the 11 examined
    thresholds for each generation noise scale.  Each setting averages 40
    runs.  The direct fitting reference fits the linear model directly on all
    $7229$ target human labels.}
  \label{fig:real_threshold}
\end{figure}

We present Gaussian simulations and real data experiments. The details of the
experiments are provided in Appendix~\ref{sec:empirical_details}.

\subsection{Gaussian Simulations}

We examine how the verification threshold and injected generation noise affect
the active coordinate variance after one post-training update under the
Gaussian design.  We set $\Sigma_1=I_d$, draw
$\Delta\sim\mathcal N(0,I_d/n)$, and take $d=20$, $n=2000$, and
$m=3\times10^6$.  We compare full fine-tuning ($s=d$) and partial
fine-tuning ($s=10$) through estimates of $\mathcal V_1+\mathcal V_2$.
Figure~\ref{fig:intro_verification} presents the three comparisons discussed
in the introduction, while
Figures~\ref{fig:full_delta}--\ref{fig:partial_sigma} display the results
across the four regimes.

The threshold comparisons illustrate how the effect of verification differs
across the four regimes.  For partial fine-tuning, the variance has a U-shaped
dependence on the threshold in Regime~II and remains near $s/n$ in
Regimes~III and~IV.  The generation noise comparisons show that nearly
deterministic generation provides limited improvement, while an intermediate
noise scale minimizes excess risk under both full and partial
fine-tuning.

\subsection{Real Data Experiments}

For the real data experiments, we consider semantic textual similarity,
where the task is to estimate a human similarity score between zero and five
for a pair of sentences.  We employ the STS2012--2015 collections
\citep{agirre2012semeval,agirre2013sem,agirre2014semeval,agirre2015semeval}
for pre-training, the STS-B training data
\citep{cer2017semeval} for post-training, and the STS-B test data for
evaluation.  For example, ``A man is playing a guitar'' and ``A man is
playing a trumpet'' have a similarity score of approximately $1.7$.  We
encode each sentence using
\texttt{all-MiniLM-L6-v2},\footnote{\url{https://huggingface.co/sentence-transformers/all-MiniLM-L6-v2}}
which is based on MiniLM \citep{wang2020minilm}.  The encoder has
approximately 22.7 million parameters and produces embeddings in
$\mathbb R^{384}$.  We construct pairwise covariates from these embeddings
and remove six nearly constant coordinates, which results in $d=762$.  In each run,
we fit the pre-trained linear estimator using a sample of size $n=2048$
from the pre-training data.

For post-training, we sample sentence pairs with replacement from the
post-training data and generate synthetic similarity scores by adding
Gaussian noise to the pre-trained predictions.  For the verification, we accept a
sample when $|\tilde y_j-y_j|\le\delta$, where $y_j$ is its human similarity
score.
Then, we  refit either all coefficients ($s=d$) or a fixed subset of $s=381$
coefficients using the accepted samples, while keeping the sentence encoder
fixed. This can be viewed as a post-training of the last linear prediction head of the model.

\begin{figure}[t]
  \centering
  \includegraphics[width=\textwidth]{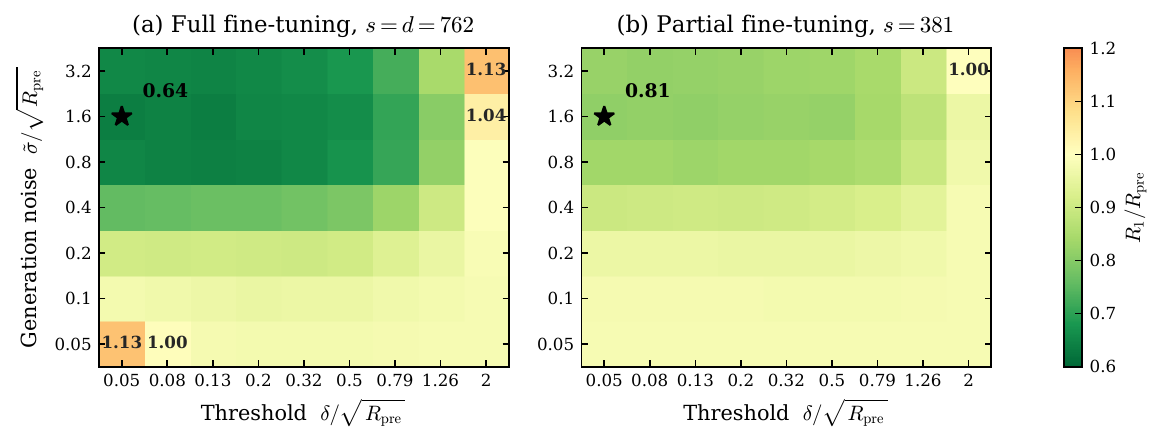}
  \caption{Test risk across verification thresholds and generation noise
    scales at a fixed accepted sample size, with $n=2048$ and $d=762$.
    Across 16 attempted runs per setting, every fitted estimator uses 3048
    accepted samples, and the star marks the minimum test risk ratio in each
    panel.}
  \label{fig:real_map}
\end{figure}

We evaluate each estimator by its mean squared error on the STS-B test data
and normalize the error by $R_{\mathrm{pre}}$, which is the average pre-training test
mean squared error.  In Figure~\ref{fig:real_threshold}, panel~(a) varies the
verification threshold over 11 logarithmically spaced values from $0.0381$ to $3.81$ at $\tilde\sigma\approx1.58\sqrt{R_{\mathrm{pre}}}$.
We observe that for both full and partial fine-tuning, the minimum test risk ratio is attained
at an intermediate verification threshold.
For the real data, we note that there exists also the effect of noisy verification, as the human similarity scores are noisy, 
and that can affect the risk behavior as well. 
For each injected generation noise scale, panel~(b) presents the best test risk ratio over the same threshold grid.  The
best ratio is attained at an intermediate generation noise scale.  Hence,
neither tighter verification nor smaller generation noise always reduces
test risk.  In Figure~\ref{fig:real_map}, when we fix the accepted sample size, we
find that the lowest test risk is attained at an intermediate scale of
generation noise with a tight verification threshold.

\FloatBarrier

%% file: sections/7conclusion.tex
\section{Conclusion}\label{sec:conclusion_main}

We have characterized
 nonasymptotic risk rates for verified post-training on self-generated synthetic data across four regimes determined by the 
verification threshold, the injected generation noise, and the pre-training error scale, under both partial and full fine-tuning.  The rates clarify how 
verification affects the error in the accepted responses, the effective accepted sample size, and the covariance of the accepted covariates, while 
partial fine-tuning introduces an additional contribution from the frozen coordinates.  
We have also investigated repeated full fine-tuning.  These results 
provide useful insights for the use of verification and synthetic data for post-training modern models, including large language models.  For future work, 
it would be interesting to extend these results to more general models and data distributions.

\paragraph*{Acknowledgments.}
The authors thank Chao Gao and Pragya Sur for helpful comments.

\paragraph*{Statement on AI use.}
We used ChatGPT to help polish the text, check grammar and the consistency of mathematical
notation, and check mathematical derivations. We also used ChatGPT
to help implement, monitor, and debug the experiments.
The authors are responsible for the final content of the paper.

%% file: Unified_appendix/app_00_preamble.tex

\paragraph{Organization of the appendix.}
In Sections~\ref{sec:common_probability_tools} and~\ref{sec:design_assumptions},
we establish the probability bounds and pre-training covariance regularity used in the appendix.
In Sections~\ref{sec:foundations} and~\ref{sec:common_var_framework}, we derive the risk and residual decompositions, establish that the active bias equals zero, and express the active variance through $\mathcal V_1$ and $\mathcal V_2$.
In Section~\ref{sec:unified_population}, we derive the population mean and covariance of the accepted samples.
In Sections~\ref{sec:common_empirical} and~\ref{sec:common_conditional_risk}, we
 establish empirical Gram matrix, conditional mean, and conditional covariance bounds for fixed $\Delta$ and 
 use a common event for the realized synthetic covariates to bound $\mathcal V_1$ and $\mathcal V_2$.
In Section~\ref{sec:full_V1_common}, we complete the full fine-tuning analysis of $\mathcal V_1$.
In Section~\ref{sec:regime_specializations}, we specialize the resulting bounds to the four regimes and prove the results.
In Section~\ref{sec:common_moment_tools}, we establish the supporting moment bounds.
In Section~\ref{sec:iterative_training_proofs}, we prove the iterative full fine-tuning results.
In Section~\ref{sec:empirical_details}, we provide further details for the Gaussian simulations and real data experiments.

For the complement event arguments in the appendix, $d+\log m$ is
bounded above by a fixed polynomial in $m$.
We use the notation and conditions introduced in
Sections~\ref{sec:problem_setting} and~\ref{section:main_results}.

%% file: Unified_appendix/app_01A_probability_tools.tex

\section{Common Probability Tools}
\label{sec:common_probability_tools}

We state the probability bounds used in the appendix.


\begin{lemma}[Gaussian maximum norm bound, {\citet[Example~2.28]{wainwright2019high}}]\label{lem:gaussian_max_design}
Let $x_1,\ldots,x_m$ be i.i.d.\ $\mathcal N(0,\Sigma)$ random vectors in  $\mathbb R^d$, 
and suppose that $\|\Sigma\|_{\mathrm{op}}\le C_1$ for a
fixed constant $C_1>0$.  There exists $C>0$, depending only on $C_1$,
such that, for every fixed coordinate subset  $S\subseteq\{1,\ldots,d\}$ 
and every $0<\eta<1$, with probability at least $1-\eta$,
\begin{align*}
\max_{1\le i\le m}\|x_{i,S}\|^2
\le C\left\{|S|+\log\left(\frac{m}{\eta}\right)\right\}.
\end{align*}
\end{lemma}

In the subsequent proof, the concentration bounds are established over $\tilde X$ for each fixed $\Delta$, whereas the conditional risk is evaluated at the fixed $\tilde X$.
The following lemma bounds the conditional probability over $\Delta$ that these bounds fail.

\begin{lemma}[Conditional probability bound]
\label{lem:fixed_covariate_conversion}
Fix $X_{\mathrm{pre}}$ and let $0<\eta\le1/2$.  Conditioned on $X_{\mathrm{pre}}$, 
suppose that $\Delta$ and $\tilde X$ are independent.
Let $\Omega_\Delta$ be determined by $\Delta$ and satisfy
$\mathbb P(\Omega_\Delta\mid X_{\mathrm{pre}})\ge1-\eta/2$.
Let $\Omega_1(\Delta,\tilde X)$ be determined by $(\Delta,\tilde X)$ and suppose 
that, for every  $\Delta\in\Omega_\Delta$,
$\mathbb P(\Omega_1(\Delta,\tilde X)\mid X_{\mathrm{pre}},\Delta) \ge1-\eta^2/4$.
Then, conditioned on $X_{\mathrm{pre}}$, with probability at least  $1-\eta/2$ over $\tilde X$,
\begin{align*}
\mathbb P_{\Delta\mid X_{\mathrm{pre}},\tilde X}\left(
\Omega_\Delta\cap\Omega_1(\Delta,\tilde X)
\right)
&\ge1-\eta.
\end{align*}
\end{lemma}

\begin{proof}
By the tower property, we have
\begin{align*}
\mathbb E_{\tilde X\mid X_{\mathrm{pre}}}\left[
\mathbb P_{\Delta\mid X_{\mathrm{pre}},\tilde X}\left(
\Omega_\Delta\cap\Omega_1(\Delta,\tilde X)^{\mathrm c}
\right)\right]
&=\mathbb P_{\Delta,\tilde X\mid X_{\mathrm{pre}}}\left(
\Omega_\Delta\cap\Omega_1(\Delta,\tilde X)^{\mathrm c}
\right)\le\frac{\eta^2}{4},
\end{align*}
where the inequality follows by conditioning on $\Delta$ and applying the
assumed conditional probability bound.
Applying Markov's inequality, we obtain that, with probability at least $1-\eta/2$ over $\tilde X$, $\mathbb P_{\Delta\mid X_{\mathrm{pre}},\tilde X}(\Omega_\Delta\cap\Omega_1(\Delta,\tilde X)^{\mathrm c})\le\eta/2$.
Conditional independence also implies $\mathbb P_{\Delta\mid X_{\mathrm{pre}},\tilde X}(\Omega_\Delta^{\mathrm c})=\mathbb P_{\Delta\mid X_{\mathrm{pre}}}(\Omega_\Delta^{\mathrm c})\le\eta/2$.
Combining these bounds, we obtain the stated conditional probability bound.
\end{proof}

\begin{lemma}[Bernstein inequality, {\citet[Proposition~2.10]{wainwright2019high}}]
\label{lem:bernstein}
Let $X_1,\ldots,X_m$ be independent centered real random variables.
Suppose that, for some $V,B>0$,
$\sum_{i=1}^m\mathbb E|X_i|^k  \le(k!/2)VB^{k-2}$ for every integer $k\ge2$.
Then, for every $t>0$,
\begin{align*}
\mathbb P\left(
\left|\sum_{i=1}^mX_i\right|
\ge\sqrt{2Vt}+Bt
\right)
&\le2e^{-t}.
\end{align*}
\end{lemma}



\begin{lemma}[Hanson\text{--}Wright inequality, {\citet[Theorem~6.2.2]{vershynin2019high}}]
\label{lem:quad_form_concentration}
Let $Z\sim\mathcal N(0,\Sigma)$ in $\mathbb R^d$, where $\Sigma\succ0$.  
There exists a universal constant $C>0$ such that, for every $0<\eta<1$, 
with probability at least $1-\eta$,
\begin{align*}
\left|\|Z\|_2^2-\tr(\Sigma)\right|
&\le C\left(
\sqrt{\tr(\Sigma^2)\log(2/\eta)} +\|\Sigma\|_{\mathrm{op}}\log(2/\eta)
\right).
\end{align*}
Moreover, for every positive integer $r$,
\begin{align*}
\left(\mathbb E\|Z\|_2^{2r}\right)^{1/r}
&\le Cr\tr(\Sigma).
\end{align*}
\end{lemma}

\begin{lemma}[Matrix Bernstein inequality, {\citet[Theorem~6.1.1]{tropp2015introduction}}]
\label{lem:bounded_matrix_bernstein}
Let $Y_1,\ldots,Y_m$ be independent centered self-adjoint $d\times d$
random matrices.  Suppose that, for some $V,R>0$,
$\max_{1\le i\le m}\|Y_i\|_{\mathrm{op}} $ $\le R$ almost surely and
$\left\|\sum_{i=1}^m\mathbb E Y_i^2\right\|_{\mathrm{op}}\le V$.
There is a universal constant $C>0$ such that, for every $0<\eta<1$, 
with probability at least $1-\eta$,
\begin{align*}
\left\|\sum_{i=1}^mY_i\right\|_{\mathrm{op}}
&\le C\left(\sqrt{V\log(2d/\eta)}
+R\log(2d/\eta)\right).
\end{align*}
\end{lemma}

For a real random variable $Y$ and $\alpha\in\{1,2\}$, define
$\|Y\|_{\psi_\alpha}:=
\inf\bigl\{t>0:\mathbb E\exp(|Y|^\alpha/t^\alpha)\le2\bigr\}$.

\begin{lemma}[Sub-exponential vector concentration]
\label{lem:vector_bernstein}
Let $m,d\ge1$, and let $X_1,\ldots,X_m\in\mathbb R^d$ be independent
centered random vectors.  Suppose that $\max_{1\le i\le m}\bigl\|\|X_i\|_2\bigr\|_{\psi_1}<\infty$.
Then, there is a universal constant $C>0$ such that, for every $0<\eta<1$,
with probability at least $1-\eta$,
\begin{align*}
\left\|\sum_{i=1}^mX_i\right\|_2^2
&\le C\left\{\log\left(\frac{m+d}{\eta}\right)
\sum_{i=1}^m\mathbb E\|X_i\|_2^2
+\left(\max_{1\le i\le m}\bigl\|\|X_i\|_2\bigr\|_{\psi_1}\right)^2
\log^4\left(\frac{m+d}{\eta}\right)\right\}.
\end{align*}
\end{lemma}

\begin{proof}
Let $L=\max_{1\le i\le m}\bigl\|\|X_i\|_2\bigr\|_{\psi_1}$.
If $L=0$, then $X_i=0$ almost surely for every $i$, and the result is immediate.  
Suppose that $L>0$.  By the definition of the $\psi_1$-norm,
$\mathbb P(\|X_i\|_2>t)\le2\exp(-ct/L)$ for $t\ge0$ and an absolute
constant $c>0$.  Since the vectors are not necessarily almost surely bounded, we
truncate them before applying matrix Bernstein.  For a sufficiently large
universal constant $C$, let $T=CL\log((m+d)/\eta)$ and  $X_i^\circ=X_i\mathbf 1\{\|X_i\|_2\le T\}$.  
Taking a  union bound, we obtain
\begin{align*}
\mathbb P\left(\max_{1\le i\le m}\|X_i\|_2>T\right)
&\le\sum_{i=1}^m\mathbb P(\|X_i\|_2>T)
\le2m\exp(-cT/L)\le\frac{\eta}{3}.
\end{align*}
Thus, with probability at least $1-\eta/3$, it holds that  $X_i^\circ=X_i$  for every $i$.
We next bound the bias introduced by truncation.  Since $\mathbb E X_i=0$, tail integration implies
\begin{align*}
\left\|\sum_{i=1}^m\mathbb E X_i^\circ\right\|_2
&\le\sum_{i=1}^m\mathbb E\left[
\|X_i\|_2\mathbf 1\{\|X_i\|_2>T\}\right]
\le L\eta.
\end{align*}
For each $i$, let
\begin{align*}
Y_i:=\begin{pmatrix}
0&(X_i^\circ-\mathbb E X_i^\circ)^\top\\
X_i^\circ-\mathbb E X_i^\circ&0
\end{pmatrix}.
\end{align*}
These $(d+1)\times(d+1)$ matrices are independent, centered, and self-adjoint, and satisfy $\|Y_i\|_{\mathrm{op}}=\|X_i^\circ-\mathbb E X_i^\circ\|_2\le2T$.
Moreover, we have
\begin{align*}
\left\|\sum_{i=1}^m\mathbb E Y_i^2\right\|_{\mathrm{op}}
&=\sum_{i=1}^m\mathbb E\|X_i^\circ-\mathbb E X_i^\circ\|_2^2
\le\sum_{i=1}^m\mathbb E\|X_i\|_2^2.
\end{align*}
Since $\|\sum_{i=1}^mY_i\|_{\mathrm{op}}=\|\sum_{i=1}^m(X_i^\circ-\mathbb E X_i^\circ)\|_2$, we apply Lemma~\ref{lem:bounded_matrix_bernstein} with failure probability $2\eta/3$ and obtain
\begin{align*}
\left\|\sum_{i=1}^m(X_i^\circ-\mathbb E X_i^\circ)\right\|_2
&\le C\left\{\left(\log\left(\frac{m+d}{\eta}\right)
\sum_{i=1}^m\mathbb E\|X_i\|_2^2\right)^{1/2}
+L\log^2\left(\frac{m+d}{\eta}\right)\right\}
\end{align*}
with probability at least $1-2\eta/3$.
By combining these bounds and squaring, we obtain the claimed bound with probability at least $1-\eta$.  This completes the proof. 
\end{proof}

%% file: Unified_appendix/app_02B_pretraining_covariance.tex

\section{Pre-Training Covariance Concentration}
\label{sec:design_assumptions}

First, we define the event on which the pre-training sample covariance is well conditioned:
\begin{equation}\label{eq:Omega_pre_def}
\Omega_{\mathrm{pre}} :=
\left\{ \frac{c_\lambda}{2}I_d
\preceq\frac{1}{n}X_{\mathrm{pre}}^\top X_{\mathrm{pre}}  \preceq2C_\lambda I_d
\right\}.
\end{equation}

\begin{lemma}[Pre-training covariance bounds, {\citet[Remark~4.7.3]{vershynin2019high}}]
\label{lem:pretraining_covariance_regularity}
Suppose that Assumption~\ref{ass:design} holds.  There exist constants $c,C>0$,
depending only on $c_\lambda$ and $C_\lambda$, such that,
whenever $n\ge Cd$, the following bounds hold on $\Omega_{\mathrm{pre}}$, where $\mathbb P(\Omega_{\mathrm{pre}})\ge1-2e^{-cn}$:
\begin{equation}\label{eq:Gamma_pre_bounds}
\frac{1}{2C_\lambda}I_d
\preceq\Gamma(X_{\mathrm{pre}})
\preceq\frac{2}{c_\lambda}I_d.
\end{equation}
Moreover, uniformly over $S\subseteq\{1,\ldots,d\}$ with $|S|=s$,
\begin{equation}\label{eq:pretraining_trace_comparisons}
\tr(\Sigma_1\Gamma)\asymp d,\qquad
\tr\left(\Sigma_{1,SS}^{-1}
(\Sigma_1\Gamma\Sigma_1)_{SS}\right)\asymp s,\qquad
\tr(\Sigma_{1,-S\mid S}\Gamma_{-S,-S})\asymp d-s\quad(s<d).
\end{equation}
\end{lemma}

\paragraph{Pre-training error scales.}
By applying the covariance bounds, we bound the total and active
coordinate pre-training errors relative to the generation noise variance.
These quantities appear in the acceptance probability in
Section~\ref{sec:unified_population} and the sample conditions and variance
bounds in Section~\ref{sec:common_conditional_risk}.
On $\Omega_{\mathrm{pre}}$, the pre-training error satisfies
$\Delta\mid X_{\mathrm{pre}}\sim
\mathcal N(0,\Gamma(X_{\mathrm{pre}})/n)$.
For every $\Delta\in\mathbb R^d$, define
$\rho_\Delta^2:=\|\Delta\|_{\Sigma_1}^2/\tilde\sigma^2$.  For a
realization of $X_{\mathrm{pre}}$, we use the notation
$\Gamma=\Gamma(X_{\mathrm{pre}})$.  The conditional mean of
$\rho_\Delta^2$ is defined as
\begin{align*}
\bar\rho^2
:=\mathbb E_\Delta[\rho_\Delta^2\mid X_{\mathrm{pre}}]
=\frac{\tr(\Sigma_1\Gamma)}{n\tilde\sigma^2}.
\end{align*}
We also define $\rho^2=d/(n\tilde\sigma^2)$.  For the active coordinates,
define
\begin{align*}
\rho_{S,\Delta}^2
:=\frac{\|(\Sigma_1\Delta)_S\|_{\Sigma_{1,SS}^{-1}}^2}{\tilde\sigma^2},
\qquad
\bar\rho_S^2
:=\mathbb E_\Delta[\rho_{S,\Delta}^2\mid X_{\mathrm{pre}}]
=\frac{\tr\left(\Sigma_{1,SS}^{-1}
(\Sigma_1\Gamma\Sigma_1)_{SS}\right)}{n\tilde\sigma^2}.
\end{align*}
On $\Omega_{\mathrm{pre}}$,
Lemma~\ref{lem:pretraining_covariance_regularity} implies $\bar\rho\asymp\rho$ and
$\bar\rho_S^2\asymp s/(n\tilde\sigma^2)$.
Moreover, we establish high probability bounds for the total and active
pre-training error scales in the following corollary.

\begin{corollary}[Pre-training error bounds]
\label{cor:pretraining_error_concentration}
Under Assumption~\ref{ass:design}, fix  $X_{\mathrm{pre}}\in\Omega_{\mathrm{pre}}$.  
There exists $C>0$ such
that, for every $0<\eta\le1/2$, conditioned on $X_{\mathrm{pre}}$, with
probability at least $1-\eta$ over $\Delta$,
\begin{align*}
\rho_\Delta^2\le
\frac{C(d+\log(4/\eta))}{n\tilde\sigma^2},\qquad
\rho_{S,\Delta}^2\le
\frac{C(s+\log(4/\eta))}{n\tilde\sigma^2}.
\end{align*}
\end{corollary}

\begin{proof}
We apply Lemma~\ref{lem:quad_form_concentration}, conditioned on
$X_{\mathrm{pre}}$, with failure probability $\eta/2$ to the Gaussian
vector $\Sigma_1^{1/2}\Delta/\tilde\sigma$.  Its squared norm is $\rho_\Delta^2$, 
and its covariance is
$\Sigma_1^{1/2}\Gamma\Sigma_1^{1/2}/(n\tilde\sigma^2)$, whose trace is $\bar\rho^2$. 
 It follows from
Lemma~\ref{lem:pretraining_covariance_regularity} and
Assumption~\ref{ass:design} that its operator norm is at most  $C/(n\tilde\sigma^2)$, 
so that its squared Frobenius norm is at most
$Cd/(n\tilde\sigma^2)^2$.  Therefore, with probability at least  $1-\eta/2$, we have
\begin{align*}
\rho_\Delta^2\le\bar\rho^2+
\frac{C(\sqrt{d\log(4/\eta)}+\log(4/\eta))}{n\tilde\sigma^2}
\le\frac{C(d+\log(4/\eta))}{n\tilde\sigma^2},
\end{align*}
where the last inequality follows from $\bar\rho^2\asymp d/(n\tilde\sigma^2)$.
The analogous analysis applied to the Gaussian vector
$\Sigma_{1,SS}^{-1/2}(\Sigma_1\Delta)_S/\tilde\sigma$ provides, with
probability at least $1-\eta/2$,
\begin{align*}
\rho_{S,\Delta}^2\le
\frac{C(s+\log(4/\eta))}{n\tilde\sigma^2}.
\end{align*}
\end{proof}

%% file: Unified_appendix/app_03C_foundational_proofs.tex

\section{Risk Decomposition and Baseline without Verification}
\label{sec:foundations}
We prove the risk decomposition in  Proposition~\ref{prop:risk_decomp}, derive the residual
representation~\eqref{eq:residual_rep_main} and the decomposition of the
active variance in~\eqref{eq:V1V2_def}, and prove the baseline without verification
in Proposition~\ref{prop:unfiltered}.

\subsection{Proof of Proposition~\ref{prop:risk_decomp}}
\begin{proof}
Conditioned on $(X_{\mathrm{pre}},\tilde X)$, we have
\begin{align*}
R_1\bigl(\hat\beta;X_{\mathrm{pre}},\tilde X\bigr)
=\mathbb E\bigl[(y-x^\top\hat\beta)^2\big|X_{\mathrm{pre}},\tilde X\bigr]
= \sigma_1^2+\mathbb E\bigl[\|\hat\beta - \beta^*\|_{\Sigma_1}^2\big|X_{\mathrm{pre}},\tilde X\bigr],
\end{align*}
where the second equality follows from the downstream model.
We apply the Schur complement to the partitioned matrix $\Sigma_1$ and obtain $\|\hat\beta-\beta^*\|_{\Sigma_1}^2=\|\hat\beta_S-\beta_S^{\mathrm{corr}}\|_{\Sigma_{1,SS}}^2+\|\Delta_{-S}\|_{\Sigma_{1,-S\mid S}}^2$,
where $\beta_S^{\mathrm{corr}}=\beta_S^*+\Sigma_{1,SS}^{-1}\Sigma_{1,S,-S}\Delta_{-S}$.
We take the conditional expectation and apply the bias and variance decomposition to the active coordinate term to obtain $\mathrm{Bias}_S^2+\Var_S$.
Since $\Delta$ and $\tilde X$ are independent given $X_{\mathrm{pre}}$ 
and  $\mathbb E[\Delta\mid X_{\mathrm{pre}}]=0$,
$\mathbb E\bigl[\|\Delta_{-S}\|_{\Sigma_{1,-S\mid S}}^2\big|
X_{\mathrm{pre}},\tilde X\bigr]
=\tr\left(\Sigma_{1,-S\mid S}\Cov(\Delta_{-S}\mid
X_{\mathrm{pre}})\right)=\Var_{-S}$.
Thus, we obtain the stated risk decomposition.
\end{proof}

\subsection{Derivation of the Residual Representation}
Recall $r_S$ and $G(\Delta)$ from
Section~\ref{ssec:risk_decomp_main}.  Define the accepted residual numerator by
\begin{equation}\label{eq:mu_def_appendix}
\mu(\Delta):=\sum_{i=1}^m\mathbb I_i\tilde x_{i,S}
\bigl(\tilde x_{i,-S}^\top\Delta_{-S}+\xi_i\bigr).
\end{equation}
On $\Omega_{\mathrm{post}}$, $G(\Delta)$ is invertible.  For the post-training estimator, we have $\hat\beta_S^{\mathrm{post}}=G(\Delta)^{-1}\sum_{i=1}^m\mathbb I_i\tilde x_{i,S}\allowbreak\bigl(\tilde y_i-\tilde x_{i,-S}^\top\hat\beta_{-S}^{\mathrm{pre}}\bigr)$.
Since $\tilde y_i=\tilde x_i^\top\beta^*+\xi_i$ and $\hat\beta_{-S}^{\mathrm{pre}}=\beta_{-S}^*-\Delta_{-S}$, we have $\tilde y_i-\tilde x_{i,-S}^\top\hat\beta_{-S}^{\mathrm{pre}}=\tilde x_{i,S}^\top\beta_S^*+\tilde x_{i,-S}^\top\Delta_{-S}+\xi_i$.
By substituting this relation into the expression for the post-training estimator, we obtain $\hat\beta_S^{\mathrm{post}}=\beta_S^*+G(\Delta)^{-1}\mu(\Delta)$.
Then, we subtract $\beta_S^{\mathrm{corr}}=\beta_S^*+\Sigma_{1,SS}^{-1}\Sigma_{1,S,-S}\Delta_{-S}$ and obtain the residual representation
\begin{equation}\label{eq:residual_rep_main}
r_S=G(\Delta)^{-1}\mu(\Delta)
-\Sigma_{1,SS}^{-1}\Sigma_{1,S,-S}\Delta_{-S}.
\end{equation}
Equivalently, since
$\tilde x_{i,-S}^\top\Delta_{-S}+\xi_i  =-\tilde x_{i,S}^\top\Delta_S+\tilde\varepsilon_i$, 
we can express the residual as
\begin{equation}\label{eq:residual_rep_generation_noise}
r_S=-\Sigma_{1,SS}^{-1}(\Sigma_1\Delta)_S
+G(\Delta)^{-1}\sum_{i=1}^m
\mathbb I_i\tilde x_{i,S}\tilde\varepsilon_i.
\end{equation}

\subsection{Derivation of the Variance Decomposition}
For the extended estimator, we have $\Var_S=\tr(\Sigma_{1,SS}\Cov(r_{S,\mathrm{ext}}\mid X_{\mathrm{pre}},\tilde X))$.
We apply the law of total covariance with respect to $\Delta$ and obtain
\begin{align*}
\Cov(r_{S,\mathrm{ext}}\mid X_{\mathrm{pre}},\tilde X)
&=
\Cov\left(
\mathbb E_{\tilde\varepsilon}
[r_{S,\mathrm{ext}}\mid\Delta,\tilde X]
\;\middle|\;X_{\mathrm{pre}},\tilde X
\right)
+
\mathbb E\left[
\Cov_{\tilde\varepsilon}
(r_{S,\mathrm{ext}}\mid\Delta,\tilde X)
\;\middle|\;X_{\mathrm{pre}},\tilde X
\right].
\end{align*}
By multiplying by $\Sigma_{1,SS}$ and taking traces, we have
$\Var_S=\mathcal V_1+\mathcal V_2$, where $\mathcal V_1$ and  $\mathcal V_2$ 
are defined in~\eqref{eq:V1V2_def}.

\subsection{Proof of Proposition~\ref{prop:unfiltered}}
\begin{proof}
When $\delta=\infty$, every generated sample is accepted.  Hence,
$\mathbb I_i=1$ for every $i$ and
$G(\Delta)=G:=\sum_{i=1}^m\tilde x_{i,S}\tilde x_{i,S}^{\top}$.
Since the vectors $\Sigma_{1,SS}^{-1/2}\tilde x_{i,S}$ are independent and follow $\mathcal N(0,I_s)$, 
we apply \citet[Remark~4.7.3]{vershynin2019high} and use $m\ge C(s+\log m)$ 
to obtain $\frac{m}{2}\Sigma_{1,SS}\preceq G\preceq2m\Sigma_{1,SS}$ with probability at least $1-m^{-c_1}$ over $\tilde X$. 
 For sufficiently large $C$ in the sample condition, the lower bound on $G$ and
Assumption~\ref{ass:design} provide
$\lambda_{\min}(G/m)\ge c_\lambda/2\ge m^{-C_G}$.  
Thus, this
event is contained in $\Omega_{\mathrm{post}}$.  On its intersection with $\Omega_{\mathrm{pre}}$, 
the post-training estimator is well defined
and agrees with the extended estimator.

On this intersection, by~\eqref{eq:residual_rep_generation_noise} and $\mathbb I_i=1$, we have $r_S=-\Sigma_{1,SS}^{-1}(\Sigma_1\Delta)_S+G^{-1}\sum_{i=1}^m\tilde x_{i,S}\tilde\varepsilon_i$.
Therefore, $\mathbb E_{\tilde\varepsilon}[r_S\mid\Delta,\tilde X]
=-\Sigma_{1,SS}^{-1}(\Sigma_1\Delta)_S$.  Conditioned on
$X_{\mathrm{pre}}$, $\Delta$ is independent of $\tilde X$ and has covariance
$\Gamma/n$.  Hence,
\begin{align*}
\mathcal V_1
&=\tr\left(\Sigma_{1,SS}
\Cov\left(
\Sigma_{1,SS}^{-1}(\Sigma_1\Delta)_S
\mid X_{\mathrm{pre}},\tilde X
\right)\right)
=\frac{1}{n}
\tr\bigl(\Sigma_{1,SS}^{-1}(\Sigma_1\Gamma\Sigma_1)_{SS}\bigr).
\end{align*}
It follows from Lemma~\ref{lem:pretraining_covariance_regularity} that $\mathcal V_1\asymp s/n$ on $\Omega_{\mathrm{pre}}$.  When $s=d$, the trace reduces to   $\tr(\Sigma_1\Gamma)/n\asymp d/n$.

For $\mathcal V_2$, we employ the independence of the generation noises and obtain $\Cov_{\tilde\varepsilon}(r_S\mid\Delta,\tilde X)=\tilde\sigma^2G^{-1}$.
This covariance does not depend on $\Delta$.  Therefore,
$\mathcal V_2=\tilde\sigma^2\tr(\Sigma_{1,SS}G^{-1})$, and the bounds on $G$
imply $\tr(\Sigma_{1,SS}G^{-1})\asymp s/m$.  Since the conditional mean of $r_S$  is centered over $\Delta$, we have $\mathrm{Bias}_S^2=0$.
Then, we apply Proposition~\ref{prop:risk_decomp} and \eqref{eq:V1V2_def} to obtain the exact risk identities in the statement.  By combining the
event on which these bounds on $G$ hold with $\Omega_{\mathrm{pre}}$, we obtain the claimed probability and rate bounds.
\end{proof}

%% file: Unified_appendix/app_04D_variance_framework.tex

\section{Variance Decomposition}
\label{sec:common_var_framework}
In this section, we analyze $\mathcal V_1$ and $\mathcal V_2$ through the conditional mean and covariance of $r_{S,\mathrm{ext}}$ over the injected generation noise $\tilde\varepsilon$, 
with $(\Delta,\tilde X)$ fixed and $0<\delta<\infty$.
We derive the residual decomposition, identify the leading variance contributions,
and bound the inverse remainder and the complement contribution.  The sharper full fine-tuning rate for $\mathcal V_1$ is established in Section~\ref{sec:full_V1_common}.

\paragraph{Conditioning convention.}
Recall $G(\Delta)$ from Section~\ref{ssec:risk_decomp_main} and $\mu(\Delta)$ from~\eqref{eq:mu_def_appendix}.
Fix $X_{\mathrm{pre}}\in\Omega_{\mathrm{pre}}$ and let $\Gamma=\Gamma(X_{\mathrm{pre}})$, for which $\Delta\mid X_{\mathrm{pre}}\sim\mathcal N(0,\Gamma/n)$.
In this section, $\mathbb E_\Delta[\cdot]$ and $\Cov_\Delta(\cdot)$ denote expectation and covariance with respect to $\Delta\mid X_{\mathrm{pre}}$, while conditioning on $\tilde X$ is written explicitly.
For fixed $(\Delta,\tilde X)$, $\mathbb E_{\tilde\varepsilon}[\cdot\mid\Delta,\tilde X]$ denotes expectation over the generation noise.
Whenever an inverse is
multiplied by an event indicator, it is evaluated on that event and
interpreted as zero outside it.

\subsection{Conditional Mean and Expected Gram Matrix Bounds}
We  introduce the conditional means of the accepted Gram matrix and
residual numerator.  These quantities determine the leading conditional mean in
$\mathcal V_1$ and center the numerator in $\mathcal V_2$.
Then, we  compare $G(\Delta)^{-1}$ with
$\bar G(\Delta)^{-1}$ and derive the residual decomposition that applies to both
variance components.

\subsubsection{Conditional means}

For fixed $(\Delta,\tilde X)$, define the conditional means of   $G(\Delta)$ 
and the numerator $\mu(\Delta)$ in~\eqref{eq:residual_rep_main} by
\begin{equation}\label{eq:common_mu_def}
\bar G(\Delta):=\mathbb E_{\tilde\varepsilon}[G(\Delta)\mid\Delta,\tilde X],
\qquad
\bar\mu(\Delta):=\mathbb E_{\tilde\varepsilon}
[\mu(\Delta)\mid\Delta,\tilde X].
\end{equation}

Define $P_i(\Delta):=\mathbb P_{\tilde\varepsilon_i}\bigl(|-\tilde x_i^\top\Delta+\tilde\varepsilon_i|\le\delta\mid\Delta,\tilde X\bigr)$.
At fixed  $(\Delta,\tilde X)$, $P_i(\Delta)$ is the acceptance probability for the
realized covariate $\tilde x_i$, and $\sum_{i=1}^mP_i(\Delta)$ is the
conditional expected number of accepted samples.

\paragraph{Population quantities.}
To compare the conditional means with the corresponding expectations over the
covariates, define the one sample population quantities by
\begin{equation}\label{eq:common_Gpop_def}
\bar G_{\mathrm{pop}}(\Delta):=
\mathbb E_{\tilde x_i}\left[
P_i(\Delta)\tilde x_{i,S}\tilde x_{i,S}^\top\mid\Delta
\right], 
\quad
\bar\mu_{\mathrm{pop}}(\Delta):=
\mathbb E_{\tilde x_i,\tilde\varepsilon_i}\left[
\mathbb I_i\tilde x_{i,S}
(\tilde x_{i,-S}^\top\Delta_{-S}+\xi_i)\mid\Delta
\right].
\end{equation}
Note that by the linearity of expectation, $\mathbb E_{\tilde X}[\bar G(\Delta)\mid\Delta]=m\bar G_{\mathrm{pop}}(\Delta)$ and $\mathbb E_{\tilde X}[\bar\mu(\Delta)\mid\Delta]=m\bar\mu_{\mathrm{pop}}(\Delta)$.

\subsubsection{Accepted Gram matrices and inverse bounds}
Conditioned on $(\Delta,\tilde X)$, the accepted Gram matrix $G(\Delta)$ is random only 
with respect to the generation noise, and its
conditional expectation is $\bar G(\Delta)$.
We employ the following conditions to control $G(\Delta)^{-1}$.
When $\bar G(\Delta)\succ0$, define the maximal relative quadratic form
\begin{align*}
\ell_{\max}(\Delta)
:=
\max_{1\le i\le m}
\tilde x_{i,S}^\top\bar G(\Delta)^{-1}\tilde x_{i,S}.
\end{align*}
For $0<\eta\le1/2$ and a sufficiently small universal constant $c\in(0,1]$, let $\Omega_{\mathrm{emp}}(\Delta)$ be an event determined by $(\Delta,\tilde X)$ on which
\begin{align}\label{eq:common_accepted_gram_conditions}
\frac{1}{2}m\bar G_{\mathrm{pop}}(\Delta)&\preceq
\bar G(\Delta)\preceq\frac{3}{2}m\bar G_{\mathrm{pop}}(\Delta),
\qquad
\ell_{\max}(\Delta)\log(m/\eta)\le c.
\end{align}
These conditions are established in Section~\ref{sec:common_empirical} under the stated sample conditions.
By Lemma~\ref{lem:common_population_covariance}, $\bar G_{\mathrm{pop}}(\Delta)$ is positive definite.
Hence, by the first condition in \eqref{eq:common_accepted_gram_conditions},
 we have $\bar G(\Delta)\succ0$ and $\bar G(\Delta)^{-1}\preceq(2/m)\bar G_{\mathrm{pop}}(\Delta)^{-1}$.
With the population bounds in
Section~\ref{sec:unified_population}, this matrix inequality provides bounds for $\tr(\Sigma_{1,SS}\bar G(\Delta)^{-1})$ and
$\|\Sigma_{1,SS}^{1/2}\bar G(\Delta)^{-1/2}\|_{\mathrm{op}}^2$.
Under the second condition in \eqref{eq:common_accepted_gram_conditions}, we show in the following proposition that
the fluctuation of $G(\Delta)$ from $\bar G(\Delta)$ is sufficiently
small with high probability over the generation noise.

\begin{proposition}[Accepted Gram matrix concentration]
\label{prop:verify_G2}
Under the model in Section~\ref{sec:problem_setting}, 
fix  $(\Delta,\tilde X)$ with $\bar G(\Delta)\succ0$.  
Define
\begin{align*}
H(\Delta)
:=
\bar G(\Delta)^{-1/2}
\bigl(G(\Delta)-\bar G(\Delta)\bigr)
\bar G(\Delta)^{-1/2},
\end{align*}
and let
$\ell_{\max}(\Delta)
:=\max_{1\le i\le m}
\tilde x_{i,S}^\top\bar G(\Delta)^{-1}\tilde x_{i,S}$.
There exists a universal constant $C>0$ such that, for every
$0<\eta\le1/2$,
\begin{align*}
\|H(\Delta)\|_{\mathrm{op}}
\le
C\Bigl(
\sqrt{\ell_{\max}(\Delta)\log(m/\eta)}
+\ell_{\max}(\Delta)\log(m/\eta)
\Bigr)
\end{align*}
with probability at least $1-\eta$ over the generation noise.
\end{proposition}

\begin{proof}
Conditioned on $(\Delta,\tilde X)$, the independence of the
generation noises implies that the indicators $\mathbb I_i$ are
independent.  By the definition of $P_i(\Delta)$, we have $\mathbb E_{\tilde\varepsilon}[\mathbb I_i\mid\Delta,\tilde X]=P_i(\Delta)$.
Let $v_i=\bar G(\Delta)^{-1/2}\tilde x_{i,S}$.
By the definitions of $\bar G(\Delta)$ and $H(\Delta)$, we have
\begin{align*}
\sum_{i=1}^mP_i(\Delta)v_iv_i^\top
&=I_s,
\qquad
H(\Delta)=\sum_{i=1}^m(\mathbb I_i-P_i(\Delta))v_iv_i^\top,
\end{align*}
where the matrices $(\mathbb I_i-P_i(\Delta))v_iv_i^\top$ are independent, centered, and self-adjoint.

By the definition of $\ell_{\max}(\Delta)$, we have $\|v_i\|_2^2\le\ell_{\max}(\Delta)$.
Since $|\mathbb I_i-P_i(\Delta)|\le1$, it follows that $\|(\mathbb I_i-P_i(\Delta))v_iv_i^\top\|_{\mathrm{op}}=|\mathbb I_i-P_i(\Delta)|\|v_i\|_2^2\le\ell_{\max}(\Delta)$ almost surely.
Next, we bound the matrix variance of $H(\Delta)$. Observe that
\begin{align*}
\sum_{i=1}^m\mathbb E_{\tilde\varepsilon}[(\mathbb I_i-P_i(\Delta))^2\mid\Delta,\tilde X](v_iv_i^\top)^2
&=\sum_{i=1}^mP_i(\Delta)(1-P_i(\Delta))\|v_i\|_2^2v_iv_i^\top
\preceq\ell_{\max}(\Delta)I_s.
\end{align*}
Then, we can apply Lemma~\ref{lem:bounded_matrix_bernstein} with $V=R=\ell_{\max}(\Delta)$ and failure probability $\eta$ and obtain the stated bound.
\end{proof}
\noindent
For this choice of $\Omega_{\mathrm{emp}}(\Delta)$, define
\begin{align}\label{eq:common_full_event_def}
\Omega(\Delta)
&:=
\Omega_{\mathrm{emp}}(\Delta)
\cap
\left\{
\|H(\Delta)\|_{\mathrm{op}}
\le C\left(
\sqrt{\ell_{\max}(\Delta)\log(m/\eta)}
+\ell_{\max}(\Delta)\log(m/\eta)
\right)
\right\},
\end{align}
On $\Omega_{\mathrm{emp}}(\Delta)$,
Proposition~\ref{prop:verify_G2} provides
\begin{align}\label{eq:common_full_event_probability}
\mathbb P_{\tilde\varepsilon}\left(
\Omega(\Delta)\middle|\Delta,\tilde X\right)&\ge1-\eta.
\end{align}
For sufficiently small $c>0$, we have $\|H(\Delta)\|_{\mathrm{op}}\le1/2$ on $\Omega(\Delta)$.
By the definition of $H(\Delta)$, this implies $G(\Delta)\succeq\bar G(\Delta)/2\succ0$.
Hence, $G(\Delta)$ is invertible.
In the following residual decomposition,
we employ the bound on $H(\Delta)$ to control the error from replacing $G(\Delta)^{-1}$ by $\bar G(\Delta)^{-1}$. 
\subsubsection{Residual decomposition}

We decompose the post-training residual to study its conditional mean and covariance over the generation noise.
Recall from~\eqref{eq:residual_rep_main} that, on $\Omega_{\mathrm{post}}$, we have
\begin{align*}
r_S
&=G(\Delta)^{-1}\mu(\Delta)
-\Sigma_{1,SS}^{-1}\Sigma_{1,S,-S}\Delta_{-S}.
\end{align*}
On $\Omega_{\mathrm{post}}$, whenever $\bar G(\Delta)$ is invertible, we compare $G(\Delta)^{-1}\mu(\Delta)$ with $\bar G(\Delta)^{-1}\bar\mu(\Delta)$ through the identity
\begin{align*}
G(\Delta)^{-1}\mu(\Delta)-\bar G(\Delta)^{-1}\bar\mu(\Delta)
&=G(\Delta)^{-1}\left(\mu(\Delta)-G(\Delta)\bar G(\Delta)^{-1}\bar\mu(\Delta)\right).
\end{align*}
By the definitions of $\bar G(\Delta)$ and $\bar\mu(\Delta)$, we have $\mathbb E_{\tilde\varepsilon}[\mu(\Delta)-G(\Delta)\bar G(\Delta)^{-1}\bar\mu(\Delta)\mid\Delta,\tilde X]=0$.
However, $G(\Delta)^{-1}$ depends on the same acceptance indicators, and its product with this numerator can have a nonzero conditional mean.
We employ $G(\Delta)^{-1}=\bar G(\Delta)^{-1}+(G(\Delta)^{-1}-\bar G(\Delta)^{-1})$ in this difference and obtain
\begin{equation}\label{eq:common_residual_decomposition}
\begin{aligned}
r_S
&=\left(\bar G(\Delta)^{-1}\bar\mu(\Delta)
-\Sigma_{1,SS}^{-1}\Sigma_{1,S,-S}\Delta_{-S}\right)
+\bar G(\Delta)^{-1}
\left(\mu(\Delta)
-G(\Delta)\bar G(\Delta)^{-1}\bar\mu(\Delta)\right)\\
&\quad+\left(G(\Delta)^{-1}-\bar G(\Delta)^{-1}\right)
\left(\mu(\Delta)
-G(\Delta)\bar G(\Delta)^{-1}\bar\mu(\Delta)\right).
\end{aligned}
\end{equation}
Conditioned on $(\Delta,\tilde X)$, the first term in~\eqref{eq:common_residual_decomposition} is fixed, and the 
second term has conditional mean zero.
The last term is the error from replacing $G(\Delta)^{-1}$ by $\bar G(\Delta)^{-1}$.
Its conditional mean and second moment contribute to the bounds for $\mathcal V_1$ and $\mathcal V_2$, respectively.
By the definition of $\mu(\Delta)$ in~\eqref{eq:mu_def_appendix} and the identity 
$\tilde x_{i,-S}^{\top}\Delta_{-S}+\xi_i=-\tilde x_{i,S}^{\top}\Delta_S+\tilde\varepsilon_i$, we also have
\begin{equation}\label{eq:common_recentered_numerator}
\mu(\Delta)-G(\Delta)\bar G(\Delta)^{-1}\bar\mu(\Delta)
=\sum_{i=1}^m\mathbb I_i\tilde x_{i,S}
\left[\tilde\varepsilon_i-\tilde x_{i,S}^{\top}
\left(\Delta_S+\bar G(\Delta)^{-1}\bar\mu(\Delta)\right)\right].
\end{equation}

The decomposition in~\eqref{eq:common_residual_decomposition} expresses the
post-training residual $r_S$.  We note that the variance bounds are stated
for the extended estimator.  By~\eqref{eq:r_ext}, the corresponding residual
$r_{S,\mathrm{ext}}$ satisfies
\begin{align*}
r_{S,\mathrm{ext}}
=r_S\mathbf 1_{\Omega_{\mathrm{post}}}
-\Sigma_{1,SS}^{-1}(\Sigma_1\Delta)_S
\mathbf 1_{\Omega_{\mathrm{post}}^{\mathrm c}}.
\end{align*}
In the applications below, we verify $\Omega(\Delta)\subseteq\Omega_{\mathrm{post}}$.  
Hence,  $r_{S,\mathrm{ext}}=r_S$ on $\Omega(\Delta)$, and
\eqref{eq:common_residual_decomposition} 
holds on the event $\Omega(\Delta)$.
For $\Omega(\Delta)$ defined in~\eqref{eq:common_full_event_def}, define the mean squared inverse remainder
on $\Omega(\Delta)$ by
\begin{equation}\label{eq:Rres_common_def}
R_{\mathrm{res}}^{\Omega}
:=\mathbb E_{\Delta,\tilde\varepsilon}\left[
\left\|\left(G(\Delta)^{-1}-\bar G(\Delta)^{-1}\right)
\left(\mu(\Delta)-G(\Delta)\bar G(\Delta)^{-1}\bar\mu(\Delta)\right)
\right\|_{\Sigma_{1,SS}}^2\mathbf 1_{\Omega(\Delta)}
\middle|X_{\mathrm{pre}},\tilde X\right].
\end{equation}
We employ this decomposition to analyze $\mathcal V_1$ and $\mathcal V_2$, with the contribution on $\Omega(\Delta)^{\mathrm c}$ bounded separately.

\subsection{Conditional Mean Variance Contribution $\mathcal V_1$}

We apply~\eqref{eq:common_residual_decomposition} to analyze the conditional mean $\mathbb E_{\tilde\varepsilon}[r_{S,\mathrm{ext}}\mid\Delta,\tilde X]$, whose variation over $\Delta$ determines $\mathcal V_1$.
Recall that
\begin{align*}
\mathcal V_1
&=\tr\left(\Sigma_{1,SS}\Cov_\Delta\left(
\mathbb E_{\tilde\varepsilon}
[r_{S,\mathrm{ext}}\mid\Delta,\tilde X]\middle|\tilde X\right)\right).
\end{align*}
The first term in~\eqref{eq:common_residual_decomposition} depends on $\tilde X$ through $\bar G(\Delta)$ and $\bar\mu(\Delta)$.
We compare this term with the population conditional mean
$\bar G_{\mathrm{pop}}(\Delta)^{-1}\bar\mu_{\mathrm{pop}}(\Delta)
-\Sigma_{1,SS}^{-1}\Sigma_{1,S,-S}\Delta_{-S}.$
By symmetry in $\Delta$, this population conditional mean has expectation zero.
Hence, we compare $\mathcal V_1$ with its second moment
\begin{align*}
\mathbb E_\Delta\left[
\left\|\bar G_{\mathrm{pop}}(\Delta)^{-1}\bar\mu_{\mathrm{pop}}(\Delta)
-\Sigma_{1,SS}^{-1}\Sigma_{1,S,-S}\Delta_{-S}
\right\|_{\Sigma_{1,SS}}^2\right].
\end{align*}


We express the difference between the conditional mean of $r_{S,\mathrm{ext}}$ and the population conditional mean to bound $\mathcal V_1$.
We consider~\eqref{eq:common_residual_decomposition} on $\Omega(\Delta)$, where the inverse difference is controlled.
Suppose that $\Omega(\Delta)\subseteq\Omega_{\mathrm{post}}$.
Since $r_{S,\mathrm{ext}}=r_S$ on $\Omega(\Delta)$, we have
\begin{align*}
\mathbb E_{\tilde\varepsilon}
[r_{S,\mathrm{ext}}\mid\Delta,\tilde X]
&=\mathbb E_{\tilde\varepsilon}
[r_S\mathbf 1_{\Omega(\Delta)}\mid\Delta,\tilde X]
+\mathbb E_{\tilde\varepsilon}
[r_{S,\mathrm{ext}}\mathbf 1_{\Omega(\Delta)^{\mathrm c}}
\mid\Delta,\tilde X].
\end{align*}

First, we analyze $\mathbb E_{\tilde\varepsilon}[r_S\mathbf 1_{\Omega(\Delta)}\mid\Delta,\tilde X]$.
The second term in~\eqref{eq:common_residual_decomposition} has conditional mean zero, but its contribution on $\Omega(\Delta)$ can be nonzero.
Therefore, on $\Omega_{\mathrm{emp}}(\Delta)$, we have
\begin{align*}
\mathbb E_{\tilde\varepsilon}\left[
\left(\mu(\Delta)-G(\Delta)\bar G(\Delta)^{-1}\bar\mu(\Delta)\right)
\mathbf 1_{\Omega(\Delta)}\mid\Delta,\tilde X\right]
=-\mathbb E_{\tilde\varepsilon}\left[
\left(\mu(\Delta)-G(\Delta)\bar G(\Delta)^{-1}\bar\mu(\Delta)\right)
\mathbf 1_{\Omega(\Delta)^{\mathrm c}}\mid\Delta,\tilde X\right].
\end{align*}
Multiplying~\eqref{eq:common_residual_decomposition} by $\mathbf 1_{\Omega(\Delta)}$ and taking conditional expectation over $\tilde\varepsilon$, we obtain
\begin{equation}\label{eq:common_V1_remainder_decomposition}
\begin{aligned}
&\mathbb E_{\tilde\varepsilon}\left[
r_S\mathbf 1_{\Omega(\Delta)}\mid\Delta,\tilde X\right]
-\left(\bar G_{\mathrm{pop}}(\Delta)^{-1}\bar\mu_{\mathrm{pop}}(\Delta)
-\Sigma_{1,SS}^{-1}\Sigma_{1,S,-S}\Delta_{-S}\right)
\mathbf 1_{\Omega_{\mathrm{emp}}(\Delta)}\\
&\quad=\mathbb P_{\tilde\varepsilon}\left(
\Omega(\Delta)\mid\Delta,\tilde X\right)
\left(\bar G(\Delta)^{-1}\bar\mu(\Delta)
-\bar G_{\mathrm{pop}}(\Delta)^{-1}\bar\mu_{\mathrm{pop}}(\Delta)\right)
\mathbf 1_{\Omega_{\mathrm{emp}}(\Delta)}\\
&\qquad-\mathbb P_{\tilde\varepsilon}\left(
\Omega(\Delta)^{\mathrm c}\mid\Delta,\tilde X\right)
\left(\bar G_{\mathrm{pop}}(\Delta)^{-1}\bar\mu_{\mathrm{pop}}(\Delta)
-\Sigma_{1,SS}^{-1}\Sigma_{1,S,-S}\Delta_{-S}\right)
\mathbf 1_{\Omega_{\mathrm{emp}}(\Delta)}\\
&\qquad-\bar G(\Delta)^{-1}
\mathbb E_{\tilde\varepsilon}\left[
\left(\mu(\Delta)
-G(\Delta)\bar G(\Delta)^{-1}\bar\mu(\Delta)\right)
\mathbf 1_{\Omega(\Delta)^{\mathrm c}}
\mid\Delta,\tilde X\right]
\mathbf 1_{\Omega_{\mathrm{emp}}(\Delta)}\\
&\qquad+\mathbb E_{\tilde\varepsilon}\left[
\left(G(\Delta)^{-1}-\bar G(\Delta)^{-1}\right)
\left(\mu(\Delta)
-G(\Delta)\bar G(\Delta)^{-1}\bar\mu(\Delta)\right)
\mathbf 1_{\Omega(\Delta)}\mid\Delta,\tilde X\right].
\end{aligned}
\end{equation}
For the first term in~\eqref{eq:common_V1_remainder_decomposition}, we bound the mean squared difference between $\bar G(\Delta)^{-1}\bar\mu(\Delta)$ and\linebreak $\bar G_{\mathrm{pop}}(\Delta)^{-1}\bar\mu_{\mathrm{pop}}(\Delta)$.
For the second and third terms, we employ~\eqref{eq:common_full_event_probability} and the conditional Cauchy--Schwarz inequality.
By conditional Jensen's inequality, the squared $\Sigma_{1,SS}$ norm of the fourth term, after taking expectation over $\Delta$, is bounded by $R_{\mathrm{res}}^\Omega$ defined in~\eqref{eq:Rres_common_def}.

After taking expectation over $\Delta$, conditional Jensen's inequality and Lemma~\ref{lem:extended_estimator_complement} also bound the contribution of $r_{S,\mathrm{ext}}$ on $\Omega(\Delta)^{\mathrm c}$.
We also bound the population conditional mean on $\Omega_{\mathrm{emp}}(\Delta)^{\mathrm c}$.
These bounds control the expected squared difference between the conditional means.
We evaluate them in Section~\ref{sec:common_conditional_risk} to obtain the upper and lower bounds for $\mathcal V_1$.
In what follows, we analyze the conditional covariance contribution $\mathcal V_2$.

\subsection{Generation Noise Variance Contribution $\mathcal V_2$}

We analyze the second variance component:
\begin{align*}
\mathcal V_2
&=\mathbb E_\Delta\left[
\tr\left(\Sigma_{1,SS}
\Cov_{\tilde\varepsilon}(r_{S,\mathrm{ext}}\mid\Delta,\tilde X)
\right)\middle|\tilde X\right].
\end{align*}

\noindent
Suppose that $\Omega(\Delta)\subseteq\Omega_{\mathrm{post}}$.
By~\eqref{eq:common_residual_decomposition} and $r_{S,\mathrm{ext}}=r_S$ on $\Omega(\Delta)$, we obtain
\begin{equation}\label{eq:common_V2_event_split}
\begin{aligned}
&r_{S,\mathrm{ext}}-
(\bar G(\Delta)^{-1}\bar\mu(\Delta)
-\Sigma_{1,SS}^{-1}\Sigma_{1,S,-S}\Delta_{-S})
\mathbf 1_{\Omega_{\mathrm{emp}}(\Delta)}\\
&\quad=\bar G(\Delta)^{-1}
(\mu(\Delta)-G(\Delta)\bar G(\Delta)^{-1}\bar\mu(\Delta))
\mathbf 1_{\Omega_{\mathrm{emp}}(\Delta)}\\
&\qquad+(G(\Delta)^{-1}-\bar G(\Delta)^{-1})
(\mu(\Delta)-G(\Delta)\bar G(\Delta)^{-1}\bar\mu(\Delta))
\mathbf 1_{\Omega(\Delta)}\\
&\qquad+\Bigl\{r_{S,\mathrm{ext}}-
(\bar G(\Delta)^{-1}\bar\mu(\Delta)
-\Sigma_{1,SS}^{-1}\Sigma_{1,S,-S}\Delta_{-S})
\mathbf 1_{\Omega_{\mathrm{emp}}(\Delta)}\\
&\qquad\quad
-\bar G(\Delta)^{-1}
(\mu(\Delta)-G(\Delta)\bar G(\Delta)^{-1}\bar\mu(\Delta))
\mathbf 1_{\Omega_{\mathrm{emp}}(\Delta)}\Bigr\}
\mathbf 1_{\Omega(\Delta)^{\mathrm c}}.
\end{aligned}
\end{equation}

We compute the conditional covariance of the first term on the right hand side of~\eqref{eq:common_V2_event_split}.
Whenever $\bar G(\Delta)$ is invertible, define
\begin{equation}\label{eq:M_def}
\begin{aligned}
M(\Delta)
&:=\Cov_{\tilde\varepsilon}\left(
\mu(\Delta)
-G(\Delta)\bar G(\Delta)^{-1}\bar\mu(\Delta)
\middle|\Delta,\tilde X\right)\\
&=\sum_{i=1}^m\tilde x_{i,S}\tilde x_{i,S}^{\top}
\Var_{\tilde\varepsilon_i}\left(
\mathbb I_i\left[\tilde\varepsilon_i-
\tilde x_{i,S}^{\top}\left(\Delta_S+\bar G(\Delta)^{-1}\bar\mu(\Delta)\right)\right]
\middle|\Delta,\tilde X\right),
\end{aligned}
\end{equation}
where the second equality follows from~\eqref{eq:common_recentered_numerator} and the independence of the generation noise across samples.
By the definitions in~\eqref{eq:common_mu_def}, the numerator has conditional mean zero. Hence,
\begin{equation}\label{eq:common_V2_leading_pointwise}
\mathbb E_{\tilde\varepsilon}\left[
\left\|\bar G(\Delta)^{-1}
\left(\mu(\Delta)
-G(\Delta)\bar G(\Delta)^{-1}\bar\mu(\Delta)\right)
\right\|_{\Sigma_{1,SS}}^2
\middle|\Delta,\tilde X\right]
=\tr\left(\Sigma_{1,SS}\bar G(\Delta)^{-1}
M(\Delta)\bar G(\Delta)^{-1}\right).
\end{equation}
This is the leading conditional covariance contribution to $\mathcal V_2$.

We compare $\mathcal V_2$ with this leading contribution.
We multiply this identity by $\mathbf 1_{\Omega_{\mathrm{emp}}(\Delta)}$ and take expectation over $\Delta$.
For the two remainder terms in~\eqref{eq:common_V2_event_split}, the conditional covariance trace of their sum is bounded by its expected squared norm.
Hence, by the triangle inequality $\left|(\mathbb E\|x\|^2)^{1/2}-(\mathbb E\|y\|^2)^{1/2}\right|^2 $  $\le\mathbb E\|x-y\|^2$ and the definition of $R_{\mathrm{res}}^\Omega$ in~\eqref{eq:Rres_common_def}, we obtain
\begin{equation}\label{eq:V2_common_comparison}
\begin{aligned}
&\left|\mathcal V_2^{1/2}
-\left(\mathbb E_\Delta\left[
\tr\left(\Sigma_{1,SS}\bar G(\Delta)^{-1}M(\Delta)
\bar G(\Delta)^{-1}\right)
\mathbf 1_{\Omega_{\mathrm{emp}}(\Delta)}
\middle|X_{\mathrm{pre}},\tilde X\right]\right)^{1/2}\right|^2\\
&\quad\le\mathbb E_\Delta\!\Biggl[
\tr\Biggl(\Sigma_{1,SS}\Cov_{\tilde\varepsilon}\!\Bigl(
r_{S,\mathrm{ext}}-\bar G(\Delta)^{-1}
(\mu(\Delta)-G(\Delta)\bar G(\Delta)^{-1}\bar\mu(\Delta))
\mathbf 1_{\Omega_{\mathrm{emp}}(\Delta)}
\Bigm|\Delta,\tilde X\Bigr)\Biggr)
\Bigm|X_{\mathrm{pre}},\tilde X\Biggr]\\
&\quad\le R_{\mathrm{res}}^\Omega
+\mathbb E_{\Delta,\tilde\varepsilon}\Biggl[
\Bigl\|r_{S,\mathrm{ext}}-
(\bar G(\Delta)^{-1}\bar\mu(\Delta)
-\Sigma_{1,SS}^{-1}\Sigma_{1,S,-S}\Delta_{-S})
\mathbf 1_{\Omega_{\mathrm{emp}}(\Delta)}\\
&\qquad\quad-\bar G(\Delta)^{-1}
(\mu(\Delta)-G(\Delta)\bar G(\Delta)^{-1}\bar\mu(\Delta))
\mathbf 1_{\Omega_{\mathrm{emp}}(\Delta)}
\Bigr\|_{\Sigma_{1,SS}}^2
\mathbf 1_{\Omega(\Delta)^{\mathrm c}}
\Bigm|X_{\mathrm{pre}},\tilde X\Biggr].
\end{aligned}
\end{equation}
This inequality provides both upper and lower bounds for $\mathcal V_2$.
The leading and remainder terms are evaluated in Section~\ref{sec:common_conditional_risk}.

\subsection{Remainder Bounds}

The following lemmas bound $R_{\mathrm{res}}^\Omega$ and the contribution of $r_{S,\mathrm{ext}}$ on $\Omega(\Delta)^{\mathrm c}$ in the analysis of $\mathcal V_1$ and $\mathcal V_2$.

\begin{lemma}[Inverse remainder bound]
\label{lem:common_centered_inverse}
Fix $0<\eta\le1/2$. Let $\Omega_{\mathrm{emp}}(\Delta)$ be an event determined by $(\Delta,\tilde X)$ on which~\eqref{eq:common_accepted_gram_conditions} holds.
Let $\Omega(\Delta)$ and $R_{\mathrm{res}}^\Omega$ be defined by~\eqref{eq:common_full_event_def} and~\eqref{eq:Rres_common_def}, respectively.
Then, it holds that
\begin{align*}
R_{\mathrm{res}}^{\Omega}
&\le C\mathbb E_\Delta\left[
\ell_{\max}(\Delta)\log(m/\eta)
\bigl\|\Sigma_{1,SS}^{1/2}\bar G(\Delta)^{-1/2}\bigr\|_{\mathrm{op}}^2
\tr\left(\bar G(\Delta)^{-1}M(\Delta)\right)
\mathbf 1_{\Omega_{\mathrm{emp}}(\Delta)}
\middle|X_{\mathrm{pre}},\tilde X\right].
\end{align*}
\end{lemma}

\begin{proof}
On $\Omega(\Delta)$, equation~\eqref{eq:common_full_event_def} and $\ell_{\max}(\Delta)\log(m/\eta)\le c$ imply $\|H(\Delta)\|_{\mathrm{op}}^2\le C\ell_{\max}(\Delta)\log(m/\eta)$.
The factorization  $G(\Delta)=\bar G(\Delta)^{1/2}(I_s+H(\Delta))\bar G(\Delta)^{1/2}$
implies
\begin{align*}
G(\Delta)^{-1}-\bar G(\Delta)^{-1}
=-\bar G(\Delta)^{-1/2}(I_s+H(\Delta))^{-1}
H(\Delta)\bar G(\Delta)^{-1/2}.
\end{align*}
Since $\|H(\Delta)\|_{\mathrm{op}}\le1/2$ on $\Omega(\Delta)$, we have $\|(I_s+H(\Delta))^{-1}\|_{\mathrm{op}}\le2$.
Hence,
\begin{align*}
&\bigl\|(G(\Delta)^{-1}-\bar G(\Delta)^{-1})
\bigl(\mu(\Delta)
-G(\Delta)\bar G(\Delta)^{-1}\bar\mu(\Delta)\bigr)
\bigr\|_{\Sigma_{1,SS}}^2\\
&\quad\le C\ell_{\max}(\Delta)\log(m/\eta)
\bigl\|\Sigma_{1,SS}^{1/2}\bar G(\Delta)^{-1/2}\bigr\|_{\mathrm{op}}^2
\bigl\|\mu(\Delta)
-G(\Delta)\bar G(\Delta)^{-1}\bar\mu(\Delta)
\bigr\|_{\bar G(\Delta)^{-1}}^2.
\end{align*}
The centered numerator has conditional mean zero and conditional covariance
$M(\Delta)$.  Consequently,
\begin{align*}
\mathbb E_{\tilde\varepsilon}\left[
\bigl\|\mu(\Delta)
-G(\Delta)\bar G(\Delta)^{-1}\bar\mu(\Delta)
\bigr\|_{\bar G(\Delta)^{-1}}^2
\middle|\Delta,\tilde X\right]
=\tr(\bar G(\Delta)^{-1}M(\Delta)).
\end{align*}
We multiply the pointwise bound by $\mathbf 1_{\Omega(\Delta)}$ and use $\mathbf 1_{\Omega(\Delta)}\le\mathbf 1_{\Omega_{\mathrm{emp}}(\Delta)}$ on its right hand side.
Conditional expectation first over $\tilde\varepsilon$ and then over $\Delta$ results in the claimed bound.
\end{proof}

We bound the contribution of $r_{S,\mathrm{ext}}$ on $\Omega(\Delta)^{\mathrm c}$ in terms of its conditional probability.

\begin{lemma}[Complement bound for the extended estimator]
\label{lem:extended_estimator_complement}
Under the model in Section~\ref{sec:problem_setting} and
Assumption~\ref{ass:design}, fix $C_G\ge2$ as in
Definition~\ref{def:ext_estimator}, and let $r_{S,\mathrm{ext}}$ be defined
by~\eqref{eq:r_ext}.
Fix $X_{\mathrm{pre}}\in\Omega_{\mathrm{pre}}$ and $\tilde X$ satisfying $\max_{1\le i\le m}\|\tilde x_i\|^2\le C(d+\log m)$.
Then, for $\Omega(\Delta)$ defined by~\eqref{eq:common_full_event_def}, it holds that
\begin{align*}
&\mathbb E_{\Delta,\tilde\varepsilon}\left[
\|r_{S,\mathrm{ext}}\|_{\Sigma_{1,SS}}^2
\mathbf 1_{\Omega(\Delta)^{\mathrm c}}
\middle|X_{\mathrm{pre}},\tilde X\right]
\le C m^{2C_G}(d+\log m)^2(d/n+\tilde\sigma^{2})
\mathbb P_{\Delta,\tilde\varepsilon}
(\Omega(\Delta)^{\mathrm c}\mid X_{\mathrm{pre}},\tilde X)^{1/2}.
\end{align*}
\end{lemma}

\begin{proof}
By~\eqref{eq:residual_rep_generation_noise} and~\eqref{eq:r_ext}, we have
\begin{align*}
r_{S,\mathrm{ext}}
&=-\Sigma_{1,SS}^{-1}(\Sigma_1\Delta)_S
+\left(G(\Delta)^{-1}\sum_{i=1}^m
\mathbb I_i\tilde x_{i,S}\tilde\varepsilon_i\right)
\mathbf 1_{\Omega_{\mathrm{post}}}.
\end{align*}
By the definition of $\Omega_{\mathrm{post}}$, we have $\|G(\Delta)^{-1}\|_{\mathrm{op}}\le m^{C_G-1}$ on $\Omega_{\mathrm{post}}$.
We apply the Cauchy--Schwarz inequality and the eigenvalue bounds in Assumption~\ref{ass:design} to obtain $\|r_{S,\mathrm{ext}}\|_{\Sigma_{1,SS}}\le C\|\Delta\|+Cm^{C_G-1/2}(d+\log m)^{1/2}\|\tilde\varepsilon\|$.

On $\Omega_{\mathrm{pre}}$, the conditional Gaussian law of $\Delta$ and Lemma~\ref{lem:pretraining_covariance_regularity} imply $\mathbb E_\Delta[\|\Delta\|^4\mid X_{\mathrm{pre}},\tilde X]\le C(d/n)^2$.
Moreover, $\mathbb E_{\tilde\varepsilon}\|\tilde\varepsilon\|^4=m(m+2)\tilde\sigma^4$.
Therefore,
\begin{align*}
\mathbb E_{\Delta,\tilde\varepsilon}\bigl[
\|r_{S,\mathrm{ext}}\|_{\Sigma_{1,SS}}^4
\big|X_{\mathrm{pre}},\tilde X\bigr]
&\le C(d/n)^2+Cm^{4C_G}(d+\log m)^2\tilde\sigma^4\\
&\le Cm^{4C_G}(d+\log m)^4(d/n+\tilde\sigma^2)^2.
\end{align*}
Finally, the Cauchy--Schwarz inequality, conditioned on $(X_{\mathrm{pre}},\tilde X)$, results in
\begin{align*}
\mathbb E[\|r_{S,\mathrm{ext}}\|_{\Sigma_{1,SS}}^2
\mathbf 1_{\Omega(\Delta)^{\mathrm c}}]
&\le
\mathbb E[\|r_{S,\mathrm{ext}}\|_{\Sigma_{1,SS}}^4]^{1/2}
\mathbb P(\Omega(\Delta)^{\mathrm c})^{1/2}.
\end{align*}
The claimed bound follows from the fourth moment bound. 
\end{proof}

\subsection{Active Coordinate Bias}
We show that the extended estimator has zero active coordinate bias.

\begin{lemma}[Active coordinate bias]
\label{lem:zero_active_bias}
Under the model in Section~\ref{sec:problem_setting}, the extended estimator
in Definition~\ref{def:ext_estimator} satisfies
$\mathbb E[r_{S,\mathrm{ext}}\mid X_{\mathrm{pre}},\tilde X]=0$ and hence
$\mathrm{Bias}_S^2=0$.
\end{lemma}

\begin{proof}
Conditioned on $(X_{\mathrm{pre}},\tilde X)$, the pairs $(\Delta,\tilde\varepsilon)$ and $(-\Delta,-\tilde\varepsilon)$ have the same distribution.
Under this sign change, $\xi_i$ changes sign, while the acceptance indicators, $G(\Delta)$, and $\Omega_{\mathrm{post}}$ remain unchanged.
By~\eqref{eq:residual_rep_generation_noise} and~\eqref{eq:r_ext}, $r_{S,\mathrm{ext}}$ also changes sign.

At fixed $(X_{\mathrm{pre}},\tilde X)$, we have $\|G(\Delta)^{-1}\|_{\mathrm{op}}\le m^{C_G-1}$ on $\Omega_{\mathrm{post}}$.
The same residual representations bound $\|r_{S,\mathrm{ext}}\|$ by a finite multiple of $\|\Delta\|+\sum_{i=1}^m|\tilde\varepsilon_i|$, which has finite conditional expectation.
Hence, symmetry results in $\mathbb E[r_{S,\mathrm{ext}}\mid X_{\mathrm{pre}},\tilde X]=0$, and $\mathrm{Bias}_S^2=0$.
\end{proof}

%% file: Unified_appendix/app_07E_accepted_population_moments.tex

\section{Population Mean and Covariance for Accepted Samples}
\label{sec:unified_population}

In Section~\ref{sec:common_var_framework}, we have
 analyzed $\mathcal V_1$ and $\mathcal V_2$ through the conditional mean and 
 covariance of $r_{S,\mathrm{ext}}$ over the generation noise, with $(\Delta,\tilde X)$ fixed.
These expressions depend on the realized synthetic covariates.
Now, we evaluate the population quantities required for these bounds, with $\Delta$ fixed.
In particular, the conditional means of the accepted Gram matrix and residual numerator satisfy
\begin{align*}
\mathbb E_{\tilde X}[\bar G(\Delta)\mid\Delta]
&=m\bar G_{\mathrm{pop}}(\Delta),\qquad
\mathbb E_{\tilde X}[\bar\mu(\Delta)\mid\Delta]
=m\bar\mu_{\mathrm{pop}}(\Delta).
\end{align*}
For $\mathcal V_1$, we evaluate the population conditional mean $\bar G_{\mathrm{pop}}(\Delta)^{-1}\bar\mu_{\mathrm{pop}}(\Delta)-\Sigma_{1,SS}^{-1}\Sigma_{1,S,-S}\Delta_{-S}$.
For $\mathcal V_2$, we replace $\bar G(\Delta)^{-1}\bar\mu(\Delta)$ in $M(\Delta)$, defined in~\eqref{eq:M_def}, by $\bar G_{\mathrm{pop}}(\Delta)^{-1}\bar\mu_{\mathrm{pop}}(\Delta)$ and take expectation over the synthetic covariates.
First, we establish the acceptance probability and conditional moment bounds used in these population formulas.
The empirical bounds are established in Section~\ref{sec:common_empirical}.

\subsection{Acceptance Probabilities and Conditional Moments}

To proceed, fix $\Delta\in\mathbb R^d$ and $0<\delta<\infty$.
Let $\tilde x^\top$ denote a generic row of $\tilde X$, where $\tilde x\sim\mathcal N(0,\Sigma_1)$, and let $\tilde\varepsilon\sim\mathcal N(0,\tilde\sigma^2)$ be its injected generation noise, independent of $\tilde x$.
Denote the generated-response error and acceptance indicator by $\xi=-\tilde x^\top\Delta+\tilde\varepsilon$ and $\mathbb I=\mathbf 1\{|\xi|\le\delta\}$, respectively.
Conditioned on $\Delta$, $\xi$ is centered Gaussian with variance $\sigma_\xi^2(\Delta):=\tilde\sigma^2+\|\Delta\|_{\Sigma_1}^2$.

For $Z\sim\mathcal N(0,1)$ and $h>0$, define
\begin{align*}
p(h)&:=\mathbb P(|Z|\le h)=2\Phi(h)-1,\qquad
\lambda_2(h):=\mathbb E[Z^2\mid |Z|\le h],\qquad
\lambda_4(h):=\mathbb E[Z^4\mid |Z|\le h].
\end{align*}
Let $u=\delta/\tilde\sigma$ and recall that $\rho_\Delta^2=\|\Delta\|_{\Sigma_1}^2/\tilde\sigma^2$.
Then, $\sigma_\xi^2(\Delta)=\tilde\sigma^2(1+\rho_\Delta^2)$, and we define
\begin{align*}
h_\Delta&:=\frac{\delta}{\sigma_\xi(\Delta)}
=\frac{u}{\sqrt{1+\rho_\Delta^2}},\qquad
\lambda_{2,\Delta}:=\lambda_2(h_\Delta),\qquad
\lambda_{4,\Delta}:=\lambda_4(h_\Delta).
\end{align*}
The acceptance probability is $p_\Delta:=\mathbb E_{\tilde x_i}[P_i(\Delta)\mid\Delta]=\mathbb P(|\xi|\le\delta\mid\Delta)=p(h_\Delta)$.
Moreover, $\lambda_{2,\Delta}$ and $\lambda_{4,\Delta}$ are the conditional second and fourth moments of $\xi/\sigma_\xi(\Delta)$ given acceptance.

\begin{lemma}[Truncated normal probability and moment bounds]
\label{lem:truncated_normal_moments}
For every $h>0$,
\begin{align*}
\lambda_2(h)&=1-\frac{2h\phi(h)}{p(h)},\qquad
\lambda_2(h)^2\le\lambda_4(h)\le3\lambda_2(h)^2.
\end{align*}
Uniformly over $h>0$,
\begin{align*}
p(h)\asymp\min(1,h),
\qquad
\lambda_2(h)\asymp\min(1,h^2),
\qquad
\lambda_4(h)\asymp\min(1,h^4).
\end{align*}
\end{lemma}

\begin{proof}
Integration by parts leads to
\begin{align*}
p(h)\lambda_2(h)
&=2\int_0^h z^2\phi(z)dz=p(h)-2h\phi(h),\\
p(h)\lambda_4(h)
&=2\int_0^h z^4\phi(z)dz
=3p(h)\lambda_2(h)-2h^3\phi(h).
\end{align*}
The first identity establishes the stated formula for $\lambda_2(h)$.
Substituting this  relation into the second, we obtain $\lambda_4(h)=3\lambda_2(h)-h^2(1-\lambda_2(h))$.

We compare the second and fourth moments.
Since $z^2$ is increasing and $\phi(z)$ is decreasing on $[0,h]$, integration of $(z^2-y^2)(\phi(z)-\phi(y))\le0$ over $(z,y)\in[0,h]^2$ results in
\begin{align*}
h\int_0^h z^2\phi(z)dz
&\le
\left(\int_0^h z^2dz\right)
\left(\int_0^h\phi(z)dz\right).
\end{align*}
Hence, $\lambda_2(h)\le h^2/3$.
Jensen's inequality implies $\lambda_4(h)\ge\lambda_2(h)^2$.
Since $0<\lambda_2(h)<1$ by its definition and formula, we also have $3\lambda_2(h)^2-\lambda_4(h)=(1-\lambda_2(h))(h^2-3\lambda_2(h))\ge0$.

Finally, for $0<h\le1$, the bounds $\phi(1)\le\phi(z)\le\phi(0)$ on $[0,h]$ imply $p(h)\asymp h$ and 
$p(h)\lambda_2(h)=2\int_0^h z^2\phi(z)dz\asymp h^3$.
For $h\ge1$, we have $p(1)\le p(h)\le1$ and $2\int_0^1 z^2\phi(z)dz\le p(h)\lambda_2(h)\le1$.
The two ranges imply the stated bounds for $p(h)$ and $\lambda_2(h)$.
The bounds for $\lambda_4(h)$ follow from $\lambda_2(h)^2\le\lambda_4(h)\le3\lambda_2(h)^2$.
\end{proof}

The acceptance probability $p_\Delta$ determines the effective accepted sample size $mp_\Delta$.
We apply Lemma~\ref{lem:truncated_normal_moments} with $h=h_\Delta=u/\sqrt{1+\rho_\Delta^2}$ and obtain, uniformly over $\Delta\in\mathbb R^d$ and $u>0$,
\begin{equation}\label{eq:marginal_acceptance_scale}
p_\Delta\asymp
\min\left(1,\frac{u}{\sqrt{1+\rho_\Delta^2}}\right).
\end{equation}


To express the population formulas in terms of the active and frozen coordinates, we note that, by the Schur complement, $\|\Delta\|_{\Sigma_1}^2=\|(\Sigma_1\Delta)_S\|_{\Sigma_{1,SS}^{-1}}^2+\|\Delta_{-S}\|_{\Sigma_{1,-S\mid S}}^2$.
As a result, $\sigma_\xi^2(\Delta)=\tilde\sigma^2+\|(\Sigma_1\Delta)_S\|_{\Sigma_{1,SS}^{-1}}^2+\|\Delta_{-S}\|_{\Sigma_{1,-S\mid S}}^2$.

\subsection{Population Accepted Covariate Second Moment}

We evaluate $\bar G_{\mathrm{pop}}(\Delta)$ to obtain the population inverse bounds required
 for $\mathcal V_1$ and $\mathcal V_2$.
By its definition and the tower property, $\bar G_{\mathrm{pop}}(\Delta)=\mathbb E_{\tilde x,\tilde\varepsilon}[\mathbb I\tilde x_S\tilde x_S^\top\mid\Delta]$.

\begin{lemma}[Population accepted covariate second moment]
\label{lem:common_population_covariance}
Under the model in Section~\ref{sec:problem_setting}, suppose that $0<\delta<\infty$ and  $\tilde\sigma>0$.  
Then, for every $\Delta\in\mathbb R^d$,
\begin{equation}\label{eq:common_population_gram}
\bar G_{\mathrm{pop}}(\Delta)=p_\Delta\left(\Sigma_{1,SS}-
\frac{1-\lambda_{2,\Delta}}{\sigma_\xi^2(\Delta)}
(\Sigma_1\Delta)_S(\Sigma_1\Delta)_S^\top\right).
\end{equation}
Moreover, the matrix $\Sigma_{1,SS}^{-1/2}\bar G_{\mathrm{pop}}(\Delta)
\Sigma_{1,SS}^{-1/2}$ has $s-1$ eigenvalues equal to $p_\Delta$, and its
remaining eigenvalue equals
\begin{align*}
p_\Delta\frac{\tilde\sigma^2+
\|\Delta_{-S}\|_{\Sigma_{1,-S\mid S}}^2+
\lambda_{2,\Delta}\|(\Sigma_1\Delta)_S\|_{\Sigma_{1,SS}^{-1}}^2}
{\sigma_\xi^2(\Delta)}.
\end{align*}
\end{lemma}

\begin{proof}
Consider a fixed $v\in\mathbb R^s$.  By the definition of conditional expectation,
$v^\top\bar G_{\mathrm{pop}}(\Delta)v=p_\Delta
\mathbb E_{\tilde x,\tilde\varepsilon}[(v^\top\tilde x_S)^2
\mid\mathbb I=1,\Delta]$.  We apply
Lemma~\ref{lem:accepted_linear_form_moments} after extending $v$ by zero outside $S$ and obtain
\begin{align*}
v^\top\bar G_{\mathrm{pop}}(\Delta)v
&=p_\Delta v^\top\left(\Sigma_{1,SS}-
\frac{1-\lambda_{2,\Delta}}{\sigma_\xi^2(\Delta)}
(\Sigma_1\Delta)_S(\Sigma_1\Delta)_S^\top\right)v.
\end{align*}
Since $v$ is arbitrary, this identity leads to
\eqref{eq:common_population_gram}.
Moreover,
 \eqref{eq:common_population_gram} shows that $\Sigma_{1,SS}^{-1/2}\bar G_{\mathrm{pop}}(\Delta)\Sigma_{1,SS}^{-1/2}$ equals $p_\Delta I_s$ minus a matrix of rank at most one.
Hence, it has $s-1$ eigenvalues equal to $p_\Delta$ and one eigenvalue equal to $p_\Delta(1-(1-\lambda_{2,\Delta})\|(\Sigma_1\Delta)_S\|_{\Sigma_{1,SS}^{-1}}^2/\sigma_\xi^2(\Delta))$.
By the Schur complement decomposition of $\sigma_\xi^2(\Delta)$, the remaining eigenvalue is equivalent to the expression in the statement.
\end{proof}

Since $p_\Delta>0$ and $\tilde\sigma>0$, it follows from Lemma~\ref{lem:common_population_covariance} that $\bar G_{\mathrm{pop}}(\Delta)$ is positive definite.
Furthermore, we quantify how the remaining eigenvalue affects the population inverse bounds in Section~\ref{sec:common_empirical}.
By the Schur complement decomposition and the truncated normal moment bounds in Lemma~\ref{lem:truncated_normal_moments}, we obtain, uniformly over $\Delta\in\mathbb R^d$ and $u>0$,
\begin{equation}\label{eq:common_population_inverse_factor_scale}
\frac{\sigma_\xi^2(\Delta)}
{\tilde\sigma^2+\|\Delta_{-S}\|_{\Sigma_{1,-S\mid S}}^2+
\lambda_{2,\Delta}\|(\Sigma_1\Delta)_S\|_{\Sigma_{1,SS}^{-1}}^2}
\asymp1+\min\left(
\frac{\|(\Sigma_1\Delta)_S\|_{\Sigma_{1,SS}^{-1}}^2}
{\tilde\sigma^2+\|\Delta_{-S}\|_{\Sigma_{1,-S\mid S}}^2},
\frac{\sigma_\xi^2(\Delta)}{\delta^2}\right).
\end{equation}
For $h_\Delta\le1$, this relation follows from $\lambda_{2,\Delta}\asymp h_\Delta^2$, the Schur complement decomposition of $\sigma_\xi^2(\Delta)$, and $h_\Delta^{-2}=\sigma_\xi^2(\Delta)/\delta^2$.
For $h_\Delta>1$, we have $\lambda_{2,\Delta}\asymp1$, and both sides are comparable to one.

\subsection{Population Accepted Residual First Moment}

We evaluate $\bar\mu_{\mathrm{pop}}(\Delta)$ to express the population conditional mean in $\mathcal V_1$.
Recall that $\bar\mu_{\mathrm{pop}}(\Delta)=\mathbb E_{\tilde x,\tilde\varepsilon}[\mathbb I\tilde x_S(\tilde x_{-S}^\top\Delta_{-S}+\xi)\mid\Delta]$.

\begin{lemma}[Population accepted residual first moment]
\label{lem:common_population_mu}
Under the assumptions of Lemma~\ref{lem:common_population_covariance}, 
let  $\Delta\in\mathbb R^d$ and define
\begin{equation}\label{eq:theta_Delta_def}
\theta_\Delta:=
\frac{\tilde\sigma^2(1-\lambda_{2,\Delta})}
{\tilde\sigma^2+\|\Delta_{-S}\|_{\Sigma_{1,-S\mid S}}^2
+\lambda_{2,\Delta}\|(\Sigma_1\Delta)_S\|_{\Sigma_{1,SS}^{-1}}^2}.
\end{equation}
Then, it holds that
\begin{equation}\label{eq:common_population_mu}
\bar G_{\mathrm{pop}}(\Delta)^{-1}\bar\mu_{\mathrm{pop}}(\Delta)
=-\Delta_S+\theta_\Delta\Sigma_{1,SS}^{-1}(\Sigma_1\Delta)_S.
\end{equation}
\end{lemma}

\begin{proof}
Recall that $\tilde x_{-S}^\top\Delta_{-S}+\xi=-\tilde x_S^\top\Delta_S+\tilde\varepsilon$.
We have $\bar\mu_{\mathrm{pop}}(\Delta)=-\bar G_{\mathrm{pop}}(\Delta)\Delta_S+\mathbb E_{\tilde x,\tilde\varepsilon}[\mathbb I\tilde x_S\tilde\varepsilon\mid\Delta]$.
Then, we need to compute $\mathbb E_{\tilde x,\tilde\varepsilon}[\mathbb I\tilde x_S\tilde\varepsilon\mid\Delta]$.
Conditioned on $\Delta$, $(\tilde x_S,\tilde\varepsilon,\xi)$ is jointly centered Gaussian, with $\Cov(\tilde x_S,\tilde\varepsilon\mid\Delta)=0$, $\Cov(\tilde x_S,\xi\mid\Delta)=-(\Sigma_1\Delta)_S$, and $\Cov(\xi,\tilde\varepsilon\mid\Delta)=\tilde\sigma^2$.
Then, it follows that $\mathbb E_{\tilde x,\tilde\varepsilon}[\tilde\varepsilon\mid\xi,\Delta]=\tilde\sigma^2\xi/\sigma_\xi^2(\Delta)$, $\Cov(\tilde x_S,\tilde\varepsilon\mid\xi,\Delta)=\tilde\sigma^2(\Sigma_1\Delta)_S/\sigma_\xi^2(\Delta)$, and $\mathbb E_{\tilde x,\tilde\varepsilon}[\tilde x_S\mid\xi,\Delta]=-\xi(\Sigma_1\Delta)_S/\sigma_\xi^2(\Delta)$.
By the definition of conditional covariance, we have
\begin{align*}
\mathbb E_{\tilde x,\tilde\varepsilon}
[\tilde x_S\tilde\varepsilon\mid\xi,\Delta]
=\Cov(\tilde x_S,\tilde\varepsilon\mid\xi,\Delta)
+\mathbb E_{\tilde x,\tilde\varepsilon}[\tilde x_S\mid\xi,\Delta]\,
\mathbb E_{\tilde x,\tilde\varepsilon}[\tilde\varepsilon\mid\xi,\Delta]
=\frac{\tilde\sigma^2}{\sigma_\xi^2(\Delta)}
\left(1-\frac{\xi^2}{\sigma_\xi^2(\Delta)}\right)(\Sigma_1\Delta)_S.
\end{align*}
By multiplying this identity by $\mathbb I$ and applying the tower property, we obtain
\begin{equation}\label{eq:common_population_accepted_noise_expectation}
\mathbb E_{\tilde x,\tilde\varepsilon}
[\mathbb I\tilde x_S\tilde\varepsilon\mid\Delta]
=\frac{p_\Delta\tilde\sigma^2(1-\lambda_{2,\Delta})}
{\sigma_\xi^2(\Delta)}(\Sigma_1\Delta)_S,
\end{equation}
where $\mathbb E_{\tilde x,\tilde\varepsilon}[\mathbb I\xi^2\mid\Delta]=p_\Delta\lambda_{2,\Delta}\sigma_\xi^2(\Delta)$ by the definition of $\lambda_{2,\Delta}$.

We apply Lemma~\ref{lem:common_population_covariance}, the definition of $\theta_\Delta$, and~\eqref{eq:common_population_accepted_noise_expectation} to obtain
\begin{align*}
\bar G_{\mathrm{pop}}(\Delta)\theta_\Delta
\Sigma_{1,SS}^{-1}(\Sigma_1\Delta)_S
&=\frac{p_\Delta\tilde\sigma^2(1-\lambda_{2,\Delta})}{\sigma_\xi^2(\Delta)}(\Sigma_1\Delta)_S
=\mathbb E_{\tilde x,\tilde\varepsilon}[\mathbb I\tilde x_S\tilde\varepsilon\mid\Delta].
\end{align*}
By applying $\bar G_{\mathrm{pop}}(\Delta)^{-1}$ to the decomposition of $\bar\mu_{\mathrm{pop}}(\Delta)$, we obtain~\eqref{eq:common_population_mu}, which completes the proof.  
\end{proof}

The term $\theta_\Delta\Sigma_{1,SS}^{-1}(\Sigma_1\Delta)_S$ is the correction
to the inherited active error $-\Delta_S$ induced by verification.
We bound $\theta_\Delta$ and $1-\theta_\Delta$ for the population conditional mean and covariance calculations.

\begin{lemma}[Bounds for $\theta_\Delta$]
\label{lem:common_population_theta_bounds}
Under the assumptions of Lemma~\ref{lem:common_population_covariance},
let $\theta_\Delta$ be defined in~\eqref{eq:theta_Delta_def}.
Then, for every $\Delta\in\mathbb R^d$, we have $0<\theta_\Delta<1$ and
\begin{align*}
1-\theta_\Delta
&\asymp
\frac{\|\Delta_{-S}\|_{\Sigma_{1,-S\mid S}}^2
+\min(\tilde\sigma^2,\delta^2)}
{\tilde\sigma^2+\|\Delta_{-S}\|_{\Sigma_{1,-S\mid S}}^2}.
\end{align*}
Moreover, there exists a universal constant $c>0$ such that
\begin{align*}
\theta_\Delta
&\le\frac{\tilde\sigma^2}
{\tilde\sigma^2+\|\Delta_{-S}\|_{\Sigma_{1,-S\mid S}}^2+c\delta^2}.
\end{align*}
\end{lemma}

\begin{proof}
By the definition of $\theta_\Delta$ and $0<\lambda_{2,\Delta}<1$, we have $0<\theta_\Delta<1$.
First, we prove the upper bound of $\theta_\Delta$.
Since $\phi(z)\ge\phi(h)$ for $0\le z\le h$, the definition of $\lambda_2(h)$ and Lemma~\ref{lem:truncated_normal_moments} imply $\lambda_2(h)\ge2h^3\phi(h)/(3p(h))=h^2(1-\lambda_2(h))/3$ for $h>0$.
Hence, by the Schur complement decomposition and $h_\Delta=\delta/\sigma_\xi(\Delta)$, we obtain
\begin{align*}
\theta_\Delta
&=\frac{\tilde\sigma^2}
{\tilde\sigma^2+\|\Delta_{-S}\|_{\Sigma_{1,-S\mid S}}^2+
\lambda_{2,\Delta}\sigma_\xi^2(\Delta)/(1-\lambda_{2,\Delta})}
\le\frac{\tilde\sigma^2}
{\tilde\sigma^2+\|\Delta_{-S}\|_{\Sigma_{1,-S\mid S}}^2+c\delta^2}.
\end{align*}

Next, by the definition of $\lambda_{2,\Delta}$, we have $\lambda_{2,\Delta}\sigma_\xi^2(\Delta)\le\delta^2$.
The definition of $\theta_\Delta$, its upper bound, and $0<1-\theta_\Delta<1$ imply
\begin{align*}
\frac{\|\Delta_{-S}\|_{\Sigma_{1,-S\mid S}}^2+c\delta^2}
{\tilde\sigma^2+\|\Delta_{-S}\|_{\Sigma_{1,-S\mid S}}^2+c\delta^2}
&\le1-\theta_\Delta
\le\min\Biggl(1,
\frac{\|\Delta_{-S}\|_{\Sigma_{1,-S\mid S}}^2+\delta^2}
{\tilde\sigma^2+\|\Delta_{-S}\|_{\Sigma_{1,-S\mid S}}^2}\Biggr).
\end{align*}
By combining these bounds, we obtain the stated rate for $1-\theta_\Delta$ and prove the lemma.
\end{proof}

\subsection{Population Conditional Mean in $\mathcal V_1$}

For $\mathcal V_1$, we evaluate the second moment of the population conditional mean $\bar G_{\mathrm{pop}}(\Delta)^{-1}\bar\mu_{\mathrm{pop}}(\Delta)-\Sigma_{1,SS}^{-1}\Sigma_{1,S,-S}\Delta_{-S}$ over $\Delta\mid X_{\mathrm{pre}}\sim\mathcal N(0,\Gamma/n)$.
Recall that $\bar\rho^2=\tr(\Sigma_1\Gamma)/(n\tilde\sigma^2)$.

For partial fine-tuning, this moment also depends on the number of frozen coordinates.
For $k\in\{1,\ldots,d-1\}$ and $z>0$, define $h_{k,d}(z)$ as  $(d/(d+z^2))^{k/2}$,
$(d/(d+z^2))^2\log(e+z^2/d)$, and
$(d/(d+kz^2))^2$ when $1\le k\le3$, $k=4$, and $k\ge5$, respectively.

\begin{corollary}[Population conditional mean scale]
\label{cor:common_population_partial_mean_scale}
Under the pre-training model in Section~\ref{sec:problem_setting} and
Assumption~\ref{ass:design}, 
fix  $X_{\mathrm{pre}}\in\Omega_{\mathrm{pre}}$ and let $u>0$.
Expectations are taken over $\Delta\mid X_{\mathrm{pre}}$.
If $1\le s<d$, then
\begin{align*}
\mathbb E_\Delta
\left\|\bar G_{\mathrm{pop}}(\Delta)^{-1}\bar\mu_{\mathrm{pop}}(\Delta)
-\Sigma_{1,SS}^{-1}\Sigma_{1,S,-S}\Delta_{-S}\right\|_{\Sigma_{1,SS}}^2
&\asymp\frac{s}{n}\left(\left(
\frac{(d-s)\bar\rho^2}{d+(d-s)\bar\rho^2}\right)^2
+\min(1,u^4)h_{d-s,d}(\bar\rho)\right).
\end{align*}
If $s=d$, then
\begin{align*}
\mathbb E_\Delta
\left\|\bar G_{\mathrm{pop}}(\Delta)^{-1}\bar\mu_{\mathrm{pop}}(\Delta)\right\|_{\Sigma_1}^2
\asymp\min(1,u^4)\frac{\tr(\Sigma_1\Gamma)}{n}.
\end{align*}
\end{corollary}

\begin{proof}
Since $\Sigma_{1,SS}^{-1}(\Sigma_1\Delta)_S=\Delta_S+\Sigma_{1,SS}^{-1}\Sigma_{1,S,-S}\Delta_{-S}$, it follows from Lemma~\ref{lem:common_population_mu} that
\begin{equation}\label{eq:common_population_mean_target}
\bar G_{\mathrm{pop}}(\Delta)^{-1}\bar\mu_{\mathrm{pop}}(\Delta)
-\Sigma_{1,SS}^{-1}\Sigma_{1,S,-S}\Delta_{-S}
=-(1-\theta_\Delta)\Sigma_{1,SS}^{-1}(\Sigma_1\Delta)_S.
\end{equation}
By Lemma~\ref{lem:common_population_theta_bounds}, the squared $\Sigma_{1,SS}$ norm satisfies, for every $u>0$ and $\Delta\in\mathbb R^d$,
\begin{equation}\label{eq:common_population_mean_scale}
\begin{aligned}
&\left\|\bar G_{\mathrm{pop}}(\Delta)^{-1}\bar\mu_{\mathrm{pop}}(\Delta)
-\Sigma_{1,SS}^{-1}\Sigma_{1,S,-S}\Delta_{-S}
\right\|_{\Sigma_{1,SS}}^2
=(1-\theta_\Delta)^2\|(\Sigma_1\Delta)_S\|_{\Sigma_{1,SS}^{-1}}^2\\
&\quad\asymp \|(\Sigma_1\Delta)_S\|_{\Sigma_{1,SS}^{-1}}^2
\left(
\frac{\|\Delta_{-S}\|_{\Sigma_{1,-S\mid S}}^2
+\min(\tilde\sigma^2,\delta^2)}
{\tilde\sigma^2+\|\Delta_{-S}\|_{\Sigma_{1,-S\mid S}}^2}
\right)^2.
\end{aligned}
\end{equation}

For $1\le s<d$, the relation $(x+y)^2\asymp x^2+y^2$ for $x,y\ge0$ and $\min(\tilde\sigma^2,\delta^2)^2/\tilde\sigma^4=\min(1,u^4)$ imply
\begin{align*}
&\mathbb E_\Delta\left[
(1-\theta_\Delta)^2\|(\Sigma_1\Delta)_S\|_{\Sigma_{1,SS}^{-1}}^2
\right]\\
&\quad\asymp\mathbb E_\Delta\left[
\|(\Sigma_1\Delta)_S\|_{\Sigma_{1,SS}^{-1}}^2
\left(
\frac{\|\Delta_{-S}\|_{\Sigma_{1,-S\mid S}}^2}
{\tilde\sigma^2+\|\Delta_{-S}\|_{\Sigma_{1,-S\mid S}}^2}
\right)^2
\right]
+\min(1,u^4)\mathbb E_\Delta\left[
\frac{\|(\Sigma_1\Delta)_S\|_{\Sigma_{1,SS}^{-1}}^2}
{\left(1+\|\Delta_{-S}\|_{\Sigma_{1,-S\mid S}}^2/\tilde\sigma^2\right)^2}
\right].
\end{align*}
By Lemma~\ref{lem:weighted_pretraining_error_moments}, the two expectations on the right hand side are comparable to $(s/n)((d-s)\bar\rho^2/(d+(d-s)\bar\rho^2))^2$ and $(s/n)h_{d-s,d}(\bar\rho)$, respectively, which establishes the result for $s<d$.

When $s=d$, the ratio in~\eqref{eq:common_population_mean_scale} equals $\min(1,u^2)$.
Since $\mathbb E_\Delta[\|\Delta\|_{\Sigma_1}^2\mid X_{\mathrm{pre}}]=\tr(\Sigma_1\Gamma)/n$, the same equation implies the stated full fine-tuning rate.
\end{proof}

\subsection{Population Conditional Covariance in $\mathcal V_2$}

We evaluate the population conditional covariance in $\mathcal V_2$.
By Lemma~\ref{lem:common_population_mu}, we have $\Delta_S+\bar G_{\mathrm{pop}}(\Delta)^{-1}$ $ \bar\mu_{\mathrm{pop}}(\Delta)=\theta_\Delta\Sigma_{1,SS}^{-1}(\Sigma_1\Delta)_S$.
Hence, after replacing $\bar G(\Delta)^{-1}\bar\mu(\Delta)$ by $\bar G_{\mathrm{pop}}(\Delta)^{-1}\bar\mu_{\mathrm{pop}}(\Delta)$ in~\eqref{eq:M_def}, the generation noise is adjusted by $\theta_\Delta\tilde x_S^\top\Sigma_{1,SS}^{-1}(\Sigma_1\Delta)_S$.
The following lemma bounds the expectation over $\tilde x$ of the resulting conditional covariance.

\begin{lemma}[Population conditional covariance of adjusted generation noise]
\label{lem:common_population_residual_covariance}
Under the model in Section~\ref{sec:problem_setting}, suppose that $0<\delta<\infty$ and $\tilde\sigma>0$.  
There exists a universal constant $c>0$ such that, for
every $\Delta\in\mathbb R^d$,
\begin{align*}
c\tilde\sigma^2(1-\theta_\Delta)\bar G_{\mathrm{pop}}(\Delta)
&\preceq\mathbb E_{\tilde x}\Bigl[
\tilde x_S\tilde x_S^\top
\Var_{\tilde\varepsilon}\Bigl(
\mathbb I\bigl(
\tilde\varepsilon-\theta_\Delta\tilde x_S^\top
\Sigma_{1,SS}^{-1}(\Sigma_1\Delta)_S
\bigr)
\Big|\Delta,\tilde x\Bigr)
\Big|\Delta\Bigr]
\preceq\tilde\sigma^2(1-\theta_\Delta)
\bar G_{\mathrm{pop}}(\Delta),
\end{align*}
where $\tilde\sigma^2(1-\theta_\Delta)\asymp
\min(\tilde\sigma^2,\delta^2)
+\tilde\sigma^2\|\Delta_{-S}\|_{\Sigma_{1,-S\mid S}}^2/
(\tilde\sigma^2+\|\Delta_{-S}\|_{\Sigma_{1,-S\mid S}}^2)$.
\end{lemma}

\begin{proof}
Fix $\Delta$ and let $W:=\tilde\varepsilon-\theta_\Delta\tilde x_S^\top\Sigma_{1,SS}^{-1}(\Sigma_1\Delta)_S$.
By the definition of conditional variance, we have
\begin{align*}
\Var_{\tilde\varepsilon}(\mathbb I W\mid\Delta,\tilde x)
&=\mathbb E_{\tilde\varepsilon}[\mathbb I W^2\mid\Delta,\tilde x]
-\bigl(\mathbb E_{\tilde\varepsilon}[\mathbb I W\mid\Delta,\tilde x]\bigr)^2.
\end{align*}
Recall that $u=\delta/\tilde\sigma$ and $p(u)=\mathbb P_{\tilde\varepsilon}(|\tilde\varepsilon|\le\delta)$.
Since $\mathbb P_{\tilde\varepsilon}(\mathbb I=1\mid\Delta,\tilde x)\le p(u)$, the Cauchy--Schwarz inequality implies $\bigl(\mathbb E_{\tilde\varepsilon}[\mathbb I W\mid\Delta,\tilde x]\bigr)^2\le p(u)\mathbb E_{\tilde\varepsilon}[\mathbb I W^2\mid\Delta,\tilde x]$.
Hence,
\begin{align}\label{eq:common_adjusted_noise_conditional_variance_bounds}
(1-p(u))\mathbb E_{\tilde\varepsilon}[\mathbb I W^2\mid\Delta,\tilde x]
&\le\Var_{\tilde\varepsilon}(\mathbb I W\mid\Delta,\tilde x)
\le\mathbb E_{\tilde\varepsilon}[\mathbb I W^2\mid\Delta,\tilde x].
\end{align}
After multiplying by $\tilde x_S\tilde x_S^\top$ and taking expectation over $\tilde x$, bounds for 
$\mathbb E_{\tilde x,\tilde\varepsilon}[\mathbb I\tilde x_S\tilde x_S^\top W^2\mid\Delta]$ establish the covariance 
upper bound for every $u>0$ and the lower bound for $0<u\le1$, since $1-p(u)\ge1-p(1)>0$.
For $u>1$, we establish the lower bound separately.

First, we evaluate $\mathbb E_{\tilde x,\tilde\varepsilon}[\mathbb I W^2\mid\Delta]/p_\Delta$, which determines the scalar factor in the covariance bounds.
By the definitions of $W$ and $\bar\mu_{\mathrm{pop}}(\Delta)$ and Lemma~\ref{lem:common_population_mu}, we have $\mathbb E_{\tilde x,\tilde\varepsilon}[\mathbb I\tilde x_SW\mid\Delta]=0$.
Hence, $\mathbb E_{\tilde x,\tilde\varepsilon}[\mathbb I W^2\mid\Delta]=\mathbb E_{\tilde x,\tilde\varepsilon}[\mathbb I\tilde\varepsilon W\mid\Delta]$.
By the definition of $W$, we have
\begin{align*}
\mathbb E_{\tilde x,\tilde\varepsilon}[\mathbb I\tilde\varepsilon W\mid\Delta]
&=\mathbb E_{\tilde x,\tilde\varepsilon}[\mathbb I\tilde\varepsilon^2\mid\Delta]
-\theta_\Delta(\Sigma_1\Delta)_S^\top\Sigma_{1,SS}^{-1}
\mathbb E_{\tilde x,\tilde\varepsilon}[\mathbb I\tilde x_S\tilde\varepsilon\mid\Delta].
\end{align*}
We have already calculated $\mathbb E_{\tilde x,\tilde\varepsilon}[\mathbb I\tilde x_S\tilde\varepsilon\mid\Delta]$ in~\eqref{eq:common_population_accepted_noise_expectation}.
Hence, it remains to evaluate $\mathbb E_{\tilde x,\tilde\varepsilon}[\mathbb I\tilde\varepsilon^2\mid\Delta]$.
Since $\mathbb I=\mathbf1\{|\xi|\le\delta\}$, we condition on $\xi$.
Conditioned on $\Delta$, $(\tilde\varepsilon,\xi)$ is jointly centered Gaussian, with $\Cov(\tilde\varepsilon,\xi\mid\Delta)=\tilde\sigma^2$ and $\Var(\xi\mid\Delta)=\sigma_\xi^2(\Delta)$.
The corresponding conditional mean and variance satisfy
\begin{align*}
\mathbb E_{\tilde x,\tilde\varepsilon}[\tilde\varepsilon\mid\xi,\Delta]
&=\frac{\tilde\sigma^2\xi}{\sigma_\xi^2(\Delta)},
&
\Var(\tilde\varepsilon\mid\xi,\Delta)
&=\tilde\sigma^2-\frac{\tilde\sigma^4}{\sigma_\xi^2(\Delta)}.
\end{align*}
By adding the conditional variance and the squared conditional mean, we obtain $\mathbb E_{\tilde x,\tilde\varepsilon}[\tilde\varepsilon^2\mid\xi,\Delta]=\tilde\sigma^2-\tilde\sigma^4/\sigma_\xi^2(\Delta)+\tilde\sigma^4\xi^2/\sigma_\xi^4(\Delta)$.
Since $\mathbb E[\mathbb I\xi^2\mid\Delta]=p_\Delta\lambda_{2,\Delta}\sigma_\xi^2(\Delta)$, the tower property implies
\begin{align*}
\frac{\mathbb E_{\tilde x,\tilde\varepsilon}
[\mathbb I\tilde\varepsilon^2\mid\Delta]}{p_\Delta}
&=\tilde\sigma^2-
\frac{(1-\lambda_{2,\Delta})\tilde\sigma^4}{\sigma_\xi^2(\Delta)}.
\end{align*}
Next, we substitute this identity and~\eqref{eq:common_population_accepted_noise_expectation} into $\mathbb E_{\tilde x,\tilde\varepsilon}[\mathbb I\tilde\varepsilon W\mid\Delta]$ and obtain
\begin{align}\label{eq:common_population_adjusted_noise_scalar_second_moment}
\frac{\mathbb E_{\tilde x,\tilde\varepsilon}
[\mathbb I W^2\mid\Delta]}{p_\Delta}
&=\tilde\sigma^2-
\frac{(1-\lambda_{2,\Delta})\tilde\sigma^2}{\sigma_\xi^2(\Delta)}
\left(\tilde\sigma^2+\theta_\Delta\|(\Sigma_1\Delta)_S\|_{\Sigma_{1,SS}^{-1}}^2\right)
=\tilde\sigma^2(1-\theta_\Delta),
\end{align}
where the last equality follows from the definition of $\theta_\Delta$ in~\eqref{eq:theta_Delta_def}.

We bound $\mathbb E_{\tilde x,\tilde\varepsilon}[\mathbb I\tilde x_S\tilde x_S^\top W^2\mid\Delta]$ through its quadratic forms.
For every $v\in\mathbb R^s$, we have $v^\top\mathbb E_{\tilde x,\tilde\varepsilon}[\mathbb I\tilde x_S\tilde x_S^\top W^2\mid\Delta]v=\mathbb E_{\tilde x,\tilde\varepsilon}[\mathbb I(v^\top\tilde x_S)^2W^2\mid\Delta]$.
By~\eqref{eq:common_population_adjusted_noise_scalar_second_moment} and the definition of $\bar G_{\mathrm{pop}}(\Delta)$, we have
\begin{align}\label{eq:common_population_second_moment_product}
\frac{\mathbb E[\mathbb I(v^\top\tilde x_S)^2\mid\Delta]\mathbb E[\mathbb I W^2\mid\Delta]}{p_\Delta}
&=\tilde\sigma^2(1-\theta_\Delta)v^\top\bar G_{\mathrm{pop}}(\Delta)v.
\end{align}
Hence, it suffices to bound $\mathbb E[\mathbb I(v^\top\tilde x_S)^2W^2\mid\Delta]$ above by $\tilde\sigma^2(1-\theta_\Delta)v^\top\bar G_{\mathrm{pop}}(\Delta)v$ and below by one half of this quantity.
We evaluate this fourth moment by conditioning on $\xi$, since $(v^\top\tilde x_S,W,\xi)$ is jointly Gaussian conditioned on $\Delta$.
Note that $\mathbb E[\mathbb I(v^\top\tilde x_S)W\mid\Delta]=0$.
By the definitions of $W$ and $\xi$ and the independence of the generation noise and covariates, we have
\begin{equation}\label{eq:common_adjusted_noise_covariance}
\Cov(W,\xi\mid\Delta)
=\tilde\sigma^2+\theta_\Delta
\|(\Sigma_1\Delta)_S\|_{\Sigma_{1,SS}^{-1}}^2.
\end{equation}
By Gaussian conditioning, we have
\begin{align*}
\mathbb E[(v^\top\tilde x_S)W\mid\xi,\Delta]
&=\Cov(v^\top\tilde x_S,W\mid\xi,\Delta)
+\frac{\xi^2\Cov(v^\top\tilde x_S,\xi\mid\Delta)
\Cov(W,\xi\mid\Delta)}{\sigma_\xi^4(\Delta)},
\end{align*}
where the second term is the product of the conditional means.
Since the conditional covariance does not depend on the value of $\xi$, expectation over $\xi$ conditioned on $\mathbb I=1$ results in
\begin{equation}\label{eq:common_accepted_cross_moment}
\begin{aligned}
0=\mathbb E[(v^\top\tilde x_S)W\mid\mathbb I=1,\Delta]
&=\Cov(v^\top\tilde x_S,W\mid\xi,\Delta)
+\lambda_{2,\Delta}
\frac{\Cov(v^\top\tilde x_S,\xi\mid\Delta)
\Cov(W,\xi\mid\Delta)}{\sigma_\xi^2(\Delta)},
\end{aligned}
\end{equation}
where $\mathbb E[\xi^2\mid\mathbb I=1,\Delta]=\lambda_{2,\Delta}\sigma_\xi^2(\Delta)$.
Similarly, Gaussian conditioning implies
\begin{align*}
\mathbb E[(v^\top\tilde x_S)^2\mid\xi,\Delta]
&=\Var(v^\top\tilde x_S\mid\xi,\Delta)
+\frac{\xi^2\Cov(v^\top\tilde x_S,\xi\mid\Delta)^2}
{\sigma_\xi^4(\Delta)}\\
&=\frac{\mathbb E
[\mathbb I(v^\top\tilde x_S)^2\mid\Delta]}{p_\Delta}
+\bigl(\xi^2-\lambda_{2,\Delta}\sigma_\xi^2(\Delta)\bigr)
\frac{\Cov(v^\top\tilde x_S,\xi\mid\Delta)^2}
{\sigma_\xi^4(\Delta)},\\
\mathbb E[W^2\mid\xi,\Delta]
&=\Var(W\mid\xi,\Delta)
+\frac{\xi^2\Cov(W,\xi\mid\Delta)^2}{\sigma_\xi^4(\Delta)}\\
&=\frac{\mathbb E
[\mathbb I W^2\mid\Delta]}{p_\Delta}
+\bigl(\xi^2-\lambda_{2,\Delta}\sigma_\xi^2(\Delta)\bigr)
\frac{\Cov(W,\xi\mid\Delta)^2}{\sigma_\xi^4(\Delta)},
\end{align*}
where the last equality in each calculation follows by averaging the first equality over $\xi$ conditioned on $\mathbb I=1$ 
and solving for the conditional variance, which is independent of $\xi$.
Since $\mathbb E[\mathbb I(\xi^2-\lambda_{2,\Delta}\sigma_\xi^2(\Delta))\mid\Delta]=0$, multiplication 
of these identities and expectation over $\xi$ result in
\begin{align*}
&\mathbb E\bigl[\mathbb I
\mathbb E[(v^\top\tilde x_S)^2\mid\xi,\Delta]
\mathbb E[W^2\mid\xi,\Delta]
\bigm|\Delta\bigr]\\*
&\quad=\frac{\mathbb E
[\mathbb I(v^\top\tilde x_S)^2\mid\Delta]
\mathbb E[\mathbb I W^2\mid\Delta]}
{p_\Delta}
+p_\Delta(\lambda_{4,\Delta}-\lambda_{2,\Delta}^2)
\frac{\Cov(v^\top\tilde x_S,\xi\mid\Delta)^2
\Cov(W,\xi\mid\Delta)^2}{\sigma_\xi^4(\Delta)},
\end{align*}
where we have used $\Var(\xi^2\mid\mathbb I=1,\Delta)=(\lambda_{4,\Delta}-\lambda_{2,\Delta}^2)\sigma_\xi^4(\Delta)$.
For jointly Gaussian random variables $X,Y$, the Gaussian fourth moment formula states that $\mathbb E[X^2Y^2]=\mathbb E[X^2]\mathbb E[Y^2]+2\Cov(X,Y)^2+4\mathbb E[X]\mathbb E[Y]\Cov(X,Y)$.
We apply this formula conditioned on $(\xi,\Delta)$ and~\eqref{eq:common_accepted_cross_moment} to obtain
\begin{align}
&\mathbb E[(v^\top\tilde x_S)^2W^2\mid\xi,\Delta]\notag\\*
&\quad=\mathbb E[(v^\top\tilde x_S)^2\mid\xi,\Delta]
\mathbb E[W^2\mid\xi,\Delta]
+\biggl(2\lambda_{2,\Delta}^2
-\frac{4\lambda_{2,\Delta}\xi^2}{\sigma_\xi^2(\Delta)}\biggr)
\frac{\Cov(v^\top\tilde x_S,\xi\mid\Delta)^2
\Cov(W,\xi\mid\Delta)^2}{\sigma_\xi^4(\Delta)}.
\label{eq:common_adjusted_noise_conditional_fourth_moment}
\end{align}
By multiplying this identity by $\mathbb I$ and taking expectation over $\xi$, we obtain
\begin{align}
&\mathbb E_{\tilde x,\tilde\varepsilon}
[\mathbb I(v^\top\tilde x_S)^2W^2\mid\Delta]\notag\\
&\quad=\mathbb E\bigl[\mathbb I
\mathbb E[(v^\top\tilde x_S)^2\mid\xi,\Delta]
\mathbb E[W^2\mid\xi,\Delta]
\bigm|\Delta\bigr]
-2p_\Delta\lambda_{2,\Delta}^2
\frac{\Cov(v^\top\tilde x_S,\xi\mid\Delta)^2
\Cov(W,\xi\mid\Delta)^2}{\sigma_\xi^4(\Delta)}\notag\\
&\quad=\tilde\sigma^2(1-\theta_\Delta)v^\top\bar G_{\mathrm{pop}}(\Delta)v
-p_\Delta\left(3-\frac{\lambda_{4,\Delta}}{\lambda_{2,\Delta}^2}\right)
\Cov(v^\top\tilde x_S,W\mid\xi,\Delta)^2,
\label{eq:common_population_adjusted_noise_fourth_moment}
\end{align}
where the first equality follows from
 $\mathbb E[\mathbb I\xi^2\mid\Delta]=p_\Delta\lambda_{2,\Delta}\sigma_\xi^2(\Delta)$, and 
 the second  from  the averaged product of the conditional second moments, \eqref{eq:common_accepted_cross_moment}, \eqref{eq:common_population_adjusted_noise_scalar_second_moment}, and the definition of $\bar G_{\mathrm{pop}}(\Delta)$.

By Lemma~\ref{lem:truncated_normal_moments}, $0\le3-\lambda_{4,\Delta}/\lambda_{2,\Delta}^2\le2$.
Hence, \eqref{eq:common_population_adjusted_noise_fourth_moment} establishes the required upper bound.
For the lower bound, we bound $\Cov(v^\top\tilde x_S,W\mid\xi,\Delta)^2$ through the conditional second moments.  Recall that
\begin{align*}
\mathbb E[(v^\top\tilde x_S)^2\mid\mathbb I=1,\Delta]
&=\Var(v^\top\tilde x_S\mid\xi,\Delta)
+\lambda_{2,\Delta}
\frac{\Cov(v^\top\tilde x_S,\xi\mid\Delta)^2}
{\sigma_\xi^2(\Delta)},\\
\mathbb E[W^2\mid\mathbb I=1,\Delta]
&=\Var(W\mid\xi,\Delta)
+\lambda_{2,\Delta}
\frac{\Cov(W,\xi\mid\Delta)^2}{\sigma_\xi^2(\Delta)}.
\end{align*}
Then, it holds that
\begin{align}
&\mathbb E[(v^\top\tilde x_S)^2\mid\mathbb I=1,\Delta]
\mathbb E[W^2\mid\mathbb I=1,\Delta]
\label{eq:common_accepted_second_moment_product}\\
&\quad\ge\biggl(
\sqrt{\Var(v^\top\tilde x_S\mid\xi,\Delta)
\Var(W\mid\xi,\Delta)}
+\bigl|\Cov(v^\top\tilde x_S,W\mid\xi,\Delta)\bigr|
\biggr)^2
\ge4\Cov(v^\top\tilde x_S,W\mid\xi,\Delta)^2,\notag
\end{align}
where the first inequality follows from $(a+b)(c+d)\ge(\sqrt{ac}+\sqrt{bd})^2$ for $a,b,c,d\ge0$ and
\eqref{eq:common_accepted_cross_moment} and the second  from
$|\Cov(v^\top\tilde x_S,W\mid\xi,\Delta)|
\le\sqrt{\Var(v^\top\tilde x_S\mid\xi,\Delta)
\Var(W\mid\xi,\Delta)}$.
By~\eqref{eq:common_accepted_second_moment_product} and~\eqref{eq:common_population_second_moment_product}, we obtain
\begin{align*}
0\le p_\Delta\left(3-\frac{\lambda_{4,\Delta}}{\lambda_{2,\Delta}^2}\right)
\Cov(v^\top\tilde x_S,W\mid\xi,\Delta)^2
\le2p_\Delta\Cov(v^\top\tilde x_S,W\mid\xi,\Delta)^2
\le\frac12\tilde\sigma^2(1-\theta_\Delta)v^\top\bar G_{\mathrm{pop}}(\Delta)v.
\end{align*}
Since $v$ is arbitrary, \eqref{eq:common_population_adjusted_noise_fourth_moment} implies
\begin{equation}\label{eq:common_population_adjusted_noise_second_moment}
\frac{1}{2}\tilde\sigma^2(1-\theta_\Delta)\bar G_{\mathrm{pop}}(\Delta)
\preceq\mathbb E_{\tilde x,\tilde\varepsilon}
[\mathbb I\tilde x_S\tilde x_S^\top W^2\mid\Delta]
\preceq\tilde\sigma^2(1-\theta_\Delta)\bar G_{\mathrm{pop}}(\Delta).
\end{equation}
By ~\eqref{eq:common_adjusted_noise_conditional_variance_bounds} and~\eqref{eq:common_population_adjusted_noise_second_moment}, 
we establish the covariance upper bound for every $u>0$ and the lower bound for $0<u\le1$.

For $u>1$, the factor $1-p(u)$ is not bounded below by a positive universal constant.
We instead obtain a lower bound on $|a|\le\delta/2$, where $a=-\tilde x^\top\Delta$.
On $|a|\le\delta/2$, we have
$|a+\tilde\varepsilon|\le\delta/2+\tilde\sigma/2\le\delta$
whenever $|\tilde\varepsilon|\le\tilde\sigma/2$.
Hence, $\mathbb I=1$ on this subset.  With expectations and variance over $\tilde\varepsilon$ conditioned on $(\Delta,\tilde x)$, we have
\begin{align*}
\Var_{\tilde\varepsilon}(\mathbb I W)
&=\inf_{t\in\mathbb R}\mathbb E_{\tilde\varepsilon}
[(\mathbb I W-t)^2]
\ge\inf_{t\in\mathbb R}\mathbb E_{\tilde\varepsilon}
[\mathbf1\{|\tilde\varepsilon|\le\tilde\sigma/2\}(W-t)^2]
=\mathbb E_{\tilde\varepsilon}
[\tilde\varepsilon^2\mathbf1\{|\tilde\varepsilon|
\le\tilde\sigma/2\}]\ge c\tilde\sigma^2,
\end{align*}
where the first inequality follows by restricting the nonnegative squared moment to $|\tilde\varepsilon|\le\tilde\sigma/2$.
The last equality follows from $\mathbb E_{\tilde\varepsilon}[\tilde\varepsilon\mathbf1\{|\tilde\varepsilon|\le\tilde\sigma/2\}]=0$ and the minimizing choice $t=-\theta_\Delta\tilde x_S^\top\Sigma_{1,SS}^{-1}(\Sigma_1\Delta)_S$.

After multiplying the variance lower bound by $\tilde x_S\tilde x_S^\top$ and taking expectation over $\tilde x$, it remains to show that $\mathbb E_{\tilde x}[\tilde x_S\tilde x_S^\top\mathbf1\{|a|\le\delta/2\}\mid\Delta]\succeq c\bar G_{\mathrm{pop}}(\Delta)$.
If $\Delta=0$, the restriction holds surely and
$\bar G_{\mathrm{pop}}(0)=p(u)\Sigma_{1,SS}$, which proves the required bound.
Suppose that $\|\Delta\|_{\Sigma_1}>0$.
By Lemma~\ref{lem:truncated_normal_moments}, 
we have $\mathbb P_{\tilde x}(|a|\le\delta/2\mid\Delta)\ge cp_\Delta$.
Lemmas~\ref{lem:accepted_linear_form_moments}
and~\ref{lem:truncated_normal_moments}, with $u>1$, also provide
\begin{align}\label{eq:common_large_u_a_second_moment_comparison}
\mathbb E[a^2\mid\mathbb I=1,\Delta]
&\le2\min(\|\Delta\|_{\Sigma_1}^2,\delta^2)
\le C\mathbb E[a^2\mid |a|\le\delta/2,\Delta].
\end{align}
Gaussian conditioning results in
\begin{equation}\label{eq:common_covariate_second_moment_given_a}
\mathbb E_{\tilde x}[\tilde x_S\tilde x_S^\top\mid a,\Delta]
=\Sigma_{1,SS}
-\frac{(\Sigma_1\Delta)_S(\Sigma_1\Delta)_S^\top}
{\|\Delta\|_{\Sigma_1}^2}
+a^2\frac{(\Sigma_1\Delta)_S(\Sigma_1\Delta)_S^\top}
{\|\Delta\|_{\Sigma_1}^4}.
\end{equation}
The constant matrix in~\eqref{eq:common_covariate_second_moment_given_a} is $\Cov(\tilde x_S\mid a,\Delta)\succeq0$, and the coefficient of $a^2$ is positive semidefinite.
Since the generation noise is independent of the covariates, $\tilde x_S$ and $\mathbb I$ are independent conditioned on $(a,\Delta)$.
Hence, the probability bound and~\eqref{eq:common_large_u_a_second_moment_comparison} imply
\begin{align*}
\mathbb E_{\tilde x}[
\tilde x_S\tilde x_S^\top\mathbf1\{|a|\le\delta/2\}\mid\Delta]
&=\mathbb P_{\tilde x}(|a|\le\delta/2\mid\Delta)
\mathbb E_{\tilde x}[
\tilde x_S\tilde x_S^\top\mid |a|\le\delta/2,\Delta]\\
&\succeq cp_\Delta\mathbb E_{\tilde x,\tilde\varepsilon}[
\tilde x_S\tilde x_S^\top\mid\mathbb I=1,\Delta]
=c\bar G_{\mathrm{pop}}(\Delta),
\end{align*}
where the last equality follows from the definition of $\bar G_{\mathrm{pop}}(\Delta)$.
Then, we multiply the variance lower bound by $\tilde x_S\tilde x_S^\top$
and take expectation over $\tilde x$ to obtain
\begin{align*}
\mathbb E_{\tilde x}[
\tilde x_S\tilde x_S^\top
\Var_{\tilde\varepsilon}(\mathbb I W\mid\Delta,\tilde x)
\mid\Delta]
&\succeq c\tilde\sigma^2\mathbb E_{\tilde x}[
\tilde x_S\tilde x_S^\top\mathbf1\{|a|\le\delta/2\}\mid\Delta]
\succeq c\tilde\sigma^2(1-\theta_\Delta)\bar G_{\mathrm{pop}}(\Delta).
\end{align*}
Finally, we apply Lemma~\ref{lem:common_population_theta_bounds} to obtain
the scalar bound, which completes the proof. 
\end{proof}

We bound the squared conditional mean of the adjusted generation
noise.  This provides the second moment bound used in Bernstein's inequality
in Corollary~\ref{cor:common_empirical_first_moment}.

\begin{lemma}[Squared conditional mean of adjusted generation noise]
\label{lem:common_population_residual_mean}
Under the model in Section~\ref{sec:problem_setting}, suppose that $0<\delta<\infty$ and  $\tilde\sigma>0$.  
There exists a universal constant $C>0$ such that, for
every $\Delta\in\mathbb R^d$,
\begin{align*}
\mathbb E_{\tilde x}\Bigl[
\tilde x_S\tilde x_S^\top
\bigl(\mathbb E_{\tilde\varepsilon}[
\mathbb I(
\tilde\varepsilon-\theta_\Delta\tilde x_S^\top
\Sigma_{1,SS}^{-1}(\Sigma_1\Delta)_S
)\mid\Delta,\tilde x]\bigr)^2
\mid\Delta\Bigr]
&\preceq C\frac{u}{1+u^2}\tilde\sigma^2(1-\theta_\Delta)
\bar G_{\mathrm{pop}}(\Delta).
\end{align*}
\end{lemma}

\begin{proof}
Fix $\Delta$ and let $W:=\tilde\varepsilon-\theta_\Delta\tilde x_S^\top\Sigma_{1,SS}^{-1}(\Sigma_1\Delta)_S$.
If $\|\Delta\|_{\Sigma_1}=0$, the conditional mean is zero by symmetry,
and the bound follows immediately.  Assume that $\|\Delta\|_{\Sigma_1}>0$.
For $0<u\le1$, equation~\eqref{eq:common_adjusted_noise_conditional_variance_bounds} implies $\bigl(\mathbb E_{\tilde\varepsilon}[\mathbb I W\mid\Delta,\tilde x]\bigr)^2\le p(u)\mathbb E_{\tilde\varepsilon}[\mathbb I W^2\mid\Delta,\tilde x]$.
We multiply by $\tilde x_S\tilde x_S^\top$, take expectation over $\tilde x$, and apply~\eqref{eq:common_population_adjusted_noise_second_moment} to obtain
\begin{align*}
\mathbb E_{\tilde x}\Bigl[
\tilde x_S\tilde x_S^\top
\bigl(\mathbb E_{\tilde\varepsilon}[\mathbb I W\mid\Delta,\tilde x]\bigr)^2
\mid\Delta\Bigr]
&\preceq p(u)\tilde\sigma^2(1-\theta_\Delta)\bar G_{\mathrm{pop}}(\Delta)
\preceq Cu\tilde\sigma^2(1-\theta_\Delta)\bar G_{\mathrm{pop}}(\Delta).
\end{align*}
Since $u/(1+u^2)\asymp u$ in this range, the claimed bound follows.

For $u>1$, we separate the two terms in the conditional mean to obtain the factor $u^{-1}$.
By the definition of $W$, we have
\begin{align}\label{eq:common_adjusted_noise_mean_decomposition}
\mathbb E_{\tilde\varepsilon}[\mathbb I W
\mid\Delta,\tilde x]
=\mathbb E_{\tilde\varepsilon}[\mathbb I\tilde\varepsilon\mid\Delta,\tilde x]
-\theta_\Delta\tilde x_S^\top\Sigma_{1,SS}^{-1}(\Sigma_1\Delta)_S
\mathbb P_{\tilde\varepsilon}(\mathbb I=1\mid\Delta,\tilde x).
\end{align}
Let $a=-\tilde x^\top\Delta$, where $a\mid\Delta\sim\mathcal N(0,\|\Delta\|_{\Sigma_1}^2)$.
Gaussian integration results in
\begin{align}\label{eq:common_generation_noise_conditional_mean}
\mathbb E_{\tilde\varepsilon}[\mathbb I\tilde\varepsilon\mid\Delta,\tilde x]
&=\tilde\sigma\bigl(
\phi((\delta+a)/\tilde\sigma)
-\phi((\delta-a)/\tilde\sigma)\bigr).
\end{align}
This conditional mean depends on $\tilde x$ only through $a$.
Recall from~\eqref{eq:common_covariate_second_moment_given_a} that
\begin{align*}
\mathbb E_{\tilde x}[\tilde x_S\tilde x_S^\top\mid a,\Delta]
&=\Sigma_{1,SS}
-\frac{(\Sigma_1\Delta)_S(\Sigma_1\Delta)_S^\top}
{\|\Delta\|_{\Sigma_1}^2}
+a^2\frac{(\Sigma_1\Delta)_S(\Sigma_1\Delta)_S^\top}
{\|\Delta\|_{\Sigma_1}^4}.
\end{align*}
Hence, we bound the squared conditional mean both without and with the factor $a^2$.
First, we apply $(x-y)^2\le2x^2+2y^2$ to~\eqref{eq:common_generation_noise_conditional_mean}.
By symmetry of $a$, it suffices to evaluate the Gaussian integrals of $e^{-(\delta-a)^2/\tilde\sigma^2}$ and $a^2e^{-(\delta-a)^2/\tilde\sigma^2}$.
Since $\delta>\tilde\sigma$ and $\sigma_\xi^2(\Delta)\le\tilde\sigma^2+2\|\Delta\|_{\Sigma_1}^2\le2\sigma_\xi^2(\Delta)$, we calculate the Gaussian integrals and obtain
\begin{align*}
\mathbb E_{\tilde x}\Bigl[
\bigl(\mathbb E_{\tilde\varepsilon}[\mathbb I\tilde\varepsilon
\mid\Delta,\tilde x]\bigr)^2\mid\Delta\Bigr]
&\le\frac{C\tilde\sigma^3}{\sigma_\xi(\Delta)}
\exp\left(-\frac{\delta^2}{2\sigma_\xi^2(\Delta)}\right),\\
\mathbb E_{\tilde x}\Bigl[
a^2\bigl(\mathbb E_{\tilde\varepsilon}[\mathbb I\tilde\varepsilon
\mid\Delta,\tilde x]\bigr)^2\mid\Delta\Bigr]
&\le\frac{C\tilde\sigma^3\delta^2\|\Delta\|_{\Sigma_1}^2}
{\sigma_\xi^3(\Delta)}
\exp\left(-\frac{\delta^2}{2\sigma_\xi^2(\Delta)}\right).
\end{align*}
To express these bounds in terms of $p_\Delta$ and $\mathbb E_{\tilde x,\tilde\varepsilon}[\mathbb I a^2\mid\Delta]$, recall that $h_\Delta=\delta/\sigma_\xi(\Delta)$ and $u=\delta/\tilde\sigma$.
In particular, $\tilde\sigma^3/\sigma_\xi(\Delta)=(\tilde\sigma^2/u)h_\Delta$ and $\tilde\sigma^3\delta^2/\sigma_\xi^3(\Delta)=(\tilde\sigma^2/u)h_\Delta^3$.
By Lemma~\ref{lem:truncated_normal_moments}, we have $h_\Delta e^{-h_\Delta^2/2}\le C\min(1,h_\Delta)\le Cp_\Delta$ and $h_\Delta^3e^{-h_\Delta^2/2}\le C\min(1,h_\Delta^3)\le Cp_\Delta\lambda_{2,\Delta}$.
Moreover, the conditional second moment formula in Lemma~\ref{lem:accepted_linear_form_moments} implies $\mathbb E_{\tilde x,\tilde\varepsilon}[\mathbb I a^2\mid\Delta]\ge p_\Delta\lambda_{2,\Delta}\|\Delta\|_{\Sigma_1}^2$.
Hence, we obtain
\begin{align}\label{eq:common_noise_mean_scalar_bounds}
\mathbb E_{\tilde x}\Bigl[
\bigl(\mathbb E_{\tilde\varepsilon}[\mathbb I\tilde\varepsilon
\mid\Delta,\tilde x]\bigr)^2\mid\Delta\Bigr]
&\le\frac{C\tilde\sigma^2p_\Delta}{u},\notag\\
\mathbb E_{\tilde x}\Bigl[
a^2\bigl(\mathbb E_{\tilde\varepsilon}[\mathbb I\tilde\varepsilon
\mid\Delta,\tilde x]\bigr)^2\mid\Delta\Bigr]
&\le\frac{C\tilde\sigma^2}{u}
\mathbb E_{\tilde x,\tilde\varepsilon}[\mathbb I a^2\mid\Delta].
\end{align}

Since $\tilde x_S$ and $\mathbb I$ are independent conditioned on $(a,\Delta)$, the tower property implies $\bar G_{\mathrm{pop}}(\Delta)=\mathbb E_{\tilde x,\tilde\varepsilon}[\mathbb I\,\mathbb E_{\tilde x}[\tilde x_S\tilde x_S^\top\mid a,\Delta]\mid\Delta]$.
Both the constant matrix and the coefficient of $a^2$ in~\eqref{eq:common_covariate_second_moment_given_a} are positive semidefinite.
Hence, we take expectation over $a$ and use the two inequalities in~\eqref{eq:common_noise_mean_scalar_bounds} to obtain the following matrix bound for the first term in~\eqref{eq:common_adjusted_noise_mean_decomposition}:
\begin{align*}
\mathbb E_{\tilde x}\Bigl[
\tilde x_S\tilde x_S^\top
\bigl(\mathbb E_{\tilde\varepsilon}[\mathbb I\tilde\varepsilon
\mid\Delta,\tilde x]\bigr)^2\mid\Delta\Bigr]
&\preceq\frac{C\tilde\sigma^2}{u}\bar G_{\mathrm{pop}}(\Delta).
\end{align*}

Next, we bound the contribution from the square of the second term in~\eqref{eq:common_adjusted_noise_mean_decomposition}.
Since $\bigl(\mathbb E_{\tilde\varepsilon}[\mathbb I\mid\Delta,\tilde x]\bigr)^2\le\mathbb E_{\tilde\varepsilon}[\mathbb I\mid\Delta,\tilde x]$, the corresponding matrix second moment has quadratic form bounded by
\begin{align*}
\theta_\Delta^2\mathbb E_{\tilde x,\tilde\varepsilon}\Bigl[
\mathbb I(v^\top\tilde x_S)^2
\bigl(\tilde x_S^\top\Sigma_{1,SS}^{-1}(\Sigma_1\Delta)_S\bigr)^2
\mid\Delta\Bigr]
\end{align*}
for every $v\in\mathbb R^s$.
To bound this expectation, we apply the Cauchy--Schwarz inequality conditioned on $(\mathbb I=1,\Delta)$, which requires the fourth moments of $v^\top\tilde x_S$ and $\tilde x_S^\top\Sigma_{1,SS}^{-1}(\Sigma_1\Delta)_S$.
For every $v\in\mathbb R^s$, we have
\begin{align*}
\mathbb E_{\tilde x}[(v^\top\tilde x_S)^4\mid\mathbb I=1,\Delta]
&\le C\|v^\top\tilde x_S\|_{\psi_2}^4
\le C\left(\mathbb E_{\tilde x}[(v^\top\tilde x_S)^2\mid\mathbb I=1,\Delta]\right)^2,
\end{align*}
where the $\psi_2$-norm is evaluated conditioned on $(\mathbb I=1,\Delta)$, and the last inequality follows from the subgaussian bound in Lemma~\ref{lem:accepted_linear_form_moments}.
We apply this inequality to $v^\top\tilde x_S$ and $\tilde x_S^\top\Sigma_{1,SS}^{-1}(\Sigma_1\Delta)_S$.
Then, the Cauchy--Schwarz inequality implies, for every $v\in\mathbb R^s$,
\begin{align*}
&\mathbb E_{\tilde x,\tilde\varepsilon}\Bigl[
\mathbb I(v^\top\tilde x_S)^2
\bigl(\tilde x_S^\top\Sigma_{1,SS}^{-1}(\Sigma_1\Delta)_S\bigr)^2
\mid\Delta\Bigr]\\
&\quad\le p_\Delta
\left(\mathbb E_{\tilde x}[(v^\top\tilde x_S)^4\mid\mathbb I=1,\Delta]\right)^{1/2}
\left(\mathbb E_{\tilde x}\left[
\bigl(\tilde x_S^\top\Sigma_{1,SS}^{-1}(\Sigma_1\Delta)_S\bigr)^4
\mid\mathbb I=1,\Delta\right]\right)^{1/2}\\
&\quad\le Cp_\Delta\,
\mathbb E_{\tilde x}\left[(v^\top\tilde x_S)^2\mid\mathbb I=1,\Delta\right]
\mathbb E_{\tilde x}\left[
\bigl(\tilde x_S^\top\Sigma_{1,SS}^{-1}(\Sigma_1\Delta)_S\bigr)^2
\mid\mathbb I=1,\Delta\right]\\
&\quad=\frac{C}{p_\Delta}
\bigl(v^\top\bar G_{\mathrm{pop}}(\Delta)v\bigr)
\|\Sigma_{1,SS}^{-1}(\Sigma_1\Delta)_S\|_{\bar G_{\mathrm{pop}}(\Delta)}^2.
\end{align*}
By~\eqref{eq:common_population_gram} and~\eqref{eq:theta_Delta_def}, we also have
\begin{align*}
(\theta_\Delta^2/p_\Delta)
\|\Sigma_{1,SS}^{-1}(\Sigma_1\Delta)_S\|_{\bar G_{\mathrm{pop}}(\Delta)}^2
&=\theta_\Delta\tilde\sigma^2(1-\lambda_{2,\Delta})
\|(\Sigma_1\Delta)_S\|_{\Sigma_{1,SS}^{-1}}^2/\sigma_\xi^2(\Delta)
\le\theta_\Delta\tilde\sigma^2,
\end{align*}
where the last inequality follows from $1-\lambda_{2,\Delta}\le1$ and $\|(\Sigma_1\Delta)_S\|_{\Sigma_{1,SS}^{-1}}^2\le\|\Delta\|_{\Sigma_1}^2\le\sigma_\xi^2(\Delta)$.
Hence, after multiplication by $\theta_\Delta^2$, the upper bound is $C\theta_\Delta\tilde\sigma^2v^\top\bar G_{\mathrm{pop}}(\Delta)v$.
Since $v\in\mathbb R^s$ is arbitrary, we obtain
\begin{align*}
\theta_\Delta^2\mathbb E_{\tilde x}\Bigl[
\tilde x_S\tilde x_S^\top
\bigl(\mathbb E_{\tilde\varepsilon}[\mathbb I\mid\Delta,\tilde x]\bigr)^2
\bigl(\tilde x_S^\top\Sigma_{1,SS}^{-1}(\Sigma_1\Delta)_S\bigr)^2
\mid\Delta\Bigr]
&\preceq C\theta_\Delta\tilde\sigma^2\bar G_{\mathrm{pop}}(\Delta)
\preceq\frac{C\tilde\sigma^2}{u^2}\bar G_{\mathrm{pop}}(\Delta),
\end{align*}
where the last inequality follows from
Lemma~\ref{lem:common_population_theta_bounds}.

Finally, we apply $(x-y)^2\le2x^2+2y^2$ to~\eqref{eq:common_adjusted_noise_mean_decomposition} and combine the two matrix bounds.
Since $u>1$, we have $u^{-2}\le u^{-1}$, and the resulting upper bound is $(C\tilde\sigma^2/u)\bar G_{\mathrm{pop}}(\Delta)$.
By Lemma~\ref{lem:common_population_theta_bounds}, $1-\theta_\Delta\asymp1$, while $u/(1+u^2)\asymp u^{-1}$ in this range.
Hence, the claimed bound follows.
\end{proof}

%% file: Unified_appendix/app_08F_empirical_comparisons.tex

\section{Empirical Mean and Covariance Bounds}
\label{sec:common_empirical}

In Section~\ref{sec:unified_population}, we have evaluated the population quantities used in $\mathcal V_1$ and $\mathcal V_2$.
In this section, we establish the empirical bounds conditioned on $\Delta$, for $0<\delta<\infty$.
First, we bound $\bar G(\Delta)$ and its inverse, then establish the conditional mean and covariance bounds used in the variance analysis.
Finally, we derive the accepted Gram matrix and conditional mean bounds under full fine-tuning.

\subsection{Conditional Expected Gram Matrix Bounds}

We fix any $\Delta\in\mathbb R^d$.
Recall that $\mathbb E_{\tilde X}[\bar G(\Delta)\mid\Delta]=m\bar G_{\mathrm{pop}}(\Delta)$.
We bound $\bar G(\Delta)$ around this expectation and obtain the inverse bounds needed for the conditional mean and covariance analysis.
To express the population inverse bounds, define
\begin{equation}\label{eq:zeta_delta_def}
\zeta_\Delta^2  :=
\frac{\sigma_\xi^2(\Delta)}
{\tilde\sigma^2+\|\Delta_{-S}\|_{\Sigma_{1,-S\mid S}}^2+
\lambda_{2,\Delta}\|(\Sigma_1\Delta)_S\|_{\Sigma_{1,SS}^{-1}}^2}.
\end{equation}
Since $0<\lambda_{2,\Delta}<1$, we have $\zeta_\Delta^2\ge1$.
By~\eqref{eq:common_population_inverse_factor_scale}, we obtain
\begin{align*}
\zeta_\Delta^2
&\asymp
1+\min\left(
\frac{\|(\Sigma_1\Delta)_S\|_{\Sigma_{1,SS}^{-1}}^2}
{\tilde\sigma^2+\|\Delta_{-S}\|_{\Sigma_{1,-S\mid S}}^2},
\frac{\sigma_\xi^2(\Delta)}{\delta^2}
\right).
\end{align*}
Then, we invoke Lemma~\ref{lem:common_population_covariance}
with  the definition of $\zeta_\Delta^2$ and obtain
\begin{align}
\tr(\Sigma_{1,SS}\bar G_{\mathrm{pop}}(\Delta)^{-1})
&=\frac{s-1+\zeta_\Delta^2}{p_\Delta},
\label{eq:common_delta_population_trace}\\
\left\|\Sigma_{1,SS}^{1/2}\bar G_{\mathrm{pop}}(\Delta)^{-1}
\Sigma_{1,SS}^{1/2}\right\|_{\mathrm{op}}
&=\frac{\zeta_\Delta^2}{p_\Delta}.
\label{eq:common_delta_population_operator}
\end{align}

\begin{corollary}[Conditional expected Gram matrix bounds]
\label{cor:common_empirical_gram}
Under the model in Section~\ref{sec:problem_setting}, fix any $\Delta\in\mathbb R^d$.
Let $0<\eta\le1/2$ and fix a sufficiently small constant $c>0$.
Suppose that, for a sufficiently large constant $C$,
\begin{align}\label{eq:common_delta_empirical_sample}
m p_\Delta
\ge C\zeta_\Delta^2\bigl(s+\log(m/\eta)\bigr)
\log^2(m/\eta).
\end{align}
Then, conditioned on $\Delta$, with probability at least $1-\eta$
over $\tilde X$, the following bounds hold simultaneously:
\begin{align}\label{eq:common_empirical_gram_comparison}
\frac{1}{2}m\bar G_{\mathrm{pop}}(\Delta)
\preceq\bar G(\Delta)
\preceq\frac{3}{2}m\bar G_{\mathrm{pop}}(\Delta).
\end{align}
The inverse satisfies
\begin{align*}
\tr(\Sigma_{1,SS}\bar G(\Delta)^{-1})
&\le\frac{C\{s+\zeta_\Delta^2\}}{m p_\Delta},\qquad
\left\|\bar G(\Delta)^{-1/2}\Sigma_{1,SS}
\bar G(\Delta)^{-1/2}\right\|_{\mathrm{op}}
\le\frac{C\zeta_\Delta^2}{m p_\Delta}.
\end{align*}
Moreover, for $\ell_{\max}(\Delta):=\max_{1\le i\le m}\tilde x_{i,S}^\top\bar G(\Delta)^{-1}\tilde x_{i,S}$, we have
\begin{align*}
\ell_{\max}(\Delta)
&\le\frac{C\{s+\zeta_\Delta^2\log(m/\eta)\}}{m p_\Delta},\qquad
\min(s,\zeta_\Delta^2)\ell_{\max}(\Delta)\log(m/\eta)\le c.
\end{align*}
\end{corollary}

\begin{proof}
Recall that $P_i(\Delta)=\mathbb E_{\tilde\varepsilon_i}[\mathbb I_i\mid\Delta,\tilde X]$ and $\bar G(\Delta)=\sum_{i=1}^mP_i(\Delta)\tilde x_{i,S}\tilde x_{i,S}^\top$.
The bound in~\eqref{eq:common_empirical_gram_comparison} is equivalent to
\begin{align*}
&\left\|\bar G_{\mathrm{pop}}(\Delta)^{-1/2}
\left(\frac{\bar G(\Delta)}{m}-\bar G_{\mathrm{pop}}(\Delta)\right)
\bar G_{\mathrm{pop}}(\Delta)^{-1/2}\right\|_{\mathrm{op}}\\
&\quad=\sup_{\|v\|=1}\left|\frac1m\sum_{i=1}^m
\left[P_i(\Delta)
\left(v^\top\bar G_{\mathrm{pop}}(\Delta)^{-1/2}\tilde x_{i,S}\right)^2-1\right]\right|
\le\frac12.
\end{align*}
For every unit vector $v\in\mathbb R^s$, the definition of $\bar G_{\mathrm{pop}}(\Delta)$ implies
\begin{align*}
\mathbb E_{\tilde x_i}\left[P_i(\Delta)
\left(v^\top\bar G_{\mathrm{pop}}(\Delta)^{-1/2}\tilde x_{i,S}\right)^2\middle|\Delta\right]
&=1,\qquad
\mathbb E_{\tilde x_i}\left[
\left(v^\top\bar G_{\mathrm{pop}}(\Delta)^{-1/2}\tilde x_{i,S}\right)^2
\middle|\mathbb I_i=1,\Delta\right]=\frac{1}{p_\Delta}.
\end{align*}
Hence, the variables $P_i(\Delta)(v^\top\bar G_{\mathrm{pop}}(\Delta)^{-1/2}\tilde x_{i,S})^2-1$ are centered conditioned on $\Delta$, and they are independent since each depends only on $\tilde x_i$.
Next, we verify the moment condition in Lemma~\ref{lem:bernstein}.
By Lemma~\ref{lem:accepted_linear_form_moments}, the $\psi_2$-norm of $v^\top\bar G_{\mathrm{pop}}(\Delta)^{-1/2}\tilde x_{i,S}$ conditioned on $(\mathbb I_i=1,\Delta)$ is at most $C/\sqrt{p_\Delta}$.
Since $P_i(\Delta)^k\le P_i(\Delta)$, the tower property implies, for every integer $k\ge2$,
\begin{align*}
\mathbb E_{\tilde x_i}\left[P_i(\Delta)^k
\left(v^\top\bar G_{\mathrm{pop}}(\Delta)^{-1/2}\tilde x_{i,S}\right)^{2k}\middle|\Delta\right]
\le p_\Delta\mathbb E_{\tilde x_i}\left[
\left(v^\top\bar G_{\mathrm{pop}}(\Delta)^{-1/2}\tilde x_{i,S}\right)^{2k}
\middle|\mathbb I_i=1,\Delta\right]
\le\frac{C^k k!}{p_\Delta^{k-1}},
\end{align*}
where the last inequality follows from the conditional $\psi_2$-bound and $t^k\le k!e^t$ for $t\ge0$.
Then, by $|x-1|^k\le2^{k-1}(|x|^k+1)$, it follows that
\begin{align*}
\sum_{i=1}^m\mathbb E_{\tilde x_i}\left[
\left|P_i(\Delta)\left(v^\top\bar G_{\mathrm{pop}}(\Delta)^{-1/2}\tilde x_{i,S}\right)^2-1\right|^k
\middle|\Delta\right]
&\le\frac{k!}{2}\frac{Cm}{p_\Delta}\left(\frac{C}{p_\Delta}\right)^{k-2}.
\end{align*}
Let $\mathcal N$ be a $1/4$-net of the unit sphere with $|\mathcal N|\le9^s$.
We apply Lemma~\ref{lem:bernstein} with $V=Cm/p_\Delta$, $B=C/p_\Delta$, and $t=\log(4|\mathcal N|/\eta)$ for each $v\in\mathcal N$ and bound the operator norm by twice the maximum absolute quadratic form over $\mathcal N$, as in~\cite{vershynin2019high}.
Then, with conditional failure probability at most $\eta/2$, we obtain
\begin{align*}
\bigg\|\bar G_{\mathrm{pop}}(\Delta)^{-1/2}
\bigg(\frac{\bar G(\Delta)}{m}-\bar G_{\mathrm{pop}}(\Delta)\bigg)
\bar G_{\mathrm{pop}}(\Delta)^{-1/2}\bigg\|_{\mathrm{op}}
\le C\left(\sqrt{\frac{s+\log(m/\eta)}{mp_\Delta}}
+\frac{s+\log(m/\eta)}{mp_\Delta}\right)\le\frac12,
\end{align*}
where the last inequality follows from~\eqref{eq:common_delta_empirical_sample}, which implies $mp_\Delta\ge C(s+\log(m/\eta))$ for a sufficiently large constant $C$.
Thus, we obtain~\eqref{eq:common_empirical_gram_comparison}.
Moreover, by~\eqref{eq:common_delta_population_trace} and~\eqref{eq:common_delta_population_operator}, we obtain the stated inverse trace and operator norm bounds.

To bound $\ell_{\max}(\Delta)$, we apply Lemma~\ref{lem:quad_form_concentration} to the Gaussian vectors $\bar G_{\mathrm{pop}}(\Delta)^{-1/2}\tilde x_{i,S}$.
By~\eqref{eq:common_delta_population_trace}, \eqref{eq:common_delta_population_operator}, and a union bound over $i$, we obtain, with conditional failure probability at most $\eta/2$,
\begin{equation}\label{eq:common_empirical_population_row}
\max_{1\le i\le m}
\tilde x_{i,S}^\top\bar G_{\mathrm{pop}}(\Delta)^{-1}\tilde x_{i,S}
\le\frac{C\bigl(s+\zeta_\Delta^2\log(m/\eta)\bigr)}{p_\Delta}.
\end{equation}
On the intersection of the events in~\eqref{eq:common_empirical_gram_comparison} and~\eqref{eq:common_empirical_population_row}, we have
\begin{align*}
\ell_{\max}(\Delta)
\le\frac{2}{m}\max_{1\le i\le m}
\tilde x_{i,S}^\top\bar G_{\mathrm{pop}}(\Delta)^{-1}\tilde x_{i,S}
\le\frac{C}{mp_\Delta}
\bigl(s+\zeta_\Delta^2\log(m/\eta)\bigr).
\end{align*}
Finally, the bound for $\ell_{\max}(\Delta)$ and~\eqref{eq:common_delta_empirical_sample} imply
\begin{align*}
\min(s,\zeta_\Delta^2)\ell_{\max}(\Delta)\log(m/\eta)
&\le\frac{C\zeta_\Delta^2(s+\log(m/\eta))\log^2(m/\eta)}{mp_\Delta}\le c.
\end{align*}
Each of~\eqref{eq:common_empirical_gram_comparison} and~\eqref{eq:common_empirical_population_row} has conditional failure probability at most $\eta/2$.
Hence, the union bound establishes all conclusions simultaneously with conditional probability at least $1-\eta$.
\end{proof}

\subsection{Empirical Conditional Mean Bounds}

The term $\bar G(\Delta)^{-1}\bar\mu(\Delta)$ is included
in both the conditional mean contribution $\mathcal V_1$ and the conditional
covariance $M(\Delta)$.  We compare this term with
$\bar G_{\mathrm{pop}}(\Delta)^{-1}\bar\mu_{\mathrm{pop}}(\Delta)$.
Whenever $\bar G(\Delta)$ is invertible, we have
\begin{align*}
\bar G(\Delta)^{-1}\bar\mu(\Delta)-\bar G_{\mathrm{pop}}(\Delta)^{-1}\bar\mu_{\mathrm{pop}}(\Delta)
&=\bar G(\Delta)^{-1}\bigl(\bar\mu(\Delta)-\bar G(\Delta)\bar G_{\mathrm{pop}}(\Delta)^{-1}\bar\mu_{\mathrm{pop}}(\Delta)\bigr).
\end{align*}
First, we bound this difference in the $\bar G(\Delta)$ norm.

\begin{corollary}[Conditional mean bound in the Gram norm]
\label{cor:common_empirical_first_moment}
Under the model in Section~\ref{sec:problem_setting}, fix any $\Delta\in\mathbb R^d$.
Let $0<\eta\le1/2$.
Suppose that \eqref{eq:common_delta_empirical_sample} holds with $C$ sufficiently large.
Then, conditioned on $\Delta$, with probability at least
$1-\eta$ over $\tilde X$, 
the matrix $\bar G(\Delta)$ is positive definite and
\begin{align*}
&\left\|\bar G(\Delta)^{-1}\bar\mu(\Delta)
-\bar G_{\mathrm{pop}}(\Delta)^{-1}\bar\mu_{\mathrm{pop}}(\Delta)
\right\|_{\bar G(\Delta)}^2\\
&\quad\le C\tilde\sigma^2(1-\theta_\Delta)
\left[\frac{u}{1+u^2}s\log(m/\eta)
+\frac{\tr(\Sigma_{1,SS}\bar G_{\mathrm{pop}}(\Delta)^{-1})
\log^4(m/\eta)}{m}\right],
\end{align*}
where $u=\delta/\tilde\sigma$ and $\theta_\Delta$ is defined in~\eqref{eq:theta_Delta_def}.
\end{corollary}

\begin{proof}
For $i=1,\ldots,m$, let $W_i=\tilde\varepsilon_i-\theta_\Delta\tilde x_{i,S}^\top\Sigma_{1,SS}^{-1}(\Sigma_1\Delta)_S$.
By Lemma~\ref{lem:common_population_mu} and the definitions of $\bar\mu(\Delta)$ and $\bar G(\Delta)$, we have
\begin{align}\label{eq:common_empirical_centered_first_moment}
\bar\mu(\Delta)-\bar G(\Delta)\bar G_{\mathrm{pop}}(\Delta)^{-1}
\bar\mu_{\mathrm{pop}}(\Delta)
&=\sum_{i=1}^m\tilde x_{i,S}
\mathbb E_{\tilde\varepsilon_i}
(\mathbb I_iW_i\mid\Delta,\tilde x_i).
\end{align}
Moreover, it holds that $\mathbb E_{\tilde x_i,\tilde\varepsilon_i}[\mathbb I_i\tilde x_{i,S}W_i\mid\Delta]
=0$.
By the tower property, the vectors $\tilde x_{i,S}\mathbb E_{\tilde\varepsilon_i}[\mathbb I_iW_i\mid\Delta,\tilde x_i]$ in~\eqref{eq:common_empirical_centered_first_moment} are centered conditioned on $\Delta$, and they are independent since each depends only on $\tilde x_i$.

To apply Lemma~\ref{lem:vector_bernstein}, we bound the second moments and $\psi_1$-norms of $\bar G_{\mathrm{pop}}(\Delta)^{-1/2}\tilde x_{i,S}\mathbb E_{\tilde\varepsilon_i}(\mathbb I_iW_i\mid\Delta,\tilde x_i)$.
By Lemma~\ref{lem:common_population_residual_mean}, we obtain
\begin{align*}
\sum_{i=1}^m\mathbb E_{\tilde x_i}\left[
\left\|\bar G_{\mathrm{pop}}(\Delta)^{-1/2}\tilde x_{i,S}
\mathbb E_{\tilde\varepsilon_i}
(\mathbb I_iW_i\mid\Delta,\tilde x_i)\right\|_2^2
\middle|\Delta\right]
&\le C\frac{u}{1+u^2}\tilde\sigma^2(1-\theta_\Delta)ms.
\end{align*}

Next, we bound the $\psi_1$-norms required by Lemma~\ref{lem:vector_bernstein}.
For this, we first establish a $\psi_2$-bound for $\mathbb E_{\tilde\varepsilon_i}(\mathbb I_iW_i\mid\Delta,\tilde x_i)$.
Conditioned on $\Delta$, $(W_i,\xi_i)$ is jointly centered Gaussian.
Hence, $W_i-\Cov(W_i,\xi_i\mid\Delta)\xi_i/\sigma_\xi^2(\Delta)$ is centered Gaussian and independent of $\xi_i$.
Conditioned on $(\mathbb I_i=1,\Delta)$, we have $\|\xi_i\|_{\psi_2}^2\le C\mathbb E[\xi_i^2\mid\mathbb I_i=1,\Delta]$, as established in the proof of Lemma~\ref{lem:accepted_linear_form_moments}.
Hence, the triangle inequality implies
\begin{align}
\|W_i\|_{\psi_2}^2
&\le C\left(\Var(W_i\mid\xi_i,\Delta)
+\frac{\Cov(W_i,\xi_i\mid\Delta)^2}{\sigma_\xi^4(\Delta)}
\mathbb E[\xi_i^2\mid\mathbb I_i=1,\Delta]\right)\notag\\
&=C\mathbb E[W_i^2\mid\mathbb I_i=1,\Delta]
=C\tilde\sigma^2(1-\theta_\Delta),\label{eq:common_accepted_W_subgaussian}
\end{align}
where the last equality follows from~\eqref{eq:common_population_adjusted_noise_scalar_second_moment} in the proof of Lemma~\ref{lem:common_population_residual_covariance}.
Since $\mathbb I_iW_i=0$ when $\mathbb I_i=0$, the same bound holds for the squared $\psi_2$-norm of $\mathbb I_iW_i$ conditioned on $\Delta$.
By conditional Jensen's inequality, we obtain, for every $u>0$,
\begin{align*}
\left\|\mathbb E_{\tilde\varepsilon_i}
(\mathbb I_iW_i\mid\Delta,\tilde x_i)\right\|_{\psi_2}^2
&\le\|\mathbb I_iW_i\|_{\psi_2}^2
\le C\tilde\sigma^2(1-\theta_\Delta).
\end{align*}
The norm of the conditional expectation and the remaining Orlicz norms are evaluated over $\tilde x_i$ conditioned on $\Delta$.
By the moment bound in Lemma~\ref{lem:quad_form_concentration}, we have $\bigl\|\|\bar G_{\mathrm{pop}}(\Delta)^{-1/2}\tilde x_{i,S}\|_2\bigr\|_{\psi_2}^2\le C\tr(\Sigma_{1,SS}\bar G_{\mathrm{pop}}(\Delta)^{-1})$.
Then, the inequality $\|xy\|_{\psi_1}\le\|x\|_{\psi_2}\|y\|_{\psi_2}$ and the bound for $\mathbb E_{\tilde\varepsilon_i}(\mathbb I_iW_i\mid\Delta,\tilde x_i)$ 
provide
\begin{align*}
\left\|\left\|
\bar G_{\mathrm{pop}}(\Delta)^{-1/2}\tilde x_{i,S}
\mathbb E_{\tilde\varepsilon_i}
(\mathbb I_iW_i\mid\Delta,\tilde x_i)
\right\|_2\right\|_{\psi_1}^2
\le C\tilde\sigma^2(1-\theta_\Delta)
\tr(\Sigma_{1,SS}\bar G_{\mathrm{pop}}(\Delta)^{-1}).
\end{align*}
Finally, we apply Lemma~\ref{lem:vector_bernstein} with failure probability $\eta/2$ to $\bar G_{\mathrm{pop}}(\Delta)^{-1/2}\tilde x_{i,S}\mathbb E_{\tilde\varepsilon_i}(\mathbb I_iW_i\mid\Delta,\tilde x_i)$, and obtain
\begin{align*}
&\left\|\bar G_{\mathrm{pop}}(\Delta)^{-1/2}
\sum_{i=1}^m\tilde x_{i,S}
\mathbb E_{\tilde\varepsilon_i}
(\mathbb I_iW_i\mid\Delta,\tilde x_i)\right\|_2^2\\
&\quad\le C\tilde\sigma^2(1-\theta_\Delta)
\left[\frac{u}{1+u^2}m s\log(m/\eta)
+\tr(\Sigma_{1,SS}\bar G_{\mathrm{pop}}(\Delta)^{-1})
\log^4(m/\eta)\right].
\end{align*}
By Corollary~\ref{cor:common_empirical_gram} and~\eqref{eq:common_delta_empirical_sample}, for sufficiently large $C$, \eqref{eq:common_empirical_gram_comparison} holds with conditional failure probability at most $\eta/2$.
On this event, $\bar G(\Delta)$ is positive definite and $\bar G(\Delta)^{-1}\preceq(2/m)\bar G_{\mathrm{pop}}(\Delta)^{-1}$.
Then, by the union bound, the bound from Lemma~\ref{lem:vector_bernstein} and~\eqref{eq:common_empirical_gram_comparison} hold simultaneously with conditional probability at least $1-\eta$.
On their intersection, it follows from \eqref{eq:common_empirical_centered_first_moment} and $\bar G(\Delta)^{-1}\preceq(2/m)\bar G_{\mathrm{pop}}(\Delta)^{-1}$ that
\begin{align*}
\|\bar G(\Delta)^{-1}\bar\mu(\Delta)
-\bar G_{\mathrm{pop}}(\Delta)^{-1}\bar\mu_{\mathrm{pop}}(\Delta)
\|_{\bar G(\Delta)}^2
&\le\frac{2}{m}\|\bar G_{\mathrm{pop}}(\Delta)^{-1/2}
(\bar\mu(\Delta)-\bar G(\Delta)\bar G_{\mathrm{pop}}(\Delta)^{-1}
\bar\mu_{\mathrm{pop}}(\Delta))\|_2^2.
\end{align*}
By substituting the bound for the sum in~\eqref{eq:common_empirical_centered_first_moment}, we complete the proof.
\end{proof}

For the conditional covariance analysis, we use
Corollary~\ref{cor:common_empirical_first_moment}  to bound the conditional mean difference at
 the scale $\tilde\sigma^2(1-\theta_\Delta)$.
For $\mathcal V_1$, we instead derive a prediction norm bound proportional to $\|\Delta\|_{\Sigma_1}^2$
and then take expectation over the pre-training error.

\begin{lemma}[Conditional mean bound in the prediction norm]
\label{lem:common_empirical_prediction_mean}
Under the model in Section~\ref{sec:problem_setting}, fix any $\Delta\in\mathbb R^d$.
Let $0<\eta\le1/2$.
Suppose that \eqref{eq:common_delta_empirical_sample} holds with $C$ sufficiently large.
Then, conditioned on $\Delta$, with probability at least $1-\eta$ over
$\tilde X$, \eqref{eq:common_empirical_gram_comparison} holds and
\begin{align*}
&\left\|\bar G(\Delta)^{-1}\bar\mu(\Delta)
-\bar G_{\mathrm{pop}}(\Delta)^{-1}\bar\mu_{\mathrm{pop}}(\Delta)
\right\|_{\Sigma_{1,SS}}^2\\
&\quad\le\frac{C\min(1,u)(s+\log(m/\eta))}{mp_\Delta}
\left[1+\min\left(
\frac{\|(\Sigma_1\Delta)_S\|_{\Sigma_{1,SS}^{-1}}^2}
{\tilde\sigma^2+\|\Delta_{-S}\|_{\Sigma_{1,-S\mid S}}^2},
\frac{\sigma_\xi^2(\Delta)}{\delta^2}
\right)\right]\|\Delta\|_{\Sigma_1}^2,
\end{align*}
where $u=\delta/\tilde\sigma$.
\end{lemma}

\begin{proof}
As in the proof of Corollary~\ref{cor:common_empirical_first_moment}, for $i=1,\ldots,m$, let $W_i=\tilde\varepsilon_i-\theta_\Delta\tilde x_{i,S}^\top\Sigma_{1,SS}^{-1}(\Sigma_1\Delta)_S$.
The vectors $\tilde x_{i,S}\mathbb E_{\tilde\varepsilon_i}[\mathbb I_iW_i\mid\Delta,\tilde x_i]$ in~\eqref{eq:common_empirical_centered_first_moment} are independent and centered conditioned on $\Delta$.

For sufficiently large $C$ in the sample condition, Corollary~\ref{cor:common_empirical_gram} implies that \eqref{eq:common_empirical_gram_comparison} holds with conditional failure probability at most $\eta/2$.
On this event, $\|(\bar G_{\mathrm{pop}}(\Delta)^{-1/2}\bar G(\Delta)\bar G_{\mathrm{pop}}(\Delta)^{-1/2})^{-1}\|_{\mathrm{op}}\le2/m$.
By~\eqref{eq:common_delta_population_operator}, we obtain
\begin{align}\label{eq:common_empirical_prediction_reduction}
&\bigl\|\bar G(\Delta)^{-1}\bar\mu(\Delta)
-\bar G_{\mathrm{pop}}(\Delta)^{-1}\bar\mu_{\mathrm{pop}}(\Delta)
\bigr\|_{\Sigma_{1,SS}}^2\\
&\quad\le\frac{C\zeta_\Delta^2}{m^2p_\Delta}
\bigl\|\bar G_{\mathrm{pop}}(\Delta)^{-1/2}
\bigl(\bar\mu(\Delta)-\bar G(\Delta)\bar G_{\mathrm{pop}}(\Delta)^{-1}
\bar\mu_{\mathrm{pop}}(\Delta)\bigr)\bigr\|_2^2.
\end{align}
To obtain a bound proportional to $\|\Delta\|_{\Sigma_1}^2$, we first bound $\mathbb E_{\tilde\varepsilon_i}[\mathbb I_iW_i\mid\Delta,\tilde x_i]$.
If $\|\Delta\|_{\Sigma_1}=0$, the right hand side of
\eqref{eq:common_empirical_centered_first_moment} is zero by symmetry, and the claim follows
from Corollary~\ref{cor:common_empirical_gram}.
Hence, suppose that $\|\Delta\|_{\Sigma_1}>0$.
Conditioned on $(\Delta,\tilde x_i)$, acceptance restricts $\tilde\varepsilon_i$ to $[\tilde x_i^\top\Delta-\delta,\tilde x_i^\top\Delta+\delta]$.
Then, by the Gaussian density, its mean conditioned on acceptance lies between $0$ and $\tilde x_i^\top\Delta$.
Hence, $|\mathbb E_{\tilde\varepsilon_i}[\mathbb I_i\tilde\varepsilon_i\mid\Delta,\tilde x_i]|\le P_i(\Delta)|\tilde x_i^\top\Delta|$.
Therefore,
\begin{align*}
\left|\mathbb E_{\tilde\varepsilon_i}
(\mathbb I_iW_i\mid\Delta,\tilde x_i)\right|
&\le P_i(\Delta)\left(
|\tilde x_i^\top\Delta|+
\left|\theta_\Delta\tilde x_{i,S}^\top\Sigma_{1,SS}^{-1}
(\Sigma_1\Delta)_S\right|\right).
\end{align*}
Recall that $u=\delta/\tilde\sigma$ and $p(h)=2\Phi(h)-1$.
We also have $P_i(\Delta)\le\mathbb P(|\tilde\varepsilon_i|\le\delta)=p(\delta/\tilde\sigma)=p(u)\le C\min(1,u)$.

For a unit vector $v\in\mathbb R^s$, the definition of $\bar G_{\mathrm{pop}}(\Delta)$ implies $\mathbb E_{\tilde x_i}[(v^\top\bar G_{\mathrm{pop}}(\Delta)^{-1/2}\tilde x_{i,S})^2\mid\mathbb I_i=1,\Delta]=p_\Delta^{-1}$.
By Lemma~\ref{lem:common_population_theta_bounds} and the Schur complement decomposition, we have $0<\theta_\Delta<1$ and $\theta_\Delta^2\|(\Sigma_1\Delta)_S\|_{\Sigma_{1,SS}^{-1}}^2\le\|\Delta\|_{\Sigma_1}^2$.
Then, conditioned on $(\mathbb I_i=1,\Delta)$, we apply Lemma~\ref{lem:accepted_linear_form_moments} to these three terms as follows.
\begin{align*}
\|v^\top\bar G_{\mathrm{pop}}(\Delta)^{-1/2}\tilde x_{i,S}\|_{\psi_2}
&\le\frac{C}{\sqrt{p_\Delta}},
&
\|\tilde x_i^\top\Delta\|_{\psi_2}+\|\theta_\Delta\tilde x_{i,S}^\top
\Sigma_{1,SS}^{-1}(\Sigma_1\Delta)_S\|_{\psi_2}
&\le C\|\Delta\|_{\Sigma_1}.
\end{align*}
By the inequality $\|xy\|_{\psi_1}\le\|x\|_{\psi_2}\|y\|_{\psi_2}$,  we obtain
\begin{align}\label{eq:common_empirical_prediction_product}
\left\|\left(v^\top\bar G_{\mathrm{pop}}(\Delta)^{-1/2}\tilde x_{i,S}\right)
\left(|\tilde x_i^\top\Delta|
+\left|\theta_\Delta\tilde x_{i,S}^\top\Sigma_{1,SS}^{-1}(\Sigma_1\Delta)_S\right|\right)\right\|_{\psi_1}
&\le\frac{C\|\Delta\|_{\Sigma_1}}{\sqrt{p_\Delta}},
\end{align}
where the norm is evaluated conditioned on $(\mathbb I_i=1,\Delta)$.
Since $P_i(\Delta)^k\le(C\min(1,u))^{k-1}P_i(\Delta)$, the tower property implies, for every integer $k\ge2$,
\begin{align*}
&\sum_{i=1}^m\mathbb E_{\tilde x_i}\left[
\left|v^\top\bar G_{\mathrm{pop}}(\Delta)^{-1/2}\tilde x_{i,S}
\mathbb E_{\tilde\varepsilon_i}
(\mathbb I_iW_i\mid\Delta,\tilde x_i)\right|^k
\middle|\Delta\right]\\
&\quad\le m(C\min(1,u))^{k-1}p_\Delta k!
\left(\frac{C\|\Delta\|_{\Sigma_1}}{\sqrt{p_\Delta}}\right)^k\\
&\quad\le\frac{k!}{2}
\left(Cm\min(1,u)\|\Delta\|_{\Sigma_1}^2\right)
\left(\frac{C\min(1,u)\|\Delta\|_{\Sigma_1}}
{\sqrt{p_\Delta}}\right)^{k-2},
\end{align*}
where the first inequality follows from~\eqref{eq:common_empirical_prediction_product} and $t^k\le k!e^t$ for $t\ge0$.
Hence, the moment condition in Lemma~\ref{lem:bernstein} holds.

Let $\mathcal N$ be a $1/2$-net of the unit sphere with $|\mathcal N|\le5^s$, as in \citet[Corollary~4.2.11]{vershynin2019high}.
We apply Lemma~\ref{lem:bernstein} with $t=\log(4|\mathcal N|/\eta)$ for each $v\in\mathcal N$ and bound the Euclidean norm by twice the maximum absolute inner product over $\mathcal N$, as in \citet[Exercise~4.35]{vershynin2019high}.
Then, with conditional failure probability at most $\eta/2$, we obtain
\begin{align*}
&\left\|\bar G_{\mathrm{pop}}(\Delta)^{-1/2}
\left(\bar\mu(\Delta)-\bar G(\Delta)
\bar G_{\mathrm{pop}}(\Delta)^{-1}\bar\mu_{\mathrm{pop}}(\Delta)\right)
\right\|_2^2\\*
&\quad\le C\|\Delta\|_{\Sigma_1}^2\left(
m\min(1,u)(s+\log(m/\eta))
+\frac{\min(1,u)^2(s+\log(m/\eta))^2}{p_\Delta}\right)\\*
&\quad\le Cm\min(1,u)\|\Delta\|_{\Sigma_1}^2
(s+\log(m/\eta)),
\end{align*}
where the last inequality follows from~\eqref{eq:common_delta_empirical_sample}, which implies $mp_\Delta\ge C\min(1,u)(s+\log(m/\eta))$.

Finally, the union bound implies that this inequality and~\eqref{eq:common_empirical_gram_comparison} hold simultaneously with conditional probability at least $1-\eta$.
We substitute the bound into~\eqref{eq:common_empirical_prediction_reduction} and apply~\eqref{eq:common_population_inverse_factor_scale} to obtain the claimed prediction norm bound.
\end{proof}

\subsection{Conditional Covariance Trace Bounds}

We bound the two traces associated with
 the empirical conditional covariance $M(\Delta)$ required to analyze $\mathcal V_2$ and the inverse remainder.
Whenever $\bar G(\Delta)$ is invertible, \eqref{eq:M_def} implies
\begin{align*}
M(\Delta)
=\sum_{i=1}^m\tilde x_{i,S}\tilde x_{i,S}^{\top}
\Var_{\tilde\varepsilon_i}\bigl(
\mathbb I_i\bigl[\tilde\varepsilon_i-\tilde x_{i,S}^{\top}
(\Delta_S+\bar G(\Delta)^{-1}\bar\mu(\Delta))\bigr]
\mid\Delta,\tilde X\bigr).
\end{align*}
By Lemma~\ref{lem:common_population_mu}, we have
$\Delta_S+\bar G_{\mathrm{pop}}(\Delta)^{-1}
\bar\mu_{\mathrm{pop}}(\Delta)
=\theta_\Delta\Sigma_{1,SS}^{-1}(\Sigma_1\Delta)_S$.
First, we replace $\Delta_S+\bar G(\Delta)^{-1}\bar\mu(\Delta)$ in $M(\Delta)$ 
by this population value and compare the resulting covariance with its expectation
over $\tilde x$.  Next, we use
Corollary~\ref{cor:common_empirical_first_moment} to bound the difference
between this covariance and $M(\Delta)$.

\begin{lemma}[Conditional covariance trace bounds]
\label{lem:common_empirical_covariance}
Under the model in Section~\ref{sec:problem_setting}, let $0<\eta\le1/2$ and fix $\Delta\in\mathbb R^d$.  
Suppose that
\eqref{eq:common_delta_empirical_sample} holds with $C$ sufficiently large.
Then, conditioned on $\Delta$, with probability at least $1-\eta$ over $\tilde X$, 
the matrix $\bar G(\Delta)$ is positive definite and
\begin{align*}
\tr\left(\Sigma_{1,SS}\bar G(\Delta)^{-1}M(\Delta)
\bar G(\Delta)^{-1}\right)
&\asymp\tilde\sigma^2(1-\theta_\Delta)
\tr(\Sigma_{1,SS}\bar G(\Delta)^{-1}),\\
\tr(\bar G(\Delta)^{-1}M(\Delta))
&\le C\tilde\sigma^2(1-\theta_\Delta)s.
\end{align*}
\end{lemma}

\begin{proof}
As in the proof of Corollary~\ref{cor:common_empirical_first_moment}, for $i=1,\ldots,m$, let $W_i:=\tilde\varepsilon_i-\theta_\Delta\tilde x_{i,S}^{\top}\Sigma_{1,SS}^{-1}(\Sigma_1\Delta)_S$.
Define
\begin{align*}
B_i:=\Var_{\tilde\varepsilon_i}(\mathbb I_iW_i\mid\Delta,\tilde x_i)
\bar G_{\mathrm{pop}}(\Delta)^{-1/2}\tilde x_{i,S}\tilde x_{i,S}^{\top}
\bar G_{\mathrm{pop}}(\Delta)^{-1/2}.
\end{align*}
Conditioned on $\Delta$, the matrices $B_i$ are independent since each depends only on $\tilde x_i$.
By Lemma~\ref{lem:common_population_residual_covariance}, we have
\begin{align*}
c\tilde\sigma^2(1-\theta_\Delta)I_s
\preceq\mathbb E(B_i\mid\Delta)
\preceq C\tilde\sigma^2(1-\theta_\Delta)I_s.
\end{align*}
To bound $\sum_{i=1}^mB_i$ around its expectation, we first bound the moments of $\Var_{\tilde\varepsilon_i}(\mathbb I_iW_i\mid\Delta,\tilde x_i)$.
By~\eqref{eq:common_accepted_W_subgaussian} in the proof of Corollary~\ref{cor:common_empirical_first_moment}, we have $\|W_i\|_{\psi_2}^2\le C\tilde\sigma^2(1-\theta_\Delta)$ conditioned on $(\mathbb I_i=1,\Delta)$.
Recall that $u=\delta/\tilde\sigma$ and $P_i(\Delta)=\mathbb P_{\tilde\varepsilon_i}(\mathbb I_i=1\mid\Delta,\tilde x_i)\le C\min(1,u)$.
It holds that, for every integer $k\ge1$,
\begin{align*}
&\mathbb E_{\tilde x_i}\left[
\Var_{\tilde\varepsilon_i}(\mathbb I_iW_i\mid\Delta,\tilde x_i)^k
\middle|\Delta\right]
\le\mathbb E_{\tilde x_i,\tilde\varepsilon_i}
\left[P_i(\Delta)^{k-1}\mathbb I_i|W_i|^{2k}\mid\Delta\right]\\
&\quad\le(C\min(1,u))^{k-1}p_\Delta
\mathbb E[|W_i|^{2k}\mid\mathbb I_i=1,\Delta]
\le\bigl(Ck\min(1,u)\tilde\sigma^2(1-\theta_\Delta)\bigr)^k,
\end{align*}
where the first inequality follows from $\Var_{\tilde\varepsilon_i}(\mathbb I_iW_i\mid\Delta,\tilde x_i)\le P_i(\Delta)\mathbb E_{\tilde\varepsilon_i}[W_i^2\mid\mathbb I_i=1,\Delta,\tilde x_i]$ and Jensen's inequality. 
The last inequality follows from the conditional $\psi_2$-bound, $t^k\le k!e^t$ for $t\ge0$, and $p_\Delta\le C\min(1,u)$.
By Markov's inequality and the moment bound, we obtain, for every $\tau\ge1$,
\begin{align}\label{eq:common_conditional_variance_tail}
\mathbb P_{\tilde x_i}\left(
\Var_{\tilde\varepsilon_i}(\mathbb I_iW_i\mid\Delta,\tilde x_i)
>C\min(1,u)\tilde\sigma^2(1-\theta_\Delta)\tau
\middle|\Delta\right)
&\le C\exp(-c\tau).
\end{align}

Next, we truncate the conditional variance to apply Lemma~\ref{lem:bernstein}.
Let $L:=C\min(1,u)\tilde\sigma^2(1-\theta_\Delta)\log(m/\eta)$ and $B_i^\circ:=B_i\mathbf 1\{\Var_{\tilde\varepsilon_i}(\mathbb I_iW_i\mid\Delta,\tilde x_i)\le L\}$.

By~\eqref{eq:common_conditional_variance_tail} and a union bound, $\mathbb P_{\tilde X}(\max_{1\le i\le m}\Var_{\tilde\varepsilon_i}(\mathbb I_iW_i\mid\Delta,\tilde x_i)>L\mid\Delta)\le\eta/6$ for sufficiently large $C$.
Since truncation changes the expectation, we also bound $\mathbb E(B_i-B_i^\circ\mid\Delta)$.
For every unit vector $v\in\mathbb R^s$, Lemma~\ref{lem:accepted_linear_form_moments} implies $\|v^\top\bar G_{\mathrm{pop}}(\Delta)^{-1/2}\tilde x_{i,S}\|_{\psi_2}^2\le C/p_\Delta$ conditioned on $(\mathbb I_i=1,\Delta)$.
By the definition of $B_i$, we have
\begin{align*}
&\mathbb E_{\tilde x_i}[(v^\top B_iv)^2\mid\Delta]\\
&\quad\le\mathbb E_{\tilde x_i,\tilde\varepsilon_i}\left[
P_i(\Delta)\mathbb I_iW_i^4
\left(v^\top\bar G_{\mathrm{pop}}(\Delta)^{-1/2}\tilde x_{i,S}\right)^4
\middle|\Delta\right]\\
&\quad\le C\min(1,u)p_\Delta
\mathbb E\left[
W_i^4\left(v^\top\bar G_{\mathrm{pop}}(\Delta)^{-1/2}\tilde x_{i,S}\right)^4
\middle|\mathbb I_i=1,\Delta\right]
\le\frac{C\min(1,u)\tilde\sigma^4(1-\theta_\Delta)^2}{p_\Delta},
\end{align*}
where the first inequality follows from the preceding conditional Jensen argument with $k=2$,
the second inequality follows from $P_i(\Delta)\le C\min(1,u)$ and conditioning on $\mathbb I_i$, and
the last inequality follows from the Cauchy--Schwarz inequality, $\mathbb E|Y|^8\le C\|Y\|_{\psi_2}^8$, and the two  $\psi_2$-bounds.
Then, since $B_i-B_i^\circ=B_i\mathbf 1\{\Var_{\tilde\varepsilon_i}(\mathbb I_iW_i\mid\Delta,\tilde x_i)>L\}$, we have
\begin{align*}
&0\le\mathbb E_{\tilde x_i}[v^\top(B_i-B_i^\circ)v\mid\Delta]
\le\left(\mathbb E_{\tilde x_i}[(v^\top B_iv)^2\mid\Delta]\right)^{1/2}
\mathbb P_{\tilde x_i}\left(
\Var_{\tilde\varepsilon_i}(\mathbb I_iW_i\mid\Delta,\tilde x_i)>L
\middle|\Delta\right)^{1/2}\\
&\le C\tilde\sigma^2(1-\theta_\Delta)\sqrt{\frac{\min(1,u)}{p_\Delta}}
\mathbb P_{\tilde x_i}\left(
\Var_{\tilde\varepsilon_i}(\mathbb I_iW_i\mid\Delta,\tilde x_i)>L
\middle|\Delta\right)^{1/2}
\le\frac{c}{4}\tilde\sigma^2(1-\theta_\Delta),
\end{align*}
where the first two upper bounds follow from the Cauchy--Schwarz inequality and the preceding  bound, respectively.
The last follows from~\eqref{eq:common_conditional_variance_tail}, $mp_\Delta\ge1$ under~\eqref{eq:common_delta_empirical_sample}, and a sufficiently large constant in $L$.
Here, the constant $c$ is same as in the lower bound for $\mathbb E(B_i\mid\Delta)$.

Now, we verify the moment condition in Lemma~\ref{lem:bernstein}. By the definition of $B_i^\circ$ and following the similar steps,  for every integer $k\ge2$, it holds that
\begin{align*}
&\mathbb E_{\tilde x_i}[(v^\top B_i^\circ v)^k\mid\Delta]\\
&\quad\le L^{k-1}p_\Delta\mathbb E\left[
W_i^2\left(v^\top\bar G_{\mathrm{pop}}(\Delta)^{-1/2}\tilde x_{i,S}\right)^{2k}
\middle|\mathbb I_i=1,\Delta\right]
\le C^k k!\tilde\sigma^2(1-\theta_\Delta)
\left(\frac{L}{p_\Delta}\right)^{k-1},
\end{align*}
where the last inequality follows from the Cauchy--Schwarz inequality and the conditional $\psi_2$-bounds.
Moreover, $|x-y|^k\le2^{k-1}(|x|^k+|y|^k)$ and Jensen's inequality imply
\begin{align*}
\sum_{i=1}^m\mathbb E\left[
\left|v^\top\{B_i^\circ-\mathbb E(B_i^\circ\mid\Delta)\}v\right|^k
\middle|\Delta\right]
&\le\frac{k!}{2}\frac{Cm\tilde\sigma^2(1-\theta_\Delta)L}{p_\Delta}
\left(\frac{CL}{p_\Delta}\right)^{k-2}.
\end{align*}
For every fixed unit vector $v$, the variables $v^\top\{B_i^\circ-\mathbb E(B_i^\circ\mid\Delta)\}v$ are independent conditioned on $\Delta$.
Hence, we apply Lemma~\ref{lem:bernstein} and a union bound over a $1/4$-net of the unit sphere, as in the proof of Corollary~\ref{cor:common_empirical_gram}.
Then, with conditional failure probability at most $\eta/6$, we obtain
\begin{align*}
\left\|\sum_{i=1}^m\{B_i^\circ-\mathbb E(B_i^\circ\mid\Delta)\}\right\|_{\mathrm{op}}
&\le C\sqrt{\frac{m\tilde\sigma^2(1-\theta_\Delta)L(s+\log(m/\eta))}{p_\Delta}}
+\frac{CL(s+\log(m/\eta))}{p_\Delta}\\
&\le\frac{c}{4}m\tilde\sigma^2(1-\theta_\Delta).
\end{align*}
The last inequality follows from~\eqref{eq:common_delta_empirical_sample}, which implies $L(s+\log(m/\eta))/(mp_\Delta)\le c_0\tilde\sigma^2(1-\theta_\Delta)$ for a sufficiently small constant $c_0>0$.
Therefore, by the bound for $\mathbb E(B_i-B_i^\circ\mid\Delta)$ and the probability of truncation, we obtain
\begin{align*}
\left\|\sum_{i=1}^m
\{B_i-\mathbb E(B_i\mid\Delta)\}\right\|_{\mathrm{op}}
&\le\frac{c}{2}m\tilde\sigma^2(1-\theta_\Delta)
\end{align*}
with conditional failure probability at most $\eta/3$.

Next, we apply Corollaries~\ref{cor:common_empirical_gram} and
\ref{cor:common_empirical_first_moment} with failure probability $\eta/3$, which is permitted by~\eqref{eq:common_delta_empirical_sample} for sufficiently large $C$.
By the union bound, these conclusions hold simultaneously with conditional probability at least $1-\eta$.
Then, the definition of $B_i$ and~\eqref{eq:common_empirical_gram_comparison} imply
\begin{equation}\label{eq:common_empirical_population_covariance}
c\tilde\sigma^2(1-\theta_\Delta)\bar G(\Delta)
\preceq\sum_{i=1}^m\tilde x_{i,S}\tilde x_{i,S}^\top
\Var_{\tilde\varepsilon_i}(\mathbb I_iW_i\mid\Delta,\tilde x_i)
\preceq C\tilde\sigma^2(1-\theta_\Delta)\bar G(\Delta).
\end{equation}

To obtain the bounds for $M(\Delta)$ from~\eqref{eq:common_empirical_population_covariance}, let $t:=\bar G(\Delta)^{-1}\bar\mu(\Delta)-\bar G_{\mathrm{pop}}(\Delta)^{-1}\bar\mu_{\mathrm{pop}}(\Delta)$.
By Lemma~\ref{lem:common_population_mu} and the definition of $W_i$, we have
\begin{align*}
\mathbb I_i\left[
\tilde\varepsilon_i-\tilde x_{i,S}^\top
\left(\Delta_S+\bar G(\Delta)^{-1}\bar\mu(\Delta)\right)
\right]
&=\mathbb I_iW_i-\mathbb I_i\tilde x_{i,S}^\top t.
\end{align*}
Let $U_i=\mathbb I_iW_i$ and $V_i=-\mathbb I_i\tilde x_{i,S}^\top t$.
Then, by~\eqref{eq:M_def},
\begin{align*}
M(\Delta)&=\sum_{i=1}^m\tilde x_{i,S}\tilde x_{i,S}^\top
\Var_{\tilde\varepsilon_i}(U_i+V_i\mid\Delta,\tilde X).
\end{align*}
Since $t$ is determined by $(\Delta,\tilde X)$, we have $\Var_{\tilde\varepsilon_i}(V_i\mid\Delta,\tilde X)=P_i(\Delta)(1-P_i(\Delta))(\tilde x_{i,S}^\top t)^2$.
By~\eqref{eq:common_empirical_population_covariance}, the sums involving $\Var_{\tilde\varepsilon_i}(U_i\mid\Delta,\tilde X)$ satisfy the required trace bounds.

To bound $\tr(\Sigma_{1,SS}\bar G(\Delta)^{-1}M(\Delta)\bar G(\Delta)^{-1})$ and $\tr(\bar G(\Delta)^{-1}M(\Delta))$, we estimate the following sums involving $\Var_{\tilde\varepsilon_i}(V_i\mid\Delta,\tilde X)$:
\begin{align*}
\sum_{i=1}^m\Var_{\tilde\varepsilon_i}(V_i\mid\Delta,\tilde X)
\|\Sigma_{1,SS}^{1/2}\bar G(\Delta)^{-1}\tilde x_{i,S}\|_2^2
&\le\|\Sigma_{1,SS}^{1/2}\bar G(\Delta)^{-1/2}\|_{\mathrm{op}}^2
\ell_{\max}(\Delta)\|t\|_{\bar G(\Delta)}^2,\\
\sum_{i=1}^m\Var_{\tilde\varepsilon_i}(V_i\mid\Delta,\tilde X)
\tilde x_{i,S}^\top\bar G(\Delta)^{-1}\tilde x_{i,S}
&\le\ell_{\max}(\Delta)\|t\|_{\bar G(\Delta)}^2,
\end{align*}
where we have employed $\ell_{\max}(\Delta)=\max_{1\le i\le m}\tilde x_{i,S}^\top\bar G(\Delta)^{-1}\tilde x_{i,S}$, $P_i(\Delta)(1-P_i(\Delta))\le P_i(\Delta)$, and $\sum_iP_i(\Delta)(\tilde x_{i,S}^\top t)^2=\|t\|_{\bar G(\Delta)}^2$.
By combining Corollary~\ref{cor:common_empirical_first_moment}, \eqref{eq:common_delta_population_trace}, and~\eqref{eq:common_delta_empirical_sample}, it holds that $\|t\|_{\bar G(\Delta)}^2\le C\tilde\sigma^2(1-\theta_\Delta)s\log(m/\eta)$.
To express the first upper bound in terms of $\tr(\Sigma_{1,SS}\bar G(\Delta)^{-1})$, we apply Corollary~\ref{cor:common_empirical_gram}, \eqref{eq:common_delta_population_trace}, and~\eqref{eq:common_delta_population_operator} and obtain
\begin{align*}
\frac{\|\Sigma_{1,SS}^{1/2}\bar G(\Delta)^{-1/2}\|_{\mathrm{op}}^2}
{\tr(\Sigma_{1,SS}\bar G(\Delta)^{-1})}
&\le\frac{C\zeta_\Delta^2}{s-1+\zeta_\Delta^2}
\le\frac{C\min\bigl(s,\zeta_\Delta^2\bigr)}{s}.
\end{align*}
Moreover, by Corollary~\ref{cor:common_empirical_gram}, $\min(s,\zeta_\Delta^2)\ell_{\max}(\Delta)\log(m/\eta)$ is sufficiently small.
Hence, we obtain
\begin{align}\label{eq:common_empirical_covariance_correction}
\|\Sigma_{1,SS}^{1/2}\bar G(\Delta)^{-1/2}\|_{\mathrm{op}}^2
\ell_{\max}(\Delta)\|t\|_{\bar G(\Delta)}^2
&\le\frac{c}{4}\tilde\sigma^2(1-\theta_\Delta)
\tr(\Sigma_{1,SS}\bar G(\Delta)^{-1}),
\end{align}
and $\ell_{\max}(\Delta)\|t\|_{\bar G(\Delta)}^2\le C\tilde\sigma^2(1-\theta_\Delta)s$.

Finally, for every $i$, we have $\frac{1}{2}\Var_{\tilde\varepsilon_i}(U_i\mid\Delta,\tilde X)-\Var_{\tilde\varepsilon_i}(V_i\mid\Delta,\tilde X)\le\Var_{\tilde\varepsilon_i}(U_i+V_i\mid\Delta,\tilde X)\le2\Var_{\tilde\varepsilon_i}(U_i\mid\Delta,\tilde X)+2\Var_{\tilde\varepsilon_i}(V_i\mid\Delta,\tilde X)$.
We apply these inequalities and use~\eqref{eq:common_empirical_population_covariance} and~\eqref{eq:common_empirical_covariance_correction} to obtain the first two sided bound.
The second trace bound follows from~\eqref{eq:common_empirical_population_covariance} and $\ell_{\max}(\Delta)\|t\|_{\bar G(\Delta)}^2\le C\tilde\sigma^2(1-\theta_\Delta)s$.
\end{proof}

\subsection{Accepted Gram Matrix and Conditional Mean Bounds under Full Fine-Tuning}

In the preceding subsections,  we first have taken expectation over the
generation noise to obtain the conditional expected Gram matrix $\bar G(\Delta)
=\mathbb E_{\tilde\varepsilon}
[G(\Delta)\mid\Delta,\tilde X]$ and residual
numerator over the generation noise $\bar\mu(\Delta)
=\mathbb E_{\tilde\varepsilon}
[\mu(\Delta)\mid\Delta,\tilde X]$.
Then, we have compared $\bar G(\Delta)^{-1}\bar\mu(\Delta)$ with $\bar G_{\mathrm{pop}}(\Delta)^{-1}\bar\mu_{\mathrm{pop}}(\Delta)$.
These results provide the common bounds for $\mathcal V_1$ and $\mathcal V_2$ for every $u>0$.  
For full fine-tuning, in particular when $0<u\le1$, we apply the
accepted first moment bound at fixed $(\mathbb I_1,\ldots,\mathbb I_m)$.
Since $\xi_i=-\tilde x_i^\top\Delta+\tilde\varepsilon_i$ and
$\mathbb I_i=1$ is equivalent to $|\xi_i|\le\delta$,
Lemma~\ref{lem:accepted_first_moment_small_window}, applied with $a=-\tilde x_i^\top\Delta$, 
leads to
\begin{align*}
\left|\mathbb E_{\tilde\varepsilon_i}
(\xi_i\mid\mathbb I_i=1,\Delta,\tilde x_i)
+\frac{u^2}{3}\tilde x_i^\top\Delta\right|
&\le Cu^4\left(1+\frac{(\tilde x_i^\top\Delta)^2}{\tilde\sigma^2}\right)
|\tilde x_i^\top\Delta|.
\end{align*}
Hence, the right hand side can be small when $u$ is small. 
For the decomposition in this case, we consider
\begin{align*}
\mathbb E_{\tilde\varepsilon}
[r_{S,\mathrm{ext}}\mid\Delta,\tilde X]
&=\mathbb E_{\tilde\varepsilon}\left[
\mathbb E_{\tilde\varepsilon}
[r_{S,\mathrm{ext}}\mid\Delta,\tilde X,
\mathbb I_1,\ldots,\mathbb I_m]
\middle|\Delta,\tilde X\right].
\end{align*}
On $\Omega_{\mathrm{post}}$, $r_{S,\mathrm{ext}}=r_S$ and conditioned on
$(\Delta,\tilde X,\mathbb I_1,\ldots,\mathbb I_m)$, $G(\Delta)$ is fixed.
When $s=d$, the residual representation implies
\begin{align*}
\mathbb E_{\tilde\varepsilon}
[r_S\mid\Delta,\tilde X,\mathbb I_1,\ldots,\mathbb I_m]
&=G(\Delta)^{-1}\sum_{i=1}^m\mathbb I_i\tilde x_i
\mathbb E_{\tilde\varepsilon_i}
(\xi_i\mid\mathbb I_i=1,\Delta,\tilde x_i) \\
&=-\frac{u^2}{3}\Delta
+G(\Delta)^{-1}\sum_{i=1}^m\mathbb I_i\tilde x_i
\left(
\mathbb E_{\tilde\varepsilon_i}
(\xi_i\mid\mathbb I_i=1,\Delta,\tilde x_i)
+\frac{u^2}{3}\tilde x_i^\top\Delta
\right).
\end{align*}
Thus, conditioning on the acceptance indicators preserves the exact accepted
Gram cancellation under full fine-tuning.  In the following proposition, we
establish the accepted Gram matrix bounds.  It bounds the conditional residual
mean around $-u^2\Delta/3$ for $0<u\le1$ and around  $-(1-\theta_\Delta)\Delta$ 
when $u$ is bounded below.

\begin{proposition}[Accepted Gram matrix and conditional mean bounds under full fine-tuning]
\label{prop:full_accepted_gram_mean}
Under the model in Section~\ref{sec:problem_setting}, let $s=d$, $u>0$, and $0<\eta\le1/2$.
Fix $\Delta\in\mathbb R^d$ and suppose that for a sufficiently large constant $C>0$,
\begin{align*}
m p_\Delta
&\ge C\{d+\log(m/\eta)\}\log^2(m/\eta).
\end{align*}
Then, conditioned on $\Delta$, there exists an event
$\Omega_{\mathrm{full}}(\Delta)$, determined by
$(\tilde X,\mathbb I_1,\ldots,\mathbb I_m)$, such that
$\mathbb P_{\tilde X,\tilde\varepsilon}
(\Omega_{\mathrm{full}}(\Delta)\mid\Delta)\ge1-\eta$.
On this event,
\begin{align*}
c m\bar G_{\mathrm{pop}}(\Delta)
&\preceq G(\Delta)\preceq C m\bar G_{\mathrm{pop}}(\Delta).
\end{align*}
If $0<u\le1$, then, on
$\Omega_{\mathrm{full}}(\Delta)\cap\Omega_{\mathrm{post}}$,
\begin{align*}
\left\|\mathbb E_{\tilde\varepsilon}
[r_S\mid\Delta,\tilde X,\mathbb I_1,\ldots,\mathbb I_m]
+\frac{u^2}{3}\Delta\right\|_{\Sigma_1}
&\le Cu^4\|\Delta\|_{\Sigma_1}.
\end{align*}
Moreover, for every fixed $u_0>0$, suppose that $u\ge u_0$ and that the
constant $C$ in the sample condition is sufficiently large depending only on $u_0$.  
Then, on $\Omega_{\mathrm{full}}(\Delta)\cap
\Omega_{\mathrm{post}}$,
\begin{align*}
\left\|\mathbb E_{\tilde\varepsilon}
[r_S\mid\Delta,\tilde X,\mathbb I_1,\ldots,\mathbb I_m]
+(1-\theta_\Delta)\Delta\right\|_{\Sigma_1}
&\le\frac{1}{4}(1-\theta_\Delta)\|\Delta\|_{\Sigma_1}.
\end{align*}
\end{proposition}

\begin{proof}
We establish $cm\bar G_{\mathrm{pop}}(\Delta)\preceq G(\Delta)\preceq Cm\bar G_{\mathrm{pop}}(\Delta)$.
By Chernoff's inequality, $mp_\Delta/2\le\sum_i\mathbb I_i\le2mp_\Delta$ with failure probability at most $\eta/8$ conditioned on $\Delta$.
The covariates $\tilde x_i$ with $\mathbb I_i=1$ are independent and have mean zero conditioned on $(\Delta,\mathbb I_1,\ldots,\mathbb I_m)$.
By the definition of $\bar G_{\mathrm{pop}}(\Delta)$, we have $\mathbb E_{\tilde X}[G(\Delta)\mid\Delta,\mathbb I_1,\ldots,\mathbb I_m]=(\sum_i\mathbb I_i/p_\Delta)\bar G_{\mathrm{pop}}(\Delta)$.
It implies, for every unit vector $v$,
\begin{align}\label{eq:full_accepted_normalized_second_moment}
\mathbb E_{\tilde x_i}\left[
\left(v^\top p_\Delta^{1/2}
\bar G_{\mathrm{pop}}(\Delta)^{-1/2}\tilde x_i\right)^2
\middle|\mathbb I_i=1,\Delta\right]
&=1.
\end{align}
The proof of~\eqref{eq:common_empirical_gram_comparison} in Corollary~\ref{cor:common_empirical_gram}  remains valid after
replacing $P_i(\Delta)$  by $\mathbb I_i$ and setting $s=d$.
By the sample condition, we obtain $cm\bar G_{\mathrm{pop}}(\Delta)\preceq G(\Delta)\preceq Cm\bar G_{\mathrm{pop}}(\Delta)$ with failure probability at most $\eta/8$ over $(\tilde X,\tilde\varepsilon)$ conditioned on $\Delta$.
By combining  these bounds, we obtain, with probability at
least $1-\eta/4$,
\begin{align}\label{eq:full_accepted_count_covariance}
\frac{1}{2}mp_\Delta
&\le\sum_{i=1}^m\mathbb I_i\le2mp_\Delta,\qquad
cm\bar G_{\mathrm{pop}}(\Delta)\preceq G(\Delta)
\preceq Cm\bar G_{\mathrm{pop}}(\Delta).
\end{align}
We fix $(\mathbb I_1,\ldots,\mathbb I_m)$ satisfying $mp_\Delta/2\le\sum_i\mathbb I_i\le2mp_\Delta$ and establish the remaining bounds conditioned on these indicators and $\Delta$.
First, we establish the conditional mean bound for $0<u\le1$.
By Lemma~\ref{lem:accepted_first_moment_small_window}, with $a=-\tilde x_i^\top\Delta$, we have, for each accepted sample,
\begin{align*}
\left|\mathbb E_{\tilde\varepsilon_i}
(\xi_i\mid\mathbb I_i=1,\Delta,\tilde x_i)
+\frac{u^2}{3}\tilde x_i^\top\Delta\right|
&\le Cu^4\left(1+\frac{(\tilde x_i^\top\Delta)^2}
{\tilde\sigma^2}\right)|\tilde x_i^\top\Delta|.
\end{align*}
To bound the sum of these remainders after multiplication by $G(\Delta)^{-1}$, we establish
\begin{align}\label{eq:full_weighted_gram_comparison}
\sum_{i=1}^m\mathbb I_i
\left(1+\frac{(\tilde x_i^\top\Delta)^2}{\tilde\sigma^2}\right)
\tilde x_i\tilde x_i^\top
&\preceq C G(\Delta).
\end{align}
By~\eqref{eq:full_accepted_normalized_second_moment} and Lemma~\ref{lem:accepted_linear_form_moments}, we have $\|p_\Delta^{1/2}v^\top\bar G_{\mathrm{pop}}(\Delta)^{-1/2}\tilde x_i\|_{\psi_2}^2\le C$ for every unit vector $v$.
Since $0<u\le1$, the lemma also implies $\mathbb E_{\tilde x_i}[(\tilde x_i^\top\Delta)^2\mid\mathbb I_i=1,\Delta]/\tilde\sigma^2\le1+u^2\le2$, and hence $\|\tilde x_i^\top\Delta/\tilde\sigma\|_{\psi_2}^2\le C$.
Note that both norms are evaluated conditioned on $(\mathbb I_i=1,\Delta)$.
Then, it follows from the Cauchy--Schwarz inequality and $\mathbb E|Y|^{2k}\le C^k k!\|Y\|_{\psi_2}^{2k}$ that, for every integer $k\ge1$,
\begin{align*}
\mathbb E_{\tilde x_i}\left[
\left(1+\frac{(\tilde x_i^\top\Delta)^2}{\tilde\sigma^2}\right)
\left(v^\top p_\Delta^{1/2}\bar G_{\mathrm{pop}}(\Delta)^{-1/2}\tilde x_i\right)^{2k}
\middle|\mathbb I_i=1,\Delta\right]
&\le C^k k!.
\end{align*}
Moreover, the subgaussian bound for $\tilde x_i^\top\Delta/\tilde\sigma$ and a union bound imply that, with conditional failure probability at most $\eta/4$, the following inequality holds for every $i$ with $\mathbb I_i=1$:
\begin{align}\label{eq:full_accepted_row_bounds}
\frac{(\tilde x_i^\top\Delta)^2}{\tilde\sigma^2}
&\le C\log(m/\eta).
\end{align}
Next, for each $i$ with $\mathbb I_i=1$, we consider the following truncated expression:
\begin{align*}
\left(1+\frac{(\tilde x_i^\top\Delta)^2}{\tilde\sigma^2}\right)
\left(v^\top p_\Delta^{1/2}\bar G_{\mathrm{pop}}(\Delta)^{-1/2}\tilde x_i\right)^2
\mathbf 1\left\{\frac{(\tilde x_i^\top\Delta)^2}{\tilde\sigma^2}\le C\log(m/\eta)\right\}.
\end{align*}
The conditional mean is bounded by $C$, while the truncation and the preceding moment bound imply that the $k$th moment is at most $k!C^k\log^{k-1}(m/\eta)$.
After centering, the moment condition in Lemma~\ref{lem:bernstein} holds with $V=Cmp_\Delta\log(m/\eta)$ and $B=C\log(m/\eta)$.
Then, we apply Lemma~\ref{lem:bernstein} over a $1/4$-net of the unit sphere and use the conditional mean bound to obtain, with conditional failure probability at most $\eta/4$,
\begin{align*}
&\sum_{i=1}^m\mathbb I_i
\left(1+\frac{(\tilde x_i^\top\Delta)^2}{\tilde\sigma^2}\right)
p_\Delta\bar G_{\mathrm{pop}}(\Delta)^{-1/2}
\tilde x_i\tilde x_i^\top\bar G_{\mathrm{pop}}(\Delta)^{-1/2}
\mathbf 1\left\{\frac{(\tilde x_i^\top\Delta)^2}{\tilde\sigma^2}\le C\log(m/\eta)\right\}\\
&\quad\preceq C\left(mp_\Delta+\sqrt{mp_\Delta\{d+\log(m/\eta)\}\log(m/\eta)}
+ \{d+\log(m/\eta)\}\log(m/\eta)\right)I_d\\
&\quad\preceq Cmp_\Delta I_d,
\end{align*}
where the last inequality follows from $mp_\Delta\ge C\{d+\log(m/\eta)\}\log^2(m/\eta)$.
Hence, by~\eqref{eq:full_accepted_row_bounds}, we obtain, with conditional failure probability at most $\eta/2$,
\begin{align}\label{eq:full_weighted_sum_concentration}
\sum_{i=1}^m\mathbb I_i
\left(1+\frac{(\tilde x_i^\top\Delta)^2}{\tilde\sigma^2}\right)
p_\Delta\bar G_{\mathrm{pop}}(\Delta)^{-1/2}
\tilde x_i\tilde x_i^\top\bar G_{\mathrm{pop}}(\Delta)^{-1/2}
&\preceq Cmp_\Delta I_d.
\end{align}
Then, we combine~\eqref{eq:full_accepted_count_covariance} and~\eqref{eq:full_weighted_sum_concentration} to obtain~\eqref{eq:full_weighted_gram_comparison}.

Now, we apply the Cauchy--Schwarz inequality
 and~\eqref{eq:full_weighted_gram_comparison} to the Taylor remainder bound and obtain, for every $v\in\mathbb R^d$,
\begin{align*}
&\left|\sum_{i=1}^m\mathbb I_i(v^\top\tilde x_i)
\left(\mathbb E_{\tilde\varepsilon_i}
(\xi_i\mid\mathbb I_i=1,\Delta,\tilde x_i)
+\frac{u^2}{3}\tilde x_i^\top\Delta\right)\right|\\
&\quad\le Cu^4
\left(\sum_{i=1}^m\mathbb I_i
\left(1+\frac{(\tilde x_i^\top\Delta)^2}{\tilde\sigma^2}\right)
(v^\top\tilde x_i)^2\right)^{1/2}
\left(\sum_{i=1}^m\mathbb I_i
\left(1+\frac{(\tilde x_i^\top\Delta)^2}{\tilde\sigma^2}\right)
(\tilde x_i^\top\Delta)^2\right)^{1/2}\\
&\quad\le Cu^4(v^\top G(\Delta)v)^{1/2}
\|G(\Delta)^{1/2}\Delta\|.
\end{align*}
With $v=G(\Delta)^{-1/2}z$, we take the supremum over $\|z\|=1$ and obtain
\begin{align}\label{eq:full_small_u_normalized_numerator}
\left\|G(\Delta)^{-1/2}\sum_{i=1}^m\mathbb I_i\tilde x_i
\left(\mathbb E_{\tilde\varepsilon_i}
(\xi_i\mid\mathbb I_i=1,\Delta,\tilde x_i)
+\frac{u^2}{3}\tilde x_i^\top\Delta\right)\right\|
&\le Cu^4\|G(\Delta)^{1/2}\Delta\|.
\end{align}
Next, we apply~\eqref{eq:full_small_u_normalized_numerator} to bound the conditional mean of $r_S$ in the $\Sigma_1$ norm.
On $\Omega_{\mathrm{post}}$, the residual representation implies
\begin{align*}
\mathbb E_{\tilde\varepsilon}
[r_S\mid\Delta,\tilde X,\mathbb I_1,\ldots,\mathbb I_m]
+\frac{u^2}{3}\Delta
&=G(\Delta)^{-1}\sum_{i=1}^m\mathbb I_i\tilde x_i
\left(\mathbb E_{\tilde\varepsilon_i}
(\xi_i\mid\mathbb I_i=1,\Delta,\tilde x_i)
+\frac{u^2}{3}\tilde x_i^\top\Delta\right).
\end{align*}
It follows from~\eqref{eq:full_small_u_normalized_numerator} that
\begin{align*}
\left\|\mathbb E_{\tilde\varepsilon}
[r_S\mid\Delta,\tilde X,\mathbb I_1,\ldots,\mathbb I_m]
+\frac{u^2}{3}\Delta\right\|_{\Sigma_1}
&\le Cu^4
\|\Sigma_1^{1/2}G(\Delta)^{-1/2}\|_{\mathrm{op}}
\|G(\Delta)^{1/2}\Delta\|.
\end{align*}
For the two norms on the right hand side, Lemma~\ref{lem:common_population_covariance} and~\eqref{eq:full_accepted_count_covariance} imply
\begin{align*}
\|\Sigma_1^{1/2}G(\Delta)^{-1/2}\|_{\mathrm{op}}^2
&\le\frac{C}{mp_\Delta}
\frac{\tilde\sigma^2+\|\Delta\|_{\Sigma_1}^2}
{\tilde\sigma^2+\lambda_{2,\Delta}\|\Delta\|_{\Sigma_1}^2},
&
\|G(\Delta)^{1/2}\Delta\|^2
&\le Cmp_\Delta
\frac{\tilde\sigma^2+\lambda_{2,\Delta}\|\Delta\|_{\Sigma_1}^2}
{\tilde\sigma^2+\|\Delta\|_{\Sigma_1}^2}
\|\Delta\|_{\Sigma_1}^2.
\end{align*}
These bounds imply $\|\Sigma_1^{1/2}G(\Delta)^{-1/2}\|_{\mathrm{op}}\|G(\Delta)^{1/2}\Delta\|\le C\|\Delta\|_{\Sigma_1}$.
Hence, we obtain
\begin{align*}
\left\|\mathbb E_{\tilde\varepsilon}
[r_S\mid\Delta,\tilde X,\mathbb I_1,\ldots,\mathbb I_m]
+\frac{u^2}{3}\Delta\right\|_{\Sigma_1}
&\le Cu^4\|\Delta\|_{\Sigma_1}.
\end{align*}

For the second conditional mean bound, we first estimate the difference from $-(1-\theta_\Delta)\Delta$ for every $u>0$.
On $\Omega_{\mathrm{post}}$, the residual representation implies
\begin{align}\label{eq:full_centered_residual_representation}
\mathbb E_{\tilde\varepsilon}
[r_S\mid\Delta,\tilde X,\mathbb I_1,\ldots,\mathbb I_m]
+(1-\theta_\Delta)\Delta
&=G(\Delta)^{-1}\sum_{i:\mathbb I_i=1}\tilde x_i
\left(
\mathbb E_{\tilde\varepsilon_i}
[\xi_i\mid\mathbb I_i=1,\Delta,\tilde x_i]
+(1-\theta_\Delta)\tilde x_i^\top\Delta
\right).
\end{align}
For $s=d$, Lemma~\ref{lem:common_population_mu} implies
\begin{align}\label{eq:full_centered_population_mean_zero}
p_\Delta\mathbb E_{\tilde x_i}\left[
\tilde x_i\left(
\mathbb E_{\tilde\varepsilon_i}
[\xi_i\mid\mathbb I_i=1,\Delta,\tilde x_i]
+(1-\theta_\Delta)\tilde x_i^\top\Delta
\right)
\middle|\mathbb I_i=1,\Delta\right]
&=\bar\mu_{\mathrm{pop}}(\Delta)
+(1-\theta_\Delta)\bar G_{\mathrm{pop}}(\Delta)\Delta=0.
\end{align}
The terms of the sum in~\eqref{eq:full_centered_residual_representation} are independent conditioned on $(\Delta,\mathbb I_1,\ldots,\mathbb I_m)$ and have mean zero by~\eqref{eq:full_centered_population_mean_zero}.

We bound the sum in~\eqref{eq:full_centered_residual_representation} using Lemma~\ref{lem:bernstein}. First, we bound $\mathbb E_{\tilde\varepsilon_i}[\xi_i\mid\mathbb I_i=1,\Delta,\tilde x_i]$.
For each fixed $\tilde x_i$, we have $\xi_i\mid(\Delta,\tilde x_i)\sim\mathcal N(-\tilde x_i^\top\Delta,\tilde\sigma^2)$, and $\mathbb I_i=1$ restricts $\xi_i$ to $[-\delta,\delta]$.
Symmetry and monotonicity of the Gaussian density imply $\min(0,-\tilde x_i^\top\Delta)\le\mathbb E_{\tilde\varepsilon_i}[\xi_i\mid\mathbb I_i=1,\Delta,\tilde x_i]\le\max(0,-\tilde x_i^\top\Delta)$.
Since $0\le1-\theta_\Delta\le1$, it follows that
\begin{align}\label{eq:full_centered_score_pointwise}
\left|
\mathbb E_{\tilde\varepsilon_i}
[\xi_i\mid\mathbb I_i=1,\Delta,\tilde x_i]
+(1-\theta_\Delta)\tilde x_i^\top\Delta
\right|
&\le|\tilde x_i^\top\Delta|.
\end{align}
To apply Lemma~\ref{lem:bernstein}, we consider the sum in~\eqref{eq:full_centered_residual_representation} after multiplication by $p_\Delta^{1/2}\bar G_{\mathrm{pop}}(\Delta)^{-1/2}$.
By~\eqref{eq:full_accepted_normalized_second_moment}, we have $\mathbb E_{\tilde x_i}[(v^\top p_\Delta^{1/2}\bar G_{\mathrm{pop}}(\Delta)^{-1/2}\tilde x_i)^2\mid\mathbb I_i=1,\Delta]=1$ for every unit vector $v$.
Hence, Lemma~\ref{lem:accepted_linear_form_moments} implies $\|v^\top p_\Delta^{1/2}\bar G_{\mathrm{pop}}(\Delta)^{-1/2}\tilde x_i\|_{\psi_2}^2\le C$ and $\|\tilde x_i^\top\Delta\|_{\psi_2}^2\le C\Delta^\top p_\Delta^{-1}\bar G_{\mathrm{pop}}(\Delta)\Delta$, where both norms are evaluated conditioned on $(\mathbb I_i=1,\Delta)$.
By~\eqref{eq:full_centered_score_pointwise}, it suffices to bound the conditional second moment and the $\psi_1$-norm of $\bigl(v^\top p_\Delta^{1/2}\bar G_{\mathrm{pop}}(\Delta)^{-1/2}\tilde x_i\bigr)(\tilde x_i^\top\Delta)$.
Then, we apply the Cauchy--Schwarz inequality
 and $\mathbb E|Y|^4\le C\|Y\|_{\psi_2}^4$, with $\sum_i\mathbb I_i\le2mp_\Delta$, to obtain
\begin{align}\label{eq:full_centered_numerator_second_moment}
\sum_{i:\mathbb I_i=1}\mathbb E_{\tilde x_i}\left[
\left(v^\top p_\Delta^{1/2}\bar G_{\mathrm{pop}}(\Delta)^{-1/2}\tilde x_i\right)^2
(\tilde x_i^\top\Delta)^2
\middle|\mathbb I_i=1,\Delta\right]
&\le Cm\Delta^\top\bar G_{\mathrm{pop}}(\Delta)\Delta.
\end{align}
Moreover, the product inequality $\|UV\|_{\psi_1}\le C\|U\|_{\psi_2}\|V\|_{\psi_2}$ results in
\begin{align}\label{eq:full_centered_numerator_orlicz}
\left\|
\left(v^\top p_\Delta^{1/2}\bar G_{\mathrm{pop}}(\Delta)^{-1/2}\tilde x_i\right)
(\tilde x_i^\top\Delta)\right\|_{\psi_1}^2
&\le C\Delta^\top p_\Delta^{-1}\bar G_{\mathrm{pop}}(\Delta)\Delta.
\end{align}
By~\eqref{eq:full_centered_score_pointwise}, the bounds in~\eqref{eq:full_centered_numerator_second_moment} and~\eqref{eq:full_centered_numerator_orlicz} also hold with $\tilde x_i^\top\Delta$ replaced by $\mathbb E_{\tilde\varepsilon_i}[\xi_i\mid\mathbb I_i=1,\Delta,\tilde x_i]+(1-\theta_\Delta)\tilde x_i^\top\Delta$.
When $\Delta=0$, the sum is zero by~\eqref{eq:full_centered_score_pointwise}.
For $\Delta\ne0$, \eqref{eq:full_centered_numerator_orlicz} implies the moment condition in Lemma~\ref{lem:bernstein} with $V=Cm\Delta^\top\bar G_{\mathrm{pop}}(\Delta)\Delta$ and $B=C(\Delta^\top p_\Delta^{-1}\bar G_{\mathrm{pop}}(\Delta)\Delta)^{1/2}$.
Then, we apply Lemma~\ref{lem:bernstein} over a $1/2$-net of the unit sphere and obtain, with conditional failure probability at most $\eta/4$,
\begin{align}\label{eq:full_centered_numerator_concentration}
&\left\|\sum_{i:\mathbb I_i=1}p_\Delta^{1/2}
\bar G_{\mathrm{pop}}(\Delta)^{-1/2}\tilde x_i
\left(
\mathbb E_{\tilde\varepsilon_i}
[\xi_i\mid\mathbb I_i=1,\Delta,\tilde x_i]
+(1-\theta_\Delta)\tilde x_i^\top\Delta
\right)\right\|\\
&\quad\le C\left(
\frac{\tilde\sigma^2+\lambda_{2,\Delta}\|\Delta\|_{\Sigma_1}^2}
{\tilde\sigma^2+\|\Delta\|_{\Sigma_1}^2}
\right)^{1/2}\|\Delta\|_{\Sigma_1}
\left(\sqrt{mp_\Delta\{d+\log(m/\eta)\}}
+d+\log(m/\eta)\right),
\end{align}
where $\Delta^\top p_\Delta^{-1}\bar G_{\mathrm{pop}}(\Delta)\Delta=(\tilde\sigma^2+\lambda_{2,\Delta}\|\Delta\|_{\Sigma_1}^2)\|\Delta\|_{\Sigma_1}^2/(\tilde\sigma^2+\|\Delta\|_{\Sigma_1}^2)$ by Lemma~\ref{lem:common_population_covariance}.
Next, we bound the difference in~\eqref{eq:full_centered_residual_representation} in the $\Sigma_1$ norm. The lower bound in~\eqref{eq:full_accepted_count_covariance} and Lemma~\ref{lem:common_population_covariance} imply
\begin{align}\label{eq:full_centered_inverse_comparison}
\|\Sigma_1^{1/2}G(\Delta)^{-1}p_\Delta^{-1/2}
\bar G_{\mathrm{pop}}(\Delta)^{1/2}\|_{\mathrm{op}}
&\le\frac{C}{mp_\Delta}\left(
\frac{\tilde\sigma^2+\|\Delta\|_{\Sigma_1}^2}
{\tilde\sigma^2+\lambda_{2,\Delta}\|\Delta\|_{\Sigma_1}^2}
\right)^{1/2}.
\end{align}
Then, we apply~\eqref{eq:full_centered_numerator_concentration} and~\eqref{eq:full_centered_inverse_comparison} to~\eqref{eq:full_centered_residual_representation} and obtain
\begin{align*}
\left\|\mathbb E_{\tilde\varepsilon}
[r_S\mid\Delta,\tilde X,\mathbb I_1,\ldots,\mathbb I_m]
+(1-\theta_\Delta)\Delta\right\|_{\Sigma_1}
&\le C\left(
\sqrt{\frac{d+\log(m/\eta)}{mp_\Delta}}
+\frac{d+\log(m/\eta)}{mp_\Delta}\right)
\|\Delta\|_{\Sigma_1}.
\end{align*}
For every fixed $u_0>0$ and $u\ge u_0$, Lemma~\ref{lem:common_population_theta_bounds} implies $1-\theta_\Delta\ge c(u_0)>0$.
The sample condition $mp_\Delta\ge C\{d+\log(m/\eta)\}\log^2(m/\eta)$, with $C$ sufficiently large depending only on $u_0$, makes the coefficient on the right hand side at most $(1-\theta_\Delta)/4$ and proves the second conditional mean bound.

Finally, let $\Omega_{\mathrm{full}}(\Delta)$ be the intersection of the
events in~\eqref{eq:full_accepted_count_covariance} and
\eqref{eq:full_centered_numerator_concentration} and, when $0<u\le1$, the event in~\eqref{eq:full_weighted_sum_concentration}.
This event is determined by $(\tilde X,\mathbb I_1,\ldots,\mathbb I_m)$.
By the preceding failure probability bounds, we take expectation over $(\mathbb I_1,\ldots,\mathbb I_m)$ and apply the union bound to obtain $\mathbb P_{\tilde X,\tilde\varepsilon}(\Omega_{\mathrm{full}}(\Delta)\mid\Delta)\ge1-\eta$.
\end{proof}

%% file: Unified_appendix/app_09aG_conditional_risk_mean.tex

\section{Variance Components for the Realized Synthetic Covariates}
\label{sec:common_conditional_risk}

In Section~\ref{sec:common_empirical}, we have established the empirical
bounds over $\tilde X$ for each fixed $\Delta$.  However, the variance
components $\mathcal V_1$ and $\mathcal V_2$ are evaluated at the realized $\tilde X$, with expectation taken
over the pre-training error.  We fix $X_{\mathrm{pre}}\in\Omega_{\mathrm{pre}}$, and expectations over $\Delta$ are conditioned on $X_{\mathrm{pre}}$.
First, we construct an event over $\tilde X$ with probability at least $1-\eta$.
For every realization of $\tilde X$ in this event, the empirical bounds hold
with conditional probability at least $1-\eta$ over $\Delta$.
Then, we derive the partial fine-tuning $\mathcal V_1$ rate, establish the full fine-tuning
conditional mean bounds used in Section~\ref{sec:full_V1_common}, and obtain the $\mathcal V_2$ rate for every $u>0$.  As in
Section~\ref{sec:common_var_framework}, an expression containing an inverse
and multiplied by an event indicator is evaluated on that event and
interpreted as zero outside it.

\subsection{A Common Event for the Realized Covariates}

The sample condition in Section~\ref{sec:common_empirical} depends on $p_\Delta$ and $\zeta_\Delta^2$.
First, we obtain a condition that does not depend on the realization of $\Delta$ and ensures that the sample condition in Corollary~\ref{cor:common_empirical_gram} holds with high probability over $\Delta$.
Recall that $\rho_\Delta^2$ measures
the total pre-training prediction error relative to the generation noise
variance, while $\rho_{S,\Delta}^2$ is its active coordinate component.
For $0<\eta\le1/2$, we exploit
Corollary~\ref{cor:pretraining_error_concentration} and obtain, with
probability at least $1-\eta$ over $\Delta$,
\begin{equation}\label{eq:common_empirical_error_bounds}
\rho_\Delta^2\le
\frac{C\{d+\log(m/\eta)\}}{n\tilde\sigma^2},\qquad
\rho_{S,\Delta}^2\le
\frac{C\{s+\log(m/\eta)\}}{n\tilde\sigma^2}.
\end{equation}
Recall that $\bar\rho^2=\mathbb E_\Delta[\rho_\Delta^2\mid X_{\mathrm{pre}}]$.
To compare $p_\Delta$ and $\zeta_\Delta^2$
with quantities determined by $X_{\mathrm{pre}}$, define
\begin{equation}\label{eq:common_conditional_scales}
p_\star:=\min\bigg(1,\frac{u}{\sqrt{1+\bar\rho^2}}\bigg),\qquad
\zeta_\star^2:=1+\min\bigg(\bar\rho^2,
\frac{1+\bar\rho^2}{u^2}\bigg).
\end{equation}
Since $X_{\mathrm{pre}}$ is fixed, these quantities do not depend on the
realized $\Delta$ or $\tilde X$.
The definitions of $\rho_{S,\Delta}^2$ and $u$ imply
$\|(\Sigma_1\Delta)_S\|_{\Sigma_{1,SS}^{-1}}^2/
(\tilde\sigma^2+\|\Delta_{-S}\|_{\Sigma_{1,-S\mid S}}^2)
\le\rho_{S,\Delta}^2$ and $\sigma_\xi^2(\Delta)/\delta^2=(1+\rho_\Delta^2)/u^2$.
For every $\Delta$ and $u>0$, \eqref{eq:zeta_delta_def} and
\eqref{eq:common_population_inverse_factor_scale} lead to
\begin{equation}\label{eq:common_population_inverse_factor_bound}
\zeta_\Delta^2
\le C\left[1+\min\left(\rho_{S,\Delta}^2,
\frac{1+\rho_\Delta^2}{u^2}\right)\right].
\end{equation}
We fix any $\Delta$ satisfying~\eqref{eq:common_empirical_error_bounds}.
The decomposition $\rho_\Delta^2=\rho_{S,\Delta}^2+ \|\Delta_{-S}\|_{\Sigma_{1,-S\mid S}}^2/\tilde\sigma^2$ 
results in $\rho_{S,\Delta}^2\le\rho_\Delta^2$.  The relation
$\bar\rho^2\asymp d/(n\tilde\sigma^2)$ and \eqref{eq:common_empirical_error_bounds} imply
$1+\rho_\Delta^2\le C(1+\bar\rho^2)\log(m/\eta)$. By combining this bound with
$\rho_{S,\Delta}^2\le\rho_\Delta^2$ and the definition of $\zeta_\star^2$, we have
\begin{align*}
1+\min(\rho_{S,\Delta}^2,(1+\rho_\Delta^2)/u^2)
&\le1+(1+\rho_\Delta^2)\min(1,u^{-2})
\le C\zeta_\star^2\log(m/\eta).
\end{align*}
Hence, by~\eqref{eq:common_population_inverse_factor_bound}, we obtain $\zeta_\Delta^2\le C\zeta_\star^2\log(m/\eta)$.
Then, by~\eqref{eq:marginal_acceptance_scale}, \eqref{eq:common_delta_population_trace}, and~\eqref{eq:common_delta_population_operator}, we obtain
\begin{align}
p_\Delta&\ge
\frac{c p_\star}{\sqrt{\log(m/\eta)}},
\label{eq:common_empirical_acceptance}\\
\tr(\Sigma_{1,SS}\bar G_{\mathrm{pop}}(\Delta)^{-1})
&\le
\frac{C\sqrt{\log(m/\eta)}}{p_\star}
\{s+\zeta_\star^2\log(m/\eta)\},
\label{eq:common_empirical_population_trace}\\
\left\|\Sigma_{1,SS}^{1/2}\bar G_{\mathrm{pop}}(\Delta)^{-1}
\Sigma_{1,SS}^{1/2}\right\|_{\mathrm{op}}
&\le
\frac{C\zeta_\star^2\{\log(m/\eta)\}^{3/2}}{p_\star}.
\label{eq:common_empirical_population_operator}
\end{align}
Suppose that the following common sample condition holds for a sufficiently large constant $C$:
\begin{align}\label{eq:common_empirical_sample}
m p_\star
\ge C\zeta_\star^2\bigl(s+\log(m/\eta)\bigr)
\log^{7/2}(m/\eta).
\end{align}
By \eqref{eq:common_empirical_acceptance} and the common sample condition~\eqref{eq:common_empirical_sample}, we have
\begin{align*}
m p_\Delta&\ge\frac{c m p_\star}{\sqrt{\log(m/\eta)}}
\ge C\zeta_\star^2\bigl(s+\log(m/\eta)\bigr)\log^3(m/\eta)
\ge C\zeta_\Delta^2\bigl(s+\log(m/\eta)\bigr)\log^2(m/\eta).
\end{align*}
Therefore, the common sample condition ensures that~\eqref{eq:common_delta_empirical_sample} holds with probability at least $1-\eta$ over $\Delta$.

We apply Lemma~\ref{lem:fixed_covariate_conversion} to convert the
fixed $\Delta$ bounds in Section~\ref{sec:common_empirical} into
bounds at the realized $\tilde X$.
For $X_{\mathrm{pre}}\in\Omega_{\mathrm{pre}}$, $u>0$, and $0<\eta\le1/2$, 
let $\Omega_{\mathrm{emp}}(\Delta)$ be the event, determined by $(\Delta,\tilde X)$, on which
\eqref{eq:common_empirical_error_bounds}, 
Corollary~\ref{cor:common_empirical_gram}, and
Lemmas~\ref{lem:common_empirical_prediction_mean}
and~\ref{lem:common_empirical_covariance}  hold simultaneously, with $\eta$
replaced by $\eta^2/12$.  For this choice of $\Omega_{\mathrm{emp}}(\Delta)$, 
define $\Omega(\Delta)$ by~\eqref{eq:common_full_event_def}.
On $\Omega_{\mathrm{emp}}(\Delta)$, we have $\mathbb P_{\tilde\varepsilon}(\Omega(\Delta)^{\mathrm c}\mid\Delta,\tilde X)\le\eta$ by~\eqref{eq:common_full_event_probability}.

\begin{corollary}[Empirical bounds at the realized covariates]
\label{cor:common_realized_covariate_event}
Under the model in Section~\ref{sec:problem_setting} and
Assumption~\ref{ass:design}, fix $C_G\ge2$ as in
Definition~\ref{def:ext_estimator}, let $X_{\mathrm{pre}}\in\Omega_{\mathrm{pre}}$, 
and take $u>0$ and $0<\eta\le1/2$.
Suppose that \eqref{eq:common_empirical_sample} holds with $\eta$ replaced by
$\eta^2$ and $C$ sufficiently large.  Then, conditioned on $X_{\mathrm{pre}}$, 
with probability at least $1-\eta$ over $\tilde X$,
\begin{align*}
\mathbb P_\Delta\left(
\Omega_{\mathrm{emp}}(\Delta)^{\mathrm c}\mid
X_{\mathrm{pre}},\tilde X\right)&\le\eta,
&
\max_{1\le i\le m}\|\tilde x_i\|^2
&\le C\{d+\log(m/\eta)\}.
\end{align*}
On this event,
$\mathbb P_{\Delta,\tilde\varepsilon}(\Omega(\Delta)^{\mathrm c}\mid
X_{\mathrm{pre}},\tilde X)\le2\eta$.
In addition, $\Omega(\Delta)\subseteq\Omega_{\mathrm{post}}$.
\end{corollary}

\begin{proof}
Let $\Omega_\Delta$ be the event in
Corollary~\ref{cor:pretraining_error_concentration} with failure probability $\eta/2$.  
For every $\Delta\in\Omega_\Delta$,
\eqref{eq:common_empirical_error_bounds} holds with $\eta$ replaced by $\eta^2/12$.  
The sample condition with sufficiently large $C$ ensures that the conditions of
Corollary~\ref{cor:common_empirical_gram} and
Lemmas~\ref{lem:common_empirical_prediction_mean}
and~\ref{lem:common_empirical_covariance} hold, each with failure probability $\eta^2/12$.  
We employ these three results and the union bound to obtain $\mathbb P_{\tilde X}(\Omega_{\mathrm{emp}}(\Delta)^{\mathrm c}\mid X_{\mathrm{pre}},\Delta)\le\eta^2/4$.
Since $\Delta$ and $\tilde X$ are independent conditioned on $X_{\mathrm{pre}}$, we apply Lemma~\ref{lem:fixed_covariate_conversion} to $\Omega_\Delta$ and $\Omega_{\mathrm{emp}}(\Delta)$ and obtain $\mathbb P_\Delta(\Omega_{\mathrm{emp}}(\Delta)^{\mathrm c}\mid X_{\mathrm{pre}},\tilde X)\le\eta$ with failure probability at most $\eta/2$ over $\tilde X$.
The maximum covariate bound follows from Lemma~\ref{lem:gaussian_max_design} with failure probability $\eta/2$.
Hence, both statements hold with probability at least $1-\eta$ over $\tilde X$.

On $\Omega_{\mathrm{emp}}(\Delta)$, it follows from Corollary~\ref{cor:common_empirical_gram} that the conditions in~\eqref{eq:common_accepted_gram_conditions} hold.
Hence, by~\eqref{eq:common_full_event_probability}, we have $\mathbb P_{\tilde\varepsilon}(\Omega(\Delta)^{\mathrm c}\mid\Delta,\tilde X)\le\eta$.
Since
$\Omega(\Delta)\subseteq\Omega_{\mathrm{emp}}(\Delta)$ by definition, the tower property results in
\begin{align*}
\mathbb P_{\Delta,\tilde\varepsilon}(\Omega(\Delta)^{\mathrm c}\mid X_{\mathrm{pre}},\tilde X)
&=\mathbb P_\Delta(\Omega_{\mathrm{emp}}(\Delta)^{\mathrm c}\mid X_{\mathrm{pre}},\tilde X)
+\mathbb E_\Delta\left[\mathbf 1_{\Omega_{\mathrm{emp}}(\Delta)}
\mathbb P_{\tilde\varepsilon}(\Omega(\Delta)^{\mathrm c}\mid\Delta,\tilde X)
\mid X_{\mathrm{pre}},\tilde X\right]\le2\eta.
\end{align*}

Finally, on $\Omega(\Delta)$, the bound $\|H(\Delta)\|_{\mathrm{op}}\le1/2$
and the lower Gram matrix bound imply $G(\Delta)/m\succeq\bar G(\Delta)/(2m) \succeq\bar G_{\mathrm{pop}}(\Delta)/4$.
Then it follows from
\eqref{eq:common_empirical_population_operator},
Assumption~\ref{ass:design}, and the sample condition that 
\begin{align*}
\lambda_{\min}\{G(\Delta)/m\}
&\ge\frac{c p_\star}{\zeta_\star^2\{\log(m/\eta^2)\}^{3/2}}
\ge m^{-1}\ge m^{-C_G}.
\end{align*}
Therefore, $\Omega(\Delta)\subseteq\Omega_{\mathrm{post}}$, which proves the
corollary.
\end{proof}

\subsection{Remainder Bounds at the Realized Covariates}

We bound $R_{\mathrm{res}}^\Omega$ and the contributions on $\Omega(\Delta)^{\mathrm c}$ in~\eqref{eq:common_V1_remainder_decomposition} and~\eqref{eq:V2_common_comparison}.
The inverse remainder is bounded using the empirical results on $\Omega_{\mathrm{emp}}(\Delta)$.
For the complement contributions, we combine the conditional probability bound in Corollary~\ref{cor:common_realized_covariate_event} with moment bounds.
Recall from~\eqref{eq:Rres_common_def} that
\begin{align*}
R_{\mathrm{res}}^\Omega
&:=\mathbb E_{\Delta,\tilde\varepsilon}\left[
\left\|\left(G(\Delta)^{-1}-\bar G(\Delta)^{-1}\right)
\left(\mu(\Delta)-G(\Delta)\bar G(\Delta)^{-1}\bar\mu(\Delta)\right)
\right\|_{\Sigma_{1,SS}}^2\mathbf 1_{\Omega(\Delta)}
\middle|X_{\mathrm{pre}},\tilde X\right].
\end{align*}

\begin{corollary}[Inverse remainder at the realized covariates]
\label{cor:common_centered_inverse_rate}
Under the model in Section~\ref{sec:problem_setting} and
Assumption~\ref{ass:design}, let $X_{\mathrm{pre}}\in\Omega_{\mathrm{pre}}$, $u>0$, and $0<\eta\le1/2$, 
and let $p_\star$ and $\zeta_\star^2$ be as in
\eqref{eq:common_conditional_scales}.  Suppose that the sample condition
\eqref{eq:common_empirical_sample}, with $\eta$ replaced by $\eta^2$,
holds with $C$ sufficiently large.  Then, conditioned on $X_{\mathrm{pre}}$, 
the following bound holds almost surely over $\tilde X$:
\begin{align*}
R_{\mathrm{res}}^\Omega
&\le C\frac{s\zeta_\star^2
\left\{s+\zeta_\star^2\bigl(\log(m/\eta)\bigr)^2\right\}
\bigl(\log(m/\eta)\bigr)^3}
{(m p_\star)^2}
\left(
\min(\delta^2,\tilde\sigma^2)
+\frac{\tilde\sigma^2(d-s)}{n\tilde\sigma^2+d-s}\right).
\end{align*}
\end{corollary}

\begin{proof}
On $\Omega_{\mathrm{emp}}(\Delta)$, the empirical conclusions hold with $\eta$ replaced by $\eta^2/12$.  
Since $\log(12m/\eta^2)\le C\log(m/\eta)$, 
Corollary~\ref{cor:common_empirical_gram}, Lemma~\ref{lem:common_empirical_covariance},
\eqref{eq:common_empirical_acceptance}, and $\zeta_\Delta^2\le C\zeta_\star^2\log(m/\eta)$ imply that 
$\ell_{\max}(\Delta)\le C\left\{s+\zeta_\star^2\bigl(\log(m/\eta)\bigr)^2\right\}\sqrt{\log(m/\eta)}/(m p_\star)$,
$\|\Sigma_{1,SS}^{1/2}\bar G(\Delta)^{-1/2}\|_{\mathrm{op}}^2\le C\zeta_\star^2\bigl(\log(m/\eta)\bigr)^{3/2}/(m p_\star)$, and $\tr(\bar G(\Delta)^{-1}M(\Delta))\le Cs\tilde\sigma^2(1-\theta_\Delta)$.
We substitute these three bounds into
Lemma~\ref{lem:common_centered_inverse} and obtain
\begin{align*}
R_{\mathrm{res}}^\Omega
&\le C\frac{s\zeta_\star^2
\left\{s+\zeta_\star^2\bigl(\log(m/\eta)\bigr)^2\right\}
\bigl(\log(m/\eta)\bigr)^3}
{(m p_\star)^2}
\mathbb E_\Delta\left[
\tilde\sigma^2(1-\theta_\Delta)
\mathbf 1_{\Omega_{\mathrm{emp}}(\Delta)}
\middle|X_{\mathrm{pre}},\tilde X\right].
\end{align*}
Then, we exploit $\mathbf 1_{\Omega_{\mathrm{emp}}(\Delta)}\le1$, the conditional
independence of $\Delta$ and $\tilde X$, and the scalar bound in
Lemma~\ref{lem:common_population_residual_covariance}.  Moreover, observe that the function
$\tilde\sigma^2x/(\tilde\sigma^2+x)$ is concave in $x$, and
$\mathbb E_\Delta[\|\Delta_{-S}\|_{\Sigma_{1,-S\mid S}}^2 \mid X_{\mathrm{pre}}]
=\tr(\Sigma_{1,-S\mid S}\Gamma_{-S,-S})/n\le C(d-s)/n$
on $\Omega_{\mathrm{pre}}$.  By Jensen's inequality, we obtain the
stated bound.
\end{proof}

For the contributions on $\Omega(\Delta)^{\mathrm c}$, we combine the conditional probability bound from Corollary~\ref{cor:common_realized_covariate_event} with bounds for the fourth moments.

\begin{lemma}[Moment bounds on $\Omega(\Delta)^{\mathrm c}$]
\label{lem:common_event_complement_moments}
Under the model in Section~\ref{sec:problem_setting} and
Assumption~\ref{ass:design}, fix $X_{\mathrm{pre}}\in\Omega_{\mathrm{pre}}$.  
Let $m\ge C$, $1\le s\le d$, $u>0$, and $c_2>0$, and take $\eta=m^{-C}$ for a
sufficiently large fixed $C$.
Suppose that the sample condition \eqref{eq:common_empirical_sample}, with $\eta$ replaced by $\eta^2$, 
holds with a sufficiently large constant.  Then, with probability at least $1-\eta$ over $\tilde X$,
the following bounds hold:
\begin{align*}
\mathbb E_{\Delta,\tilde\varepsilon}\left[
(\|\Delta\|_{\Sigma_1}^2+\tilde\sigma^2)
\mathbf 1_{\Omega(\Delta)^{\mathrm c}}
\middle|X_{\mathrm{pre}},\tilde X\right]
\le C(d/n+\tilde\sigma^2)m^{-c_2},
\end{align*}
and
\begin{align*}
&\mathbb E_{\Delta,\tilde\varepsilon}\left[
\left\|\bar G(\Delta)^{-1}\bar\mu(\Delta)
-\Sigma_{1,SS}^{-1}\Sigma_{1,S,-S}\Delta_{-S}
\right\|_{\Sigma_{1,SS}}^2
\mathbf 1_{\Omega_{\mathrm{emp}}(\Delta)\cap
\Omega(\Delta)^{\mathrm c}}
\middle|X_{\mathrm{pre}},\tilde X\right]\\
&\quad+\mathbb E_{\Delta,\tilde\varepsilon}\left[
\left\|\bar G(\Delta)^{-1}
\left(\mu(\Delta)
-G(\Delta)\bar G(\Delta)^{-1}\bar\mu(\Delta)\right)
\right\|_{\Sigma_{1,SS}}^2
\mathbf 1_{\Omega_{\mathrm{emp}}(\Delta)\cap
\Omega(\Delta)^{\mathrm c}}
\middle|X_{\mathrm{pre}},\tilde X\right]\\
&\quad\le C(d/n+\tilde\sigma^2)m^{-c_2}.
\end{align*}
\end{lemma}

\begin{proof}
Consider a fixed $\tilde X$ in the event from
Corollary~\ref{cor:common_realized_covariate_event}.  On this event,
$\mathbb P_{\Delta,\tilde\varepsilon}(\Omega(\Delta)^{\mathrm c}\mid
X_{\mathrm{pre}},\tilde X)\le2\eta$ and $\max_{1\le i\le m}\|\tilde x_i\|^2\le C(d+\log(m/\eta))$.

On $\Omega_{\mathrm{emp}}(\Delta)$, we apply
Lemma~\ref{lem:common_empirical_prediction_mean} and use
\eqref{eq:common_empirical_acceptance},
\eqref{eq:common_population_inverse_factor_scale}, and the common sample condition~\eqref{eq:common_empirical_sample}
to obtain
\begin{align*}
\left\|\bar G(\Delta)^{-1}\bar\mu(\Delta)
-\bar G_{\mathrm{pop}}(\Delta)^{-1}\bar\mu_{\mathrm{pop}}(\Delta)
\right\|_{\Sigma_{1,SS}}^2
&\le C\|\Delta\|_{\Sigma_1}^2.
\end{align*}
By~\eqref{eq:common_population_mean_target} and $0\le1-\theta_\Delta\le1$,
$\left\|\bar G_{\mathrm{pop}}(\Delta)^{-1}\bar\mu_{\mathrm{pop}}(\Delta)
-\Sigma_{1,SS}^{-1}\Sigma_{1,S,-S}\Delta_{-S}
\right\|_{\Sigma_{1,SS}}^2
\le\|(\Sigma_1\Delta)_S\|_{\Sigma_{1,SS}^{-1}}^2
\le\|\Delta\|_{\Sigma_1}^2$.
Therefore, the triangle inequality implies
$\left\|\bar G(\Delta)^{-1}\bar\mu(\Delta)
-\Sigma_{1,SS}^{-1}\Sigma_{1,S,-S}\Delta_{-S}
\right\|_{\Sigma_{1,SS}}^2
\le C\|\Delta\|_{\Sigma_1}^2$.

For the centered numerator, we apply~\eqref{eq:common_recentered_numerator}.
Since
$\Sigma_{1,SS}^{-1}(\Sigma_1\Delta)_S
=\Delta_S+\Sigma_{1,SS}^{-1}\Sigma_{1,S,-S}\Delta_{-S}$,
the preceding bound and the eigenvalue bounds in
Assumption~\ref{ass:design} imply
$\|\Delta_S+\bar G(\Delta)^{-1}\bar\mu(\Delta)\|^2
\le C\|\Delta\|_{\Sigma_1}^2$.
Corollary~\ref{cor:common_empirical_gram}
and the sample condition also imply
$\|\Sigma_{1,SS}^{1/2}\bar G(\Delta)^{-1}\|_{\mathrm{op}}\le C$.
By combining these bounds and the maximum covariate bound, we obtain
\begin{align*}
&\left\|\bar G(\Delta)^{-1}
\left(\mu(\Delta)
-G(\Delta)\bar G(\Delta)^{-1}\bar\mu(\Delta)\right)
\right\|_{\Sigma_{1,SS}}\\
&\quad\le Cm\left((d+\log(m/\eta))^{1/2}
\max_{1\le i\le m}|\tilde\varepsilon_i|
+(d+\log(m/\eta))\|\Delta\|_{\Sigma_1}\right).
\end{align*}
By Lemmas~\ref{lem:quad_form_concentration} and~\ref{lem:pretraining_covariance_regularity}, we have $\mathbb E_\Delta[\|\Delta\|_{\Sigma_1}^4\mid X_{\mathrm{pre}},\tilde X]\le C(d/n)^2$, 
which also leads to
 $\mathbb E_\Delta[(\|\Delta\|_{\Sigma_1}^2+\tilde\sigma^2)^2\mid X_{\mathrm{pre}},\tilde X]\le C(d/n+\tilde\sigma^2)^2$.
Lemma~\ref{lem:gaussian_max_design} also implies $\mathbb E_{\tilde\varepsilon}[\max_{1\le i\le m}|\tilde\varepsilon_i|^4]\le C\tilde\sigma^4(\log m)^2$.
Hence, the pointwise bound for $\bar G(\Delta)^{-1}(\mu(\Delta)-G(\Delta)\bar G(\Delta)^{-1}\bar\mu(\Delta))$ implies that its fourth moment restricted to $\Omega_{\mathrm{emp}}(\Delta)$ is at most $Cm^4(d+\log(m/\eta))^4(d/n+\tilde\sigma^2)^2$.

By the Cauchy--Schwarz inequality and the conditional probability bound from
Corollary~\ref{cor:common_realized_covariate_event}, each of the three
expectations in the statement is bounded by
\begin{align*}
C m^2(d+\log(m/\eta))^2(d/n+\tilde\sigma^2)\eta^{1/2}
\le C(d/n+\tilde\sigma^2)m^{-c_2}.
\end{align*}
The last inequality follows from $\eta=m^{-C}$ and the polynomial growth condition on $d+\log m$, with $C$ sufficiently large.
\end{proof}

\subsection{Conditional Mean Bounds for $\mathcal V_1$}

The variation over $\Delta$ of $\mathbb E_{\tilde\varepsilon}[r_{S,\mathrm{ext}}\mid\Delta,\tilde X]$ determines $\mathcal V_1$.
We have bounded 
the inverse remainder and the contributions on $\Omega(\Delta)^{\mathrm c}$ in~\eqref{eq:common_V1_remainder_decomposition}
in the preceding subsection.
It remains to take expectation the conditional mean bound in Lemma~\ref{lem:common_empirical_prediction_mean} over $\Delta$ and combine it with these remainder bounds.

\begin{corollary}[Mean squared approximation error]
\label{cor:common_empirical_prediction_mean_average}
Under the model in Section~\ref{sec:problem_setting}, fix $X_{\mathrm{pre}}\in\Omega_{\mathrm{pre}}$, 
and let $u>0$ and $0<\eta\le1/2$.  Suppose that the sample condition
\eqref{eq:common_empirical_sample} holds with $\eta$ replaced by
$\eta^2$ and $C$ sufficiently large.  Then, for every realization of
$\tilde X$,
\begin{align*}
&\mathbb E_\Delta\left[
\left\|\bar G(\Delta)^{-1}\bar\mu(\Delta)
-\bar G_{\mathrm{pop}}(\Delta)^{-1}\bar\mu_{\mathrm{pop}}(\Delta)
\right\|_{\Sigma_{1,SS}}^2
\mathbf 1_{\Omega_{\mathrm{emp}}(\Delta)}
\middle|X_{\mathrm{pre}},\tilde X\right]\\
&\quad\le C\frac{s+\log(m/\eta)}{m}
\frac{\tr(\Sigma_1\Gamma)}{n}
\max\left(1,\frac{\sqrt{1+\bar\rho^2}}{\max(1,u)}\right)
\left[1+\min\left(\bar\rho_S^2,
\frac{1+\bar\rho^2}{u^2}\right)\right].
\end{align*}
\end{corollary}

\begin{proof}
Consider a fixed realization of $\tilde X$.
On $\Omega_{\mathrm{emp}}(\Delta)$, we apply Lemma~\ref{lem:common_empirical_prediction_mean} with $\eta$ replaced by $\eta^2/12$.

By~\eqref{eq:marginal_acceptance_scale}, $p_\Delta\asymp\min(1,u/\sqrt{1+\rho_\Delta^2})$.
Hence, $\min(1,u)/p_\Delta\le C\max(1,\sqrt{1+\rho_\Delta^2}/\max(1,u))$.
The definitions of $\rho_{S,\Delta}$ and $u$ also imply $\|(\Sigma_1\Delta)_S\|_{\Sigma_{1,SS}^{-1}}^2/(\tilde\sigma^2+\|\Delta_{-S}\|_{\Sigma_{1,-S\mid S}}^2)\le\rho_{S,\Delta}^2$ and $\sigma_\xi^2(\Delta)/\delta^2=(1+\rho_\Delta^2)/u^2$.
We substitute these bounds into Lemma~\ref{lem:common_empirical_prediction_mean}, multiply by $\mathbf 1_{\Omega_{\mathrm{emp}}(\Delta)}$, and take conditional expectation over $\Delta$.

We apply the moment bound in Lemma~\ref{lem:quad_form_concentration} to $\tilde\sigma^{-1}\Sigma_1^{1/2}\Delta$ and $\tilde\sigma^{-1}\Sigma_{1,SS}^{-1/2}(\Sigma_1\Delta)_S$.
For $q\in\{1,2,3\}$, we obtain $\{\mathbb E_\Delta[\rho_\Delta^{2q}\mid X_{\mathrm{pre}}]\}^{1/q}\le C\bar\rho^2$ and $\{\mathbb E_\Delta[\rho_{S,\Delta}^{2q}\mid X_{\mathrm{pre}}]\}^{1/q}\le C\bar\rho_S^2$.
Consequently, we obtain
\begin{align*}
&\mathbb E_\Delta\left[
\|\Delta\|_{\Sigma_1}^2
\max\left(1,\frac{\sqrt{1+\rho_\Delta^2}}{\max(1,u)}\right)
\left(1+\min\left(\rho_{S,\Delta}^2,
\frac{1+\rho_\Delta^2}{u^2}\right)\right)
\middle|X_{\mathrm{pre}}\right]\\
&\quad\le C\frac{\tr(\Sigma_1\Gamma)}{n}
\max\left(1,\frac{\sqrt{1+\bar\rho^2}}{\max(1,u)}\right)
\left[1+\min\left(\bar\rho_S^2,
\frac{1+\bar\rho^2}{u^2}\right)\right],
\end{align*}
where we have used $\tilde\sigma^2\bar\rho^2=\tr(\Sigma_1\Gamma)/n$.
We substitute this inequality into the bound after taking expectation over $\Delta$ from Lemma~\ref{lem:common_empirical_prediction_mean} and obtain the claimed result.
\end{proof}

We combine the conditional mean and remainder bounds in
\eqref{eq:common_V1_remainder_decomposition} with the contribution of $r_{S,\mathrm{ext}}$ on $\Omega(\Delta)^{\mathrm c}$.
Then, we evaluate the population conditional mean to obtain the partial fine-tuning $\mathcal V_1$ rate for every $u>0$.

\begin{corollary}[Partial fine-tuning rate for $\mathcal V_1$]
\label{cor:common_partial_V1_extended}
Under the model in Section~\ref{sec:problem_setting} and
Assumption~\ref{ass:design}, fix $C_G\ge2$ as in
Definition~\ref{def:ext_estimator} and $X_{\mathrm{pre}}\in\Omega_{\mathrm{pre}}$.
Let $m\ge C$, $1\le s<d$, and $u>0$, and let $p_\star$ and $\zeta_\star^2$ 
be defined by~\eqref{eq:common_conditional_scales}.
Let $\eta=m^{-C}$ for a sufficiently large fixed $C$.  Suppose that the
sample condition~\eqref{eq:common_empirical_sample}, with $\eta$ replaced by $\eta^2$, 
holds with a sufficiently large constant.
Conditioned on $X_{\mathrm{pre}}$, with probability at least $1-\eta$ over $\tilde X$, the
extended estimator satisfies
\begin{align*}
\mathcal V_1
&\lesssim
\frac{s}{n}\left(
\left(
\frac{(d-s)\bar\rho^2}{d+(d-s)\bar\rho^2}
\right)^2
+\min(1,u^4)h_{d-s,d}(\bar\rho)
\right)+C\mathrm{Rem}_{\mathcal V_1},\\
\mathcal V_1
&\gtrsim
\frac{s}{n}\left(
\left(
\frac{(d-s)\bar\rho^2}{d+(d-s)\bar\rho^2}
\right)^2
+\min(1,u^4)h_{d-s,d}(\bar\rho)
\right)-C\mathrm{Rem}_{\mathcal V_1},
\end{align*}
where
\begin{align*}
\mathrm{Rem}_{\mathcal V_1}
&:=
\frac{s+\log m}{m}\frac{\tr(\Sigma_1\Gamma)}{n}
\max\left(
1,\frac{\sqrt{1+\bar\rho^2}}{\max(1,u)}
\right)
\left[
1+\min\left(
\bar\rho_S^2,\frac{1+\bar\rho^2}{u^2}
\right)
\right]\\
&\quad+
\frac{s\zeta_\star^2
\left(s+\zeta_\star^2(\log m)^2\right)(\log m)^3}
{(m p_\star)^2}
\left(
\min(\delta^2,\tilde\sigma^2)
+\frac{\tilde\sigma^2(d-s)}{n\tilde\sigma^2+d-s}
\right).
\end{align*}
The function $h_{d-s,d}$ is as in Corollary~\ref{cor:common_population_partial_mean_scale}.
\end{corollary}

\begin{proof}
Consider a fixed realization of $\tilde X$ in the event from Corollary~\ref{cor:common_realized_covariate_event}, on which $\Omega(\Delta)\subseteq\Omega_{\mathrm{post}}$.
In~\eqref{eq:common_V1_remainder_decomposition}, we take the squared $\Sigma_{1,SS}$ norm and then take expectation over $\Delta$.
By $\|\sum_{j=1}^4a_j\|^2\le4\sum_{j=1}^4\|a_j\|^2$, it suffices to bound the four terms separately.
We use Corollary~\ref{cor:common_empirical_prediction_mean_average} to bound the difference between the empirical and population approximations to the conditional mean.
For the inverse remainder, we use Corollary~\ref{cor:common_centered_inverse_rate} with conditional Jensen's inequality.
For the third term in~\eqref{eq:common_V1_remainder_decomposition}, we use the second conclusion of Lemma~\ref{lem:common_event_complement_moments} with conditional Jensen's inequality to bound its expected squared norm by $C(d/n+\tilde\sigma^2)m^{-2}$.

By~\eqref{eq:common_population_mean_target}, the squared norm of the population conditional mean is $(1-\theta_\Delta)^2\|(\Sigma_1\Delta)_S\|_{\Sigma_{1,SS}^{-1}}^2\le\|\Delta\|_{\Sigma_1}^2$, since $0\le1-\theta_\Delta\le1$.
We also have $\Omega_{\mathrm{emp}}(\Delta)^{\mathrm c}\subseteq\Omega(\Delta)^{\mathrm c}$.
Therefore, 
we use the scalar bound in Lemma~\ref{lem:common_event_complement_moments} to control the expected squared norms 
of the population conditional mean on $\Omega_{\mathrm{emp}}(\Delta)^{\mathrm c}$ and the complement-probability term 
in~\eqref{eq:common_V1_remainder_decomposition} by $C(d/n+\tilde\sigma^2)m^{-2}$.
Corollary~\ref{cor:common_realized_covariate_event} and Lemma~\ref{lem:extended_estimator_complement}, with $\eta=m^{-C}$, imply $\mathbb E_{\Delta,\tilde\varepsilon}[\|r_{S,\mathrm{ext}}\|_{\Sigma_{1,SS}}^2\mathbf 1_{\Omega(\Delta)^{\mathrm c}}\mid X_{\mathrm{pre}},\tilde X]\le C(d/n+\tilde\sigma^2)m^{-2}$.
Conditional Jensen's inequality provides the same bound after taking the conditional mean over $\tilde\varepsilon$.
Hence, by the definition of $\mathrm{Rem}_{\mathcal V_1}$, we absorb the complement contributions into $C\mathrm{Rem}_{\mathcal V_1}$.

It follows from Lemma~\ref{lem:zero_active_bias} that $\mathcal V_1$ equals the conditional second moment of $\mathbb E_{\tilde\varepsilon}[r_{S,\mathrm{ext}}\mid\Delta,\tilde X]$.
Finally, we apply the triangle inequality and combine these bounds to obtain 
\begin{align*}
&\left|\mathcal V_1^{1/2}-
\left(\mathbb E_\Delta\left\|
\bar G_{\mathrm{pop}}(\Delta)^{-1}\bar\mu_{\mathrm{pop}}(\Delta)
-\Sigma_{1,SS}^{-1}\Sigma_{1,S,-S}\Delta_{-S}
\right\|_{\Sigma_{1,SS}}^2\right)^{1/2}\right|\\
&\quad\le
\left(\mathbb E_\Delta\left\|
\mathbb E_{\tilde\varepsilon}
[r_{S,\mathrm{ext}}\mid\Delta,\tilde X]
-\bar G_{\mathrm{pop}}(\Delta)^{-1}\bar\mu_{\mathrm{pop}}(\Delta)
+\Sigma_{1,SS}^{-1}\Sigma_{1,S,-S}\Delta_{-S}
\right\|_{\Sigma_{1,SS}}^2\right)^{1/2}
\le C\mathrm{Rem}_{\mathcal V_1}^{1/2}.
\end{align*}
By Corollary~\ref{cor:common_population_partial_mean_scale} and $(x+y)^2\le2x^2+2y^2$, we obtain the stated upper and lower bounds for $\mathcal V_1$.
\end{proof}

\subsection{Full Fine-Tuning Conditional Mean at the Realized Covariates}

For partial fine-tuning, the result above involves $\bar G(\Delta)^{-1}\bar\mu(\Delta)$.
When $s=d$, we employ the conditional mean bounds in Proposition~\ref{prop:full_accepted_gram_mean} and take expectation over the acceptance indicators.
To bound the contribution on $\Omega_{\mathrm{full}}(\Delta)^{\mathrm c}$, we first establish a bound valid for every realization of these indicators.

Since $\xi_i\mid(\Delta,\tilde x_i)\sim \mathcal N(-\tilde x_i^\top\Delta,\tilde\sigma^2)$ and
$\mathbb I_i=1$ is equivalent to $|\xi_i|\le\delta$, symmetry and
monotonicity of the Gaussian density imply
$\left|\mathbb E_{\tilde\varepsilon_i} (\xi_i\mid\mathbb I_i=1,\Delta,\tilde x_i)\right| \le|\tilde x_i^\top\Delta|$.
On $\Omega_{\mathrm{post}}$, we have $r_{S,\mathrm{ext}}=r_S$ and
$\sum_i\mathbb I_iG(\Delta)^{-1/2}\tilde x_i\tilde x_i^\top
G(\Delta)^{-1/2}=I_d$.  Therefore, \eqref{eq:residual_rep_main} and the Cauchy--Schwarz inequality imply
\begin{equation}\label{eq:full_indicator_conditional_mean_bound}
\left\|\mathbb E_{\tilde\varepsilon}
[r_{S,\mathrm{ext}}\mid\Delta,\tilde X,
\mathbb I_1,\ldots,\mathbb I_m]\right\|_{\Sigma_1}
\le
\|\Sigma_1^{1/2}G(\Delta)^{-1/2}\|_{\mathrm{op}}
\|G(\Delta)^{1/2}\Delta\|.
\end{equation}
On $\Omega_{\mathrm{post}}^{\mathrm c}$, $r_{S,\mathrm{ext}}=-\Delta$.

\begin{corollary}[Full fine-tuning conditional mean at the realized covariates]
\label{cor:full_conditional_mean_realized_covariates}
Under the model in Section~\ref{sec:problem_setting} and
Assumption~\ref{ass:design}, fix $C_G\ge2$ and $X_{\mathrm{pre}}\in\Omega_{\mathrm{pre}}$. 
Let $m\ge C$, $s=d$, and $u>0$, and take $\eta=m^{-C}$ 
for a sufficiently large fixed $C$.
Suppose that the sample condition
\eqref{eq:common_empirical_sample}, with $\eta$ replaced by $\eta^2$,
holds with a sufficiently large constant.  Conditioned on $X_{\mathrm{pre}}$, 
with probability at least $1-\eta$ over $\tilde X$,
for every $\Delta\in\mathbb R^d$, it holds that
\begin{equation}\label{eq:full_conditional_mean_global_bound}
\left\|\mathbb E_{\tilde\varepsilon}
[r_{S,\mathrm{ext}}\mid\Delta,\tilde X]\right\|_{\Sigma_1}
\le C m^{C_G/2}\sqrt{d+\log m}\|\Delta\|_{\Sigma_1}.
\end{equation}
Moreover, for $0<u\le1$, conditioned on
$(X_{\mathrm{pre}},\tilde X)$, with probability at least $1-\eta$ over $\Delta$,
\begin{equation}\label{eq:full_small_u_conditional_mean}
\left\|\mathbb E_{\tilde\varepsilon}
[r_{S,\mathrm{ext}}\mid\Delta,\tilde X]
+\frac{u^2}{3}\Delta\right\|_{\Sigma_1}
\le Cu^4\|\Delta\|_{\Sigma_1}.
\end{equation}
For every fixed $u_0>0$, if $u\ge u_0$ and the constants in the stated conditions are sufficiently large depending on $u_0$, then, conditioned on
$(X_{\mathrm{pre}},\tilde X)$, with probability at least $1-\eta$ over
$\Delta$,
\begin{equation}\label{eq:full_conditional_mean_general_bound}
\left\|\mathbb E_{\tilde\varepsilon}
[r_{S,\mathrm{ext}}\mid\Delta,\tilde X]
+(1-\theta_\Delta)\Delta\right\|_{\Sigma_1}
\le\frac{1}{2}(1-\theta_\Delta)\|\Delta\|_{\Sigma_1}.
\end{equation}
\end{corollary}

\begin{proof}
Let $\Omega_\Delta$ be the event in
Corollary~\ref{cor:pretraining_error_concentration} with failure probability $\eta/2$.  
For every $\Delta\in\Omega_\Delta$, the sample condition and
\eqref{eq:common_empirical_acceptance} imply  $m p_\Delta\ge C\{d+\log(8m/\eta^3)\}\log^2(8m/\eta^3)$.
For such $\Delta$, we apply
Proposition~\ref{prop:full_accepted_gram_mean} with failure probability
$\eta^3/8$.  For the resulting event $\Omega_{\mathrm{full}}(\Delta)$, we
have
$\mathbb E_{\tilde X}[\mathbb P_{\tilde\varepsilon}
(\Omega_{\mathrm{full}}(\Delta)^{\mathrm c}\mid\Delta,\tilde X)
\mid\Delta]\le\eta^3/8$.  Markov's inequality implies
\begin{align*}
\mathbb P_{\tilde X}\left\{
\mathbb P_{\tilde\varepsilon}
(\Omega_{\mathrm{full}}(\Delta)^{\mathrm c}\mid\Delta,\tilde X)
>\frac{\eta}{2}\middle|\Delta\right\}
&\le\frac{\eta^2}{4}.
\end{align*}
Since $\Delta$ and $\tilde X$ are independent conditioned on $X_{\mathrm{pre}}$, we apply Lemma~\ref{lem:fixed_covariate_conversion} to $\Omega_\Delta$ and the event $\{\mathbb P_{\tilde\varepsilon}(\Omega_{\mathrm{full}}(\Delta)^{\mathrm c}\mid\Delta,\tilde X)\le\eta/2\}$.
Then, with
probability at least $1-\eta/2$ over $\tilde X$, the properties
$\Delta\in\Omega_\Delta$ and
$\mathbb P_{\tilde\varepsilon}
(\Omega_{\mathrm{full}}(\Delta)^{\mathrm c}\mid\Delta,\tilde X)
\le\eta/2$ hold with conditional probability at least $1-\eta$ over
$\Delta$.  We also apply Lemma~\ref{lem:gaussian_max_design} with failure
probability $\eta/2$.  Therefore, the union bound shows that these properties
and $\max_{1\le i\le m}\|\tilde x_i\|^2\le C(d+\log m)$ hold with probability
at least $1-\eta$ over $\tilde X$.

On $\Omega_{\mathrm{post}}$, \eqref{eq:full_indicator_conditional_mean_bound}
and the definition of $\Omega_{\mathrm{post}}$ imply
$\|\Sigma_1^{1/2}G(\Delta)^{-1/2}\|_{\mathrm{op}} \le Cm^{(C_G-1)/2}$, 
while the covariate maximum and Assumption~\ref{ass:design} imply
$\|G(\Delta)^{1/2}\Delta\| \le C\sqrt{m(d+\log m)}\|\Delta\|_{\Sigma_1}$.
On $\Omega_{\mathrm{post}}^{\mathrm c}$, the extended residual equals $-\Delta$.  
Hence, for every realization of the acceptance indicators,
$\left\|\mathbb E_{\tilde\varepsilon}
[r_{S,\mathrm{ext}}\mid\Delta,\tilde X,
\mathbb I_1,\ldots,\mathbb I_m]\right\|_{\Sigma_1}
\le C m^{C_G/2}\sqrt{d+\log m}\|\Delta\|_{\Sigma_1}$.
Then, we take expectation over the acceptance indicators and apply the triangle inequality to obtain~\eqref{eq:full_conditional_mean_global_bound}.

For the realizations of $\Delta$ satisfying the two properties above, we apply
the lower bound on Gram matrices in
Proposition~\ref{prop:full_accepted_gram_mean} and
\eqref{eq:common_empirical_population_operator}.  On  $\Omega_{\mathrm{full}}(\Delta)$, we have
$\lambda_{\min}\{G(\Delta)/m\}\ge c p_\star/[\zeta_\star^2\{\log(m/\eta)\}^{3/2}]
\ge m^{-C_G}$, where the second inequality follows from the sample condition
and $C_G\ge2$.  Hence, $\Omega_{\mathrm{full}}(\Delta)\subseteq\Omega_{\mathrm{post}}$.

Suppose that $0<u\le1$.
For fixed $(\Delta,\tilde X)$, the event $\Omega_{\mathrm{full}}(\Delta)$ is determined by the acceptance indicators.
We apply Proposition~\ref{prop:full_accepted_gram_mean} on this event and the bound valid for every realization of the indicators on its complement.
Then, we take expectation over the indicators and obtain
\begin{align*}
\left\|\mathbb E_{\tilde\varepsilon}
[r_{S,\mathrm{ext}}\mid\Delta,\tilde X]
+\frac{u^2}{3}\Delta\right\|_{\Sigma_1}
&\le C\left(u^4+\eta m^{C_G/2}\sqrt{d+\log m}\right)
\|\Delta\|_{\Sigma_1}.
\end{align*}
We take the exponent in $\eta=m^{-C}$ sufficiently large and the constant in the sample condition sufficiently large to obtain
\eqref{eq:full_small_u_conditional_mean}.

Fix $u_0>0$ and suppose that $u\ge u_0$.
The analogous averaging argument, with the second conditional mean bound in Proposition~\ref{prop:full_accepted_gram_mean}, bounds the left hand side of~\eqref{eq:full_conditional_mean_general_bound} by $\{(1-\theta_\Delta)/4+C\eta m^{C_G/2}\sqrt{d+\log m}\}\|\Delta\|_{\Sigma_1}$.
By Lemma~\ref{lem:common_population_theta_bounds}, $1-\theta_\Delta\ge c(u_0)>0$.
Hence, for a sufficiently large exponent in $\eta=m^{-C}$, the complement contribution is at most $(1-\theta_\Delta)\|\Delta\|_{\Sigma_1}/4$, which proves~\eqref{eq:full_conditional_mean_general_bound}.
\end{proof}

%% file: Unified_appendix/app_09bG5_conditional_risk_variance.tex

\subsection{Conditional Covariance Contribution to $\mathcal V_2$}

The conditional covariance over generation noise determines $\mathcal V_2$.
By~\eqref{eq:V2_common_comparison}, it suffices to bound
\begin{align*}
\mathbb E_\Delta\left[
\tr\left(\Sigma_{1,SS}\bar G(\Delta)^{-1}M(\Delta)
\bar G(\Delta)^{-1}\right)
\mathbf 1_{\Omega_{\mathrm{emp}}(\Delta)}
\middle|X_{\mathrm{pre}},\tilde X\right],
\end{align*}
the inverse remainder $R_{\mathrm{res}}^\Omega$, and the contribution
on $\Omega(\Delta)^{\mathrm c}$.
First, we express this expectation in terms of $\bar G_{\mathrm{pop}}(\Delta)$ and bound $R_{\mathrm{res}}^\Omega$ by a small multiple of it.

\begin{proposition}[Conditional covariance trace and inverse remainder]
\label{prop:common_V2_leading}
Under the model in Section~\ref{sec:problem_setting}, fix $X_{\mathrm{pre}}\in\Omega_{\mathrm{pre}}$, and let $u>0$ and $0<\eta\le1/2$.  
Suppose that the sample condition
\eqref{eq:common_empirical_sample} holds with $\eta$ replaced by $\eta^2$ 
and $C$ sufficiently large.  Then, it holds that 
\begin{align*}
&\mathbb E_\Delta\left[
\tr\left(\Sigma_{1,SS}\bar G(\Delta)^{-1}M(\Delta)
\bar G(\Delta)^{-1}\right)
\mathbf 1_{\Omega_{\mathrm{emp}}(\Delta)}
\middle|X_{\mathrm{pre}},\tilde X\right]\\
&\quad\asymp\frac{1}{m}\mathbb E_\Delta\left[
\left(\min(\delta^2,\tilde\sigma^2)
+\frac{\tilde\sigma^2
\|\Delta_{-S}\|_{\Sigma_{1,-S\mid S}}^2}
{\tilde\sigma^2+
\|\Delta_{-S}\|_{\Sigma_{1,-S\mid S}}^2}\right)
\tr\left(\Sigma_{1,SS}
\bar G_{\mathrm{pop}}(\Delta)^{-1}\right)
\mathbf 1_{\Omega_{\mathrm{emp}}(\Delta)}
\middle|X_{\mathrm{pre}},\tilde X\right].
\end{align*}
Moreover, the remainder $R_{\mathrm{res}}^\Omega$ defined
in~\eqref{eq:Rres_common_def} satisfies
\begin{align*}
R_{\mathrm{res}}^\Omega\le\frac{1}{16}
\mathbb E_\Delta\left[
\tr\left(\Sigma_{1,SS}\bar G(\Delta)^{-1}M(\Delta)
\bar G(\Delta)^{-1}\right)
\mathbf 1_{\Omega_{\mathrm{emp}}(\Delta)}
\middle|X_{\mathrm{pre}},\tilde X\right].
\end{align*}
\end{proposition}

\begin{proof}
Consider a fixed realization of $\tilde X$.  On $\Omega_{\mathrm{emp}}(\Delta)$, 
we apply Lemma~\ref{lem:common_empirical_covariance} and the Gram matrix bounds in
Corollary~\ref{cor:common_empirical_gram} to have
\begin{align*}
\tr\left(\Sigma_{1,SS}\bar G(\Delta)^{-1}M(\Delta)
\bar G(\Delta)^{-1}\right)
\asymp\frac{\tilde\sigma^2(1-\theta_\Delta)}{m}
\tr\left(\Sigma_{1,SS}\bar G_{\mathrm{pop}}(\Delta)^{-1}\right).
\end{align*}
We substitute the scalar bound in
Lemma~\ref{lem:common_population_residual_covariance}, multiply by
$\mathbf 1_{\Omega_{\mathrm{emp}}(\Delta)}$, and take conditional
expectation over $\Delta$ to obtain the first statement.

For the remainder, the two trace bounds in
Lemma~\ref{lem:common_empirical_covariance} imply, on $\Omega_{\mathrm{emp}}(\Delta)$,
\begin{align*}
\frac{
\|\Sigma_{1,SS}^{1/2}\bar G(\Delta)^{-1/2}\|_{\mathrm{op}}^2
\tr\left(\bar G(\Delta)^{-1}M(\Delta)\right)}
{\tr\left(\Sigma_{1,SS}\bar G(\Delta)^{-1}M(\Delta)
\bar G(\Delta)^{-1}\right)}
\quad\le
C\frac{s\|\Sigma_{1,SS}^{1/2}\bar G(\Delta)^{-1/2}\|_{\mathrm{op}}^2}
{\tr\left(\Sigma_{1,SS}\bar G(\Delta)^{-1}\right)}
\le C\frac{s\zeta_\Delta^2}{s-1+\zeta_\Delta^2}
\le C\min\bigl(s,\zeta_\Delta^2\bigr),
\end{align*}
where the last two inequalities follow from the Gram matrix bounds in
Corollary~\ref{cor:common_empirical_gram}, \eqref{eq:common_delta_population_trace},
\eqref{eq:common_delta_population_operator}, and $\zeta_\Delta^2\ge1$.

On $\Omega_{\mathrm{emp}}(\Delta)$, we invoke
Corollary~\ref{cor:common_empirical_gram} with failure probability $\eta^2/12$, and obtain 
 $\min(s,\zeta_\Delta^2) \ell_{\max}(\Delta)$ $\log(m/\eta)\le c.$ 
We substitute this bound and the ratio bound into
Lemma~\ref{lem:common_centered_inverse}.  By choosing the constant in the sample
condition sufficiently large, we have
\begin{align*}
R_{\mathrm{res}}^\Omega
&\le\frac{1}{16}\mathbb E_\Delta\left[
\tr\left(\Sigma_{1,SS}\bar G(\Delta)^{-1}M(\Delta)
\bar G(\Delta)^{-1}\right)
\mathbf 1_{\Omega_{\mathrm{emp}}(\Delta)}
\middle|X_{\mathrm{pre}},\tilde X\right].
\end{align*}
\end{proof}

Next, we bound the contribution on $\Omega(\Delta)^{\mathrm c}$ and remove the indicator $\mathbf 1_{\Omega_{\mathrm{emp}}(\Delta)}$ to express the bounds for $\mathcal V_2$ through population moments.

\begin{corollary}[Population bounds for $\mathcal V_2$]
\label{cor:common_V2_population_representation}
Under the model in Section~\ref{sec:problem_setting} and
Assumption~\ref{ass:design}, fix $C_G\ge2$ as in
Definition~\ref{def:ext_estimator} and $X_{\mathrm{pre}}\in\Omega_{\mathrm{pre}}$.
Let $m\ge C$, $1\le s\le d$, $u>0$, and $c_2>0$, and let $p_\star$ and $\zeta_\star^2$ 
be defined by~\eqref{eq:common_conditional_scales}.
Let $\eta=m^{-C}$ for a sufficiently large fixed $C$.  Suppose that the
sample condition~\eqref{eq:common_empirical_sample}, with $\eta$ replaced by $\eta^2$, 
holds with a sufficiently large constant.  Conditioned on $X_{\mathrm{pre}}$, 
with probability at least $1-\eta$ over $\tilde X$, the
extended estimator satisfies
\begin{align*}
\mathcal V_2&\lesssim\frac{1}{m}\mathbb E_\Delta\Bigg[
\left(\min(\delta^2,\tilde\sigma^2)
+\frac{\tilde\sigma^2\|\Delta_{-S}\|_{\Sigma_{1,-S\mid S}}^2}
{\tilde\sigma^2+\|\Delta_{-S}\|_{\Sigma_{1,-S\mid S}}^2}\right)
\tr\left(\Sigma_{1,SS}\bar G_{\mathrm{pop}}(\Delta)^{-1}\right)
\Bigm|X_{\mathrm{pre}}\Bigg]\\
&\quad+C(d/n+\tilde\sigma^2)m^{-c_2},\\
\mathcal V_2&\gtrsim\frac{1}{m}\mathbb E_\Delta\Bigg[
\left(\min(\delta^2,\tilde\sigma^2)
+\frac{\tilde\sigma^2\|\Delta_{-S}\|_{\Sigma_{1,-S\mid S}}^2}
{\tilde\sigma^2+\|\Delta_{-S}\|_{\Sigma_{1,-S\mid S}}^2}\right)
\tr\left(\Sigma_{1,SS}\bar G_{\mathrm{pop}}(\Delta)^{-1}\right)
\Bigm|X_{\mathrm{pre}}\Bigg]\\
&\quad-C(d/n+\tilde\sigma^2)m^{-c_2}.
\end{align*}
\end{corollary}

\begin{proof}
Consider a fixed realization of $\tilde X$ in the event from Corollary~\ref{cor:common_realized_covariate_event}.
We use Lemmas~\ref{lem:common_event_complement_moments} and~\ref{lem:extended_estimator_complement} to bound the complement contribution in~\eqref{eq:V2_common_comparison}. This leads to
\begin{align*}
&\mathbb E_{\Delta,\tilde\varepsilon}\Biggl[
\Bigl\|r_{S,\mathrm{ext}}-
(\bar G(\Delta)^{-1}\bar\mu(\Delta)
-\Sigma_{1,SS}^{-1}\Sigma_{1,S,-S}\Delta_{-S})
\mathbf 1_{\Omega_{\mathrm{emp}}(\Delta)}\\
&\qquad\quad-\bar G(\Delta)^{-1}
(\mu(\Delta)-G(\Delta)\bar G(\Delta)^{-1}\bar\mu(\Delta))
\mathbf 1_{\Omega_{\mathrm{emp}}(\Delta)}
\Bigr\|_{\Sigma_{1,SS}}^2
\mathbf 1_{\Omega(\Delta)^{\mathrm c}}
\Bigm|X_{\mathrm{pre}},\tilde X\Biggr]\le C(d/n+\tilde\sigma^2)m^{-c_2}.
\end{align*}
We substitute this bound into~\eqref{eq:V2_common_comparison} and apply both conclusions of Proposition~\ref{prop:common_V2_leading}.
Then, we obtain
\begin{align*}
\mathcal V_2&\lesssim\frac{1}{m}\mathbb E_\Delta\Bigg[
\left(\min(\delta^2,\tilde\sigma^2)
+\frac{\tilde\sigma^2\|\Delta_{-S}\|_{\Sigma_{1,-S\mid S}}^2}
{\tilde\sigma^2+\|\Delta_{-S}\|_{\Sigma_{1,-S\mid S}}^2}\right)
\tr\left(\Sigma_{1,SS}\bar G_{\mathrm{pop}}(\Delta)^{-1}\right)
\mathbf 1_{\Omega_{\mathrm{emp}}(\Delta)}
\Bigm|X_{\mathrm{pre}},\tilde X\Bigg]\\
&\quad+C(d/n+\tilde\sigma^2)m^{-c_2},\\
\mathcal V_2&\gtrsim\frac{1}{m}\mathbb E_\Delta\Bigg[
\left(\min(\delta^2,\tilde\sigma^2)
+\frac{\tilde\sigma^2\|\Delta_{-S}\|_{\Sigma_{1,-S\mid S}}^2}
{\tilde\sigma^2+\|\Delta_{-S}\|_{\Sigma_{1,-S\mid S}}^2}\right)
\tr\left(\Sigma_{1,SS}\bar G_{\mathrm{pop}}(\Delta)^{-1}\right)
\mathbf 1_{\Omega_{\mathrm{emp}}(\Delta)}
\Bigm|X_{\mathrm{pre}},\tilde X\Bigg]\\
&\quad-C(d/n+\tilde\sigma^2)m^{-c_2}.
\end{align*}

It remains to remove $\mathbf 1_{\Omega_{\mathrm{emp}}(\Delta)}$ from the
population expectation.  Corollary~\ref{cor:common_realized_covariate_event}
implies that $\Omega_{\mathrm{emp}}(\Delta)$ holds with conditional
probability at least $1-\eta$ over $\Delta$.
By~\eqref{eq:common_delta_population_trace},
\eqref{eq:common_population_inverse_factor_bound}, and $\rho_{S,\Delta}^2\le\rho_\Delta^2$, 
we have $\tr(\Sigma_{1,SS}\bar G_{\mathrm{pop}}(\Delta)^{-1})
\le C(s+\rho_\Delta^2)/p_\Delta.$
The scalar weight in the population expectation is at most $2\tilde\sigma^2$.  
By~\eqref{eq:marginal_acceptance_scale}, we have $p_\Delta\asymp\min(1,u/\sqrt{1+\rho_\Delta^2})$.
We apply the moment bound in Lemma~\ref{lem:quad_form_concentration} to obtain
$\left(\mathbb E_\Delta\left[
\frac{(s+\rho_\Delta^2)^2}{p_\Delta^2}
\middle|X_{\mathrm{pre}}\right]\right)^{1/2}
\le \frac{C(s+\bar\rho^2)}{p_\star}.$
Then, the Cauchy--Schwarz inequality bounds the contribution on $\Omega_{\mathrm{emp}}(\Delta)^{\mathrm c}$ by
\begin{align*}
\frac{C\tilde\sigma^2(s+\bar\rho^2)\eta^{1/2}}{mp_\star}
&\le C\left(\frac{d}{n}+\tilde\sigma^2\right)m^{-c_2},
\end{align*}
where the last inequality follows from the sample condition and $\eta=m^{-C}$ for sufficiently large $C$.
Hence, we remove the indicator and obtain the stated bounds.
\end{proof}

%% file: Unified_appendix/app_09cG6_weighted_population_variance.tex

\subsection{Weighted Population Inverse Trace}

We evaluate the expectation over $\Delta$ in
Corollary~\ref{cor:common_V2_population_representation} for partial and full fine-tuning.

\begin{lemma}[Weighted population inverse trace]
\label{lem:common_weighted_population_inverse_trace}
Under the pre-training model in Section~\ref{sec:problem_setting} and
Assumption~\ref{ass:design}, fix
$X_{\mathrm{pre}}\in\Omega_{\mathrm{pre}}$ and let $u>0$.
Let $p_\star$ and $\zeta_\star^2$ be defined by
\eqref{eq:common_conditional_scales}.
Expectations are taken over $\Delta\mid X_{\mathrm{pre}}$.

If $1\le s<d$, let $k=d-s$.  Recall that, for $z>0$, $g_{k,d}(z)$ equals
$1+z/\sqrt{d}$, $\log(e+z^2/d)$, and $1$ when $k=1$, $k=2$, and
$k\ge3$, respectively.  Then,
\begin{align*}
&\mathbb E_\Delta\Bigg[
\left(
\min(\delta^2,\tilde\sigma^2)
+\frac{\tilde\sigma^2
\|\Delta_{-S}\|_{\Sigma_{1,-S\mid S}}^2}
{\tilde\sigma^2+\|\Delta_{-S}\|_{\Sigma_{1,-S\mid S}}^2}
\right)
\tr\left(\Sigma_{1,SS}\bar G_{\mathrm{pop}}(\Delta)^{-1}\right)
\Bigg] \asymp
\frac{s\tilde\sigma^2}{p_\star}
\left(
\min(1,u^2)+\frac{k\bar\rho^2}{d+k\bar\rho^2}
\right)g_{k,d}(\zeta_\star).
\end{align*}
If $s=d$, then
\begin{align*}
\mathbb E_\Delta\left[\min(\delta^2,\tilde\sigma^2)
\tr\left(\Sigma_1\bar G_{\mathrm{pop}}(\Delta)^{-1}\right)\right]
&\asymp\frac{\tilde\sigma^2\min(1,u^2)}{p_\star}
\left(d-1+\zeta_\star^2\right).
\end{align*}
\end{lemma}

\begin{proof}
For brevity, we omit the fixed conditioning on $X_{\mathrm{pre}}$ in the expectation in the proof.
Suppose that
$1\le s<d$, let $k=d-s$, and define the normalized squared frozen and active errors by
\begin{align*}
x:=\frac{\|\Delta_{-S}\|_{\Sigma_{1,-S\mid S}}^2}{\tilde\sigma^2},
\qquad y:=\frac{\|(\Sigma_1\Delta)_S\|_{\Sigma_{1,SS}^{-1}}^2}{\tilde\sigma^2},
\end{align*}
where it follows that $\rho_\Delta^2=x+y$.  The population inverse trace in
Lemma~\ref{lem:common_population_covariance} can be expressed as follows: 
\begin{align*}
\tr\left(\Sigma_{1,SS}\bar G_{\mathrm{pop}}(\Delta)^{-1}\right)
&=\frac{1}{p_\Delta}\left(s-1+
\frac{1+x+y}{1+x+\lambda_{2,\Delta}y}\right).
\end{align*}
Also the marginal acceptance bound in
\eqref{eq:marginal_acceptance_scale} is written as
$p_\Delta^{-1}\asymp1+\sqrt{1+x+y}/u$, while
\eqref{eq:common_population_inverse_factor_scale} implies $(1+x+y)/(1+x+\lambda_{2,\Delta}y)\asymp1+\min(y/(1+x),(1+x+y)/u^2)$.
For every $u>0$, we have
\begin{align*}
1+\frac{y}{1+x+u^2}
&\le1+\min\left(\frac{y}{1+x},\frac{1+x+y}{u^2}\right)
\le2\left(1+\frac{y}{1+x+u^2}\right).
\end{align*}
We substitute these bounds into the population trace identity and obtain, uniformly over $u>0$,
\begin{align} 
\tr\left(\Sigma_{1,SS}\bar G_{\mathrm{pop}}(\Delta)^{-1}\right) \asymp
\left(1+\frac{\sqrt{1+x+y}}{u}\right) \left(s+\frac{y}{1+u^2+x}\right). \label{eq:trace_comparison}
\end{align}
Moreover, the covariance scale in the expectation is  represented as $\tilde\sigma^2(\min(1,u^2)+x/(1+x))$. 
This factor depends only on $x$, whereas the population inverse trace also depends on the active error through $y$. 
Therefore, we condition on $\Delta_{-S}$ to take expectation over $y$ first and
then take expectation over $x$.  

Conditioned on $\Delta_{-S}$, the vector $\tilde\sigma^{-1}\Sigma_{1,SS}^{-1/2}(\Sigma_1\Delta)_S$ is Gaussian.  It follows from the
Gaussian conditioning formula, Lemma~\ref{lem:pretraining_covariance_regularity},
and Assumption~\ref{ass:design} that 
\begin{align*}
\frac{c\bar\rho^2}{d}I_s
\preceq
\Cov_\Delta\left(
\frac{\Sigma_{1,SS}^{-1/2}(\Sigma_1\Delta)_S}{\tilde\sigma}
\middle|\Delta_{-S}\right)
\preceq\frac{C\bar\rho^2}{d}I_s,
\qquad
\left\|
\mathbb E_\Delta\left[
\frac{\Sigma_{1,SS}^{-1/2}(\Sigma_1\Delta)_S}{\tilde\sigma}
\middle|\Delta_{-S}\right]\right\|_2^2
\le Cx.
\end{align*}
The Gaussian second and fourth moment formulas result in 
\begin{align*}
\frac{cs\bar\rho^2}{d}
\le\mathbb E_\Delta(y\mid\Delta_{-S})
\le C\left(\frac{s\bar\rho^2}{d}+x\right),
\qquad
\mathbb E_\Delta(y^2\mid\Delta_{-S})
\le C\left(\frac{s\bar\rho^2}{d}+x\right)^2.
\end{align*}
Also, we note that $\mathbb E_\Delta(\sqrt{1+x+y}\mid\Delta_{-S})\asymp\sqrt{1+s\bar\rho^2/d+x}$.  Since $y$ and
$\sqrt{1+x+y}$ are nondecreasing functions of $y$, their conditional
covariance is nonnegative and the Cauchy--Schwarz inequality implies
\begin{align*}
\frac{cs\bar\rho^2}{d}
\sqrt{1+\frac{s\bar\rho^2}{d}+x}
&\le
\mathbb E_\Delta\left[
y\sqrt{1+x+y}\middle|\Delta_{-S}\right]
\le
C\left(\frac{s\bar\rho^2}{d}+x\right)
\sqrt{1+\frac{s\bar\rho^2}{d}+x}.
\end{align*}
Then, we expand the conditional expectation of the right hand side in \eqref{eq:trace_comparison}, and have
\begin{align*}
&\mathbb E_\Delta\left[
\left(1+\frac{\sqrt{1+x+y}}{u}\right)
\left(s+\frac{y}{1+u^2+x}\right)
\middle|\Delta_{-S}\right]\\
&\quad=
s\left(1+\frac{
\mathbb E_\Delta(\sqrt{1+x+y}\mid\Delta_{-S})}{u}\right)
+\frac{
\mathbb E_\Delta(y\mid\Delta_{-S})
+u^{-1}\mathbb E_\Delta(y\sqrt{1+x+y}\mid\Delta_{-S})}
{1+u^2+x}\\
&\quad\asymp
\left(1+\frac{\sqrt{1+s\bar\rho^2/d+x}}{u}\right)
\left(s+\frac{s\bar\rho^2/d}{1+u^2+x}\right),
\end{align*}
where the terms proportional to $x$ in the upper bound are absorbed by $x/(1+u^2+x)\le1$.
By the tower property, after dividing by $\tilde\sigma^2$, the expectation in
the partial fine-tuning statement has the same order as
\begin{align}
&
s\mathbb E_\Delta\left[
\left(\min(1,u^2)+\frac{x}{1+x}\right)
\left(1+\frac{\sqrt{1+s\bar\rho^2/d+x}}{u}\right)\right]\notag\\
&\quad + \frac{s\bar\rho^2}{d}\mathbb E_\Delta\left[
\frac{\min(1,u^2)+x/(1+x)}{1+u^2+x}
\left(1+\frac{\sqrt{1+s\bar\rho^2/d+x}}{u}\right)\right].
\label{eq:weighted_population_trace_terms}
\end{align}

First, we evaluate the first term in \eqref{eq:weighted_population_trace_terms}.
Since $c\bar\rho^2X_k/d\le x\le C\bar\rho^2X_k/d$ for $X_k\sim\chi_k^2$, the moment bounds in Lemma~\ref{lem:frozen_denominator_marginal}, the covariance eigenvalue bounds, and Gaussian norm moments imply, for each $r\in\{1,2\}$,
\begin{align*}
\mathbb E_\Delta\left[
\left(\min(1,u^2)+\frac{x}{1+x}\right)^r\right]
&\asymp
\left(\min(1,u^2)+\frac{k\bar\rho^2}{d+k\bar\rho^2}\right)^r,\\
\mathbb E_\Delta\left[
\left(1+\frac{\sqrt{1+s\bar\rho^2/d+x}}{u}\right)^r\right]
&\asymp
\left(1+\frac{\sqrt{1+\bar\rho^2}}{u}\right)^r.
\end{align*}
Since $x/(1+x)$ and $\sqrt{1+s\bar\rho^2/d+x}$ are nondecreasing in $x$, the covariance of the two factors in the first term of \eqref{eq:weighted_population_trace_terms} is nonnegative.
Hence, by the above moment bounds, we obtain 
\begin{align*}
\mathbb E_\Delta\left[
\left(\min(1,u^2)+\frac{x}{1+x}\right)
\left(1+\frac{\sqrt{1+s\bar\rho^2/d+x}}{u}\right)\right] \asymp
\left(\min(1,u^2)+\frac{k\bar\rho^2}{d+k\bar\rho^2}\right)
\left(1+\frac{\sqrt{1+\bar\rho^2}}{u}\right).
\end{align*}

Next, we evaluate the second term in \eqref{eq:weighted_population_trace_terms}.
For $0<u\le1$, we have $1+u^2+x\asymp1+x$, and the final bound in Lemma~\ref{lem:frozen_denominator_marginal} applies after expanding the numerator.  For $u>1$, both $\min(1,u^2)+x/(1+x)$ and $\min(1,u^2)+k\bar\rho^2/(d+k\bar\rho^2)$ are in $[1,2]$.  Hence,
\begin{align*}
\mathbb E_\Delta\left[
\frac{\min(1,u^2)+x/(1+x)}{1+u^2+x}\right]
&\asymp
\left(\min(1,u^2)+\frac{k\bar\rho^2}{d+k\bar\rho^2}\right)
\mathbb E_\Delta\left(\frac{1}{1+u^2+x}\right).
\end{align*}
For the remaining factor, we use $\sqrt{1+s\bar\rho^2/d+x}\le\sqrt{1+s\bar\rho^2/d}+x/(2\sqrt{1+s\bar\rho^2/d})$ to obtain
\begin{align*}
&\left(1+\frac{\sqrt{1+s\bar\rho^2/d}}{u}\right)
\mathbb E_\Delta\left[
\frac{\min(1,u^2)+x/(1+x)}{1+u^2+x}\right]\\
&\quad\le
\mathbb E_\Delta\left[
\frac{\min(1,u^2)+x/(1+x)}{1+u^2+x}
\left(1+\frac{\sqrt{1+s\bar\rho^2/d+x}}{u}\right)\right]\\
&\quad\le
\left(1+\frac{\sqrt{1+s\bar\rho^2/d}}{u}\right)
\mathbb E_\Delta\left[
\frac{\min(1,u^2)+x/(1+x)}{1+u^2+x}\right]\\
&\qquad+
\frac{1}{2u\sqrt{1+s\bar\rho^2/d}}
\mathbb E_\Delta\left[
\frac{x\left(\min(1,u^2)+x/(1+x)\right)}{1+u^2+x}\right].
\end{align*}
Since $x/(1+u^2+x)\le1$, the last expectation term satisfies
\begin{align*}
\mathbb E_\Delta\left[
\frac{x\left(\min(1,u^2)+x/(1+x)\right)}{1+u^2+x}\right]
&\le \mathbb E_\Delta\left[\min(1,u^2)+\frac{x}{1+x}\right]
\asymp \min(1,u^2)+\frac{k\bar\rho^2}{d+k\bar\rho^2}.
\end{align*}
After multiplication by $s\bar\rho^2/d$, its contribution is at most
\begin{align*}
\frac{Cs\bar\rho^2}{du\sqrt{1+s\bar\rho^2/d}}
\left(\min(1,u^2)+\frac{k\bar\rho^2}{d+k\bar\rho^2}\right)
&\le
Cs\left(\min(1,u^2)+\frac{k\bar\rho^2}{d+k\bar\rho^2}\right)
\left(1+\frac{\sqrt{1+\bar\rho^2}}{u}\right).
\end{align*}
Thus, the additional term in the upper bound is bounded by the first term in \eqref{eq:weighted_population_trace_terms}.  These bounds imply
\begin{align}
&\mathbb E_\Delta\Bigg[
\left(\min(\delta^2,\tilde\sigma^2)
+\frac{\tilde\sigma^2\|\Delta_{-S}\|_{\Sigma_{1,-S\mid S}}^2}
{\tilde\sigma^2+\|\Delta_{-S}\|_{\Sigma_{1,-S\mid S}}^2}\right)
\tr\left(\Sigma_{1,SS}\bar G_{\mathrm{pop}}(\Delta)^{-1}\right)
\Bigg]\notag\\
&\quad\asymp
s\tilde\sigma^2
\left(\min(1,u^2)+\frac{k\bar\rho^2}{d+k\bar\rho^2}\right)
\left(1+\frac{\sqrt{1+\bar\rho^2}}{u}\right)
\left[
1+\frac{\bar\rho^2}{d}
\frac{1+\sqrt{1+s\bar\rho^2/d}/u}
{1+\sqrt{1+\bar\rho^2}/u}
\mathbb E_\Delta\left(\frac{1}{1+u^2+x}\right)
\right].
\label{eq:weighted_population_trace_assembly}
\end{align}

It remains to evaluate $\mathbb E_\Delta[(1+u^2+x)^{-1}]$ in the last factor of \eqref{eq:weighted_population_trace_assembly}.
Since $s=d-k$, $(1+\sqrt{1+s\bar\rho^2/d}/u)/(1+\sqrt{1+\bar\rho^2}/u)\asymp1$ for $k\in\{1,2\}$, while $(1+\sqrt{1+s\bar\rho^2/d}/u)/(1+\sqrt{1+\bar\rho^2}/u)\le1$ for $k\ge3$.
Since $c\bar\rho^2X_k/d\le x\le C\bar\rho^2X_k/d$ for $X_k\sim\chi_k^2$, Lemma~\ref{lem:chisq_denom_uniform} implies
\begin{align*}
&1+\frac{\bar\rho^2}{d}
\frac{1+\sqrt{1+s\bar\rho^2/d}/u}
{1+\sqrt{1+\bar\rho^2}/u}
\mathbb E_\Delta\left(\frac{1}{1+u^2+x}\right)
\asymp
\begin{cases}
\displaystyle
1+\frac{\bar\rho}{\sqrt{d(1+u^2)}}, & k=1,\\
\displaystyle
\log\left(e+\frac{\bar\rho^2}{d(1+u^2)}\right), & k=2,\\
1, & k\ge3.
\end{cases}
\end{align*}
Recall that $p_\star^{-1}\asymp1+\sqrt{1+\bar\rho^2}/u$ and $\zeta_\star^2\asymp1+\bar\rho^2/(1+u^2)$.  
We substitute these relations into the three cases and \eqref{eq:weighted_population_trace_assembly}  to obtain
 the partial fine-tuning result.

Finally, suppose that $s=d$.  By~\eqref{eq:trace_comparison},
with $x=0$ and $y=\rho_\Delta^2$, we obtain
\begin{align*}
&\min(\delta^2,\tilde\sigma^2)
\tr\left(\Sigma_1\bar G_{\mathrm{pop}}(\Delta)^{-1}\right)\asymp
\tilde\sigma^2\min(1,u^2)
\left(1+\frac{\sqrt{1+\rho_\Delta^2}}{u}\right)
\left(d+\frac{\rho_\Delta^2}{1+u^2}\right).
\end{align*}
Also, observe that the covariance bounds in Lemma~\ref{lem:pretraining_covariance_regularity}
and Gaussian moments imply $\mathbb E_\Delta\sqrt{1+\rho_\Delta^2}\asymp\sqrt{1+\bar\rho^2},$ and  $\mathbb E_\Delta\left[
\rho_\Delta^2\sqrt{1+\rho_\Delta^2}\right] \asymp\bar\rho^2\sqrt{1+\bar\rho^2}.$
Then, we take expectation over $\Delta$ in the pointwise bound and use the definitions of $p_\star$ and $\zeta_\star$.
We have full fine-tuning result:
\begin{align*}
\mathbb E_\Delta\left[
\min(\delta^2,\tilde\sigma^2)
\tr\left(\Sigma_1\bar G_{\mathrm{pop}}(\Delta)^{-1}\right)\right]
&\asymp\frac{\tilde\sigma^2\min(1,u^2)}{p_\star}
\left(d-1+\zeta_\star^2\right).
\end{align*}
\end{proof}

\begin{corollary}[Extended estimator rate for $\mathcal V_2$]
\label{cor:common_V2_extended}
Under the model in Section~\ref{sec:problem_setting} and
Assumption~\ref{ass:design}, fix $C_G\ge2$ as in
Definition~\ref{def:ext_estimator} and
$X_{\mathrm{pre}}\in\Omega_{\mathrm{pre}}$.
Let $m\ge C$, $1\le s\le d$, $u>0$, and $c_2>0$, and let
$p_\star$ and $\zeta_\star^2$ be defined by
\eqref{eq:common_conditional_scales}.
Let $\eta=m^{-C}$ for a sufficiently large fixed $C$.
Suppose that the sample condition
\eqref{eq:common_empirical_sample}, with $\eta$ replaced by $\eta^2$,
holds with a sufficiently large constant.
Conditioned on $X_{\mathrm{pre}}$, with probability at least $1-\eta$
over $\tilde X$, the extended estimator satisfies the following rates.

If $1\le s<d$, let $k=d-s$.  Then,
\begin{align*}
\mathcal V_2
&\lesssim
\frac{s\tilde\sigma^2}{mp_\star}
\left(
\min(1,u^2)+\frac{k\bar\rho^2}{d+k\bar\rho^2}
\right)g_{k,d}(\zeta_\star)
+C\left(\frac{d}{n}+\tilde\sigma^2\right)m^{-c_2},\\
\mathcal V_2
&\gtrsim
\frac{s\tilde\sigma^2}{mp_\star}
\left(
\min(1,u^2)+\frac{k\bar\rho^2}{d+k\bar\rho^2}
\right)g_{k,d}(\zeta_\star)
-C\left(\frac{d}{n}+\tilde\sigma^2\right)m^{-c_2}.
\end{align*}
If $s=d$, then
\begin{align*}
\mathcal V_2
&\lesssim
\frac{\tilde\sigma^2\min(1,u^2)}{mp_\star}
\left(d-1+\zeta_\star^2\right)
+C\left(\frac{d}{n}+\tilde\sigma^2\right)m^{-c_2},\\
\mathcal V_2
&\gtrsim
\frac{\tilde\sigma^2\min(1,u^2)}{mp_\star}
\left(d-1+\zeta_\star^2\right)
-C\left(\frac{d}{n}+\tilde\sigma^2\right)m^{-c_2}.
\end{align*}
\end{corollary}

\begin{proof}
We apply Corollary~\ref{cor:common_V2_population_representation} to
express $\mathcal V_2$ through the weighted population inverse trace.  We
evaluate this expectation using
Lemma~\ref{lem:common_weighted_population_inverse_trace} and prove the
corollary.
\end{proof}

%% file: Unified_appendix/app_10H1_small_u_analysis.tex

\section{Full Fine-Tuning Rate for $\mathcal V_1$}
\label{sec:full_V1_common}

In Section~\ref{sec:common_conditional_risk}, we have established the partial fine-tuning bound for $\mathcal V_1$ and the $\mathcal V_2$ rate for partial and full fine-tuning.
We complete the full fine-tuning analysis of $\mathcal V_1$ using the conditional mean bounds in Corollary~\ref{cor:full_conditional_mean_realized_covariates}.
First, we derive an expansion for $0<u\le1$, and then obtain the rate for every $u>0$.

We take expectation over $\Delta$ of the squared $\Sigma_1$ norm of the conditional residual mean, 
using~\eqref{eq:full_small_u_conditional_mean}.

\begin{corollary}[Full fine-tuning $\mathcal V_1$ expansion]
\label{cor:full_small_u_V1_extended}
Under the model in Section~\ref{sec:problem_setting} and
Assumption~\ref{ass:design}, fix $C_G\ge2$ as in
Definition~\ref{def:ext_estimator} and
$X_{\mathrm{pre}}\in\Omega_{\mathrm{pre}}$.
Let $m\ge C$, $s=d$, and $0<u\le1$, and let $p_\star$ and
$\zeta_\star^2$ be defined by~\eqref{eq:common_conditional_scales}.
Let $\eta=m^{-C}$ for a sufficiently large fixed $C$.  Suppose that the
sample condition~\eqref{eq:common_empirical_sample}, with $\eta$ replaced by
$\eta^2$, holds with a sufficiently large constant.  Conditioned on
$X_{\mathrm{pre}}$, with probability at least $1-\eta$ over $\tilde X$, the
extended estimator satisfies
\begin{align*}
\mathcal V_1
&=\frac{u^4}{9}\frac{\tr(\Sigma_1\Gamma)}{n}\{1+O(u^2)\}.
\end{align*}
\end{corollary}

\begin{proof}
Fix a realization of $\tilde X$ for which Corollary~\ref{cor:full_conditional_mean_realized_covariates} holds.
It follows from Lemma~\ref{lem:zero_active_bias} that $\mathcal V_1$ is the expectation over $\Delta$ of the squared $\Sigma_1$ norm of $\mathbb E_{\tilde\varepsilon}[r_{S,\mathrm{ext}}\mid\Delta,\tilde X]$.
We apply~\eqref{eq:full_small_u_conditional_mean}  to obtain, with conditional probability at least $1-\eta$ over $\Delta$,
\begin{align*}
\left|\left\|\mathbb E_{\tilde\varepsilon}
[r_{S,\mathrm{ext}}\mid\Delta,\tilde X]\right\|_{\Sigma_1}^2
-\frac{u^4}{9}\|\Delta\|_{\Sigma_1}^2\right|
&\le Cu^6\|\Delta\|_{\Sigma_1}^2.
\end{align*}

On the complementary set of $\Delta$ realizations, we apply~\eqref{eq:full_conditional_mean_global_bound}, 
the Cauchy--Schwarz inequality, and the Gaussian fourth moment bound in Lemma~\ref{lem:quad_form_concentration}.
The contributions of $\|\mathbb E_{\tilde\varepsilon}[r_{S,\mathrm{ext}}\mid\Delta,\tilde X]\|_{\Sigma_1}^2$ and $u^4\|\Delta\|_{\Sigma_1}^2/9$ after taking expectation over $\Delta$ are each at most $C m^{C_G}(d+\log m)\eta^{1/2}\tr(\Sigma_1\Gamma)/n\le C m^{-6}\tr(\Sigma_1\Gamma)/n$, for a sufficiently large exponent in $\eta=m^{-C}$.
Since $p_\star\le u$, the sample condition~\eqref{eq:common_empirical_sample} implies $m^{-6}\lesssim u^6$.
Finally, we take expectation over $\Delta$ and employ $\mathbb E_\Delta\|\Delta\|_{\Sigma_1}^2=\tr(\Sigma_1\Gamma)/n$ to obtain the claim.
\end{proof}

%% file: Unified_appendix/app_11H2_large_u_analysis.tex

For sufficiently small $u$, we employ the expansion in
Corollary~\ref{cor:full_small_u_V1_extended}.  When $u$ is bounded below by
a positive constant, we exploit
Corollary~\ref{cor:full_conditional_mean_realized_covariates}.  The following
corollary combines these two cases.

\begin{corollary}[Full fine-tuning rate for $\mathcal V_1$]
\label{cor:full_V1_direct}
Under the model in Section~\ref{sec:problem_setting} and
Assumption~\ref{ass:design}, fix $C_G\ge2$ as in
Definition~\ref{def:ext_estimator} and
$X_{\mathrm{pre}}\in\Omega_{\mathrm{pre}}$.
Let $m\ge C$, $s=d$, and $u>0$, and let $p_\star$ and
$\zeta_\star^2$ be defined by~\eqref{eq:common_conditional_scales}.
Let $\eta=m^{-C}$ for a sufficiently large fixed $C$.  Suppose that the
sample condition~\eqref{eq:common_empirical_sample}, with $\eta$ replaced by
$\eta^2$, holds with a sufficiently large constant.  Conditioned on
$X_{\mathrm{pre}}$, with probability at least $1-\eta$ over $\tilde X$, the
extended estimator satisfies
\begin{align*}
\mathcal V_1
&\asymp\min(1,u^4)\frac{\tr(\Sigma_1\Gamma)}{n}.
\end{align*}
\end{corollary}

\begin{proof}
Fix a sufficiently small $u_0\in(0,1]$.  For $0<u\le u_0$,
we obtain the stated bound from the expansion in Corollary~\ref{cor:full_small_u_V1_extended}. 
Suppose that $u>u_0$ and fix a realization of $\tilde X$ for which Corollary~\ref{cor:full_conditional_mean_realized_covariates} holds.
We apply~\eqref{eq:full_conditional_mean_general_bound} and the triangle inequality to obtain, with conditional probability at least $1-\eta$ over $\Delta$,
\begin{align*}
\frac{1}{2}(1-\theta_\Delta)\|\Delta\|_{\Sigma_1}
&\le
\left\|\mathbb E_{\tilde\varepsilon}
[r_{S,\mathrm{ext}}\mid\Delta,\tilde X]\right\|_{\Sigma_1}
\le\frac{3}{2}(1-\theta_\Delta)\|\Delta\|_{\Sigma_1}.
\end{align*}
For $s=d$, \eqref{eq:common_population_mean_target} and Corollary~\ref{cor:common_population_partial_mean_scale} imply $\mathbb E_\Delta[(1-\theta_\Delta)^2\|\Delta\|_{\Sigma_1}^2\mid X_{\mathrm{pre}}]\asymp\min(1,u^4)\tr(\Sigma_1\Gamma)/n$.

On the complementary set of $\Delta$ realizations, we apply~\eqref{eq:full_conditional_mean_global_bound} and the Cauchy--Schwarz inequality as in the proof of Corollary~\ref{cor:full_small_u_V1_extended}.
Since $0<1-\theta_\Delta<1$, the contributions of $\|\mathbb E_{\tilde\varepsilon} [r_{S,\mathrm{ext}}\mid\Delta,\tilde X]\|_{\Sigma_1}^2$ and $(1-\theta_\Delta)^2\|\Delta\|_{\Sigma_1}^2$, after taking expectation over $\Delta$, are each at most $C\tr(\Sigma_1\Gamma)m^{-2}/n$.
For sufficiently large $C$ in $m\ge C$, this contribution is bounded by a small fixed multiple 
of $\min(1,u^4)\tr(\Sigma_1\Gamma)/n$ since $u>u_0$.
Finally, we square the conditional mean bounds and take expectation over $\Delta$ to obtain the results.
\end{proof}

%% file: Unified_appendix/app_12aI_regime_i_ii_results.tex

\section{Regime Specializations and Main Results}
\label{sec:regime_specializations}

We prove the partial and full fine-tuning results in
Section~\ref{section:main_results} using the bounds for $\mathcal V_1$ and $\mathcal V_2$ in Sections~\ref{sec:common_conditional_risk} and~\ref{sec:full_V1_common}. 
First, we relate the quantities depending on $X_{\mathrm{pre}}$ to the deterministic scales.

\subsection{Regime Overview}

Recall that $\rho^2=d/(n\tilde\sigma^2)$.
On $\Omega_{\mathrm{pre}}$, it follows from Lemma~\ref{lem:pretraining_covariance_regularity} that
\begin{equation}\label{eq:conditional_deterministic_scale_transfer}
\bar\rho\asymp\rho,\qquad
\bar\rho_S^2\asymp\frac{s}{n\tilde\sigma^2},\qquad
\frac{\tr(\Sigma_1\Gamma)}{n}\asymp\frac{d}{n}.
\end{equation}
Moreover, note that $\tr(\Sigma_{1,-S\mid S}\Gamma_{-S,-S})/n\asymp(d-s)/n$ for $s<d$, and this trace is zero when $s=d$.

For partial fine-tuning with $1\le s<d$, recall $h_{d-s,d}$ from
Corollary~\ref{cor:common_population_partial_mean_scale} and $g_{d-s,d}$ from
Lemma~\ref{lem:common_weighted_population_inverse_trace}.  The relation
$\bar\rho\asymp\rho$ also implies
\begin{equation}\label{eq:conditional_deterministic_partial_transfer}
g_{d-s,d}(\bar\rho)\asymp g_{d-s,d}(\rho),\qquad
\frac{(d-s)\bar\rho^2}{d+(d-s)\bar\rho^2}
\asymp\frac{(d-s)\rho^2}{d+(d-s)\rho^2}.
\end{equation}
Then, we substitute $\bar\rho\asymp\rho$ into~\eqref{eq:common_conditional_scales} and obtain the following scales in each regime.
{\small
\begin{equation}\label{eq:conditional_deterministic_acceptance_transfer}
\begin{array}{c|ccc}
\text{Regime}&\text{Regime condition}&p_\star&\zeta_\star\\
\hline
I&\sqrt{d/n}\lesssim\delta&
p_\star\asymp p_{\regimeid{1}}=\min(1,u)&\zeta_\star\asymp1\\
II&\delta\lesssim\sqrt{d/n}\lesssim\tilde\sigma&
p_\star\asymp p_{\regimeid{2}}=u&\zeta_\star\asymp1\\
III&\delta\lesssim\tilde\sigma\lesssim\sqrt{d/n}&
p_\star\asymp p_{\regimeid{3}}=u/\rho&
\zeta_\star\asymp\sqrt{1+\rho^2}\\
IV&\tilde\sigma\lesssim\delta\lesssim\sqrt{d/n}&
p_\star\asymp p_{\regimeid{4}}=u/\rho&
\zeta_\star\asymp\rho/u
\end{array}
\end{equation}}
The following tables summarize the leading terms in the bounds for $\mathcal V_1$ and $\mathcal V_2$.
\paragraph{Partial fine-tuning.}
For $1\le s<d$, we apply Corollaries~\ref{cor:common_partial_V1_extended} and~\ref{cor:common_V2_extended} to obtain the following leading terms.
\begin{center}
\scriptsize
\setlength{\tabcolsep}{3pt}
\renewcommand{\arraystretch}{1.55}
\begin{tabular}{>{\raggedright\arraybackslash}p{0.08\linewidth}
>{\raggedright\arraybackslash}p{0.37\linewidth}
>{\raggedright\arraybackslash}p{0.41\linewidth}}
\toprule
\textbf{Regime}&\textbf{$\mathcal V_1$}&\textbf{$\mathcal V_2$}\\
\midrule
I&
$\displaystyle\frac{s}{n}\min(1,u^4)$&
$\displaystyle\frac{s\min(\delta^2,\tilde\sigma^2)}
{m p_{\regimeid{1}}}$\\[10pt]
II&
$\displaystyle\frac{s}{n}\left(
\left(\frac{(d-s)\rho^2}{d+(d-s)\rho^2}\right)^{2}+u^4\right)$&
$\displaystyle\frac{s}{m p_{\regimeid{2}}}\left(\delta^2+
\frac{\tilde\sigma^2(d-s)}{n\tilde\sigma^2+d-s}\right)$\\[12pt]
III&
$\displaystyle\frac{s}{n}\left(
\left(\frac{(d-s)\rho^2}{d+(d-s)\rho^2}\right)^{2}
+u^4\right)$&
$\displaystyle\frac{s\tilde\sigma^2}{m p_{\regimeid{3}}}
\left(u^2+\frac{(d-s)\rho^2}{d+(d-s)\rho^2}\right)g_{d-s,d}(\rho)$\\[18pt]
IV&
$\displaystyle\frac{s}{n}$&
$\displaystyle\frac{s\tilde\sigma^2}{m p_{\regimeid{4}}}
g_{d-s,d}(\rho/u)$\\
\bottomrule
\end{tabular}
\normalsize
\end{center}

\paragraph{Full fine-tuning.}
For $s=d$, we apply Corollaries~\ref{cor:full_V1_direct} and~\ref{cor:common_V2_extended} to obtain the following leading terms.
\begin{center}
\scriptsize
\setlength{\tabcolsep}{3pt}
\renewcommand{\arraystretch}{1.45}
\begin{tabular}{>{\raggedright\arraybackslash}p{0.08\linewidth}
>{\raggedright\arraybackslash}p{0.38\linewidth}
>{\raggedright\arraybackslash}p{0.38\linewidth}}
\toprule
\textbf{Regime}&\textbf{$\mathcal V_1$}&\textbf{$\mathcal V_2$}\\
\midrule
I&
$\displaystyle\frac{d}{n}\min(1,u^4)$&
$\displaystyle\frac{d\min(\delta^2,\tilde\sigma^2)}
{m p_{\regimeid{1}}}$\\[10pt]
II&
$\displaystyle\frac{d}{n} u^4$&
$\displaystyle\frac{d\delta^2}{m p_{\regimeid{2}}}$\\[10pt]
III&
$\displaystyle\frac{d}{n} u^4$&
$\displaystyle\frac{d\delta^2(1+1/(n\tilde\sigma^2))}{m p_{\regimeid{3}}}$\\[12pt]
IV&
$\displaystyle\frac{d}{n}$&
$\displaystyle\frac{d\tilde\sigma^2(1+1/(n\delta^2))}{m p_{\regimeid{4}}}$\\
\bottomrule
\end{tabular}
\normalsize
\end{center}

For the proofs below, take $m\ge C$ and $\eta=m^{-C}$, with $C$ sufficiently large for the prescribed $c_1,c_2$.
By~\eqref{eq:conditional_deterministic_acceptance_transfer} and $\log(m/\eta^2)\asymp\log m$, each regime sample condition implies~\eqref{eq:common_empirical_sample} with $\eta$ replaced by $\eta^2$.
Then, the required bounds  hold simultaneously with conditional probability at least $1-m^{-c_1}$ over $\tilde X$ for every $X_{\mathrm{pre}}\in\Omega_{\mathrm{pre}}$.
Let $\mathcal E$ denote this event in the regime under consideration.
By Lemma~\ref{lem:pretraining_covariance_regularity}, we obtain
\begin{equation}\label{eq:regime_joint_probability}
\mathbb P(\mathcal E^{\mathrm c})
\le\mathbb P(\Omega_{\mathrm{pre}}^{\mathrm c})
+\mathbb E\left[\mathbf 1_{\Omega_{\mathrm{pre}}}
\mathbb P(\mathcal E^{\mathrm c}\mid X_{\mathrm{pre}})\right]
\le2e^{-cn}+m^{-c_1}.
\end{equation}

\subsection{Regime~I}

We present the partial and full fine-tuning results in Regime~I.

\begin{theorem}[Partial fine-tuning in Regime~I]
Under the model in Section~\ref{sec:problem_setting} and
Assumption~\ref{ass:design}, let $1\le s<d$.
Suppose that \eqref{eq:I_conditions} holds and $m p_{\regimeid{1}}\ge C(s+\log m)(\log m)^{7/2}$.  Then, with
probability at least $1-2e^{-cn}-m^{-c_1}$ jointly over
$(X_{\mathrm{pre}},\tilde X)$, the extended estimator satisfies
\begin{align*}
\mathcal V_1
&\lesssim\frac{s}{n}\min(1,u^4)+C\mathrm{Rem}_{\mathcal V_1},
&\mathcal V_1
&\gtrsim\frac{s}{n}\min(1,u^4)-C\mathrm{Rem}_{\mathcal V_1},\\
\mathcal V_2
&\lesssim\frac{s\min(\delta^2,\tilde\sigma^2)}
{m p_{\regimeid{1}}}
+C\left(\frac{d}{n}+\tilde\sigma^2\right)m^{-c_2},
&\mathcal V_2
&\gtrsim\frac{s\min(\delta^2,\tilde\sigma^2)}
{m p_{\regimeid{1}}}
-C\left(\frac{d}{n}+\tilde\sigma^2\right)m^{-c_2},
\end{align*}
where $\mathrm{Rem}_{\mathcal V_1}:=d(s+\log m)/(mn)+s(s+(\log m)^2)(\log m)^3\min(\delta^2,\tilde\sigma^2)/(m p_{\regimeid{1}})^2$.
\end{theorem}

\begin{proof}
Consider a fixed $X_{\mathrm{pre}}\in\Omega_{\mathrm{pre}}$.
It follows from \eqref{eq:I_conditions},
\eqref{eq:conditional_deterministic_scale_transfer}, and
\eqref{eq:conditional_deterministic_acceptance_transfer} that
$\bar\rho\lesssim u$,
$p_\star\asymp p_{\regimeid{1}}=\min(1,u)$, and
$\zeta_\star\asymp1$.
For $\mathcal V_1$, the definition of $h_{d-s,d}$ in Corollary~\ref{cor:common_population_partial_mean_scale} implies $(d/(d+(d-s)\bar\rho^2))^2\le h_{d-s,d}(\bar\rho)\le1$.
Since $\bar\rho\lesssim u$, we obtain
\begin{align*}
\left(\frac{(d-s)\bar\rho^2}{d+(d-s)\bar\rho^2}\right)^2
+\min(1,u^4)h_{d-s,d}(\bar\rho)
&\asymp\min(1,u^4).
\end{align*}
For the remainder in Corollary~\ref{cor:common_partial_V1_extended}, $\bar\rho\lesssim u$ and $\bar\rho_S^2\le C\bar\rho^2$ imply $\max(1,\sqrt{1+\bar\rho^2}/\max(1,u))[1+\min(\bar\rho_S^2,(1+\bar\rho^2)/u^2)]\asymp1$.
Under \eqref{eq:I_conditions}, $\tilde\sigma^2(d-s)/(n\tilde\sigma^2+d-s)\le\min(\tilde\sigma^2,(d-s)/n)\lesssim\min(\delta^2,\tilde\sigma^2)$.
Then, we apply Corollary~\ref{cor:common_partial_V1_extended} and use $\tr(\Sigma_1\Gamma)/n\asymp d/n$ to obtain the stated upper and lower bounds for $\mathcal V_1$.

For $\mathcal V_2$, note that $\zeta_\star\asymp1$ and $\bar\rho\lesssim u$ imply $g_{d-s,d}(\zeta_\star)\asymp1$ and $\min(1,u^2)+(d-s)\bar\rho^2/(d+(d-s)\bar\rho^2)\asymp\min(1,u^2)$.
We apply Corollary~\ref{cor:common_V2_extended} and use $\tilde\sigma^2\min(1,u^2)=\min(\delta^2,\tilde\sigma^2)$ to obtain the stated upper and lower bounds for $\mathcal V_2$.
Finally, we apply~\eqref{eq:regime_joint_probability} and complete the proof.
\end{proof}

\begin{theorem}[Full fine-tuning in Regime~I]
Under the model in Section~\ref{sec:problem_setting} and
Assumption~\ref{ass:design}, suppose that $s=d$ and that
\eqref{eq:I_conditions} holds and $m p_{\regimeid{1}}\ge C(d+\log m)(\log m)^{7/2}$.  Then, with
probability at least $1-2e^{-cn}-m^{-c_1}$ jointly over
$(X_{\mathrm{pre}},\tilde X)$, the extended estimator satisfies
\begin{align*}
\mathcal V_1
&\asymp\min(1,u^4)\frac{d}{n},\\
\mathcal V_2
&\lesssim\frac{d\min(\delta^2,\tilde\sigma^2)}
{m p_{\regimeid{1}}}
+C\left(\frac{d}{n}+\tilde\sigma^2\right)m^{-c_2},
&\mathcal V_2
&\gtrsim\frac{d\min(\delta^2,\tilde\sigma^2)}
{m p_{\regimeid{1}}}
-C\left(\frac{d}{n}+\tilde\sigma^2\right)m^{-c_2}.
\end{align*}
\end{theorem}

\begin{proof}
Consider a fixed $X_{\mathrm{pre}}\in\Omega_{\mathrm{pre}}$.
It follows from \eqref{eq:conditional_deterministic_acceptance_transfer} that
$p_\star\asymp p_{\regimeid{1}}=\min(1,u)$ and
$\zeta_\star\asymp1$.
For $\mathcal V_1$, we apply Corollary~\ref{cor:full_V1_direct} and utilize 
$\tr(\Sigma_1\Gamma)/n\asymp d/n$ from
\eqref{eq:conditional_deterministic_scale_transfer} to obtain the
stated result.

For $\mathcal V_2$, $d-1+\zeta_\star^2\asymp d$ and $\tilde\sigma^2\min(1,u^2)=\min(\delta^2,\tilde\sigma^2)$.
We employ Corollary~\ref{cor:common_V2_extended} and obtain the stated result.
Finally, we apply~\eqref{eq:regime_joint_probability} to prove the theorem.
\end{proof}

\subsection{Regime~II}

We present the partial and full fine-tuning results in Regime~II.

\begin{theorem}[Partial fine-tuning in Regime~II]
Under the model in Section~\ref{sec:problem_setting} and
Assumption~\ref{ass:design}, let $1\le s<d$.
Suppose that
\eqref{eq:II_conditions} holds and $m p_{\regimeid{2}}\ge C(s+\log m)(\log m)^{7/2}$.  Then, with probability
at least $1-2e^{-cn}-m^{-c_1}$ jointly over
$(X_{\mathrm{pre}},\tilde X)$, the extended estimator satisfies
\begin{align*}
\mathcal V_1
&\lesssim\frac{s}{n}\left(
\left(\frac{(d-s)\rho^2}{d+(d-s)\rho^2}\right)^2+u^4
\right)+C\mathrm{Rem}_{\mathcal V_1},
&\mathcal V_1
&\gtrsim\frac{s}{n}\left(
\left(\frac{(d-s)\rho^2}{d+(d-s)\rho^2}\right)^2+u^4
\right)-C\mathrm{Rem}_{\mathcal V_1},\\
\mathcal V_2
&\asymp\frac{s}{m p_{\regimeid{2}}}\left(
\delta^2+\frac{\tilde\sigma^2(d-s)}{n\tilde\sigma^2+d-s}\right),
\end{align*}
where $\mathrm{Rem}_{\mathcal V_1}:=d(s+\log m)/(mn)+s(s+(\log m)^2)(\log m)^3(\delta^2+\tilde\sigma^2(d-s)/(n\tilde\sigma^2+d-s))/(m p_{\regimeid{2}})^2$.
\end{theorem}

\begin{proof}
Consider a fixed $X_{\mathrm{pre}}\in\Omega_{\mathrm{pre}}$.  It follows from \eqref{eq:II_conditions}, \eqref{eq:conditional_deterministic_scale_transfer}, and \eqref{eq:conditional_deterministic_acceptance_transfer} that $u\lesssim\bar\rho\lesssim1$, $p_\star\asymp p_{\regimeid{2}}=u$, and $\zeta_\star\asymp1$.
For $\mathcal V_1$, the definition of $h_{d-s,d}$ and $\bar\rho\lesssim1$ imply $h_{d-s,d}(\bar\rho)\asymp1$, while $u\lesssim1$ implies $\min(1,u^4)\asymp u^4$.
Then, by~\eqref{eq:conditional_deterministic_partial_transfer}, we obtain
\begin{align*}
\left(\frac{(d-s)\bar\rho^2}{d+(d-s)\bar\rho^2}\right)^2
+\min(1,u^4)h_{d-s,d}(\bar\rho)
&\asymp
\left(\frac{(d-s)\rho^2}{d+(d-s)\rho^2}\right)^2+u^4.
\end{align*}
For the remainder in Corollary~\ref{cor:common_partial_V1_extended}, $\bar\rho_S^2\le C\bar\rho^2\lesssim1$, which implies $\max(1,\sqrt{1+\bar\rho^2}/\max(1,u))[1+\min(\bar\rho_S^2,(1+\bar\rho^2)/u^2)]\asymp1$.
Under \eqref{eq:II_conditions}, it holds that $\min(\delta^2,\tilde\sigma^2)\asymp\delta^2$.
We apply Corollary~\ref{cor:common_partial_V1_extended} and use $\tr(\Sigma_1\Gamma)/n\asymp d/n$ to obtain the stated upper and lower bounds for $\mathcal V_1$.

For $\mathcal V_2$, note that $g_{d-s,d}(\zeta_\star)\asymp1$, $\tilde\sigma^2\min(1,u^2)\asymp\delta^2$, and $\tilde\sigma^2(d-s)\bar\rho^2/(d+(d-s)\bar\rho^2)\asymp\tilde\sigma^2(d-s)/(n\tilde\sigma^2+d-s)$.
We apply Corollary~\ref{cor:common_V2_extended} with a sufficiently large polynomial decay exponent.
Since $d/n\lesssim\tilde\sigma^2$ and $\delta^2=p_{\regimeid{2}}^2\tilde\sigma^2$, the sample condition implies that the complement remainder is bounded by $C\delta^2/(m^2p_{\regimeid{2}})$.
For sufficiently large $C$ in $m\ge C$, this is a sufficiently small multiple of the leading term, and we obtain the stated $\mathcal V_2$ rate.
Finally, we apply~\eqref{eq:regime_joint_probability} to obtain the claim.
\end{proof}

\begin{theorem}[Full fine-tuning in Regime~II]
Under the model in Section~\ref{sec:problem_setting} and
Assumption~\ref{ass:design}, suppose that $s=d$ and that
\eqref{eq:II_conditions} holds and $m p_{\regimeid{2}}\ge C(d+\log m)(\log m)^{7/2}$.  Then, with
probability at least $1-2e^{-cn}-m^{-c_1}$ jointly over
$(X_{\mathrm{pre}},\tilde X)$, the extended estimator satisfies
\begin{align*}
\mathcal V_1
\asymp u^4\frac{d}{n},\quad 
\mathcal V_2
\asymp\frac{d\delta^2}{m p_{\regimeid{2}}}.
\end{align*}
\end{theorem}

\begin{proof}
Consider a fixed $X_{\mathrm{pre}}\in\Omega_{\mathrm{pre}}$.  It follows from \eqref{eq:conditional_deterministic_acceptance_transfer} that $p_\star\asymp p_{\regimeid{2}}=u$ and $\zeta_\star\asymp1$.
For $\mathcal V_1$, we have $\min(1,u^4)\asymp u^4$.
Then, we invoke Corollary~\ref{cor:full_V1_direct}, $\tr(\Sigma_1\Gamma)/n\asymp d/n$ from~\eqref{eq:conditional_deterministic_scale_transfer}, and obtain the stated result.

For $\mathcal V_2$, note that $d-1+\zeta_\star^2\asymp d$ and $\tilde\sigma^2\min(1,u^2)\asymp\delta^2$. 
We apply Corollary~\ref{cor:common_V2_extended} and absorb the complement remainder by the same argument as in the partial fine-tuning proof to obtain the stated result.
Then, we apply~\eqref{eq:regime_joint_probability} and complete the proof.
\end{proof}

%% file: Unified_appendix/app_12bI_regime_iii_results.tex

\subsection{Regime~III}

We present the partial and full fine-tuning results in Regime~III.

\begin{theorem}[Partial fine-tuning in Regime~III]
Under the model in Section~\ref{sec:problem_setting} and
Assumption~\ref{ass:design}, let $1\le s<d$.
Suppose that \eqref{eq:III_conditions} holds and $m p_{\regimeid{3}}\ge C(1+d/(n\tilde\sigma^2))(s+\log m)(\log m)^{7/2}$.  Then, with probability
at least $1-2e^{-cn}-m^{-c_1}$ jointly over
$(X_{\mathrm{pre}},\tilde X)$, the extended estimator satisfies
\begin{align*}
\mathcal V_1
&\lesssim\frac{s}{n}\left(
\left(\frac{(d-s)\rho^2}{d+(d-s)\rho^2}\right)^2
+u^4\right)+C\mathrm{Rem}_{\mathcal V_1},
&\mathcal V_1
&\gtrsim\frac{s}{n}\left(
\left(\frac{(d-s)\rho^2}{d+(d-s)\rho^2}\right)^2
+u^4\right)-C\mathrm{Rem}_{\mathcal V_1},\\
\mathcal V_2
&\asymp\frac{s\tilde\sigma^2}{m p_{\regimeid{3}}}
\left(u^2+\frac{(d-s)\rho^2}{d+(d-s)\rho^2}\right)g_{d-s,d}(\rho),
\end{align*}
where
\begin{align*}
\mathrm{Rem}_{\mathcal V_1}
&:=\frac{d(s+\log m)}{mn}\sqrt{1+\rho^2}
\left(1+\frac{s\rho^2}{d}\right)\\
&\quad
+\frac{s(1+\rho^2)
\bigl(s+(1+\rho^2)(\log m)^2\bigr)(\log m)^3}
{(m p_{\regimeid{3}})^2}
\left(\delta^2+
\tilde\sigma^2\frac{(d-s)\rho^2}{d+(d-s)\rho^2}\right).
\end{align*}
\end{theorem}

\begin{proof}
Consider a fixed $X_{\mathrm{pre}}\in\Omega_{\mathrm{pre}}$.  It follows from \eqref{eq:III_conditions}, \eqref{eq:conditional_deterministic_scale_transfer}, and \eqref{eq:conditional_deterministic_acceptance_transfer} that $\bar\rho\asymp\rho$, $u\lesssim1\lesssim\bar\rho$, $p_\star\asymp p_{\regimeid{3}}=u/\rho$, and $\zeta_\star^2\asymp1+\rho^2$.
For $\mathcal V_1$, observe that $\min(1,u^4)\asymp u^4$.  The definition of $h_{d-s,d}$ in Corollary~\ref{cor:common_population_partial_mean_scale} implies $(1+(d-s)\bar\rho^2/d)^{-2}\le h_{d-s,d}(\bar\rho)\le1$.
When $(d-s)\bar\rho^2/d\le1$, we have $h_{d-s,d}(\bar\rho)\asymp1$.  When $(d-s)\bar\rho^2/d\ge1$, we have $((d-s)\bar\rho^2/(d+(d-s)\bar\rho^2))^2\asymp1$.
Since $\min(1,u^4)\le1$, it follows from \eqref{eq:conditional_deterministic_partial_transfer} that
\begin{align*}
\left(\frac{(d-s)\bar\rho^2}{d+(d-s)\bar\rho^2}\right)^2+\min(1,u^4)h_{d-s,d}(\bar\rho)
&\asymp\left(\frac{(d-s)\rho^2}{d+(d-s)\rho^2}\right)^2+u^4.
\end{align*}
For the remainder in Corollary~\ref{cor:common_partial_V1_extended}, \eqref{eq:conditional_deterministic_scale_transfer} implies $\bar\rho_S^2\le Cs\bar\rho^2/d$.
Since $\max(1,u)\asymp1$, we obtain
\begin{align*}
\max\left(1,\frac{\sqrt{1+\bar\rho^2}}{\max(1,u)}\right)
\left[1+\min\left(\bar\rho_S^2,\frac{1+\bar\rho^2}{u^2}\right)\right]
&\lesssim\sqrt{1+\bar\rho^2}\left(1+\frac{s\bar\rho^2}{d}\right).
\end{align*}
Under \eqref{eq:III_conditions}, we have $\min(\delta^2,\tilde\sigma^2)\asymp\delta^2$.
Moreover, $\tilde\sigma^2(d-s)/(n\tilde\sigma^2+d-s)=\tilde\sigma^2(d-s)\rho^2/(d+(d-s)\rho^2)$.
We employ Corollary~\ref{cor:common_partial_V1_extended}, $\bar\rho\asymp\rho$ and $\tr(\Sigma_1\Gamma)/n\asymp d/n$ to obtain the stated upper and lower bounds for $\mathcal V_1$.

For $\mathcal V_2$, we have $\min(1,u^2)\asymp u^2$ and $\zeta_\star\asymp\sqrt{1+\rho^2}\asymp\rho$, which imply $g_{d-s,d}(\zeta_\star)\asymp g_{d-s,d}(\rho)$.
We exploit Corollary~\ref{cor:common_V2_extended} and~\eqref{eq:conditional_deterministic_partial_transfer}
and choose  the  polynomial decay exponent sufficiently large.
Since $d/n+\tilde\sigma^2\asymp d/n=\delta^2/p_{\regimeid{3}}^2$, the sample condition implies that the complement remainder is bounded by $C\delta^2/(m^2p_{\regimeid{3}})$.
For sufficiently large $C$ in $m\ge C$, this is a sufficiently small multiple of the leading term, and we obtain the stated $\mathcal V_2$ rate.
Finally, we apply~\eqref{eq:regime_joint_probability} to obtain the claim.
\end{proof}

\begin{theorem}[Full fine-tuning in Regime~III]
Under the model in Section~\ref{sec:problem_setting} and
Assumption~\ref{ass:design}, suppose that $s=d$ and that
\eqref{eq:III_conditions} holds and $m p_{\regimeid{3}}\ge C(1+d/(n\tilde\sigma^2))(d+\log m)(\log m)^{7/2}$.  Then, with
probability at least $1-2e^{-cn}-m^{-c_1}$ jointly over
$(X_{\mathrm{pre}},\tilde X)$, the extended estimator satisfies
\begin{align*}
\mathcal V_1
\asymp u^4\frac{d}{n},\quad 
\mathcal V_2
\asymp\frac{\delta^2(d+\rho^2)}{m p_{\regimeid{3}}}.
\end{align*}
\end{theorem}

\begin{proof}
Consider a fixed $X_{\mathrm{pre}}\in\Omega_{\mathrm{pre}}$.  
It follows from \eqref{eq:III_conditions}, \eqref{eq:conditional_deterministic_scale_transfer},
 and \eqref{eq:conditional_deterministic_acceptance_transfer} that $u\lesssim1$,
$\bar\rho\asymp\rho$, $p_\star\asymp p_{\regimeid{3}}=u/\rho$, and $\zeta_\star^2\asymp1+\rho^2$.  
For $\mathcal V_1$, $\min(1,u^4)\asymp u^4$.  
Using Corollary~\ref{cor:full_V1_direct} and $\tr(\Sigma_1\Gamma)/n\asymp d/n$ from \eqref{eq:conditional_deterministic_scale_transfer}, 
we obtain the stated result.
For $\mathcal V_2$, $\min(1,u^2)\asymp u^2$, $d-1+\zeta_\star^2\asymp d+\rho^2$, and $\tilde\sigma^2u^2=\delta^2$.  
By invoking Corollary~\ref{cor:common_V2_extended} with $c_2=3$, we  obtain the stated result.  
Since $d/n+\tilde\sigma^2\asymp d/n=\delta^2/p_{\regimeid{3}}^2$, the sample condition implies $(d/n+\tilde\sigma^2)m^{-3}/\{\delta^2/(m p_{\regimeid{3}})\}\lesssim m^{-1}$.  Therefore, for sufficiently large $C$ in $m\ge C$, the complement term is a sufficiently small fixed multiple of the stated $\mathcal V_2$ rate.  
By applying \eqref{eq:regime_joint_probability} to the
 $\mathcal V_1$ and $\mathcal V_2$ results, we prove the theorem.
\end{proof}

%% file: Unified_appendix/app_12cI_regime_iv_results.tex

\subsection{Regime~IV}

We present the partial and full fine-tuning results in Regime~IV.

\begin{theorem}[Partial fine-tuning in Regime~IV]
Under the model in Section~\ref{sec:problem_setting} and
Assumption~\ref{ass:design}, let $1\le s<d$.
Suppose that \eqref{eq:IV_conditions} holds and $m p_{\regimeid{4}}\ge C d(s+\log m)(\log m)^{7/2}/(n\delta^2)$.  Then, with probability
at least $1-2e^{-cn}-m^{-c_1}$ jointly over
$(X_{\mathrm{pre}},\tilde X)$, the extended estimator satisfies
\begin{align*}
\mathcal V_1
&\lesssim\frac{s}{n}+C\mathrm{Rem}_{\mathcal V_1},
&\mathcal V_1
&\gtrsim\frac{s}{n}-C\mathrm{Rem}_{\mathcal V_1},\\
\mathcal V_2
&\lesssim\frac{s\tilde\sigma^2}{m p_{\regimeid{4}}}
g_{d-s,d}(\rho/u)
+C\left(\frac{d}{n}+\tilde\sigma^2\right)m^{-c_2},
&\mathcal V_2
&\gtrsim\frac{s\tilde\sigma^2}{m p_{\regimeid{4}}}
g_{d-s,d}(\rho/u)
-C\left(\frac{d}{n}+\tilde\sigma^2\right)m^{-c_2},
\end{align*}
where
\begin{align*}
\mathrm{Rem}_{\mathcal V_1}
:=\frac{d(s+\log m)}{mn}
\frac{\rho}{u}
\left[1+\min\left(\frac{s}{n\tilde\sigma^2},
\frac{\rho^2}{u^2}\right)\right]
+\frac{s(\rho^2/u^2)
\bigl(s+(\rho^2/u^2)(\log m)^2\bigr)(\log m)^3}
{(m p_{\regimeid{4}})^2}\tilde\sigma^2.
\end{align*}
\end{theorem}

\begin{proof}
Consider a fixed $X_{\mathrm{pre}}\in\Omega_{\mathrm{pre}}$.  It follows from \eqref{eq:IV_conditions}, \eqref{eq:conditional_deterministic_scale_transfer}, and \eqref{eq:conditional_deterministic_acceptance_transfer} that $1\lesssim u\lesssim\bar\rho$, $\bar\rho\asymp\rho$, $p_\star\asymp p_{\regimeid{4}}=u/\rho$, and $\zeta_\star^2\asymp\rho^2/u^2$.

For $\mathcal V_1$, we have $\min(1,u^4)\asymp1$.  As in the Regime~III proof, $(1+(d-s)\bar\rho^2/d)^{-2}\le h_{d-s,d}(\bar\rho)\le1$.
Hence, we obtain
\begin{align*}
\left(\frac{(d-s)\bar\rho^2}{d+(d-s)\bar\rho^2}\right)^2+\min(1,u^4)h_{d-s,d}(\bar\rho)&\asymp1.
\end{align*}
For the remainder in Corollary~\ref{cor:common_partial_V1_extended}, \eqref{eq:conditional_deterministic_scale_transfer} implies $\bar\rho_S^2\le Cs/(n\tilde\sigma^2)$.
Since $\max(1,u)\asymp u$, we obtain
\begin{align*}
\max\left(1,\frac{\sqrt{1+\bar\rho^2}}{\max(1,u)}\right)
\left[1+\min\left(\bar\rho_S^2,\frac{1+\bar\rho^2}{u^2}\right)\right]\lesssim\frac{\bar\rho}{u}
\left[1+\min\left(\frac{s}{n\tilde\sigma^2},\frac{\bar\rho^2}{u^2}\right)\right].
\end{align*}
Moreover, $\min(\delta^2,\tilde\sigma^2)+\tilde\sigma^2(d-s)/(n\tilde\sigma^2+d-s)\asymp\tilde\sigma^2$.
We apply Corollary~\ref{cor:common_partial_V1_extended} and use $\bar\rho\asymp\rho$ and $\tr(\Sigma_1\Gamma)/n\asymp d/n$ to obtain the stated upper and lower bounds for $\mathcal V_1$.

For $\mathcal V_2$, $\min(1,u^2)+(d-s)\bar\rho^2/(d+(d-s)\bar\rho^2)\asymp1$, while $\zeta_\star\asymp\rho/u$ and the definition of $g_{d-s,d}$ imply
$g_{d-s,d}(\zeta_\star)\asymp g_{d-s,d}(\rho/u)$.
We use Corollary~\ref{cor:common_V2_extended} to have
the stated $\mathcal V_2$ result.
By applying \eqref{eq:regime_joint_probability} to the $\mathcal V_1$ and $\mathcal V_2$ results, we prove the theorem.
\end{proof}

\begin{theorem}[Full fine-tuning in Regime~IV]
Under the model in Section~\ref{sec:problem_setting} and
Assumption~\ref{ass:design}, suppose that $s=d$ and that
\eqref{eq:IV_conditions} holds and $m p_{\regimeid{4}}\ge C d(d+\log m)(\log m)^{7/2}/(n\delta^2)$.  Then, with
probability at least $1-2e^{-cn}-m^{-c_1}$ jointly over
$(X_{\mathrm{pre}},\tilde X)$, the extended estimator satisfies
\begin{align*}
\mathcal V_1
&\asymp\frac{d}{n},\\
\mathcal V_2
&\lesssim\frac{\tilde\sigma^2(d+\rho^2/u^2)}{m p_{\regimeid{4}}}
+C\left(\frac{d}{n}+\tilde\sigma^2\right)m^{-c_2},
&\mathcal V_2
&\gtrsim\frac{\tilde\sigma^2(d+\rho^2/u^2)}{m p_{\regimeid{4}}}
-C\left(\frac{d}{n}+\tilde\sigma^2\right)m^{-c_2}.
\end{align*}
\end{theorem}

\begin{proof}
Consider a fixed $X_{\mathrm{pre}}\in\Omega_{\mathrm{pre}}$.  It follows 
from \eqref{eq:IV_conditions}, \eqref{eq:conditional_deterministic_scale_transfer}, and \eqref{eq:conditional_deterministic_acceptance_transfer} that $1\lesssim u$, 
$p_\star\asymp p_{\regimeid{4}}=u/\rho$, and $\zeta_\star^2\asymp\rho^2/u^2$.
For $\mathcal V_1$, $\min(1,u^4)\asymp1$.  We apply Corollary~\ref{cor:full_V1_direct} and utilize $\tr(\Sigma_1\Gamma)/n\asymp d/n$ from \eqref{eq:conditional_deterministic_scale_transfer}. Then, we obtain the stated result.

For $\mathcal V_2$, $\min(1,u^2)\asymp1$ and, since $\rho^2/u^2\gtrsim1$, $d-1+\zeta_\star^2\asymp d+\rho^2/u^2$.
From Corollary~\ref{cor:common_V2_extended}, the stated result follows.
By applying \eqref{eq:regime_joint_probability} to the $\mathcal V_1$ and $\mathcal V_2$ results, we prove the theorem.
\end{proof}

%% file: Unified_appendix/app_15J_moment_tools.tex

\section{Moment Bounds for Accepted Samples and Pre-Training Error}
\label{sec:common_moment_tools}
In this section, we establish the moment bounds used in the population and empirical analysis.
First, we consider moments conditioned on acceptance.
Then, we derive chi-square moment bounds and apply them to the Gaussian pre-training error.
The accepted linear form bound is also used in
the iterative analysis in Section~\ref{sec:iterative_training_proofs}.

\subsection{Accepted Sample Moments}

For fixed $\Delta$, we evaluate the second moments of $a=-\tilde x^\top\Delta$ and $v^\top\tilde x$ conditioned on $|\xi|\le\delta$.
We also bound the conditional $\psi_2$-norm of $v^\top\tilde x$ in terms of its conditional second moment.

\begin{lemma}[Accepted linear form moments]
\label{lem:accepted_linear_form_moments}
Let $\delta,\tilde\sigma>0$,  $\tilde x\sim\mathcal N(0,\Sigma_1)$, 
and
$\tilde\varepsilon\sim\mathcal N(0,\tilde\sigma^2)$ be independent, and
fix $\Delta\in\mathbb R^d$.  Let $a:=-\tilde x^\top\Delta$,
$\xi:=a+\tilde\varepsilon$, and
$\sigma_\xi^2(\Delta):=\|\Delta\|_{\Sigma_1}^2+\tilde\sigma^2$.
Define
$\lambda_{2,\Delta}:=\mathbb E[\xi^2\mid |\xi|\le\delta,\Delta]/
\sigma_\xi^2(\Delta)$.  Then,
\begin{align*}
\mathbb E[a^2\mid |\xi|\le\delta,\Delta]
&=\|\Delta\|_{\Sigma_1}^2
\frac{\tilde\sigma^2+\lambda_{2,\Delta}\|\Delta\|_{\Sigma_1}^2}
{\sigma_\xi^2(\Delta)}
\le\min\left(\|\Delta\|_{\Sigma_1}^2,
\tilde\sigma^2+\delta^2\right).
\end{align*}
Moreover, there exists a universal constant $C>0$ such that, for every
$v\in\mathbb R^d$,
\begin{align*}
\mathbb E[(v^\top\tilde x)^2\mid |\xi|\le\delta,\Delta]
&=v^\top\Sigma_1v-
\frac{(1-\lambda_{2,\Delta})(v^\top\Sigma_1\Delta)^2}
{\sigma_\xi^2(\Delta)},\\
\|v^\top\tilde x\|_{\psi_2}^2
&\le C\mathbb E[(v^\top\tilde x)^2\mid |\xi|\le\delta,\Delta]
\le C v^\top\Sigma_1v,
\end{align*}
where the $\psi_2$-norm is evaluated conditioned on $|\xi|\le\delta$
and $\Delta$.
\end{lemma}

\begin{proof}
Conditioned on $\Delta$, the pair $(a,\xi)$ is centered jointly Gaussian.
Since $\xi=a+\tilde\varepsilon$, we have  $\Cov(a,\xi\mid\Delta)=\|\Delta\|_{\Sigma_1}^2$ 
and
$\Var(\xi\mid\Delta)=\sigma_\xi^2(\Delta)$. 
 Hence,
\begin{align*}
a\mid\xi,\Delta
&\sim\mathcal N\biggl(
\frac{\|\Delta\|_{\Sigma_1}^2}{\sigma_\xi^2(\Delta)}\xi,
\frac{\|\Delta\|_{\Sigma_1}^2\tilde\sigma^2}
{\sigma_\xi^2(\Delta)}
\biggr).
\end{align*}
We take the expectation of this conditional second moment over $\xi$ conditioned
on $|\xi|\le\delta$, and obtain
\begin{align*}
\mathbb E[a^2\mid |\xi|\le\delta,\Delta]
=\frac{\|\Delta\|_{\Sigma_1}^2\tilde\sigma^2}
{\sigma_\xi^2(\Delta)}
+\frac{\|\Delta\|_{\Sigma_1}^4}{\sigma_\xi^4(\Delta)}
\mathbb E[\xi^2\mid |\xi|\le\delta,\Delta]
=\|\Delta\|_{\Sigma_1}^2
\frac{\tilde\sigma^2+\lambda_{2,\Delta}
\|\Delta\|_{\Sigma_1}^2}{\sigma_\xi^2(\Delta)},
\end{align*}
where the last equality follows from the definition of $\lambda_{2,\Delta}$.
Since Lemma~\ref{lem:truncated_normal_moments} provides
$\lambda_{2,\Delta}\le1$, this identity leads to
$\mathbb E[a^2\mid |\xi|\le\delta,\Delta]
\le\|\Delta\|_{\Sigma_1}^2$.
The two terms in this decomposition are bounded by $\tilde\sigma^2$ and
$\mathbb E[\xi^2\mid |\xi|\le\delta,\Delta]\le\delta^2$, respectively.
Hence, $\mathbb E[a^2\mid |\xi|\le\delta,\Delta] \le\tilde\sigma^2+\delta^2$.

Since $\Cov(v^\top\tilde x,\xi\mid\Delta)=-v^\top\Sigma_1\Delta$, Gaussian conditioning implies
\begin{align}\label{eq:accepted_linear_conditional_second_moment}
\mathbb E[(v^\top\tilde x)^2\mid\xi,\Delta]
&=v^\top\Sigma_1v-
\frac{(v^\top\Sigma_1\Delta)^2}{\sigma_\xi^2(\Delta)}
+\frac{(v^\top\Sigma_1\Delta)^2}{\sigma_\xi^4(\Delta)}\xi^2.
\end{align}
We take expectation over $\xi$ conditioned on $(|\xi|\le\delta,\Delta)$ in~\eqref{eq:accepted_linear_conditional_second_moment} and obtain the stated second moment formula.
Its upper bound follows from $\lambda_{2,\Delta}\le1$.

It remains to prove the subgaussian bound.  Since $|\xi|\le\delta$ under acceptance, we have $\|\xi\|_{\psi_2}^2\le C\delta^2$.
Moreover, since $e^{\xi^2/(4\sigma_\xi^2(\Delta))}$ is increasing in $|\xi|$, we have $\mathbb E[e^{\xi^2/(4\sigma_\xi^2(\Delta))}\mid|\xi|\le\delta,\Delta]\le\mathbb E[e^{\xi^2/(4\sigma_\xi^2(\Delta))}\mid\Delta]=\sqrt2<2$.
Hence, $\|\xi\|_{\psi_2}^2\le4\sigma_\xi^2(\Delta)$.
Then, the bounds for $\lambda_2(h)$ in Lemma~\ref{lem:truncated_normal_moments} lead to
\begin{align}\label{eq:accepted_xi_subgaussian_bound}
\|\xi\|_{\psi_2}^2
&\le C\min(\delta^2,\sigma_\xi^2(\Delta))
\le C\lambda_{2,\Delta}\sigma_\xi^2(\Delta)
=C\mathbb E[\xi^2\mid |\xi|\le\delta,\Delta].
\end{align}

Similarly, by Gaussian conditioning, $v^\top\tilde x=-(v^\top\Sigma_1\Delta)\xi/\sigma_\xi^2(\Delta)+
\left(v^\top\tilde x+(v^\top\Sigma_1\Delta)\xi/\sigma_\xi^2(\Delta)\right)$, where the second term 
is centered Gaussian and independent of $\xi$ given $\Delta$, 
with variance $v^\top\Sigma_1v-(v^\top\Sigma_1\Delta)^2/\sigma_\xi^2(\Delta)$.
Finally, by the triangle inequality, \eqref{eq:accepted_linear_conditional_second_moment}, and \eqref{eq:accepted_xi_subgaussian_bound}, 
it follows that
\begin{align*}
\|v^\top\tilde x\|_{\psi_2}^2
&\le C\left(v^\top\Sigma_1v-\frac{(v^\top\Sigma_1\Delta)^2}{\sigma_\xi^2(\Delta)}
+\frac{(v^\top\Sigma_1\Delta)^2}{\sigma_\xi^4(\Delta)}\|\xi\|_{\psi_2}^2\right)\\
&\le C\left(v^\top\Sigma_1v-\frac{(v^\top\Sigma_1\Delta)^2}{\sigma_\xi^2(\Delta)}
+\frac{(v^\top\Sigma_1\Delta)^2}{\sigma_\xi^4(\Delta)}\mathbb E[\xi^2\mid|\xi|\le\delta,\Delta]\right)\\
&=C\mathbb E[(v^\top\tilde x)^2\mid|\xi|\le\delta,\Delta]
\le C v^\top\Sigma_1v.
\end{align*}
\end{proof}

Next, we derive the conditional mean expansion for $\xi=a+\tilde\varepsilon$ with $a$ fixed and $0<u=\delta/\tilde\sigma\le1$.
This expansion is used in Proposition~\ref{prop:full_accepted_gram_mean} with $a=-\tilde x_i^\top\Delta$.

\begin{lemma}[Accepted first moment for $0<u\le1$]
\label{lem:accepted_first_moment_small_window}
Let $\tilde\sigma>0$, $0<\delta\le\tilde\sigma$, and
$\tilde\varepsilon\sim\mathcal N(0,\tilde\sigma^2)$.  For fixed $a\in\mathbb R$, let $\xi=a+\tilde\varepsilon$ and 
$u=\delta/\tilde\sigma$.
There exists a universal constant $C>0$ such that
\begin{align*}
\left|\mathbb E_{\tilde\varepsilon}(\xi\mid |\xi|\le\delta,a)
-\frac{u^2}{3}a\right|
&\le Cu^4\left(1+\frac{a^2}{\tilde\sigma^2}\right)|a|.
\end{align*}
\end{lemma}

\begin{proof}
We substitute $x=\tilde\sigma uz$ into the integrals defining the conditional mean and obtain
\begin{align}\label{eq:accepted_first_moment_representation}
\mathbb E_{\tilde\varepsilon}
(\xi\mid |\xi|\le\delta,a)
&=\tilde\sigma u
\frac{\int_{-1}^1 z e^{auz/\tilde\sigma}e^{-u^2z^2/2}dz}
{\int_{-1}^1 e^{auz/\tilde\sigma}e^{-u^2z^2/2}dz}.
\end{align}
First, suppose that $|au/\tilde\sigma|\le1$.  Then by the Taylor expansion for $|z|\le1$,
we have 
\begin{align*}
e^{auz/\tilde\sigma} &=1+\frac{au}{\tilde\sigma}z
+\frac{a^2u^2}{2\tilde\sigma^2}z^2 +O\left(\frac{|a|^3u^3}{\tilde\sigma^3}|z|^3\right).
\end{align*}
Since $e^{-u^2z^2/2}$ is even and $e^{-u^2z^2/2}=1+O(u^2z^2)$, the integration results in
\begin{align}\label{eq:accepted_first_moment_integral_expansions}
\begin{aligned}
\int_{-1}^1 z e^{auz/\tilde\sigma}e^{-u^2z^2/2}dz
&=\frac{2au}{3\tilde\sigma}
+O\left(\frac{u^3|a|}{\tilde\sigma}
\left(1+\frac{a^2}{\tilde\sigma^2}\right)\right),\\
\int_{-1}^1 e^{auz/\tilde\sigma}e^{-u^2z^2/2}dz
&=2+O\left(u^2\left(1+\frac{a^2}{\tilde\sigma^2}\right)\right).
\end{aligned}
\end{align}
Since $u\le1$ and $|au/\tilde\sigma|\le1$, we have $\int_{-1}^1 e^{auz/\tilde\sigma}e^{-u^2z^2/2}dz\ge c$.
Hence, by \eqref{eq:accepted_first_moment_integral_expansions} it holds that the reciprocal of this integral is $1/2+O(u^2(1+a^2/\tilde\sigma^2))$.
Moreover, note that $u^2(1+a^2/\tilde\sigma^2)=u^2+(au/\tilde\sigma)^2\le2$.
Therefore, \eqref{eq:accepted_first_moment_representation} and \eqref{eq:accepted_first_moment_integral_expansions} imply
\begin{align*}
\mathbb E_{\tilde\varepsilon}(\xi\mid |\xi|\le\delta,a)
=\tilde\sigma u\left\{\frac{au}{3\tilde\sigma}
+O\left(\frac{u^3|a|}{\tilde\sigma}\left(1+\frac{a^2}{\tilde\sigma^2}\right)\right)\right\}
=\frac{u^2}{3}a +O\left(u^4\left(1+\frac{a^2}{\tilde\sigma^2}\right)|a|\right).
\end{align*}
Finally, suppose that $|au/\tilde\sigma|>1$.  Since $|\xi|\le\delta$ under acceptance,
\begin{align*}
\left|\mathbb E_{\tilde\varepsilon}(\xi\mid |\xi|\le\delta,a)
-\frac{u^2}{3}a\right|
&\le\tilde\sigma u+\frac{u^2}{3}|a|
\le\frac{Cu^4|a|^3}{\tilde\sigma^2}
\le Cu^4\left(1+\frac{a^2}{\tilde\sigma^2}\right)|a|.
\end{align*}
Combining the two cases, we obtain the claimed bound.
\end{proof}

\subsection{Chi-Square Moment Bounds}

The following lemma provides the chi-square moment bounds used for the
pre-training error.

\begin{lemma}[Chi-square moment bounds]
\label{lem:chisq_denom_uniform}
Let $k\ge1$ be an integer and $X_k\sim\chi_k^2$.
For every $t>0$ and $r\in\{1,2\}$,
\begin{align*}
\mathbb E[(1+tX_k)^{-r}]\asymp
\begin{cases}
(1+t)^{-k/2}, & k<2r,\\
(1+t)^{-r}\log(e+t), & k=2r\\
(1+kt)^{-r}, & k>2r
\end{cases},
\quad \mathbb E\left[\left(\frac{tX_k}{1+tX_k}\right)^r\right]
\asymp\left(\frac{kt}{1+kt}\right)^r.
\end{align*}
Moreover, for every $t>0$,
\begin{align*}
\mathbb E\left[\frac{tX_k}{(1+tX_k)^2}\right]
&\asymp\frac{kt}{1+kt}\mathbb E[(1+tX_k)^{-1}].
\end{align*}
\end{lemma}

\begin{proof}
First, we bound $\mathbb E[(1+tX_k)^{-r}]$.  Suppose that $k\le2r$.
For $0<t\le1$, by the Jensen's inequality and $(1+tX_k)^{-r}\le1$, it holds that $(1+kt)^{-r}\le\mathbb E[(1+tX_k)^{-r}]\le1$.  
Since $1\le k\le2r\le4$, it follows that $\mathbb E[(1+tX_k)^{-r}]\asymp1$, which agrees with the stated rate.
For $t\ge1$, integration against the chi-square density and the substitution $y=tx$ imply
\begin{align*}
\mathbb E[(1+tX_k)^{-r}]
&=\frac{t^{-k/2}}{2^{k/2}\Gamma(k/2)}
\int_0^\infty
\frac{y^{k/2-1}e^{-y/(2t)}}{(1+y)^r}dy.
\end{align*}
Note that integral over $(0,1)$ is bounded above and below by universal positive constants and on 
$(1,t)$, $y^{k/2-1}e^{-y/(2t)}/(1+y)^r\asymp y^{k/2-r-1}$.  
Moreover, on $(t,\infty)$, the substitution $y=tx$ and $k\le2r$ bound the integral by $Ct^{k/2-r}\le C$.  Hence,
\begin{align*}
\int_0^\infty\frac{y^{k/2-1}e^{-y/(2t)}}{(1+y)^r}dy
&\asymp1+\int_1^t y^{k/2-r-1}dy
\asymp
\begin{cases}
1, & k<2r,\\
\log(e+t), & k=2r.
\end{cases}
\end{align*}
This proves the stated bound for $k\le2r$.

Now, suppose that $k>2r$.  Similar to the previous case, Jensen's inequality and $\mathbb E[X_k^{-r}]=\prod_{j=1}^r(k-2j)^{-1}\asymp k^{-r}$ 
result in
\begin{align*}
(1+kt)^{-r}&\le\mathbb E[(1+tX_k)^{-r}]
\le\min\left(1,t^{-r}\mathbb E[X_k^{-r}]\right)
\le C(1+kt)^{-r}.
\end{align*}
Thus, the first stated bound holds for every $k\ge1$.
For the rest, let $X_{k+2}\sim\chi_{k+2}^2$ and $X_{k+4}\sim\chi_{k+4}^2$.  Integration against the chi-square densities implies
\begin{align*}
\mathbb E\left[\frac{tX_k}{1+tX_k}\right]
&=kt\mathbb E[(1+tX_{k+2})^{-1}]
\asymp\frac{kt}{1+(k+2)t}\asymp\frac{kt}{1+kt},\\
\mathbb E\left[\left(\frac{tX_k}{1+tX_k}\right)^2\right]
&=k(k+2)t^2\mathbb E[(1+tX_{k+4})^{-2}]
\asymp\frac{k(k+2)t^2}{(1+(k+4)t)^2}
\asymp\left(\frac{kt}{1+kt}\right)^2,
\end{align*}
where we use the first statement with $k+2,r=1$ and $k+4,r=2$, respectively.

Finally, we note that integration against the chi-square density implies $\mathbb E[tX_k/(1+tX_k)^2]=kt\mathbb E[(1+tX_{k+2})^{-2}]$.  
We use the first statement with $k+2$ in place of $k$ and $r=2$, and obtain
\begin{align*}
\mathbb E\left[\frac{tX_k}{(1+tX_k)^2}\right]
\asymp
\begin{cases}
t(1+t)^{-3/2}, & k=1,\\
t(1+t)^{-2}\log(e+t), & k=2,\\
kt(1+kt)^{-2}, & k\ge3,
\end{cases}
\asymp\frac{kt}{1+kt}\mathbb E[(1+tX_k)^{-1}],
\end{align*}
where the last relation follows from the first statement with $r=1$ in the same three cases.
Thus, we prove all the stated bounds.
\end{proof}

\subsection{Pre-Training Error Moments}

Conditioned on $X_{\mathrm{pre}}$, $\Delta_{-S}$ is centered Gaussian, 
and we evaluate the marginal moments of 
$\|\Delta_{-S}\|_{\Sigma_{1,-S\mid S}}^2$ and its joint moments with 
$\|(\Sigma_1\Delta)_S\|_{\Sigma_{1,SS}^{-1}}^2$.  In this subsection, 
$\mathbb E_\Delta$ and $\Cov_\Delta$ denote expectation and covariance 
over $\Delta$ given $X_{\mathrm{pre}}$.

\begin{lemma}[Marginal moments for frozen coordinates]
\label{lem:frozen_denominator_marginal}
Under the pre-training model in Section~\ref{sec:problem_setting} and
Assumption~\ref{ass:design}, fix $X_{\mathrm{pre}}\in\Omega_{\mathrm{pre}}$
and suppose that $1\le s<d$.  Let $k=d-s$.  Then, for each
$r\in\{1,2\}$,
\begin{align*}
\mathbb E_\Delta\left[
\left(1+\frac{\|\Delta_{-S}\|_{\Sigma_{1,-S\mid S}}^2}
{\tilde\sigma^2}\right)^{-r}\right]
\asymp
\begin{cases}
(1+\bar\rho^2/d)^{-k/2}, & k<2r,\\
(1+\bar\rho^2/d)^{-r}\log(e+\bar\rho^2/d), & k=2r,\\
(1+k\bar\rho^2/d)^{-r}, & k>2r.
\end{cases}
\end{align*}
Moreover, for each $r\in\{1,2\}$,
\begin{align*}
\mathbb E_\Delta\left[
\left(\frac{\|\Delta_{-S}\|_{\Sigma_{1,-S\mid S}}^2/\tilde\sigma^2}
{1+\|\Delta_{-S}\|_{\Sigma_{1,-S\mid S}}^2/\tilde\sigma^2}\right)^r
\right]
\asymp
\left(\frac{k\bar\rho^2}{d+k\bar\rho^2}\right)^r.
\end{align*}
Finally, we have
\begin{align*}
\mathbb E_\Delta\left[
\frac{\|\Delta_{-S}\|_{\Sigma_{1,-S\mid S}}^2/\tilde\sigma^2}
{(1+\|\Delta_{-S}\|_{\Sigma_{1,-S\mid S}}^2/\tilde\sigma^2)^2}
\right]
\asymp
\frac{k\bar\rho^2}{d+k\bar\rho^2}
\mathbb E_\Delta\left[
\frac{1}{1+\|\Delta_{-S}\|_{\Sigma_{1,-S\mid S}}^2/\tilde\sigma^2}
\right].
\end{align*}
\end{lemma}

\begin{proof}
First, we express the normalized squared frozen coordinate error as a Gaussian quadratic form.
Conditioned on $X_{\mathrm{pre}}$, the Gaussian pre-training model implies
\begin{align*}
\tilde\sigma^{-1}\Sigma_{1,-S\mid S}^{1/2}\Delta_{-S}
\sim\mathcal N\left(0,\frac{1}{n\tilde\sigma^2}
\Sigma_{1,-S\mid S}^{1/2}\Gamma_{-S,-S}
\Sigma_{1,-S\mid S}^{1/2}\right).
\end{align*}
Let $\nu_1,\ldots,\nu_k$ denote the eigenvalues of this covariance matrix.  On $\Omega_{\mathrm{pre}}$, Lemma~\ref{lem:pretraining_covariance_regularity}, Assumption~\ref{ass:design}, and $\bar\rho^2=\tr(\Sigma_1\Gamma)/(n\tilde\sigma^2)$ 
result in $c\bar\rho^2/d\le\nu_j\le C\bar\rho^2/d$ for $1\le j\le k$.
Let $Z_1,\ldots,Z_k$ be independent standard normal variables and $X_k=\sum_{j=1}^kZ_j^2\sim\chi_k^2$.  Then, we have $\|\Delta_{-S}\|_{\Sigma_{1,-S\mid S}}^2/\tilde\sigma^2=\sum_{j=1}^k\nu_jZ_j^2$ and $c\bar\rho^2X_k/d\le\sum_{j=1}^k\nu_jZ_j^2\le C\bar\rho^2X_k/d$.
For each $r\in\{1,2\}$, we have $(1+\sum_{j=1}^k\nu_jZ_j^2)^{-r}\asymp(1+\bar\rho^2X_k/d)^{-r}$ and $\left((\sum_{j=1}^k\nu_jZ_j^2)/(1+\sum_{j=1}^k\nu_jZ_j^2)\right)^r\asymp\left((\bar\rho^2X_k/d)/(1+\bar\rho^2X_k/d)\right)^r$.
We apply Lemma~\ref{lem:chisq_denom_uniform} and obtain the first two statements.

For the last statement, we bound the numerator and denominator separately and obtain 
$(\sum_{j=1}^k\nu_jZ_j^2)$ $/ (1+\sum_{j=1}^k\nu_jZ_j^2)^2  \asymp(\bar\rho^2X_k/d)/(1+\bar\rho^2X_k/d)^2$.
We apply the last conclusion of Lemma~\ref{lem:chisq_denom_uniform} and its first conclusion with $r=1$ and obtain
\begin{align*}
\mathbb E_\Delta\left[
\frac{\|\Delta_{-S}\|_{\Sigma_{1,-S\mid S}}^2/\tilde\sigma^2}
{(1+\|\Delta_{-S}\|_{\Sigma_{1,-S\mid S}}^2/\tilde\sigma^2)^2}
\right]
&\asymp\mathbb E\left[
\frac{(\bar\rho^2/d)X_k}{(1+(\bar\rho^2/d)X_k)^2}
\right]
\asymp\frac{k\bar\rho^2}{d+k\bar\rho^2}
\mathbb E[(1+(\bar\rho^2/d)X_k)^{-1}]\\
&\asymp\frac{k\bar\rho^2}{d+k\bar\rho^2}
\mathbb E_\Delta\left[
\frac{1}{1+\|\Delta_{-S}\|_{\Sigma_{1,-S\mid S}}^2/\tilde\sigma^2}
\right].
\end{align*}
\end{proof}

For the proof of Corollary~\ref{cor:common_population_partial_mean_scale}, we need 
two moments of the active quadratic form $\|(\Sigma_1\Delta)_S\|_{\Sigma_{1,SS}^{-1}}^2$ 
weighted by the frozen coordinate error.  
Since $(\Sigma_1\Delta)_S$ and $\Delta_{-S}$ 
are not necessarily independent, the following lemma provides a joint evaluation of their moments.

\begin{lemma}[Weighted moments of the active quadratic form]
\label{lem:weighted_pretraining_error_moments}
Under the pre-training model in Section~\ref{sec:problem_setting} and
Assumption~\ref{ass:design}, fix $X_{\mathrm{pre}}\in\Omega_{\mathrm{pre}}$
and suppose that $1\le s<d$.  Let $k=d-s$.  Then,
\begin{align*}
\mathbb E_\Delta\left[
\frac{\|(\Sigma_1\Delta)_S\|_{\Sigma_{1,SS}^{-1}}^2}
{\left(1+\|\Delta_{-S}\|_{\Sigma_{1,-S\mid S}}^2/\tilde\sigma^2\right)^2}
\right]
&\asymp \frac{s}{n}h_{k,d}(\bar\rho),
\end{align*}
where, for $z>0$, $h_{k,d}(z)=(d/(d+z^2))^{k/2}$ if $1\le k\le3$, $h_{4,d}(z)=(d/(d+z^2))^2\log(e+z^2/d)$, and $h_{k,d}(z)=(d/(d+kz^2))^2$ if $k\ge5$.
Moreover,
\begin{align*}
\mathbb E_\Delta\left[
\|(\Sigma_1\Delta)_S\|_{\Sigma_{1,SS}^{-1}}^2
\left(
\frac{\|\Delta_{-S}\|_{\Sigma_{1,-S\mid S}}^2/\tilde\sigma^2}
{1+\|\Delta_{-S}\|_{\Sigma_{1,-S\mid S}}^2/\tilde\sigma^2}
\right)^2
\right]
&\asymp \frac{s}{n}\left(\frac{k\bar\rho^2}{d+k\bar\rho^2}\right)^2.
\end{align*}
\end{lemma}

\begin{proof}
Note that $\|\Sigma_{1,SS}^{-1/2}(\Sigma_1\Delta)_S\|_2^2+\|\Delta_{-S}\|_{\Sigma_{1,-S\mid S}}^2=\|\Delta\|_{\Sigma_1}^2.$
Since $\Delta\mid X_{\mathrm{pre}}\sim\mathcal N(0,\Gamma/n)$, Lemma~\ref{lem:pretraining_covariance_regularity}, and
Assumption~\ref{ass:design} imply that the joint covariance of $\Sigma_{1,SS}^{-1/2}(\Sigma_1\Delta)_S$ and $\Sigma_{1,-S\mid S}^{1/2}\Delta_{-S}$ is 
bounded above and below by constant multiples of $I_d/n$.  Thus, we have 
\begin{align}\label{eq:weighted_active_conditional_covariance}
\frac{c}{n}I_s\preceq\Cov_\Delta\left(\Sigma_{1,SS}^{-1/2}(\Sigma_1\Delta)_S\mid\Delta_{-S}\right)\preceq\frac{C}{n}I_s.
\end{align}
Conditioned on $\Delta_{-S}$, we also have 
\begin{align}\label{eq:weighted_active_conditional_second_moment}
\hspace*{-6mm}
\mathbb E_\Delta[\|(\Sigma_1\Delta)_S\|_{\Sigma_{1,SS}^{-1}}^2\mid\Delta_{-S}]
=\tr\left(\Cov_\Delta\left(
\Sigma_{1,SS}^{-1/2}(\Sigma_1\Delta)_S\mid\Delta_{-S}\right)\right) +\left\|\mathbb E_\Delta\left[
\Sigma_{1,SS}^{-1/2}(\Sigma_1\Delta)_S\mid\Delta_{-S}
\right]\right\|_2^2.
\end{align}

First, we prove the first conclusion.  In \eqref{eq:weighted_active_conditional_second_moment},
 we multiply both terms on the right hand side by $(1+\|\Delta_{-S}\|_{\Sigma_{1,-S\mid S}}^2/\tilde\sigma^2)^{-2}$ 
 and take expectation over $\Delta_{-S}$.  Equation~\eqref{eq:weighted_active_conditional_covariance} 
 and Lemma~\ref{lem:frozen_denominator_marginal} imply that the expectation of the first term is comparable
  with $(s/n)h_{k,d}(\bar\rho)$ which serves as the lower bound. 
It remains to bound the weighted squared norm of the conditional mean in~\eqref{eq:weighted_active_conditional_second_moment}.  For this purpose, we express $\mathbb E_\Delta[\Sigma_{1,SS}^{-1/2}(\Sigma_1\Delta)_S\mid\Delta_{-S}]$ and $\|\Delta_{-S}\|_{\Sigma_{1,-S\mid S}}^2$ using the same independent standard Gaussian variables.

Let $\nu_1,\ldots,\nu_k$ and $v_1,\ldots,v_k$ denote the eigenvalues and orthonormal eigenvectors, respectively, of the covariance matrix of $\Sigma_{1,-S\mid S}^{1/2}\Delta_{-S}$.  Define $Z_j=\nu_j^{-1/2}v_j^\top\Sigma_{1,-S\mid S}^{1/2}\Delta_{-S}$ for $1\le j\le k$.  These variables are independent standard normal, and
\begin{align*}
\Sigma_{1,-S\mid S}^{1/2}\Delta_{-S}&=\sum_{j=1}^k\sqrt{\nu_j}v_jZ_j,\qquad \|\Delta_{-S}\|_{\Sigma_{1,-S\mid S}}^2=\sum_{j=1}^k\nu_jZ_j^2.
\end{align*}
Then, the Gaussian regression formula  implies
\begin{align}\label{eq:weighted_active_gaussian_regression}
\mathbb E_\Delta\left[\Sigma_{1,SS}^{-1/2}(\Sigma_1\Delta)_S\mid\Delta_{-S}\right]
=\sum_{j=1}^k\frac{\Cov_\Delta\left(\Sigma_{1,SS}^{-1/2}(\Sigma_1\Delta)_S,\Sigma_{1,-S\mid S}^{1/2}\Delta_{-S}\right)v_j}{\sqrt{\nu_j}}Z_j.
\end{align}
By joint Gaussianity, the conditional covariance does not depend on $\Delta_{-S}$.  Hence, the law of total covariance implies
\begin{align*}
\Cov_\Delta\left(\mathbb E_\Delta\left[\Sigma_{1,SS}^{-1/2}(\Sigma_1\Delta)_S\mid\Delta_{-S}\right]\right)
=\Cov_\Delta\left(\Sigma_{1,SS}^{-1/2}(\Sigma_1\Delta)_S\right)-\Cov_\Delta\left(\Sigma_{1,SS}^{-1/2}(\Sigma_1\Delta)_S\mid\Delta_{-S}\right)\preceq\frac{C}{n}I_s.
\end{align*}
By taking traces and using the Gaussian regression representation, we obtain
\begin{align}\label{eq:weighted_active_regression_coefficients}
\hspace*{-3mm}
\!\!\!\!\sum_{j=1}^k\left\|\frac{\Cov_\Delta\left(\Sigma_{1,SS}^{-1/2}(\Sigma_1\Delta)_S,\Sigma_{1,-S\mid S}^{1/2}\Delta_{-S}\right)v_j}{\sqrt{\nu_j}}\right\|_2^2
\!=\!\tr\left(\Cov_\Delta\left(\mathbb E_\Delta\left[\Sigma_{1,SS}^{-1/2}(\Sigma_1\Delta)_S\mid\Delta_{-S}\right]\right)\right)\le\frac{Cs}{n}.
\end{align}
By expanding the squared norm in~\eqref{eq:weighted_active_gaussian_regression}, we
evaluate $\mathbb E_\Delta[Z_iZ_j(1+\sum_{\ell=1}^k\nu_\ell Z_\ell^2/
\tilde\sigma^2)^{-2}]$ for $1\le i,j\le k$.  For $i\ne j$, the 
integrand is odd in $Z_i$, and hence its expectation is zero.  For 
$i=j$, conditioned on $(Z_\ell)_{\ell\ne j}$, the function $(1+\sum_
{\ell\ne j}\nu_\ell Z_\ell^2/\tilde\sigma^2+\nu_jz/\tilde\sigma^2)^{-2}
$ is nonincreasing in $z$ for $z\ge0$.  Since $\mathbb E Z_j^2=1$, it 
follows that for $1\le j\le k$,
\begin{align*}
\mathbb E_\Delta\left[Z_j^2\left(1+\sum_{\ell=1}^k\frac{\nu_\ell Z_\ell^2}{\tilde\sigma^2}\right)^{-2}\right]&\le\mathbb E_\Delta\left[\left(1+\sum_{\ell=1}^k\frac{\nu_\ell Z_\ell^2}{\tilde\sigma^2}\right)^{-2}\right].
\end{align*}
By combining these relations with \eqref{eq:weighted_active_regression_coefficients}, we have
\begin{align*}
\mathbb E_\Delta\left[
\left\|\mathbb E_\Delta\left[
\Sigma_{1,SS}^{-1/2}(\Sigma_1\Delta)_S\mid\Delta_{-S}
\right]\right\|_2^2
\left(1+\frac{\|\Delta_{-S}\|_{\Sigma_{1,-S\mid S}}^2}
{\tilde\sigma^2}\right)^{-2}\right]
\le\frac{Cs}{n}\mathbb E_\Delta\left[
\left(1+\frac{\|\Delta_{-S}\|_{\Sigma_{1,-S\mid S}}^2}
{\tilde\sigma^2}\right)^{-2}\right].
\end{align*}
By combining this bound with the first term in \eqref{eq:weighted_active_conditional_second_moment}, we obtain
\begin{align*}
\mathbb E_\Delta\left[\frac{\|(\Sigma_1\Delta)_S\|_{\Sigma_{1,SS}^{-1}}^2}{(1+\|\Delta_{-S}\|_{\Sigma_{1,-S\mid S}}^2/\tilde\sigma^2)^2}\right]\asymp\frac{s}{n}\mathbb E_\Delta\left[(1+\|\Delta_{-S}\|_{\Sigma_{1,-S\mid S}}^2/\tilde\sigma^2)^{-2}\right].
\end{align*}
Then, we apply Lemma~\ref{lem:frozen_denominator_marginal} with $r=2$ and obtain the first conclusion.

Similarly, we prove the second conclusion.  We multiply \eqref{eq:weighted_active_conditional_second_moment} by the squared ratio in the second conclusion and take expectation over $\Delta_{-S}$.  Equation~\eqref{eq:weighted_active_conditional_covariance} and Lemma~\ref{lem:frozen_denominator_marginal} with $r=2$ imply that the first term satisfies
\begin{align*}
\mathbb E_\Delta\left[\tr\left(\Cov_\Delta\left(\Sigma_{1,SS}^{-1/2}(\Sigma_1\Delta)_S\mid\Delta_{-S}\right)\right)
\left(\frac{\|\Delta_{-S}\|_{\Sigma_{1,-S\mid S}}^2/\tilde\sigma^2}{1+\|\Delta_{-S}\|_{\Sigma_{1,-S\mid S}}^2/\tilde\sigma^2}\right)^2\right]
\asymp\frac{s}{n}\left(\frac{k\bar\rho^2}{d+k\bar\rho^2}\right)^2.
\end{align*}
It remains to bound the second term in \eqref{eq:weighted_active_conditional_second_moment}.  
The Gaussian representation above and the covariance trace bounds imply 
$\mathbb E_\Delta\|\mathbb E_\Delta[\Sigma_{1,SS}^{-1/2}(\Sigma_1\Delta)_S\mid\Delta_{-S}]\|_2^4\le Cs^2/n^2$ and $\mathbb E_\Delta\|\Delta_{-S}\|_{\Sigma_{1,-S\mid S}}^8\le Ck^4/n^4$.  Since $(x/(1+x))^2\le\min(x^2,1)$ for $x\ge0$, 
it follows from the Cauchy--Schwarz inequality that
\begin{align*}
&\mathbb E_\Delta\left[
\left\|\mathbb E_\Delta\left[
\Sigma_{1,SS}^{-1/2}(\Sigma_1\Delta)_S\mid\Delta_{-S}
\right]\right\|_2^2
\left(
\frac{\|\Delta_{-S}\|_{\Sigma_{1,-S\mid S}}^2/\tilde\sigma^2}
{1+\|\Delta_{-S}\|_{\Sigma_{1,-S\mid S}}^2/\tilde\sigma^2}
\right)^2\right]\\
&\le\min\Biggl\{\mathbb E_\Delta\left\|\mathbb E_\Delta\left[\Sigma_{1,SS}^{-1/2}(\Sigma_1\Delta)_S\mid\Delta_{-S}\right]\right\|_2^2,\frac{1}{\tilde\sigma^4}\left(\mathbb E_\Delta\left\|\mathbb E_\Delta\left[\Sigma_{1,SS}^{-1/2}(\Sigma_1\Delta)_S\mid\Delta_{-S}\right]\right\|_2^4\mathbb E_\Delta\|\Delta_{-S}\|_{\Sigma_{1,-S\mid S}}^8\right)^{1/2}\Biggr\}\\
&\le\frac{Cs}{n}\min\left\{1,\left(\frac{k}{n\tilde\sigma^2}\right)^2\right\}\le\frac{Cs}{n}\left(\frac{k\bar\rho^2}{d+k\bar\rho^2}\right)^2,
\end{align*}
where the last inequality follows from $k/(n\tilde\sigma^2)\asymp k\bar\rho^2/d$ on $\Omega_{\mathrm{pre}}$ and 
$\min(x^2,1)\asymp(x/(1+x))^2$ for $x\ge0$. Hence, we obtain the second conclusion and prove the lemma.
\end{proof}

%% file: Unified_appendix/app_16K_iterative_training.tex

\section{Proofs for Iterative Full Fine-Tuning}\label{sec:iterative_training_proofs}

In this section, we prove Theorems~\ref{thm:iterative_full_finetuning} and~\ref{thm:iterative_full_finetuning_lower_bound}.
First, we establish bounds for the expected squared error after one generation.
Then, we apply these bounds successively to obtain the upper and lower bounds over multiple generations.

\subsection{Auxiliary Bounds}

Let $\mathcal F_t$ be the sigma field generated by the initial error, all synthetic covariates, 
and generation noises through generation $t$.
At generation $t$, we exploit the one-generation notation with $\Delta=\Delta^{(t-1)}$ and $r_{S,\mathrm{ext}}=-\Delta^{(t)}$.
Conditioned on $\mathcal F_{t-1}$, the preceding error is fixed and the new covariates and 
generation noises have the synthetic distribution described in Section~\ref{sec:problem_setting}.
The following deterministic bound holds at every generation.

\begin{lemma}[Uniform error bound over generations]
\label{lem:iterative_error_bound}
Under the iterative training model in Section~\ref{sec:iterative_training}, 
suppose that $s=d$, $\delta>0$, $C_G\ge2$, and $\Sigma_1\preceq C_\lambda I_d$.
Then, for every initial error $\Delta^{(0)}$, every integer $t\ge0$, and every realization of the synthetic 
samples through generation $t$, the extended estimator satisfies
\begin{align*}
\|\Delta^{(t)}\|_{\Sigma_1}^2&\le\max\bigl(\|\Delta^{(0)}\|_{\Sigma_1}^2,C_\lambda\delta^2m^{C_G}\bigr).
\end{align*}
\end{lemma}

\begin{proof}
Consider a generation $t\ge1$ and let $\Delta=\Delta^{(t-1)}$.  On $\Omega_{\mathrm{post}}$, we have $r_{S,\mathrm{ext}}=r_S=-\Delta^{(t)}$.
Since $s=d$, the residual representation~\eqref{eq:residual_rep_main} implies $G(\Delta)r_S=\sum_{i=1}^m\mathbb I_i\tilde x_i\xi_i$.  
By the Cauchy--Schwarz inequality, it follows that
\begin{align*}
\|r_S\|_{G(\Delta)}^2&=\sum_{i=1}^m\mathbb I_i\tilde x_i^\top r_S\xi_i\le\|r_S\|_{G(\Delta)}\biggl(\sum_{i=1}^m\mathbb I_i\xi_i^2\biggr)^{1/2}.
\end{align*}
Therefore, on $\Omega_{\mathrm{post}}$, we have
\begin{align}\label{eq:iterative_residual_bound}
\|r_{S,\mathrm{ext}}\|_{G(\Delta)}^2&\le\sum_{i=1}^m\mathbb I_i\xi_i^2.
\end{align}
On this event, $G(\Delta)\succeq m^{1-C_G}I_d$, while $\mathbb I_i=1$ implies $|\xi_i|\le\delta$.  Hence,
$\|\Delta^{(t)}\|_{\Sigma_1}^2=\|r_{S,\mathrm{ext}}\|_{\Sigma_1}^2\le C_\lambda m^{C_G-1}\|r_{S,\mathrm{ext}}\|_{G(\Delta)}^2\le C_\lambda\delta^2m^{C_G}.$
On $\Omega_{\mathrm{post}}^{\mathrm c}$, the extended estimator retains the preceding estimate, and hence 
$\Delta^{(t)}=\Delta^{(t-1)}$.  Therefore, every realization satisfies 
$\|\Delta^{(t)}\|_{\Sigma_1}^2\le\max\bigl(\|\Delta^{(t-1)}\|_{\Sigma_1}^2,C_\lambda\delta^2m^{C_G}\bigr)$ and 
induction on $t$ proves the claimed bound.
\end{proof}

In the following corollary, we bound the conditional error after one generation when $\|\Delta\|_{\Sigma_1}\le\tilde\sigma$.
This bound provides the contraction step in the iterative argument.

\begin{corollary}[Conditional squared error after one generation] \label{cor:iterative_one_generation_error}
Under the iterative training model in Section~\ref{sec:iterative_training} with 
$c_\lambda I_d\preceq\Sigma_1\preceq C_\lambda I_d$, let $s=d$ and fix $C_G\ge2$ as in Definition~\ref{def:ext_estimator}.
There exist constants $C>0$ and $u_0\in(0,1]$, depending only on $c_\lambda$, $C_\lambda$, and $C_G$, 
such that the following holds.
Suppose that $m\ge C$, $0<u=\delta/\tilde\sigma\le u_0$, and $mu\ge C(d+\log m)\log^2m$.
Then, for every $\Delta\in\mathbb R^d$ satisfying $\|\Delta\|_{\Sigma_1}^2\le\tilde\sigma^2$, 
the extended residual satisfies
\begin{align*}
\mathbb E_{\tilde X,\tilde\varepsilon}[\|r_{S,\mathrm{ext}}\|_{\Sigma_1}^2\mid\Delta]&\le\frac{2u^4}{9}\|\Delta\|_{\Sigma_1}^2+\frac{Cd\delta^2}{mu}.
\end{align*}
\end{corollary}

\begin{proof}
Consider a fixed $\Delta$ satisfying $\|\Delta\|_{\Sigma_1}^2\le\tilde\sigma^2$, and let $\eta=m^{-(C_G+1)}$.  
Since $\rho_\Delta^2\le1$, it follows from~\eqref{eq:marginal_acceptance_scale} and~\eqref{eq:zeta_delta_def} 
that $p_\Delta\ge cu$ and $\zeta_\Delta^2\le2$.
Moreover, the sample condition in the statement implies
$mp_\Delta \ge C\bigl(d+\log(m/\eta)\bigr)\log^2(m/\eta).$

Since $0<\eta\le1/2$, we apply Proposition~\ref{prop:full_accepted_gram_mean} and obtain 
an event $\Omega_{\mathrm{full}}(\Delta)$, determined by $(\tilde X,\mathbb I_1,\ldots,\mathbb I_m)$, with conditional probability at least $1-\eta$.
On this event, the lower Gram matrix bound in 
Proposition~\ref{prop:full_accepted_gram_mean} and~\eqref{eq:common_delta_population_operator} result in
\begin{align*}
\frac{G(\Delta)}m&\succeq c\bar G_{\mathrm{pop}}(\Delta)\succeq\frac{cp_\Delta}{\zeta_\Delta^2}\Sigma_1\succeq cu\Sigma_1.
\end{align*}
The lower covariance bound, $C_G\ge2$, and the sample condition imply $cu\Sigma_1\succeq m^{-C_G}I_d$ 
for sufficiently large $C$.  Therefore, $\Omega_{\mathrm{full}}(\Delta)\subseteq\Omega_{\mathrm{post}}$ 
and $r_{S,\mathrm{ext}}=r_S$ on $\Omega_{\mathrm{full}}(\Delta)$.
To bound the expected squared error on $\Omega_{\mathrm{full}}(\Delta)$, we upper bound the conditional mean and covariance of $r_S$:
\begin{align*}
\mathbb E_{\tilde\varepsilon}[\|r_S\|_{\Sigma_1}^2\mid\Delta,\tilde X,\mathbb I_1,\ldots,\mathbb I_m]
=\bigl\|\mathbb E_{\tilde\varepsilon}[r_S\mid\Delta,\tilde X,\mathbb I_1,\ldots,\mathbb I_m]\bigr\|_{\Sigma_1}^2
+\tr\bigl(\Sigma_1\Cov_{\tilde\varepsilon}(r_S\mid\Delta,\tilde X,\mathbb I_1,\ldots,\mathbb I_m)\bigr).
\end{align*}
The conditional mean bound in Proposition~\ref{prop:full_accepted_gram_mean} and the triangle inequality imply
\begin{align*}
\bigl\|\mathbb E_{\tilde\varepsilon}
[r_S\mid\Delta,\tilde X,\mathbb I_1,\ldots,\mathbb I_m]\bigr\|_{\Sigma_1}
&\le\biggl(\frac{u^2}{3}+Cu^4\biggr)\|\Delta\|_{\Sigma_1}.
\end{align*}

Next, conditioned on $(\Delta,\tilde X,\mathbb I_1,\ldots,\mathbb I_m)$, the variables $\xi_1,\ldots,\xi_m$ are
 independent, and $\mathbb I_i=1$ implies $\Var_{\tilde\varepsilon_i}(\xi_i\mid\Delta,\tilde x_i,\mathbb I_i=1)\le\delta^2$.  
 The residual representation~\eqref{eq:residual_rep_main} with $s=d$ leads to 
\begin{align*}
\Cov_{\tilde\varepsilon}(r_S\mid\Delta,\tilde X,\mathbb I_1,\ldots,\mathbb I_m)=G(\Delta)^{-1}
\biggl(\sum_{i=1}^m\mathbb I_i\tilde x_i\tilde x_i^\top
\Var_{\tilde\varepsilon_i}(\xi_i\mid\Delta,\tilde x_i,\mathbb I_i=1)\biggr)
G(\Delta)^{-1}\preceq\delta^2G(\Delta)^{-1},
\end{align*}
Using $G(\Delta)\succeq cmu\Sigma_1$ in this matrix bound, we obtain
\begin{align*}
\tr\bigl(\Sigma_1\Cov_{\tilde\varepsilon}(r_S\mid\Delta,\tilde X,\mathbb I_1,\ldots,\mathbb I_m)\bigr)&\le\delta^2\tr\bigl(\Sigma_1G(\Delta)^{-1}\bigr)\le\frac{Cd\delta^2}{mu}.
\end{align*}

Finally, from the sample condition and $C_G\ge2$, we have
 $\tilde\sigma^2=\delta^2/u^2\le C_\lambda\delta^2m^{C_G}$ for sufficiently large $C$.  
 We apply Lemma~\ref{lem:iterative_error_bound} with initial error $\Delta$ and obtain $\|r_{S,\mathrm{ext}}\|_{\Sigma_1}^2\le C_\lambda\delta^2m^{C_G}$ for every realization.
Since $\mathbb P_{\tilde X,\tilde\varepsilon}(\Omega_{\mathrm{full}}(\Delta)^{\mathrm c}\mid\Delta)\le\eta$, we obtain
$\mathbb E_{\tilde X,\tilde\varepsilon}[\|r_{S,\mathrm{ext}}\|_{\Sigma_1}^2\mathbf 1_{\Omega_{\mathrm{full}}(\Delta)^{\mathrm c}}\mid\Delta]\le C_\lambda\delta^2m^{C_G}\eta=C_\lambda\delta^2/m$.
We take expectation over $(\tilde X,\mathbb I_1,\ldots,\mathbb I_m)$ conditioned on $\Delta$ and add the contribution on $\Omega_{\mathrm{full}}(\Delta)^{\mathrm c}$ to obtain
\begin{align*}
\mathbb E_{\tilde X,\tilde\varepsilon}[\|r_{S,\mathrm{ext}}\|_{\Sigma_1}^2\mid\Delta]
&\le\biggl(\frac{u^2}{3}+Cu^4\biggr)^2\|\Delta\|_{\Sigma_1}^2+\frac{Cd\delta^2}{mu}+\frac{C_\lambda\delta^2}{m}\le\frac{2u^4}{9}\|\Delta\|_{\Sigma_1}^2+\frac{Cd\delta^2}{mu}.
\end{align*}
The last inequality holds for sufficiently small $u_0$.
\end{proof}

The preceding corollary requires $\|\Delta\|_{\Sigma_1}\le\tilde\sigma$, which is not guaranteed after each generation.
The following lemma bounds the probability of a large post-training error on $\Omega_{\mathrm{post}}$ for an arbitrary preceding error.

\begin{lemma}[Tail bound for the post-training error]
\label{lem:iterative_accepted_update_error}
Under the iterative training model in Section~\ref{sec:iterative_training}, let $s=d$, $\delta,\tilde\sigma>0$, and fix $C_G\ge2$ as in Definition~\ref{def:ext_estimator}.  There exist universal constants $c,C>0$ such that the following holds.  For every $\Delta\in\mathbb R^d$, let $p_\Delta=\mathbb P(|\xi_i|\le\delta\mid\Delta)$ and
$\zeta_\Delta^2=(\tilde\sigma^2+\|\Delta\|_{\Sigma_1}^2)/(\tilde\sigma^2+\lambda_{2,\Delta}\|\Delta\|_{\Sigma_1}^2)$.
Then,
\begin{align*}
\mathbb P_{\tilde X,\tilde\varepsilon}
\bigl(\Omega_{\mathrm{post}}\cap
\{\|r_{S,\mathrm{ext}}\|_{\Sigma_1}^2>2\delta^2\zeta_\Delta^2\}\mid\Delta\bigr)
&\le2\exp(Cd-cmp_\Delta).
\end{align*}
\end{lemma}

\begin{proof}
Consider a fixed $\Delta\in\mathbb R^d$.  Conditioned on $(\Delta,\mathbb I_1,\ldots,\mathbb I_m)$, 
the accepted covariates are independent since each acceptance indicator depends only on its corresponding 
covariate and generation noise.  By the definition of $\bar G_{\mathrm{pop}}(\Delta)$, 
we have $\mathbb E_{\tilde X}[G(\Delta)\mid\Delta,\mathbb I_1,\ldots,\mathbb I_m]=(\sum_i\mathbb I_i/p_\Delta)\bar G_{\mathrm{pop}}(\Delta)$.

For every unit vector $v$, \eqref{eq:full_accepted_normalized_second_moment} and Lemma~\ref{lem:accepted_linear_form_moments} imply $\mathbb E[(v^\top p_\Delta^{1/2}\bar G_{\mathrm{pop}}(\Delta)^{-1/2}\tilde x_i)^2\mid\mathbb I_i=1,\Delta]=1$ and $\|v^\top p_\Delta^{1/2}\bar G_{\mathrm{pop}}(\Delta)^{-1/2}\tilde x_i\|_{\psi_2}^2\le C$, where the norm is evaluated conditioned on $(\mathbb I_i=1,\Delta)$.
Then, we apply Lemma~\ref{lem:bernstein} to $\sum_{i=1}^m\mathbb I_i[p_\Delta(v^\top\bar G_{\mathrm{pop}}(\Delta)^{-1/2}\tilde x_i)^2-1]$
and utilize the $1/4$-net argument to obtain
\begin{align}
\mathbb P_{\tilde X}\biggl(
G(\Delta)\succeq
\frac{\sum_i\mathbb I_i}{2p_\Delta}\bar G_{\mathrm{pop}}(\Delta)
\biggm|\Delta,\mathbb I_1,\ldots,\mathbb I_m
\biggr)
&\ge1-2e^{Cd-c\sum_i\mathbb I_i}.
\label{eq:iterative_accepted_gram_probability}
\end{align}
Note that the identity~\eqref{eq:common_delta_population_operator} implies $\bar G_{\mathrm{pop}}(\Delta)\succeq p_\Delta\Sigma_1/\zeta_\Delta^2$.  
Hence, on $\Omega_{\mathrm{post}}$ and the matrix event in~\eqref{eq:iterative_accepted_gram_probability}, we have $\sum_i\mathbb I_i\ge1$ 
and $G(\Delta)\succeq(\sum_i\mathbb I_i)\Sigma_1/(2\zeta_\Delta^2)$.  
Then, the residual bound~\eqref{eq:iterative_residual_bound}  implies
$\|r_{S,\mathrm{ext}}\|_{\Sigma_1}^2\le2\zeta_\Delta^2(\sum_i\mathbb I_i)^{-1}\sum_i\mathbb I_i\xi_i^2\le2\delta^2\zeta_\Delta^2$,
where the last inequality uses $|\xi_i|\le\delta$ for every accepted sample. 
Consequently, on $\Omega_{\mathrm{post}}$, the event $\{\|r_{S,\mathrm{ext}}\|_{\Sigma_1}^2>2\delta^2\zeta_\Delta^2\}$ can occur only when the matrix bound in~\eqref{eq:iterative_accepted_gram_probability} fails.

Conditioned on $\Delta$, the indicators are independent Bernoulli variables with parameter $p_\Delta$.  
By taking expectation of the conditional failure probability in~\eqref{eq:iterative_accepted_gram_probability} over the indicators, we have
\begin{align*}
&\mathbb P_{\tilde X,\tilde\varepsilon}
\bigl(\Omega_{\mathrm{post}}\cap\{\|r_{S,\mathrm{ext}}\|_{\Sigma_1}^2>2\delta^2\zeta_\Delta^2\}\mid\Delta\bigr)\\
&\quad \le2e^{Cd}\mathbb E_{\tilde X,\tilde\varepsilon}\bigl[e^{-c\sum_i\mathbb I_i}\mid\Delta\bigr]
=2e^{Cd}(1-p_\Delta+p_\Delta e^{-c})^m
\le2e^{Cd-(1-e^{-c})mp_\Delta}.
\end{align*}
\end{proof}

To iterate Corollary~\ref{cor:iterative_one_generation_error}, we bound the squared error contribution on $\{\|\Delta^{(t)}\|_{\Sigma_1}>\tilde\sigma\}$ uniformly over all generations.

\begin{lemma}[Squared error contribution on $\{\|\Delta^{(t)}\|_{\Sigma_1}>\tilde\sigma\}$]
\label{lem:iterative_error_outside_radius}
Under the iterative training model in Section~\ref{sec:iterative_training} with $c_\lambda I_d\preceq\Sigma_1\preceq C_\lambda I_d$, let $s=d$, $\delta,\tilde\sigma>0$, and let $C_G\ge2$ be the exponent in Definition~\ref{def:ext_estimator}.
There exist constants $c,C>0$ and $u_0\in(0,1]$, depending only on $c_\lambda$, $C_\lambda$, and $C_G$, such that, if $m\ge C$, $u=\delta/\tilde\sigma\le u_0$, and $mu\ge Cd^2\log m$, then, for every $\Delta^{(0)}\in\mathbb R^d$ satisfying $\|\Delta^{(0)}\|_{\Sigma_1}^2\le\tilde\sigma^2$ and every integer $t\ge0$,
\begin{align*}
\mathbb E\bigl[
\|\Delta^{(t)}\|_{\Sigma_1}^2
\mathbf 1\{\|\Delta^{(t)}\|_{\Sigma_1}^2>\tilde\sigma^2\}
\mid\Delta^{(0)}\bigr]
&\le C\delta^2e^{-cmu}.
\end{align*}
\end{lemma}

\begin{proof}
Consider an initial error $\Delta^{(0)}$ satisfying $\|\Delta^{(0)}\|_{\Sigma_1}^2\le\tilde\sigma^2$. 
The sample condition and $C_G\ge2$ imply $C_\lambda u^2m^{C_G}=C_\lambda(mu)^2m^{C_G-2}\ge1$, 
and hence $C_\lambda\delta^2m^{C_G}\ge\tilde\sigma^2$.  It follows from Lemma~\ref{lem:iterative_error_bound} that
$\|\Delta^{(t)}\|_{\Sigma_1}^2\le C_\lambda\delta^2m^{C_G}$ for every $t\ge0$.
To obtain a bound uniform in $t$, we bound the probability of leaving $\{\|\Delta\|_{\Sigma_1}\le\tilde\sigma\}$ 
from above and the probability of returning to this set from below.

First, consider a preceding error $\Delta$ satisfying $\|\Delta\|_{\Sigma_1}^2\le\tilde\sigma^2$.  
Since $p_\Delta=2\Phi(u/\sqrt{1+\rho_\Delta^2})-1$ and $\rho_\Delta^2\le1$, we have $p_\Delta\ge2\Phi(u/\sqrt2)-1\ge cu$.  
By \eqref{eq:zeta_delta_def}, we also have $2\delta^2\zeta_\Delta^2\le4\delta^2\le\tilde\sigma^2$ for $u_0\le1/2$.  
On $\Omega_{\mathrm{post}}^{\mathrm c}$, the update retains $\Delta$ and cannot satisfy $\|r_{S,\mathrm{ext}}\|_{\Sigma_1}>\tilde\sigma$. 
From Lemma~\ref{lem:iterative_accepted_update_error}, we have for a constant $c_0>0$,
\begin{align}\label{eq:iterative_departure_sigma}
\mathbb P_{\tilde X,\tilde\varepsilon}
\bigl(\|r_{S,\mathrm{ext}}\|_{\Sigma_1}^2>\tilde\sigma^2\mid\Delta\bigr)
&\le2\exp(Cd-c_0mu).
\end{align}

Next, for every $\Delta$ satisfying $\tilde\sigma^2<\|\Delta\|_{\Sigma_1}^2\le C_\lambda\delta^2m^{C_G}$, we establish the uniform lower bound
\begin{align}\label{eq:iterative_return_sigma}
\mathbb P_{\tilde X,\tilde\varepsilon}
\bigl(\|r_{S,\mathrm{ext}}\|_{\Sigma_1}^2\le\tilde\sigma^2\mid\Delta\bigr)
&\ge m^{-Cd^2}\exp(-c_0mu/2).
\end{align}
Let $C_0>1$ be sufficiently large, and take $u_0$ sufficiently small depending on $C_0$.
First, suppose that $\tilde\sigma^2<\|\Delta\|_{\Sigma_1}^2\le C_0\tilde\sigma^2$.
Then, $p_\Delta\ge cu/\sqrt{1+C_0}$ and $\zeta_\Delta^2\le1+C_0$.
It follows from Chernoff's inequality, the conditional Gram matrix 
bound~\eqref{eq:iterative_accepted_gram_probability}, and~\eqref{eq:common_delta_population_operator} that
$\sum_{i=1}^m\mathbb I_i\ge mp_\Delta/2$ and $G(\Delta)\succeq(2\zeta_\Delta^2)^{-1}\sum_i\mathbb I_i\,\Sigma_1$
with conditional probability at least $1-C\exp(Cd-cmu)\ge1/2$, by the sample condition.
On this event, $G(\Delta)/m\succeq cu I_d\succeq m^{-C_G}I_d$, and hence $\Omega_{\mathrm{post}}$ holds.
The lower bound for $G(\Delta)$ and~\eqref{eq:iterative_residual_bound} imply
$\|r_{S,\mathrm{ext}}\|_{\Sigma_1}^2\le2\delta^2\zeta_\Delta^2\le2u^2(1+C_0)\tilde\sigma^2\le\tilde\sigma^2$ 
by the choice of $u_0$.
Therefore, $\mathbb P_{\tilde X,\tilde\varepsilon}(\|r_{S,\mathrm{ext}}\|_{\Sigma_1}^2\le\tilde\sigma^2\mid\Delta)\ge1/2$.

For the remaining range, suppose that $C_0\tilde\sigma^2<\|\Delta\|_{\Sigma_1}^2\le C_\lambda\delta^2m^{C_G}$.
We construct an event on which $G(\Delta)$ satisfies the condition defining $\Omega_{\mathrm{post}}$ 
and the accepted response errors are small enough to yield $\|r_{S,\mathrm{ext}}\|_{\Sigma_1}\le\tilde\sigma$.
Let $k=\lceil C_1d\rceil$, where $C_1$ is sufficiently large depending only on the covariance bounds.
The sample condition implies $k\le m$.
Consider the event on which the first $k$ synthetic samples satisfy the following Gram matrix 
and residual bounds and all remaining samples are rejected:
\begin{align}\label{eq:iterative_return_construction}
m^{1-C_G}I_d
\preceq\sum_{i=1}^k\tilde x_i\tilde x_i^\top
\preceq\frac{4C_\lambda}{c_\lambda}m^{1-C_G}I_d, \quad
|\xi_i|\le\delta m^{-C_G/2},\quad 1\le i\le k,
\end{align}
where $\mathbb I_i=0$ for $k<i\le m$.
On this event, exactly the first $k$ synthetic samples are accepted, and their Gram matrix satisfies 
the condition defining $\Omega_{\mathrm{post}}$.
Hence,~\eqref{eq:iterative_residual_bound} implies
$\|r_{S,\mathrm{ext}}\|_{\Sigma_1}^2\le C_\lambda m^{C_G-1}\sum_{i=1}^k\xi_i^2\le C_\lambda(k/m)\delta^2\le\tilde\sigma^2$.

To bound the probability of the Gram condition, we apply Lemma~\ref{lem:pretraining_covariance_regularity} with $n=k$ and $\Sigma=\Sigma_1$ to the covariates $\tilde x_1,\ldots,\tilde x_k$.
We obtain $(c_\lambda k/2)I_d\preceq\sum_{i=1}^k\tilde x_i\tilde x_i^\top\preceq2C_\lambda kI_d$ with probability at least $1/2$.
Scaling  each covariate by $(2m^{1-C_G}/(c_\lambda k))^{1/2}$ transforms
 this event to the Gram condition in~\eqref{eq:iterative_return_construction}.
Since the $k$ covariates contain $kd$ scalar coordinates, the volume of the event is multiplied by $(2m^{1-C_G}/(c_\lambda k))^{kd/2}$.  Therefore,
\begin{align*}
\mathbb P_{\tilde X}\biggl(
 m^{1-C_G}I_d
\preceq\sum_{i=1}^k\tilde x_i\tilde x_i^\top
\preceq\frac{4C_\lambda}{c_\lambda}m^{1-C_G}I_d
\biggm|\Delta\biggr)&\ge\frac12\biggl(\frac{2m^{1-C_G}}{c_\lambda k}\biggr)^{kd/2}
\ge m^{-Cd^2}.
\end{align*}

For covariates satisfying these Gram matrix bounds, we bound the conditional probability that the response error conditions in~\eqref{eq:iterative_return_construction} also hold.
Since $\xi_i=-\tilde x_i^\top\Delta+\tilde\varepsilon_i$ and the generation noises are independent Gaussian variables, the response errors are independent conditioned on $(\Delta,\tilde x_1,\ldots,\tilde x_k)$, with $\xi_i\sim\mathcal N(-\tilde x_i^\top\Delta,\tilde\sigma^2)$.
The upper Gram matrix bound and $\Sigma_1\succeq c_\lambda I_d$ imply $\sum_{i=1}^k(\tilde x_i^\top\Delta)^2=\Delta^\top(\sum_{i=1}^k\tilde x_i\tilde x_i^\top)\Delta\le Cm^{1-C_G}\|\Delta\|_{\Sigma_1}^2$.
For $Z\sim\mathcal N(-a,\tilde\sigma^2)$ and $h>0$, observe that $\mathbb P(|Z|\le h)\ge c(h/\tilde\sigma)\exp\{-C(a^2+h^2)/\tilde\sigma^2\}$.
We apply this inequality with $a=\tilde x_i^\top\Delta$ and $h=\delta m^{-C_G/2}$ to obtain
\begin{align*}
&\mathbb P_{\tilde\varepsilon}\biggl(
|\xi_i|\le\delta m^{-C_G/2}
\text{ for }1\le i\le k
\biggm|\Delta,\tilde x_1,\ldots,\tilde x_k\biggr)\\
&\quad\ge(cu m^{-C_G/2})^k
\exp\biggl[-C\biggl(ku^2m^{-C_G}
+m^{1-C_G}\frac{\|\Delta\|_{\Sigma_1}^2}{\tilde\sigma^2}\biggr)\biggr]\\
&\quad\ge m^{-Cd^2}\exp(-Cmu^2),
\end{align*}
where the last inequality follows from $u\ge1/m$, $k\le(C_1+1)d$, $k\le m$, 
and $\|\Delta\|_{\Sigma_1}^2\le C_\lambda\delta^2m^{C_G}$.

Conditioned on $\Delta$, the samples indexed by $k+1,\ldots,m$ are independent of the first $k$ samples.
By the choice of $C_0$, we have $p_\Delta\le Cu/\sqrt{1+C_0}\le1/2$. Hence,
\begin{align*}
\mathbb P_{\tilde X,\tilde\varepsilon}
(\mathbb I_i=0\text{ for }k<i\le m\mid\Delta)
&=(1-p_\Delta)^{m-k}
\ge\exp\{-2(m-k)p_\Delta\}
\ge\exp\left\{-\frac{Cmu}{\sqrt{1+C_0}}\right\}.
\end{align*}
By taking expectation of the conditional probability bound for the first $k$ samples over $(\tilde x_1,\ldots,\tilde x_k)$
on the Gram condition and multiplying by this rejection probability, we obtain
\begin{align*}
\mathbb P_{\tilde X,\tilde\varepsilon}
\bigl(\|r_{S,\mathrm{ext}}\|_{\Sigma_1}^2\le\tilde\sigma^2\mid\Delta\bigr)
&\ge m^{-Cd^2}\exp\biggl(-Cmu^2-\frac{Cmu}{\sqrt{1+C_0}}\biggr).
\end{align*}
For sufficiently large $C_0$, sufficiently small $u_0$, and a sufficiently large constant in the sample condition, both lower bounds are at least $m^{-Cd^2}\exp(-c_0mu/2)$.  Hence,~\eqref{eq:iterative_return_sigma}
holds throughout $\tilde\sigma^2<\|\Delta\|_{\Sigma_1}^2\le C_\lambda\delta^2m^{C_G}$.

For each $t\ge1$, we apply~\eqref{eq:iterative_departure_sigma} and~\eqref{eq:iterative_return_sigma} conditioned on $\mathcal F_{t-1}$ and take expectation conditioned on $\Delta^{(0)}$ to obtain
\begin{align*}
\mathbb P\bigl(\|\Delta^{(t)}\|_{\Sigma_1}^2>\tilde\sigma^2\mid\Delta^{(0)}\bigr)
&=\mathbb E\biggl[
\mathbf 1\{\|\Delta^{(t-1)}\|_{\Sigma_1}^2\le\tilde\sigma^2\}
\mathbb P\bigl(\|\Delta^{(t)}\|_{\Sigma_1}^2>\tilde\sigma^2\mid\mathcal F_{t-1}\bigr)
\biggm|\Delta^{(0)}\biggr]\\
&+\mathbb E\biggl[
\mathbf 1\{\|\Delta^{(t-1)}\|_{\Sigma_1}^2>\tilde\sigma^2\}
\bigl\{1-\mathbb P\bigl(\|\Delta^{(t)}\|_{\Sigma_1}^2\le\tilde\sigma^2\mid\mathcal F_{t-1}\bigr)\bigr\}
\biggm|\Delta^{(0)}\biggr]\\
&\le2e^{Cd-c_0mu}
+\bigl(1-m^{-Cd^2}e^{-c_0mu/2}\bigr)
\mathbb P\bigl(\|\Delta^{(t-1)}\|_{\Sigma_1}^2>\tilde\sigma^2\mid\Delta^{(0)}\bigr).
\end{align*}
Since $\|\Delta^{(0)}\|_{\Sigma_1}\le\tilde\sigma$ by assumption, induction on $t$ results in
\begin{align*}
\mathbb P\bigl(\|\Delta^{(t)}\|_{\Sigma_1}^2>\tilde\sigma^2\mid\Delta^{(0)}\bigr)
&\le\frac{2e^{Cd-c_0mu}}{m^{-Cd^2}e^{-c_0mu/2}}
=2m^{Cd^2}e^{Cd-c_0mu/2}.
\end{align*}
Then, the  deterministic bound from Lemma~\ref{lem:iterative_error_bound} and this probability estimate imply
\begin{align*}
\mathbb E\bigl[\|\Delta^{(t)}\|_{\Sigma_1}^2
\mathbf 1\{\|\Delta^{(t)}\|_{\Sigma_1}^2>\tilde\sigma^2\}
\mid\Delta^{(0)}\bigr]&\le C\delta^2m^{C_G+Cd^2}e^{Cd-c_0mu/2}
\le C\delta^2e^{-cmu}.
\end{align*}
For a sufficiently large constant in the sample condition, we have $(C_G+Cd^2)\log m+Cd\le c_0mu/4$, which leads  to  the last inequality.
This proves the stated bound for every $t\ge0$.
\end{proof}

\subsection{Proof of Theorem~\ref{thm:iterative_full_finetuning}}

\begin{proof}
Fix $\Delta^{(0)}$ with $\|\Delta^{(0)}\|_{\Sigma_1}^2\le\tilde\sigma^2$, and let $t\ge1$.
When $\|\Delta^{(t-1)}\|_{\Sigma_1}^2\le\tilde\sigma^2$, we invoke Corollary~\ref{cor:iterative_one_generation_error} conditioned on $\mathcal F_{t-1}$ to obtain
\begin{align}\label{eq:iterative_conditional_recursion}
\mathbb E[\|\Delta^{(t)}\|_{\Sigma_1}^2\mid\mathcal F_{t-1}]
&\le\frac{2u^4}{9}\|\Delta^{(t-1)}\|_{\Sigma_1}^2
+\frac{Cd\delta^2}{mu}.
\end{align}

Next, we bound the contribution on $\{\|\Delta^{(t-1)}\|_{\Sigma_1}>\tilde\sigma\}$.
By splitting according to whether $\|\Delta^{(t)}\|_{\Sigma_1}>\tilde\sigma$, we have
\begin{align*}
\|\Delta^{(t)}\|_{\Sigma_1}^2
\mathbf 1\{\|\Delta^{(t-1)}\|_{\Sigma_1}^2>\tilde\sigma^2\}&\le\|\Delta^{(t)}\|_{\Sigma_1}^2
\mathbf 1\{\|\Delta^{(t)}\|_{\Sigma_1}^2>\tilde\sigma^2\}+\|\Delta^{(t-1)}\|_{\Sigma_1}^2
\mathbf 1\{\|\Delta^{(t-1)}\|_{\Sigma_1}^2>\tilde\sigma^2\}.
\end{align*}
We take expectation conditioned on $\Delta^{(0)}$ and use Lemma~\ref{lem:iterative_error_outside_radius} 
at $t$ and $t-1$ to obtain
\begin{align*}
\mathbb E\bigl[\|\Delta^{(t)}\|_{\Sigma_1}^2
\mathbf 1\{\|\Delta^{(t-1)}\|_{\Sigma_1}^2>\tilde\sigma^2\}
\mid\Delta^{(0)}\bigr]
&\le C\delta^2e^{-cmu}.
\end{align*}
We multiply~\eqref{eq:iterative_conditional_recursion} by $\mathbf 1\{\|\Delta^{(t-1)}\|_{\Sigma_1}\le\tilde\sigma\}$, take expectation conditioned on $\Delta^{(0)}$, and add the complementary contribution.
Since $e^{-cmu}\le C/(mu)$ and $d\ge1$, we obtain
\begin{align*}
\mathbb E[\|\Delta^{(t)}\|_{\Sigma_1}^2\mid\Delta^{(0)}]
&\le\frac{2u^4}{9}\mathbb E[\|\Delta^{(t-1)}\|_{\Sigma_1}^2\mid\Delta^{(0)}]
+\frac{Cd\delta^2}{mu}.
\end{align*}
Formally, we iterate this recursion and obtain
\begin{align*}
\mathbb E[\|\Delta^{(T)}\|_{\Sigma_1}^2\mid\Delta^{(0)}]
&\le\left(\frac{2u^4}{9}\right)^T\|\Delta^{(0)}\|_{\Sigma_1}^2
+\frac{1-(2u^4/9)^T}{1-2u^4/9}\frac{Cd\delta\tilde\sigma}{m}.
\end{align*}
\end{proof}

\subsection{Lower Bound}

For the lower bound, we bound the expected update in the direction $\Sigma_1\Delta^{(t-1)}$.
The following monotonicity property is used to obtain this bound.

\begin{lemma}[Monotonicity of the normalized conditional generation noise mean]
\label{lem:iterative_generation_noise_mean_monotonicity}
Let $\delta,\tilde\sigma>0$ and $\tilde\varepsilon\sim\mathcal N(0,\tilde\sigma^2)$.
For $a>0$, define $f(a):=\mathbb E[\tilde\varepsilon\mid|\tilde\varepsilon-a|\le\delta]/a$.
Then $f$ is nondecreasing on $(0,\infty)$ and satisfies $0\le f(a)\le1$ for every $a>0$.
\end{lemma}

\begin{proof}
For $a>0$, let $\xi=a-\tilde\varepsilon$.
Then $\xi\sim\mathcal N(a,\tilde\sigma^2)$, and $f(a)=\mathbb E[a-\xi\mid|\xi|\le\delta]/a$.
We calculate the Gaussian integrals and obtain
\begin{align*}
\int_{-\delta}^{\delta}(a-x)e^{-(x-a)^2/(2\tilde\sigma^2)}\,dx
&=\tilde\sigma^2\bigl(e^{-(\delta-a)^2/(2\tilde\sigma^2)}-e^{-(\delta+a)^2/(2\tilde\sigma^2)}\bigr)\\
&=a\int_{-\delta}^{\delta}e^{(x^2-\delta^2)/(2\tilde\sigma^2)}e^{-(x-a)^2/(2\tilde\sigma^2)}\,dx.
\end{align*}
After dividing by $a\int_{-\delta}^{\delta}e^{-(x-a)^2/(2\tilde\sigma^2)}\,dx$, we have
\begin{align*}
f(a)&=\mathbb E\biggl[\exp\biggl(\frac{\xi^2-\delta^2}{2\tilde\sigma^2}\biggr)\biggm|\,|\xi|\le\delta\biggr].
\end{align*}
Since $0<\exp((\xi^2-\delta^2)/(2\tilde\sigma^2))\le1$ when $|\xi|\le\delta$, this proves $0\le f(a)\le1$.

Next, we differentiate the numerator and denominator of this conditional expectation with respect to $a$ and obtain
\begin{align*}
f'(a)
&=\frac1{\tilde\sigma^2}\mathbb E\biggl[
e^{(\xi^2-\delta^2)/(2\tilde\sigma^2)}
\bigl(\xi-\mathbb E[\xi\mid|\xi|\le\delta]\bigr)
\biggm|\,|\xi|\le\delta\biggr]\\
&=\frac1{\tilde\sigma^2}\Cov\biggl(
e^{(\xi^2-\delta^2)/(2\tilde\sigma^2)},\xi
\biggm|\,|\xi|\le\delta\biggr).
\end{align*}
To determine the sign of this covariance, we condition further on $|\xi|$.
For $0<x\le\delta$, the density of $|\xi|$ at $x$ is the sum of the Gaussian densities of $\xi$ at $x$ and $-x$.
Hence,
\begin{align*}
\mathbb P(\xi>0\mid|\xi|=x)
&=\frac{e^{-(x-a)^2/(2\tilde\sigma^2)}}
{e^{-(x-a)^2/(2\tilde\sigma^2)}+e^{-(x+a)^2/(2\tilde\sigma^2)}}.
\end{align*}
Since $\xi$ takes the values $x$ and $-x$ conditioned on $|\xi|=x$, we obtain
\begin{align*}
\mathbb E[\xi\mid|\xi|=x]
&=x\bigl(2\mathbb P(\xi>0\mid|\xi|=x)-1\bigr)\\
&=x\frac{e^{-(x-a)^2/(2\tilde\sigma^2)}-e^{-(x+a)^2/(2\tilde\sigma^2)}}{e^{-(x-a)^2/(2\tilde\sigma^2)}+e^{-(x+a)^2/(2\tilde\sigma^2)}}
=x\frac{e^{2ax/\tilde\sigma^2}-1}{e^{2ax/\tilde\sigma^2}+1}.
\end{align*}
Note that the exponential in the covariance depends only on $|\xi|$.
Hence, taking conditional expectation over the sign of $\xi$, we obtain
\begin{align*}
f'(a)&=\frac1{\tilde\sigma^2}\Cov\biggl(
e^{(\xi^2-\delta^2)/(2\tilde\sigma^2)},
|\xi|\frac{e^{2a|\xi|/\tilde\sigma^2}-1}{e^{2a|\xi|/\tilde\sigma^2}+1}
\biggm|\,|\xi|\le\delta\biggr)\ge0,
\end{align*}
where the inequality follows since both expressions are nondecreasing functions of $|\xi|$.
Therefore, $f$ is nondecreasing on $(0,\infty)$, which completes the proof.
\end{proof}

We apply this monotonicity to the expected update $\Delta^{(t-1)}-\Delta^{(t)}$ and obtain the following conditional squared error lower bound.

\begin{lemma}[Conditional squared error lower bound after one generation]
\label{lem:iterative_empirical_error_lower_bound}
Under the iterative training model in Section~\ref{sec:iterative_training}, suppose that $s=d$.  
For $\Delta\in\mathbb R^d$, let $\theta_\Delta=\tilde\sigma^2(1-\lambda_{2,\Delta})/\{\tilde\sigma^2+\lambda_{2,\Delta}\|\Delta\|_{\Sigma_1}^2\}$.  
Then, for every integer $t\ge1$, almost surely,
\begin{align*}
\mathbb E[\|\Delta^{(t)}\|_{\Sigma_1}^2\mid\mathcal F_{t-1}]
&\ge(1-\theta_{\Delta^{(t-1)}})^2\|\Delta^{(t-1)}\|_{\Sigma_1}^2.
\end{align*}
\end{lemma}

\begin{proof}
Fix an integer $t\ge1$, condition on $\mathcal F_{t-1}$, and let $\Delta=\Delta^{(t-1)}$.
The claim is trivial when $\Delta=0$, and we assume $\Delta\ne0$.
It follows from Lemma~\ref{lem:iterative_error_bound} that $\mathbb E[\|\Delta^{(t)}\|_{\Sigma_1}^2\mid\mathcal F_{t-1}]<\infty$.
By Jensen's inequality and the Cauchy--Schwarz inequality, we have
\begin{align}
\mathbb E[\|\Delta^{(t)}\|_{\Sigma_1}^2\mid\mathcal F_{t-1}]
&\ge\bigl\|\mathbb E[\Delta^{(t)}\mid\mathcal F_{t-1}]\bigr\|_{\Sigma_1}^2
\ge\frac{\bigl(\Delta^\top\Sigma_1\mathbb E[\Delta^{(t)}\mid\mathcal F_{t-1}]\bigr)^2}{\|\Delta\|_{\Sigma_1}^2}\notag\\
&=\frac{\bigl(\|\Delta\|_{\Sigma_1}^2-\Delta^\top\Sigma_1\mathbb E[\Delta-\Delta^{(t)}\mid\mathcal F_{t-1}]\bigr)^2}{\|\Delta\|_{\Sigma_1}^2}.\label{eq:iterative_error_projection}
\end{align}
Since $0\le\theta_\Delta\le1$, equation~\eqref{eq:iterative_error_projection} shows that it suffices to prove
\begin{align}
\Delta^\top\Sigma_1\mathbb E[\Delta-\Delta^{(t)}\mid\mathcal F_{t-1}]
&\le\theta_\Delta\|\Delta\|_{\Sigma_1}^2.\label{eq:iterative_directional_target}
\end{align}
Recall the residual representation~\eqref{eq:residual_rep_generation_noise}.  By the definition of the extended estimator, we have
\begin{align}
\Delta-\Delta^{(t)}
&=\mathbf 1_{\Omega_{\mathrm{post}}}G(\Delta)^{-1}
\sum_{i=1}^m\mathbb I_i\tilde x_i\tilde\varepsilon_i,\label{eq:iterative_error_update}
\end{align}
where $\mathbf 1_{\Omega_{\mathrm{post}}}G(\Delta)^{-1}$ is interpreted as zero on $\Omega_{\mathrm{post}}^{\mathrm c}$.
By substituting~\eqref{eq:iterative_error_update} into the left hand side of~\eqref{eq:iterative_directional_target}, we obtain
\begin{align}
\Delta^\top\Sigma_1\mathbb E[\Delta-\Delta^{(t)}\mid\mathcal F_{t-1}]
&=\mathbb E\left[\mathbf 1_{\Omega_{\mathrm{post}}}\sum_{i=1}^m\mathbb I_i
(\tilde x_i^\top G(\Delta)^{-1}\Sigma_1\Delta)\tilde\varepsilon_i\middle|\mathcal F_{t-1}\right].\label{eq:iterative_directional_correction}
\end{align}

To evaluate the expectation in~\eqref{eq:iterative_directional_correction}, we condition further on $(\mathbb I_1,\ldots,\mathbb I_m)$ and let $k=\sum_{i=1}^m\mathbb I_i$.
If $k=0$, then $\Delta-\Delta^{(t)}=0$.
For $k\ge1$, we relabel the accepted samples as $1,\ldots,k$ and let $a_i=-\tilde x_i^\top\Delta$.
For the accepted samples, the pairs $(\tilde x_i,\tilde\varepsilon_i)$ are independent with common distribution equal to that of $(\tilde x,\tilde\varepsilon)$ conditioned on $(\mathbb I=1,\Delta)$.
Hence, $a_1,\ldots,a_k$ are independent and identically distributed and, since $\Delta\ne0$, are nonzero almost surely.

Recall from Lemma~\ref{lem:iterative_generation_noise_mean_monotonicity} that $f(a):=\mathbb E[\tilde\varepsilon\mid|\tilde\varepsilon-a|\le\delta]/a$ for $a>0$.
Since $\xi_i=a_i+\tilde\varepsilon_i$, the event $\mathbb I_i=1$ is equivalent to $|\tilde\varepsilon_i+a_i|\le\delta$.
By symmetry of the generation noise, we obtain, for $1\le i\le k$,
\begin{align}
\mathbb E[\tilde\varepsilon_i\mid a_i,\mathbb I_i=1,\Delta]
=\mathbb E[\tilde\varepsilon_i\mid|\tilde\varepsilon_i+a_i|\le\delta,a_i,\Delta]
=-\frac{a_i}{|a_i|}\mathbb E[\tilde\varepsilon\mid|\tilde\varepsilon-|a_i||\le\delta]=-a_i f(|a_i|).\label{eq:iterative_accepted_noise_mean}
\end{align}
Next, we evaluate $\mathbb E[\mathbf 1_{\Omega_{\mathrm{post}}}\tilde x_i^\top G(\Delta)^{-1}\Sigma_1\Delta\mid a_1,\ldots,a_k,\mathbb I_1,\ldots,\mathbb I_m,\Delta]$ for $1\le i\le k$.
Gaussian conditioning implies
\begin{align*}
\mathbb E[\tilde x_i\mid a_i,\mathbb I_i=1,\Delta]
&=-\frac{a_i\Sigma_1\Delta}{\|\Delta\|_{\Sigma_1}^2},
&\Cov(\tilde x_i\mid a_i,\mathbb I_i=1,\Delta)
&=\Sigma_1-\frac{\Sigma_1\Delta\Delta^\top\Sigma_1}{\|\Delta\|_{\Sigma_1}^2}.
\end{align*}
Conditioned on $a_i$, the indicator $\mathbb I_i$ depends only on $\tilde\varepsilon_i$, 
while $\tilde x_i+a_i\Sigma_1\Delta/\|\Delta\|_{\Sigma_1}^2$ is independent of $(a_i,\tilde\varepsilon_i)$.  
Thus, conditioned on $(a_1,\ldots,a_k,\mathbb I_1,\ldots,\mathbb I_m,\Delta)$, 
the vectors $(\tilde x_i+a_i\Sigma_1\Delta/\|\Delta\|_{\Sigma_1}^2)_{i=1}^k$ 
are independent centered Gaussian vectors with the covariance above and are independent of $(\tilde\varepsilon_i)_{i=1}^k$.

The definition of $\Omega_{\mathrm{post}}$ implies $\mathbf 1_{\Omega_{\mathrm{post}}}\sum_{i=1}^k(\tilde x_i^\top G(\Delta)^{-1}\Sigma_1\Delta)^2\le m^{C_G-1}\|\Sigma_1\Delta\|^2$. Hence, these conditional expectations are finite.
We show that $\mathbb E[\mathbf 1_{\Omega_{\mathrm{post}}}\tilde x_i^\top G(\Delta)^{-1}\Sigma_1\Delta\mid a_1,\ldots,a_k,\mathbb I_1,\ldots,\mathbb I_m,\Delta]$ is proportional to $a_i$, with the same proportionality constant for every $1\le i\le k$.

Consider an orthogonal matrix $U\in\mathbb R^{k\times k}$ satisfying $U(a_1,\ldots,a_k)^\top=(a_1,\ldots,a_k)^\top$.
Conditioned on $(a_1,\ldots,a_k,\mathbb I_1,\ldots,\mathbb I_m,\Delta)$, the conditional Gaussian distribution implies $(\sum_{j=1}^kU_{ij}\tilde x_j)_{i=1}^k\stackrel{d}{=}(\tilde x_i)_{i=1}^k$.
By orthogonality it holds that $\sum_{i,j,\ell=1}^kU_{ij}U_{i\ell}\tilde x_j\tilde x_\ell^\top=\sum_{j=1}^k\tilde x_j\tilde x_j^\top=G(\Delta)$, and hence $\Omega_{\mathrm{post}}$ is unchanged.
Since $G(\Delta)$ is unchanged, we have $(\sum_{j=1}^kU_{ij}\tilde x_j)^\top G(\Delta)^{-1}\Sigma_1\Delta=\sum_{j=1}^kU_{ij}(\tilde x_j^\top G(\Delta)^{-1}\Sigma_1\Delta)$.
Thus, the conditional expectation of $(\mathbf 1_{\Omega_{\mathrm{post}}}\tilde x_i^\top G(\Delta)^{-1}\Sigma_1\Delta)_{i=1}^k$ is unchanged by every orthogonal matrix fixing $(a_1,\ldots,a_k)^\top$.
Hence, the stated proportionality follows.
On $\Omega_{\mathrm{post}}$, we have
\begin{align*}
\sum_{i=1}^k a_i\tilde x_i^\top G(\Delta)^{-1}\Sigma_1\Delta
&=-\|\Delta\|_{\Sigma_1}^2.
\end{align*}
By multiplying this identity by $\mathbf 1_{\Omega_{\mathrm{post}}}$, taking conditional expectation, 
and using this proportionality, we obtain, for $1\le i\le k$,
\begin{align}
\mathbb E\left[\mathbf 1_{\Omega_{\mathrm{post}}}\tilde x_i^\top G(\Delta)^{-1}\Sigma_1\Delta
\middle|a_1,\ldots,a_k,\mathbb I_1,\ldots,\mathbb I_m,\Delta\right]
&=-\mathbb P(\Omega_{\mathrm{post}}\mid a_1,\ldots,a_k,\mathbb I_1,\ldots,\mathbb I_m,\Delta)
\frac{a_i\|\Delta\|_{\Sigma_1}^2}{\sum_{j=1}^k a_j^2}.\label{eq:iterative_inverse_gram_direction}
\end{align}

We take the inner product of~\eqref{eq:iterative_error_update} with $\Sigma_1\Delta$ and condition on
 $(a_1,\ldots,a_k,\mathbb I_1,\ldots,\mathbb I_m,\Delta)$.   
 Then, equations~\eqref{eq:iterative_accepted_noise_mean} and~\eqref{eq:iterative_inverse_gram_direction} result in
\begin{align}
&\Delta^\top\Sigma_1\mathbb E[\Delta-\Delta^{(t)}
\mid a_1,\ldots,a_k,\mathbb I_1,\ldots,\mathbb I_m,\Delta]\notag\\
&\quad=-\sum_{i=1}^k a_i f(|a_i|)\mathbb E\left[\mathbf 1_{\Omega_{\mathrm{post}}}
\tilde x_i^\top G(\Delta)^{-1}\Sigma_1\Delta
\middle|a_1,\ldots,a_k,\mathbb I_1,\ldots,\mathbb I_m,\Delta\right]\notag\\
&\quad=\|\Delta\|_{\Sigma_1}^2
\mathbb P(\Omega_{\mathrm{post}}\mid a_1,\ldots,a_k,\mathbb I_1,\ldots,\mathbb I_m,\Delta)
\frac{\sum_{i=1}^k a_i^2f(|a_i|)}{\sum_{i=1}^k a_i^2}\notag\\
&\quad\le\|\Delta\|_{\Sigma_1}^2
\frac{\sum_{i=1}^k a_i^2f(|a_i|)}{\sum_{i=1}^k a_i^2}. \label{eq:iterative_conditional_correction_ratio}
\end{align}
Thus, to prove~\eqref{eq:iterative_directional_target}, it remains to show that, for every realization of the indicators with $k\ge1$,
\begin{align}
\mathbb E\left[\left.\frac{\sum_{i=1}^k a_i^2f(|a_i|)}{\sum_{i=1}^k a_i^2}\right|\mathbb I_1,\ldots,\mathbb I_m,\Delta\right]
&\le\theta_\Delta.\label{eq:iterative_weighted_noise_target}
\end{align}

We express $\theta_\Delta$ using the conditional moments of $a_i$.  For $s=d$, the definition of $\bar\mu_{\mathrm{pop}}(\Delta)$ and Lemma~\ref{lem:common_population_mu} provide
\begin{align}
\mathbb E_{\tilde x,\tilde\varepsilon}[\mathbb I\tilde x\tilde\varepsilon\mid\Delta]
&=\bar\mu_{\mathrm{pop}}(\Delta)+\bar G_{\mathrm{pop}}(\Delta)\Delta=\theta_\Delta\bar G_{\mathrm{pop}}(\Delta)\Delta.\label{eq:iterative_population_direction}
\end{align}
We take the inner product in~\eqref{eq:iterative_population_direction} with $\Delta$ and apply~\eqref{eq:iterative_accepted_noise_mean} to obtain
\begin{align*}
\mathbb E_{\tilde x,\tilde\varepsilon}[\mathbb I(\tilde x^\top\Delta)^2f(|\tilde x^\top\Delta|)\mid\Delta]
&=\Delta^\top\mathbb E_{\tilde x,\tilde\varepsilon}[\mathbb I\tilde x\tilde\varepsilon\mid\Delta]
=\theta_\Delta\mathbb E_{\tilde x,\tilde\varepsilon}[\mathbb I(\tilde x^\top\Delta)^2\mid\Delta].
\end{align*}
Since $p_\Delta>0$, we divide the preceding identity by $p_\Delta$.  Conditioned on the indicators and $\Delta$, 
each accepted $a_i$ has the distribution of $-\tilde x^\top\Delta$ given $(\mathbb I=1,\Delta)$.  Hence, for every $1\le i\le k$,
\begin{align}
\mathbb E[a_i^2f(|a_i|)\mid\mathbb I_1,\ldots,\mathbb I_m,\Delta]
&=\theta_\Delta\mathbb E[a_i^2\mid\mathbb I_1,\ldots,\mathbb I_m,\Delta].\label{eq:iterative_accepted_projection_moment}
\end{align}

For fixed $a_2,\ldots,a_k$, we consider the conditional distribution of $a_1$ with density multiplied by $a_1^2/\mathbb E[a_1^2\mid\mathbb I_1,\ldots,\mathbb I_m,\Delta]$.
By~\eqref{eq:iterative_accepted_projection_moment}, the expectation of $f(|a_1|)$ under this distribution equals $\theta_\Delta$.
Moreover, Lemma~\ref{lem:iterative_generation_noise_mean_monotonicity} implies that $f(|a_1|)$ is nondecreasing in $a_1^2$, whereas $(a_1^2+\sum_{i=2}^k a_i^2)^{-1}$ is nonincreasing.
Hence, these two functions have nonpositive covariance. We obtain
\begin{align*}
&\mathbb E\left[\left.\frac{a_1^2f(|a_1|)}{\sum_{i=1}^k a_i^2}\right|a_2,\ldots,a_k,\mathbb I_1,\ldots,\mathbb I_m,\Delta\right]\\
&\quad\le\frac{\mathbb E[a_1^2f(|a_1|)\mid\mathbb I_1,\ldots,\mathbb I_m,\Delta]}{\mathbb E[a_1^2\mid\mathbb I_1,\ldots,\mathbb I_m,\Delta]}
\mathbb E\left[\left.\frac{a_1^2}{\sum_{i=1}^k a_i^2}\right|a_2,\ldots,a_k,\mathbb I_1,\ldots,\mathbb I_m,\Delta\right]\\
&\quad=\theta_\Delta\mathbb E\left[\left.\frac{a_1^2}{\sum_{i=1}^k a_i^2}\right|a_2,\ldots,a_k,\mathbb I_1,\ldots,\mathbb I_m,\Delta\right].
\end{align*}
The same calculation applies to every $1\le i\le k$.  By taking expectation over the remaining $a_i$ 
and summing over $i$, we have
\begin{align*}
\mathbb E\left[\left.\frac{\sum_{i=1}^k a_i^2f(|a_i|)}{\sum_{j=1}^k a_j^2}\right|\mathbb I_1,\ldots,\mathbb I_m,\Delta\right]
\le\theta_\Delta\sum_{i=1}^k\mathbb E\left[\left.\frac{a_i^2}{\sum_{j=1}^k a_j^2}\right|\mathbb I_1,\ldots,\mathbb I_m,\Delta\right]=\theta_\Delta.
\end{align*}
Thus, \eqref{eq:iterative_weighted_noise_target} holds.
We take expectation in~\eqref{eq:iterative_conditional_correction_ratio} over $a_1,\ldots,a_k$ and then the indicators
to prove \eqref{eq:iterative_directional_target}, with the case $k=0$ contributing zero.  
Substituting \eqref{eq:iterative_directional_target} into~\eqref{eq:iterative_error_projection} proves the lemma.
\end{proof}

\subsection{Proof of Theorem~\ref{thm:iterative_full_finetuning_lower_bound}}

\begin{proof}
For $T=0$, the stated bound holds with equality.  Consider $T\ge1$ and an initial error $\Delta^{(0)}\in\mathbb R^d$.
Since $s=d$, we have $\Delta_{-S}=0$.  
Lemma~\ref{lem:common_population_theta_bounds} implies, for every $\Delta\in\mathbb R^d$,
\begin{align*}
\theta_\Delta&\le\frac{\tilde\sigma^2}{\tilde\sigma^2+c\delta^2},\qquad 1-\theta_\Delta\ge\frac{c\delta^2}{\tilde\sigma^2+c\delta^2}=\frac{u^2}{u^2+C},
\end{align*}
where $C=1/c$ is universal.
For every $t\ge1$, Lemma~\ref{lem:iterative_empirical_error_lower_bound} implies
\begin{align*}
\mathbb E[\|\Delta^{(t)}\|_{\Sigma_1}^2\mid\mathcal F_{t-1}]
&\ge\biggl(\frac{u^2}{u^2+C}\biggr)^2\|\Delta^{(t-1)}\|_{\Sigma_1}^2.
\end{align*}
By taking expectation conditioned on $\Delta^{(0)}$ and applying the tower property, we have
\begin{align*}
\mathbb E[\|\Delta^{(t)}\|_{\Sigma_1}^2\mid\Delta^{(0)}]
&\ge\biggl(\frac{u^2}{u^2+C}\biggr)^2\mathbb E[\|\Delta^{(t-1)}\|_{\Sigma_1}^2\mid\Delta^{(0)}].
\end{align*}
Iterating this inequality for $t=1,\ldots,T$ proves the stated bound.
\end{proof}

%% file: Unified_appendix/app_17L_empirical_details.tex
\section{Experimental Details}\label{sec:empirical_details}

We provide additional details for the Gaussian simulations and real data
experiments in Section~\ref{sec:empirical_studies}.

\subsection{Gaussian Simulations}

In each run, we sample $\Delta\sim\mathcal N(0,I_d/n)$ directly rather than
fitting a pre-trained estimator from a generated dataset.  Conditioned on
$\Delta$, we first draw the number of accepted samples from
$\mathrm{Binomial}(m,p_\Delta)$ and then generate the accepted samples
from their joint conditional distribution given verification.  This procedure
has the same distribution as generating all $m$ candidate samples and
applying the verification rule.  Then, we fit the accepted responses by least
squares and  fix the frozen coordinates at their pre-trained values under
partial fine-tuning.

For Figure~\ref{fig:intro_verification}, panel~(a) varies $\delta$ over
$[3\times10^{-4},3]$ at $\tilde\sigma=1$, panel~(b) varies $\tilde\sigma$
over $[10^{-4},30]$ at $\delta=0.003$, and panel~(c) varies $\delta$ over
$[3\times10^{-5},0.05]$ with $\tilde\sigma=5\times10^{-4}$.  
For each setting, we determine the number of runs from the expected accepted
sample size.  Each point averages at least 50 runs, with at least 100 runs for
the threshold comparisons, and additional runs are used when fewer
samples are expected to be accepted.  Figure~\ref{fig:full_delta} excludes
five settings with the smallest thresholds for which at least one run had too
few accepted samples to fit the estimator.

\subsection{Real Data Experiments}

For unit norm sentence embeddings $v_1,v_2\in\mathbb R^{384}$, we construct
each pair covariate by concatenating $v_1\odot v_2$ and $|v_1-v_2|$, where both
operations are coordinatewise. We remove the six coordinates whose empirical
variance is below $10^{-12}$ times the largest coordinate variance, and the resulting dimension is
$d=762$.  For the data, we identify duplicates across the three datasets by comparing
covariates rounded to six decimal places.  Then, we remove from the
post-training data any sentence pairs that appear in the test data and from
the pre-training data any sentence pairs that appear in either the
post-training or test data.  After removing these duplicates, the pre-training,
post-training, and test data contain 4989, 7229, and 1379 sentence pairs,
respectively.
We center the covariates and responses using their means over the combined
pre-training and post-training data.  Therefore, human labels from the
post-training data are used in the common response centering, while test labels are
used only to evaluate the fitted estimators.  

For partial fine-tuning, we select 381 active coordinates using a fixed random
permutation.  We generate synthetic responses by adding Gaussian
noise to the pre-trained predictions without restricting the resulting scores
to $[0,5]$.  For each generation noise scale and fine-tuning setting, we reuse
the candidate covariates across runs, and within each run, we use the same
generated samples for all threshold comparisons.  For
Figure~\ref{fig:real_map}, we sample sentence pairs with replacement from the
post-training data and generate responses until 3048 samples pass verification.
The direct fitting reference fits the linear model directly to all 7229 human
labels in the post-training data using the ridge penalty coefficient
$10^{-8}\tr(X^\top X)/d$, where $X$ is the centered post-training covariate
matrix.  All fits in Figure~\ref{fig:real_map} were successful except under full fine-tuning at
$\delta/\sqrt{R_{\mathrm{pre}}}=\tilde\sigma/\sqrt{R_{\mathrm{pre}}}=0.05$.
At this setting, the minimum eigenvalue of the normalized active coordinate
Gram matrix exceeds $10^{-10}$ in seven of the 16 attempted fits, and the
average is computed over these seven fits.

%% file: reference.bib
@article{vershynin2019high,
  title={High-dimensional probability},
  author={Vershynin, Roman},
  journal={Cambridge Series in Statistical and Probabilistic Mathematics},
  volume={47},
  year={2019}
}

@book{wainwright2019high,
  title={High-dimensional statistics: A non-asymptotic viewpoint},
  author={Wainwright, Martin J},
  volume={48},
  year={2019},
  publisher={Cambridge university press}
}

@article{tropp2015introduction,
  title={An introduction to matrix concentration inequalities},
  author={Tropp, Joel A},
  journal={Foundations and trends in machine learning},
  volume={8},
  number={1-2},
  pages={1--230},
  year={2015},
  publisher={Emerald Publishing Limited}
}

@inproceedings{song2025mind,
  title={Mind the gap: Examining the self-improvement capabilities of large language models},
  author={Song, Yuda and Zhang, Hanlin and Eisenach, Carson and Kakade, Sham and Foster, Dean and Ghai, Udaya},
  booktitle={International Conference on Learning Representations},
  volume={2025},
  pages={39894--39931},
  year={2025}
}

@article{zelikman2022star,
  title={Star: Bootstrapping reasoning with reasoning},
  author={Zelikman, Eric and Wu, Yuhuai and Mu, Jesse and Goodman, Noah},
  journal={Advances in Neural Information Processing Systems},
  volume={35},
  pages={15476--15488},
  year={2022}
}

@article{shumailov2024ai,
  title={AI models collapse when trained on recursively generated data},
  author={Shumailov, Ilia and Shumaylov, Zakhar and Zhao, Yiren and Papernot, Nicolas and Anderson, Ross and Gal, Yarin},
  journal={Nature},
  volume={631},
  number={8022},
  pages={755--759},
  year={2024},
  publisher={Nature Publishing Group UK, London}
}

@article{gerstgrasser2024model,
  title={Is model collapse inevitable? breaking the curse of recursion by accumulating real and synthetic data},
  author={Gerstgrasser, Matthias and Schaeffer, Rylan and Dey, Apratim and Rafailov, Rafael and Sleight, Henry and Hughes, John and Korbak, Tomasz and Agrawal, Rajashree and Pai, Dhruv and Gromov, Andrey and others},
  journal={arXiv preprint arXiv:2404.01413},
  year={2024}
}

@inproceedings{dohmatob2025strong,
  title={Strong model collapse},
  author={Dohmatob, Elvis and Feng, Yunzhen and Subramonian, Arjun and Kempe, Julia},
  booktitle={International Conference on Learning Representations},
  volume={2025},
  pages={15656--15691},
  year={2025}
}

@article{dey2024universality,
  title={Universality of the $\pi^2/6$ pathway in avoiding model collapse},
  author={Dey, Apratim and Donoho, David},
  journal={arXiv preprint arXiv:2410.22812},
  year={2024}
}

@article{yang2026spend,
  title={Spend wisely: Maximizing post-training gains in iterative synthetic data bootstrapping},
  author={Yang, Pu and Feng, Yunzhen and Chen, Ziyuan and Wu, Yuhang and Li, Zhuoyuan},
  journal={Advances in Neural Information Processing Systems},
  volume={38},
  pages={158602--158639},
  year={2026}
}

@article{kanabar2025model,
  title={Model non-collapse: Minimax bounds for recursive discrete distribution estimation},
  author={Kanabar, Millen and Gastpar, Michael},
  journal={IEEE Transactions on Information Theory},
  year={2025},
  publisher={IEEE}
}

@article{amin2026escaping,
  title={Escaping collapse: The strength of weak data for large language model training},
  author={Amin, Kareem and Babakniya, Sara and Bie, Alex and Kong, Weiwei and Syed, Umar and Vassilvitskii, Sergei},
  journal={Advances in Neural Information Processing Systems},
  volume={38},
  pages={29345--29378},
  year={2026}
}

@article{he2026recursive,
  title={Recursive learning without collapse: a weighting-based stabilization framework},
  author={He, Hengzhi and Xu, Shirong and Cheng, Guang},
  journal={Journal of the Royal Statistical Society Series B: Statistical Methodology},
  pages={qkag099},
  year={2026},
  publisher={Oxford University Press, UK}
}

@article{xu2025probabilistic,
  title={A probabilistic perspective on model collapse},
  author={Xu, Shirong and He, Hengzhi and Cheng, Guang},
  journal={arXiv preprint arXiv:2505.13947},
  year={2025}
}

@inproceedings{garg2026preventing,
  title={Preventing model collapse under overparametrization: Optimal mixing ratios for interpolation learning and ridge regression},
  author={Garg, Anvit and Bhattacharya, Sohom and Sur, Pragya},
  booktitle={International Conference on Learning Representations},
  volume={2026},
  pages={146481--146509},
  year={2026}
}

@inproceedings{frei2022self,
  title={Self-training converts weak learners to strong learners in mixture models},
  author={Frei, Spencer and Zou, Difan and Chen, Zixiang and Gu, Quanquan},
  booktitle={International Conference on Artificial Intelligence and Statistics},
  pages={8003--8021},
  year={2022},
  organization={PMLR}
}

@article{das2024retraining,
  title={Retraining with predicted hard labels provably increases model accuracy},
  author={Das, Rudrajit and Dhillon, Inderjit S and Epasto, Alessandro and Javanmard, Adel and Mao, Jieming and Mirrokni, Vahab and Sanghavi, Sujay and Zhong, Peilin},
  journal={arXiv preprint arXiv:2406.11206},
  year={2024}
}

@inproceedings{feng2025beyond,
  title={Beyond model collapse: Scaling up with synthesized data requires verification},
  author={Feng, Yunzhen and Dohmatob, Elvis and Yang, Pu and Charton, Francois and Kempe, Julia},
  booktitle={International Conference on Learning Representations},
  volume={2025},
  pages={89702--89730},
  year={2025}
}

@article{ferbach2024self,
  title={Self-consuming generative models with curated data provably optimize human preferences},
  author={Ferbach, Damien and Bertrand, Quentin and Bose, Avishek Joey and Gidel, Gauthier},
  journal={arXiv preprint arXiv:2407.09499},
  year={2024}
}

@inproceedings{el2025maximizing,
  title={Maximizing the potential of synthetic data: Insights from random matrix theory},
  author={El Firdoussi, Aymane and Seddik, Mohamed El Amine and Hayou, Soufiane and Alami, Reda and Alzubaidi, Ahmed and Hacid, Hakim},
  booktitle={International Conference on Learning Representations},
  volume={2025},
  pages={46780--46815},
  year={2025}
}

@inproceedings{yi2026escaping,
  title={Escaping model collapse via synthetic data verification: Near-term improvements and long-term convergence},
  author={Yi, Bingji and Liu, Qiyuan and Cheng, Yuwei and Xu, Haifeng},
  booktitle={International Conference on Learning Representations},
  volume={2026},
  pages={24231--24260},
  year={2026}
}

@article{fu2026self,
  title={Self-verification provably prevents model collapse in recursive synthetic training},
  author={Fu, Shi and Wang, Yingjie and Chen, Yuzhu and Shen, Li and Tao, Dacheng},
  journal={Advances in Neural Information Processing Systems},
  volume={38},
  pages={36101--36154},
  year={2026}
}

@article{yuan2023scaling,
  title={Scaling relationship on learning mathematical reasoning with large language models},
  author={Yuan, Zheng and Yuan, Hongyi and Li, Chengpeng and Dong, Guanting and Lu, Keming and Tan, Chuanqi and Zhou, Chang and Zhou, Jingren},
  journal={arXiv preprint arXiv:2308.01825},
  year={2023}
}

@inproceedings{renze2024effect,
  title={The effect of sampling temperature on problem solving in large language models},
  author={Renze, Matthew},
  booktitle={Findings of the association for computational linguistics: EMNLP 2024},
  pages={7346--7356},
  year={2024}
}

@article{gulcehre2023reinforced,
  title={Reinforced self-training (rest) for language modeling},
  author={Gulcehre, Caglar and Paine, Tom Le and Srinivasan, Srivatsan and Konyushkova, Ksenia and Weerts, Lotte and Sharma, Abhishek and Siddhant, Aditya and Ahern, Alex and Wang, Miaosen and Gu, Chenjie and others},
  journal={arXiv preprint arXiv:2308.08998},
  year={2023}
}

@article{setlur2024rl,
  title={Rl on incorrect synthetic data scales the efficiency of llm math reasoning by eight-fold},
  author={Setlur, Amrith and Garg, Saurabh and Geng, Xinyang and Garg, Naman and Smith, Virginia and Kumar, Aviral},
  journal={Advances in Neural Information Processing Systems},
  volume={37},
  pages={43000--43031},
  year={2024}
}

@article{zhao2026self,
  title={Self-distilled reasoner: On-policy self-distillation for large language models},
  author={Zhao, Siyan and Xie, Zhihui and Liu, Mengchen and Huang, Jing and Pang, Guan and Chen, Feiyu and Grover, Aditya},
  journal={arXiv preprint arXiv:2601.18734},
  year={2026}
}

@article{zhao2026absolute,
  title={Absolute zero: Reinforced self-play reasoning with zero data},
  author={Zhao, Andrew and Wu, Yiran and Wu, Tong and Xu, Quentin and Yue, Yang and Lin, Matthieu and Wang, Shenzhi and Wu, Qingyun and Zheng, Zilong and Huang, Gao},
  journal={Advances in Neural Information Processing Systems},
  volume={38},
  pages={105816--105879},
  year={2026}
}

@inproceedings{wuself,
  title={Why Self-Distillation Helps and Hurts: Denoising vs. Signal Forgetting},
  author={Wu, Mingqi and Yang, Archer Yi and Sun, Qiang},
  booktitle={Forty-third International Conference on Machine Learning},
  year={2026}
}

@article{moniri2026mechanisms,
  title={On the mechanisms of weak-to-strong generalization: A theoretical perspective},
  author={Moniri, Behrad and Hassani, Hamed},
  journal={Advances in Neural Information Processing Systems},
  volume={38},
  pages={121409--121444},
  year={2026}
}

@inproceedings{das2023understanding,
  title={Understanding self-distillation in the presence of label noise},
  author={Das, Rudrajit and Sanghavi, Sujay},
  booktitle={International Conference on Machine Learning},
  pages={7102--7140},
  year={2023},
  organization={PMLR}
}

@article{lecoiu2026self,
  title={Self-Distillation is Optimal Among Spectral Shrinkage Estimators in Spiked Covariance Models},
  author={Lecoiu, Radu and Mukherjee, Debarghya and Sur, Pragya},
  journal={arXiv preprint arXiv:2605.17778},
  year={2026}
}

@article{dang2026prediction,
  title={Prediction-Only Distillation in Linear and Logistic Regression},
  author={Dang, Hien and Patil, Pratik and Rinaldo, Alessandro},
  journal={arXiv preprint arXiv:2607.15450},
  year={2026}
}

@article{zheng2023judging,
  title={Judging llm-as-a-judge with mt-bench and chatbot arena},
  author={Zheng, Lianmin and Chiang, Wei-Lin and Sheng, Ying and Zhuang, Siyuan and Wu, Zhanghao and Zhuang, Yonghao and Lin, Zi and Li, Zhuohan and Li, Dacheng and Xing, Eric and others},
  journal={Advances in neural information processing systems},
  volume={36},
  pages={46595--46623},
  year={2023}
}

@inproceedings{alemohammad2024self,
  title={Self-consuming generative models go mad},
  author={Alemohammad, Sina and Casco-Rodriguez, Josue and Luzi, Lorenzo and Humayun, Ahmed Imtiaz and Babaei, Hossein and LeJeune, Daniel and Siahkoohi, Ali and Baraniuk, Richard},
  booktitle={International Conference on Learning Representations},
  volume={2024},
  pages={53581--53608},
  year={2024}
}

@article{dohmatob2024model,
  title={Model collapse demystified: The case of regression},
  author={Dohmatob, Elvis and Feng, Yunzhen and Kempe, Julia},
  journal={Advances in Neural Information Processing Systems},
  volume={37},
  pages={46979--47013},
  year={2024}
}

@article{gu2026survey,
  title={A survey on llm-as-a-judge},
  author={Gu, Jiawei and Jiang, Xuhui and Shi, Zhichao and Tan, Hexiang and Zhai, Xuehao and Xu, Chengjin and Li, Wei and Shen, Yinghan and Ma, Shengjie and Liu, Honghao and others},
  journal={The Innovation},
  volume={7},
  number={6},
  year={2026},
  publisher={Elsevier}
}

@article{wang2020minilm,
  title={Minilm: Deep self-attention distillation for task-agnostic compression of pre-trained transformers},
  author={Wang, Wenhui and Wei, Furu and Dong, Li and Bao, Hangbo and Yang, Nan and Zhou, Ming},
  journal={Advances in neural information processing systems},
  volume={33},
  pages={5776--5788},
  year={2020}
}

@inproceedings{cer2017semeval,
  title={SemEval-2017 task 1: Semantic textual similarity multilingual and crosslingual focused evaluation},
  author={Cer, Daniel and Diab, Mona and Agirre, Eneko and Lopez-Gazpio, Inigo and Specia, Lucia},
  booktitle={Proceedings of the 11th international workshop on semantic evaluation (SemEval-2017)},
  pages={1--14},
  year={2017}
}

@inproceedings{agirre2012semeval,
    title = "{S}em{E}val-2012 Task 6: A Pilot on Semantic Textual Similarity",
    author = "Agirre, Eneko  and
      Cer, Daniel  and
      Diab, Mona  and
      Gonzalez-Agirre, Aitor",
    editor = "Agirre, Eneko  and
      Bos, Johan  and
      Diab, Mona  and
      Manandhar, Suresh  and
      Marton, Yuval  and
      Yuret, Deniz",
    booktitle = "*{SEM} 2012: The First Joint Conference on Lexical and Computational Semantics {--} Volume 1: Proceedings of the main conference and the shared task, and Volume 2: Proceedings of the Sixth International Workshop on Semantic Evaluation ({S}em{E}val 2012)",
    month = "7-8 " # jun,
    year = "2012",
    address = "Montr{\'e}al, Canada",
    publisher = "Association for Computational Linguistics",
    url = "https://aclanthology.org/S12-1051/",
    pages = "385--393"
}

@inproceedings{agirre2013sem,
  title={* SEM 2013 shared task: Semantic textual similarity},
  author={Agirre, Eneko and Cer, Daniel and Diab, Mona and Gonzalez-Agirre, Aitor and Guo, Weiwei},
  booktitle={Second joint conference on lexical and computational semantics (* SEM), volume 1: proceedings of the Main conference and the shared task: semantic textual similarity},
  pages={32--43},
  year={2013}
}

@inproceedings{agirre2014semeval,
  title={Semeval-2014 task 10: Multilingual semantic textual similarity},
  author={Agirre, Eneko and Banea, Carmen and Cardie, Claire and Cer, Daniel and Diab, Mona and Gonzalez-Agirre, Aitor and Guo, Weiwei and Mihalcea, Rada and Rigau, German and Wiebe, Janyce},
  booktitle={Proceedings of the 8th international workshop on semantic evaluation (SemEval 2014)},
  pages={81--91},
  year={2014}
}

@inproceedings{agirre2015semeval,
  title={Semeval-2015 task 2: Semantic textual similarity, english, spanish and pilot on interpretability},
  author={Agirre, Eneko and Banea, Carmen and Cardie, Claire and Cer, Daniel and Diab, Mona and Gonzalez-Agirre, Aitor and Guo, Weiwei and Lopez-Gazpio, Inigo and Maritxalar, Montse and Mihalcea, Rada and others},
  booktitle={Proceedings of the 9th international workshop on semantic evaluation (SemEval 2015)},
  pages={252--263},
  year={2015}
}

@inproceedings{houlsby2019parameter,
  title={Parameter-efficient transfer learning for NLP},
  author={Houlsby, Neil and Giurgiu, Andrei and Jastrzebski, Stanislaw and Morrone, Bruna and De Laroussilhe, Quentin and Gesmundo, Andrea and Attariyan, Mona and Gelly, Sylvain},
  booktitle={International conference on machine learning},
  pages={2790--2799},
  year={2019},
  organization={PMLR}
}

@inproceedings{rezaei2026high,
  title={High-dimensional analysis of synthetic data selection},
  author={Rezaei, Parham and Kova{\v{c}}evi{\'c}, Filip and Locatello, Francesco and Mondelli, Marco},
  booktitle={International Conference on Learning Representations},
  volume={2026},
  pages={94117--94168},
  year={2026}
}
